\documentclass{article}

\usepackage{microtype}
\usepackage{graphicx}
\usepackage{subfigure}
\usepackage{booktabs} 

\usepackage{hyperref}

\usepackage[accepted]{icml2023}

\usepackage{amsmath}
\usepackage{amssymb}
\usepackage{mathtools}
\usepackage{amsthm}

\usepackage[capitalize,noabbrev]{cleveref}

\usepackage{paralist}
\graphicspath{{figs/}{figs/mf}{figs/randomness}{figs/appendix_sweeps}{figs/single_seeds}{figs/mf/brax}{figs/mf/gym}{figs/mf/minigrid}{figs/randomness/brax}{figs/randomness/gym}{figs/randomness/minigrid}}

\usepackage{amsmath}
\usepackage{mathtools}  
\usepackage{amsfonts}

\DeclareMathOperator*{\expectation}{\mathbb{E}} 
\DeclareMathOperator*{\argmin}{arg\,min}        

\newcommand{\confspace}[0]{\pmb{\Lambda}}       
\newcommand{\conf}[0]{\pmb{\lambda}}            
\newcommand{\confopt}[0]{{\conf}^*}             

\newcommand{\cost}[0]{c}                        

\newcommand{\algo}[0]{A}                        

\newcommand{\statespace}[0]{\mathcal{S}}                

\newcommand{\policies}[0]{\mathbf{\Pi}}                 
\newcommand{\policy}[0]{\pi}                            
\newcommand{\policyopt}[0]{{\policy}^*}                 

\newcommand{\inst}[0]{i}                    
\newcommand{\insts}[0]{\mathcal{I}}         

\usepackage{pifont}
\newcommand{\cmark}{\ding{51}}%
\newcommand{\xmark}{\ding{55}}%
\usepackage[font=small,skip=4pt]{caption}
\usepackage{wrapfig}
\usepackage{multirow}

\theoremstyle{plain}
\newtheorem{theorem}{Theorem}[section]

\theoremstyle{definition}
\newtheorem{definition}[theorem]{Definition}

\theoremstyle{remark}

\usepackage[textsize=tiny]{todonotes}

\icmltitlerunning{\hfill Hyperparameters in RL and How To Tune Them \hfill \thepage}

\begin{document}

\twocolumn[
\icmltitle{Hyperparameters in Reinforcement Learning and How To Tune Them}



\icmlsetsymbol{equal}{*}

\begin{icmlauthorlist}
\icmlauthor{Theresa Eimer}{luh,equal}
\icmlauthor{Marius Lindauer}{luh}
\icmlauthor{Roberta Raileanu}{fair}
\end{icmlauthorlist}

\icmlaffiliation{luh}{Leibniz University Hannover}
\icmlaffiliation{fair}{Meta AI}

\icmlcorrespondingauthor{Theresa Eimer}{t.eimer@ai.uni-hannover.de}

\icmlkeywords{AutoRL, AutoML, Reinforcement Learning, Reproducibility, Hyperparameter Optimization}

\vskip 0.3in
]



\newcommand{\internshipnotice}{\textsuperscript{*}Work was done during an internship at Meta AI.}
\printAffiliationsAndNotice{\internshipnotice}  

\begin{abstract}
In order to improve reproducibility, deep reinforcement learning (RL) has been adopting better scientific practices such as standardized evaluation metrics and reporting. 
However, the process of hyperparameter optimization still varies widely across papers, which makes it challenging to compare RL algorithms fairly.  
In this paper, we show that hyperparameter choices in RL can significantly affect the agent's final performance and sample efficiency, and that the hyperparameter landscape can strongly depend on the tuning seed which may lead to overfitting. 
We therefore propose adopting established best practices from AutoML, such as the separation of tuning and testing seeds, as well as principled hyperparameter optimization (HPO) across a broad search space. 
We support this by comparing multiple state-of-the-art HPO tools on a range of RL algorithms and environments to their hand-tuned counterparts, demonstrating that HPO approaches often have higher performance and lower compute overhead. 
As a result of our findings, we recommend a set of best practices for the RL community, which should result in stronger empirical results with fewer computational costs, better reproducibility, and thus faster progress.
In order to encourage the adoption of these practices, we provide plug-and-play implementations of the tuning algorithms used in this paper at \url{https://github.com/facebookresearch/how-to-autorl}.
\end{abstract}

\section{Introduction}
Deep reinforcement Learning (RL) algorithms contain a number of design decisions and hyperparameter settings, many of which have a critical influence on the learning speed and success of the algorithm.
While design decisions and implementation details have received greater attention in the last years~\cite{henderson-aaai18a,engstrom-iclr20,hsu-corr20,andrychowicz-iclr21,obandoceron-icml21}, the same is less true of RL hyperparameters.
Progress in self-adapting algorithms~\cite{zahavy-neurips20}, RL-specific hyperparameter optimization tools~\cite{franke-arxiv20a,wan-automl22}, and meta-learnt hyperparameters~\cite{flennerhag-iclr22} has not yet been adopted by RL practitioners. 
In fact, most papers only report final model hyperparameters or grid search sweeps known to be suboptimal and costly compared to even simple Hyperparameter Optimization (HPO) baselines like random search~\cite{bergstra-jmlr12a}. In addition, the seeds used for tuning and evaluation are rarely reported, leaving it unclear if the hyperparameters were tuned on the test seeds, which is -- as we will show -- a major reproducibility issue.
In this paper, we aim to lay out and address the potential causes for the lack of adoption of HPO methods in the RL community.

\begin{figure}
    \centering
    \includegraphics[width=0.4\textwidth]{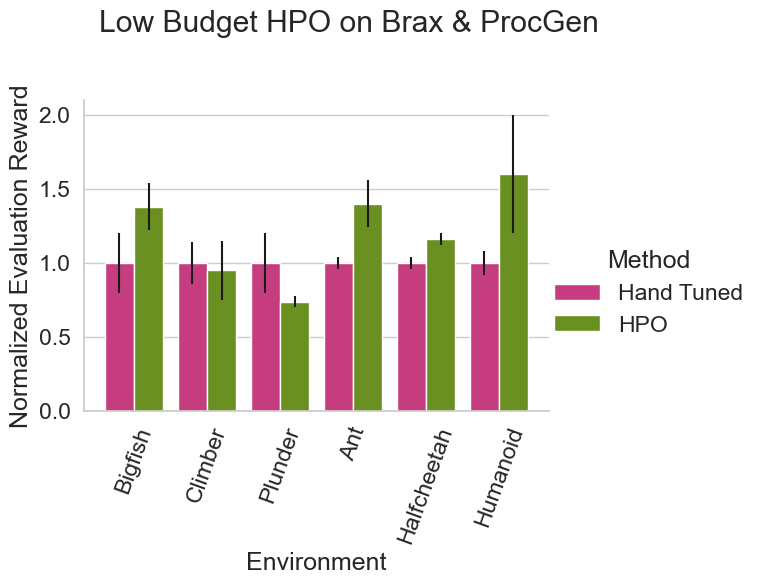}
    \caption{\textbf{Comparison of Hyperparameter Tuning Approaches:} state-of-the-art hyperparameter optimization packages match or outperform hand tuning via grid search, while using less than 1/12 of the budget.}
    \label{fig:demo_fig}
\end{figure}

\textbf{Underestimation of Hyperparameter Influence}
While it has been previously shown that hyperparameters are important to an RL algorithm's success~\citep{henderson-aaai18a, engstrom-iclr20, andrychowicz-iclr21}, the impact of even seemingly irrelevant hyperparameters is still underestimated by the community, as indicated by the fact that many papers tune only two or three hyperparameters~\cite{schulman-arxiv17a,berner-corr19,badia-icml20,hambro-neurips22}.
We show that even often overlooked hyperparameters can make or break an algorithm's success, meaning that careful consideration is necessary for a broad range of hyperparameters. 
This is especially important for as-of-yet unexplored domains, as pointed out by~\citet{zhang-aistats21}. 
Furthermore, hyperparameters cause different algorithm behaviors depending on the random seed which is a well-known fact in AutoML~\cite{eggensperger-jair19a,lindauer-jmlr20a} but has not yet factored widely into RL research, negatively impacting reproducibility.


\textbf{Fractured State of the Art in AutoRL}
Even though HPO approaches have succeeded in tuning RL algorithms~\cite{franke-iclr21,awad-ijcai21,zhang-aistats21,wan-automl22}, the costs and benefits of HPO are relatively unknown in the community.
AutoRL papers often compare only a few HPO methods, are limited to single domains or toy problems, or use a single RL algorithm~\cite{jaderberg-arxiv17a,parkerholder-neurips20,awad-ijcai21,kiran-corr22,wan-automl22}. 
In this work, we aim to understand the need for and challenges of AutoRL by comparing multiple HPO methods across various state-of-the-art RL algorithms on challenging environments. Our results demonstrate that HPO approaches have better performance and less compute overhead than hyperparameter sweeps or grid searches which are typically used in the RL community (see ~\cref{fig:demo_fig}).

\textbf{Ease of Use}
State-of-the-art AutoML tools are often released as research papers rather than standalone packages. In addition, they are not immediately compatible with standard RL code, while easy to use solutions like Optuna~\cite{optuna} or Ax~\cite{Ax} only provide a limited selection of HPO aproaches.
To improve the availability of these tools, we provide Hydra sweepers~\cite{hydra} for several variations of population-based methods, such as standard PBT~\cite{jaderberg-arxiv17a}, PB2~\cite{parkerholder-neurips20} and BGT~\cite{wan-automl22}, as well as the evolutionary algorithm DEHB~\cite{awad-ijcai21}. Note that all of these have shown to improve over random search for tuning RL agents.
As black-box methods, they are compatible with any RL algorithm or environment and due to Hydra, users do not have to change their implementation besides returning a success metric like the reward once training is finished.
Based on our empirical insights, we provide best practice guidelines on how to use HPO for RL.

In this paper, we demonstrate that \textbf{compared to tuning hyperparameters by hand, existing HPO tools are capable of producing better performing, more stable, and more easily comparable RL agents, while using fewer computational resources}. 
We believe widespread adoption of HPO protocols within the RL community will therefore result in more accurate and fair comparisons across RL methods and in the end to faster progress.

To summarize, \textbf{our contributions} are:
\begin{enumerate}
    \item Exploration of the hyperparameter landscape for commonly-used RL algorithms and environments;
    \item Comparison of different types of HPO methods on state-of-the-art RL algorithms and challenging RL environments;
    \item Open-source implementations of advanced HPO methods that can easily be used with any RL algorithm and environment; and
    \item Best practice recommendations for HPO in RL.
\end{enumerate}

\section{The Hyperparameter Optimization Problem}
We provide an overview of the most relevant formalizations of HPO in RL, Algorithm Configuration~\cite{schede-jair22} and Dynamic Algorithm Configuration~\cite{adriaensen-jair22}.
Algorithm Configuration (AC) is a popular paradigm for optimizing hyperparameters of several different kinds of algorithms~\citep{eggensperger-jair19a}.
\begin{definition}[AC]
Given an algorithm $\algo$, a hyperparameter space $\confspace$, as well as a distribution of environments or environment instances $\insts$, and a cost function $\cost$, find the optimal configuration $\confopt \in \confspace$ across possible tasks s.t.:

\centering
$\confopt \in \argmin_{\conf \in \confspace} \expectation_{\inst \sim \insts} \left[\cost(\algo(\inst; \conf))\right]$.
\end{definition}
The cost function could be the negative of the agent's reward or a failure indicator across a distribution of tasks.
Thus it is quite flexible and can accommodate a diverse set of possible goals for algorithm performance.
This definition is not restricted to one train and test setting but aims to achieve the best possible performance across a range of environments or environment instances.
AC approaches thus strive to avoid overfitting the hyperparameters to a specific scenario.
Even for RL problems focusing on generalization, AC is therefore a suitable framework.
Commonly, the HPO process is terminated before we have found the true $\confopt$ via an optimization budget (e.g. the runtime or number of training steps). The best found hyperparameter configuration found by the optimization process is called the incumbent.

Another relevant paradigm for tuning RL is Dynamic Algorithm Configuration (DAC) \cite{biedenkapp-ecai20,adriaensen-jair22}.
DAC is a generalization of AC that does not search for a single optimal hyperparameter value per algorithm run but instead for a sequence of values.
\begin{definition}[DAC]
Given an algorithm $\algo$, a hyperparameter space $\confspace$ as well as a distribution of environments or environment instances $\insts$ with state space $\statespace$, cost function $\cost$ and a space of dynamic configuration policies $\policies$ with each $\policy \in \policies : \statespace \times \insts \to \confspace$, find $\policyopt \in \policies$ s.t.:

\centering
$\policyopt \in \argmin_{\policy \in \policies} \expectation_{\inst \sim \insts}\cost(\algo(\inst; \policy))$
\end{definition}
As RL is a dynamic optimization process, it can benefit from dynamic changes in the hyperparameter values such as learning rate schedules~\cite{zhang-aistats21,parkerholder-jair22}.
Thus HPO tools developed specifically for RL have been following the DAC paradigm in order to tailor the hyperparameter values closer to the training progress \cite{franke-arxiv20a,zhang-aistats21,wan-automl22}.

It is worth noting that while the model architecture can be defined by a set of hyperparameters like the number of layers, architecture search is generally more complex and thus separately defined as the NAS problem or combined with HPO to form the general AutoDL problem \cite{zimmer-tpami21}.
While some tools include options for optimizing architecture hyperparameters, insights into how to find good architectures for RL are out of scope for this paper.

\section{Related Work}
While RL as a field has seen many innovations in the last years, small changes to the algorithm or its implementation can have a big impact on its results~\cite{henderson-aaai18a,andrychowicz-iclr21,engstrom-iclr20}.
In an effort to consolidate these innovations, several papers have examined the effect of smaller design decisions like the loss function or policy regularization for on-policy algorithms~\cite{hsu-corr20,andrychowicz-iclr21}, DQN~\cite{obandoceron-icml21} and offline RL~\cite{zhang-neurips21}.
AutoRL methods, on the other hand, have focused on automating and abstracting some of these decisions~\cite{parkerholder-jair22} by using data-driven approaches to learn various algorithmic components~\cite{bechtle-icpr20,xu-neurips20,metz-corr22} or even entire RL algorithms~\cite{wang-icml16a,duan-corr16,coreyes-iclr21,lu-corr22}.

While overall there has been less interest in hyperparameter optimization, some RL-specific HPO algorithms have been developed.
STACX~\cite{zahavy-neurips20} is an example of a self-tuning algorithm, using meta-gradients~\cite{xu-neurips18} to optimize its hyperparameters during runtime. 
This idea has recently been generalized to bootstrapped meta-learning, enabling the use of meta-gradients to learn any combination of hyperparameters on most RL algorithms on the fly~\cite{flennerhag-iclr22}. 
Such gradient-based approaches are fairly general and have shown a lot of promise~\cite{paul-neurips19}. 
However, they require access to the algorithm's gradients, thus limiting their use and incurring a larger compute overhead.
In this paper, we focus on purely black-box methods for their ease of use in any RL setting.

Extensions of population-based training (PBT)~\cite{jaderberg-arxiv17a,li-kdd19} improvements like BO kernels~\cite{parkerholder-neurips20} or added NAS components~\cite{franke-arxiv20a,wan-automl22} have led to significant performance and efficiency gains, offering a RL-specific way of optimizing hyperparameters during training.
A benefit of PBT methods is that they implicitly find a schedule of hyperparameter settings instead of a fixed value.

Beyond PBT methods, many general AC algorithms have proven to perform well on ML and RL tasks~\cite{schede-jair22}. A few such examples are SMAC~\cite{lindauer-jmlr22} and DEHB~\cite{awad-ijcai21} which are based on Bayesian Optimization and evolutionary algorithms, respectively.
SMAC is model-based (i.e. it learns a model of the hyperparameter landscape using a Gaussian process) and both are multi-fidelity methods (i.e. they utilize shorter training runs to test many different configurations, only progressing the best ones).
While these algorithms have rarely been used in RL so far, there is no evidence to suggest they perform any worse than RL-specific optimization approaches.
In fact, a possible advantage of multi-fidelity approaches over population-based ones is 
that given the same budget, multi-fidelity methods see a larger number of total configurations, while population-based ones see a smaller number of configurations trained for a longer time.

\section{The Hyperparameter Landscape of RL}
\label{sec:exp1}
Before comparing HPO algorithms, we empirically motivate why using dedicated tuning tools is important in RL.
To this end we study the effect of hyperparameters as well as that of the random seed on the final performance of RL algorithms. We also investigate the smoothness of the hyperparameter space. 
The goal of this section is not to achieve the best possible results on each task but to gather insights into how hyperparameters affect RL algorithms and how we can optimize them effectively.

\textbf{Experimental Setup}
To gain robust insights into the impact of hyperparameters on the performance of an RL agent, we consider a range of widely-used environments and algorithms. We use basic gym environments such as OpenAI's Pendulum and Acrobot~\cite{gym}, gridworld with an exploration component such as MiniGrid's Empty and DoorKey 5x5~\cite{minigrid}, as well as robot locomotion tasks such as Brax's Ant, Halfcheetah and Humanoid~\cite{freeman-neurips21}.
We use PPO~\cite{schulman-arxiv17a} and DQN~\cite{mnih-nature13} for the discrete environments, and PPO as well as SAC~\cite{haarnoja-icml18} for the continuous ones, all in their \texttt{StableBaselines3} implementations~\cite{sb3}. 
This selection is representative of the main classes of model-free RL algorithms (i.e. on-policy policy-optimization, off-policy value-based, and off-policy actor-critic) and covers a diverse set of tasks posing different challenges (i.e. discrete and continuous control), allowing us to draw meaningful and generalizable conclusions.

For each environment, we sweep over $8$ hyperparameters for DQN, $7$ for SAC and $11$ for PPO (for a full list, see Appendix~\ref{app:configurations}).
We run each combination of hyperparameter setting, algorithm and environment for $5$ different random seeds. For brevity's sake, we focus on the PPO results in the main paper. The results on the other algorithms lead to similar conclusions and can be found in Appendix~\ref{app:more_pointplots}.


For the tuning insights in this section, we use random search (RS) in its Optuna implementation~\cite{optuna}, a multi-fidelity method called DEHB~\cite{awad-ijcai21} and a PBT approach called PB2~\cite{parkerholder-neurips20}. 
Although grid search is certainly more commonly-used in RL than RS, we do not include it as a baseline due to its major disadvantages relative to RS such as its poor scaling with the size of the search space and heavy reliance on domain knowledge~\citep{bergstra-jmlr12a}.
We choose DEHB and PB2 as two standard incarnations of multi-fidelity and PBT methods without any extensions like run initialization~\cite{wan-automl22} or configuration racing~\cite{lindauer-jmlr22} because we want to test how well lightweight vanilla versions of these algorithm classes perform on RL.
 We use a total budget of $10$ full RL runs for all methods. 
For more background on these methods as well as their own hyperparameter settings, see Appendix~\ref{app:more_background}.
A complete overview of search spaces and experiment settings can be found in Appendix~\ref{app:search_spaces}.
The code for all experiments in this paper can be found at \url{https://github.com/facebookresearch/how-to-autorl}.

\subsection{Which RL Hyperparameters Should Be Tuned?}
Our goal is not to find good default hyperparameter settings (see Appendix~\ref{app:configurations} for our reasoning) or gain insights into why some configurations perform a certain way. Instead, we are interested in their general relevance, i.e., the effect size for hyperparameter tuning.
Thus, we run sweeps over our chosen hyperparameters for each environment and algorithm to get an impression of which hyperparameters are important in each setting. See Appendix \ref{app:more_sweeps} for the full results.

\begin{figure}
    \centering
    \includegraphics[width=0.23\textwidth]{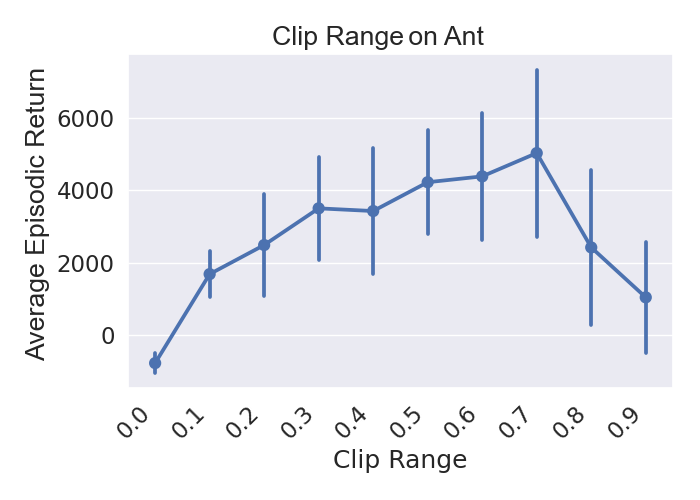}
    \includegraphics[width=0.23\textwidth]{boxplot_halfcheetah_PPO_algorithm.model_kwargs.ent_coef}
    \includegraphics[width=0.23\textwidth]{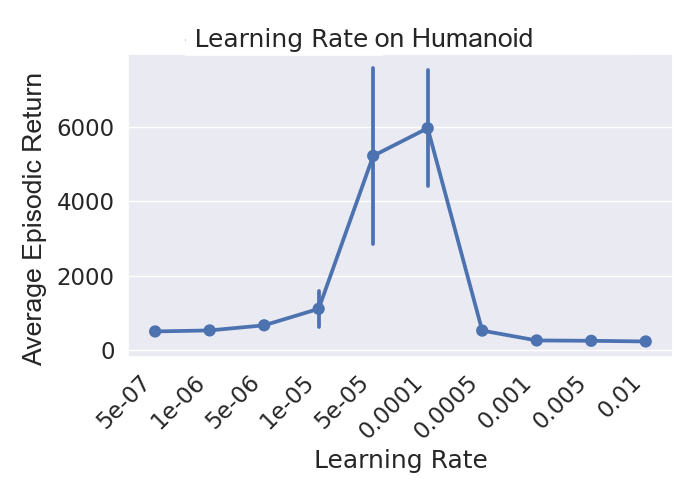}
    \includegraphics[width=0.23\textwidth]{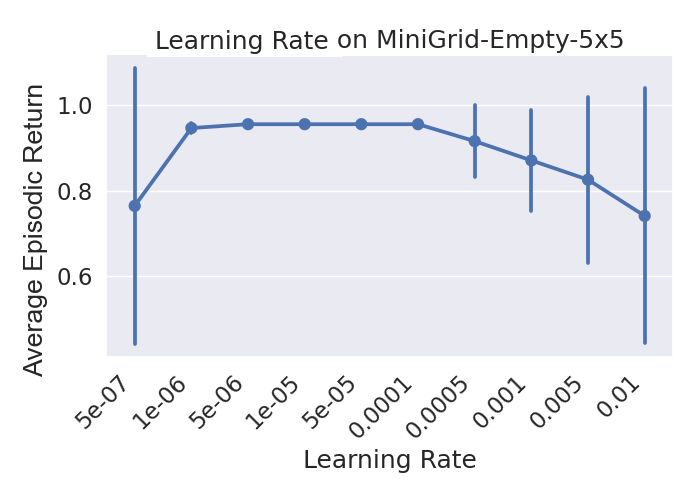}
    \caption{Hyperparameter landscapes of learning rate, clip range and entropy coefficient for PPO on Brax and MiniGrid. For each hyperparameter value, we report the average final return and standard deviation across 5 seeds.}
    \label{fig:ppo_boxplots}
\end{figure}
In Figure~\ref{fig:ppo_boxplots}, we can see a \textit{large influence on the final performance of almost every hyperparameter we sweep over for each environment.}
Even the rarely tuned clip range can be a deciding factor between an agent succeeding or failing in an environment such as in Ant. 
In many cases, hyperparameters can also have a large effect on the algorithm's stability throughout training. 
In total, we observed only the worst hyperparameter choice being within the best choice's standard deviation $7$ times out of $126$ settings and only $13$ times the median performance dropping by less than $20\%$. 
At the same time, hyperparameter importance analysis using fANOVA~\cite{hutter-icml14a} shows that one or two hyperparameters monopolize the importance on each environment - though they tend to vary from environment to environment (e.g.fro PPO learning rate on Acrobot, clip range on Pendulum and the GAE lambda on MiniGrid - see Appendix~\ref{app:importance} for full results).
Additionally, Partial Dependency Plots on Pendulum and Acrobot (see Appendix~\ref{app:pdps}) show that there are almost no complex interaction patterns between the hyperparameters which would increase the difficulty when tuning all of them at the same time.
Since most hyperparameters have significant influences on performance, their importance varies across environments and there are only few interference effects, we recommend tuning as many hyperparameters as possible -- as is best practice in the AutoML community~\cite{eggensperger-jair19a}.

This result suggests that common grid search approaches are likely suboptimal as good search space coverage along many  dimensions is highly expensive~\cite{bergstra-jmlr12a}. 
In order to empirically test if current HPO tools are well suited to such a set of diverse hyperparameters, we tune our algorithms using differently sized search spaces: (i) only the learning rate (which could be hand-tuned), (ii) a small space with three hyperparameters (which would be expensive but possible to tune manually) and (iii) the full search space of $7$ hyperparameters for SAC, $9$ for DQN, and $11$ for PPO (which is too large to feasibly tune by hand - sweeping $7$ hyperparameters with only three values amounts to a grid search of $2187$ runs).

\def\arraystretch{1}
\setlength\tabcolsep{1pt}
\begin{table}
\centering
        \caption{Tuning PPO on Acrobot (top) and SAC on Pendulum (bottom) across different search space sizes (i.e. only learning rate, $\{$learning rate, entropy coefficient, training epochs$\}$, and full search space). Shown is the negative evaluation reward across $5$ tuning runs. Lower numbers are better, best performance on each environment is highlighted. The best final performance on a single seed from our sweeps is also reported.}
    \begin{tabular}{ll|ccc|c}
        \toprule
             & & DEHB Inc. &  PB2 Inc. & RS Inc. &  Sweep \\
         \midrule
         \parbox[t]{3mm}{\multirow{3}{*}{\rotatebox[origin=c]{90}{\small Acrobot}}} & LR Only &$\mathbf{71 \pm 1}$ & $94 \pm 22$ & $78\pm 5$ & $81$\\
         & Small & $\mathbf{72 \pm 1}$ & $193 \pm 160$ & $80 \pm 6$ & \\
         & Full & $\mathbf{71 \pm 3}$ & $305 \pm 186$ & $83 \pm 5$ &\\
         \midrule
         \parbox[t]{3mm}{\multirow{3}{*}{\rotatebox[origin=c]{90}{\small Pendulum}}}& LR Only & $\mathbf{71 \pm 12}$& $207 \pm 126$ & $89 \pm 25$ & $117 $\\
         & Small & $119 \pm 12$ & $106 \pm 12$ & $401\pm 363$ &\\
         & Full & $112 \pm 24$ & $78 \pm 19$ & $144\pm 48$ & \\
         \bottomrule
    \end{tabular}
    \label{tab:search_spaces_table}
\end{table}

In Table~\ref{tab:search_spaces_table} we see that RS performs well on Acrobot but it falls short on Pendulum, displaying large discrepancies across seeds, some performing well, and some failing to find a good configuration. 
While this is a typical failure case of RS, this does not mean RS is a weak candidate, ranking second overall by outperforming PB2 in several cases.
PB2 is also quite unreliable: on Acrobot, its performance decreases with the size of the search space; on Pendulum, however, it improves with the size of the search space. 
As with RS, part of the underlying issue is the inconsistent performance of PB2. 
Note that the incumbent configuration is fairly static across all PB2 runs for the larger search spaces.
In most cases, the configuration changes at most once during training, showing that PB2 currently does not take full advantage of its ability to find dynamic schedules.
DEHB is the most stable in terms of standard deviation across seeds, even though we see a slight decrease in performance on Pendulum with larger search spaces.

\begin{figure}
    \centering
     \includegraphics[width=0.23\textwidth]{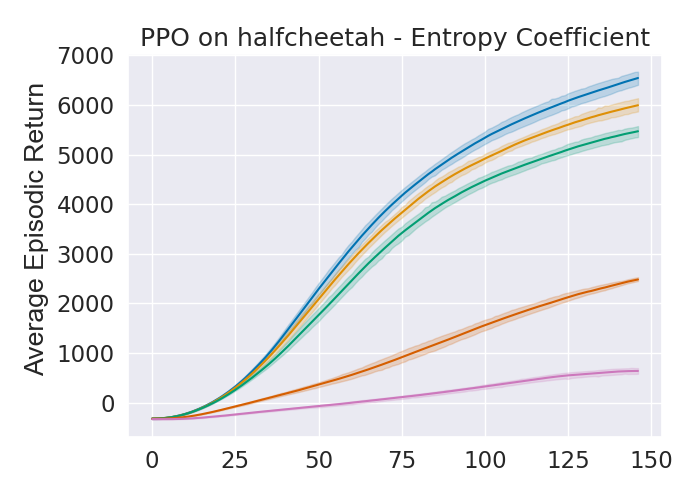}
    \includegraphics[width=0.23\textwidth]{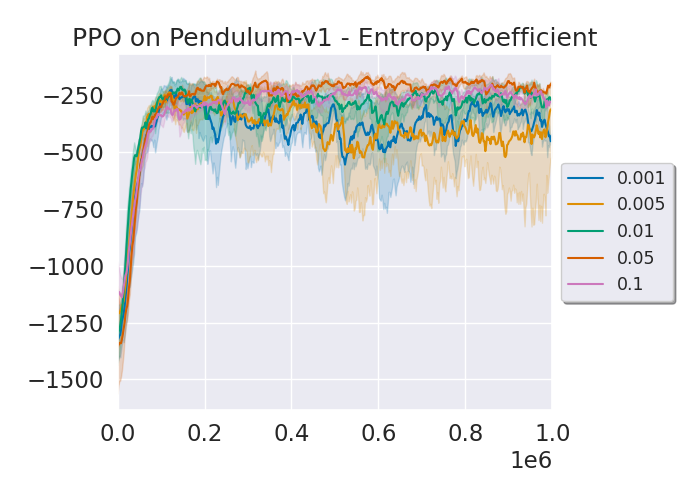}
    \includegraphics[width=0.23\textwidth]{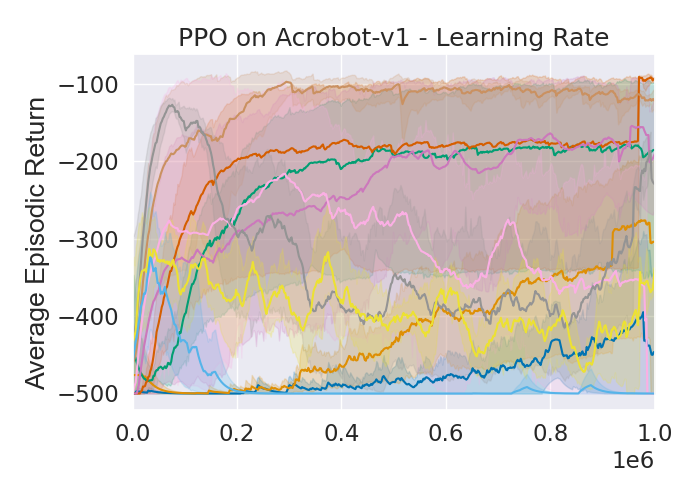}
    \includegraphics[width=0.23\textwidth]{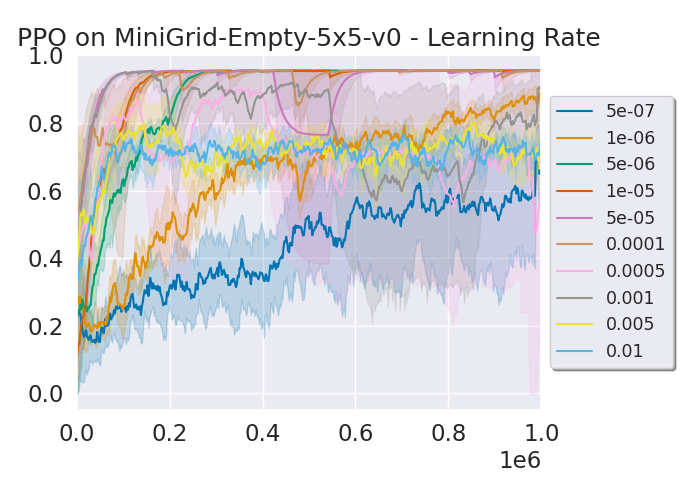}
    \includegraphics[width=0.23\textwidth]{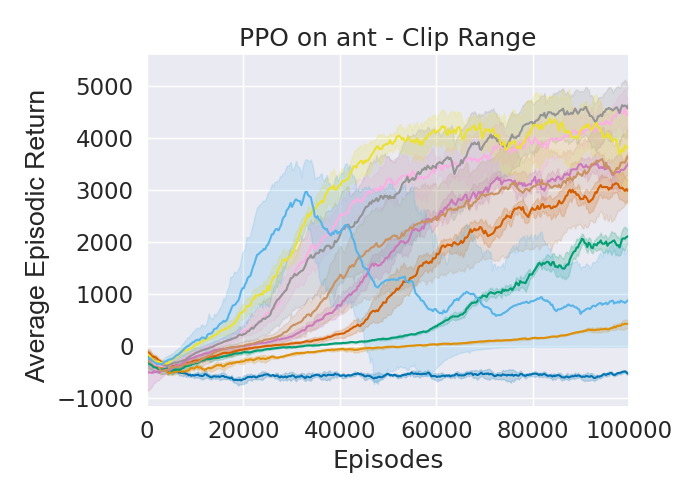}
    \includegraphics[width=0.23\textwidth]{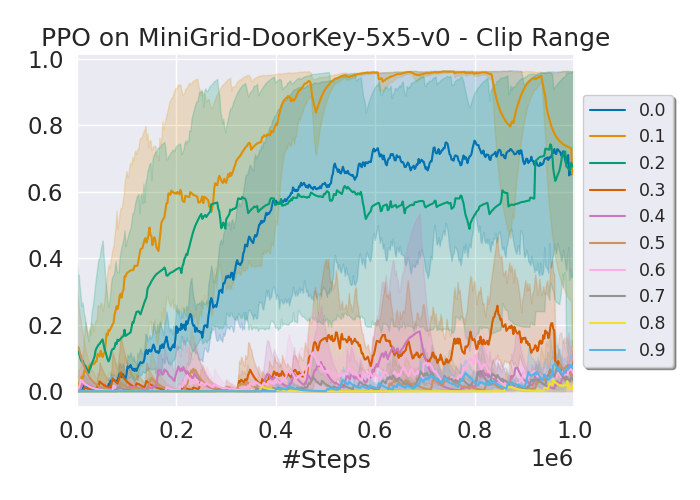}
    \caption{Hyperparameter Sweeps for PPO across learning rates, entropy coefficients and clip ranges on various environments. The mean and standard deviation are computed across $5$ seeds.}
    \label{fig:ppo_sweeps}
\end{figure}

Overall we see that finding well performing configurations across large search spaces is usually possible even with a simple algorithm like RS. 
All methods deliver reasonable hyperparameter configurations across a large search space, especially given all of them use only $10$ full training runs. 
On this small budget, they are able to match or outperform the single best seeds in all our sweep runs which use a total of $125$ runs per environment.
Our experiments show that \textit{automatically tuning a large variety of hyperparameters is both beneficial and efficient using even simple algorithms like RS or vanilla instantiations of multi-fidelity and population-based methods.}

\subsection{Are Hyperparameters in RL Well Behaved?}
In addition to the large number of hyperparameters contributing to an algorithm's performance, how an algorithm behaves with respect to changing hyperparameter values is an important factor in tuning algorithms.
Ideally, we want the algorithm's performance to be predictable, i.e., if the hyperparameter value is close to the optimum, we want the agent to perform well and then become progressively worse the farther we move away -- in essence, a smooth optimization landscape~\cite{pushak-acm22}. 
As we can see in Figure \ref{fig:ppo_boxplots}, the transitions between different parts of the search space are fairly smooth. 
The configurations perform in the order we would expect them to given the best values, with the drops in performance being mostly gradual instead of sudden.
Figure~\ref{fig:ppo_sweeps} shows configurations also performing consistently with regards to one another during the runtime, i.e., good configurations tend to learn quickly and bad configurations decay soon after training begins.
This means HPO approaches utilizing partial algorithm runs to measure the quality of configurations like multi-fidelity methods or PBT should not face major issues tuning RL algorithms.

\begin{figure}
    \centering
    \includegraphics[width=0.23\textwidth]{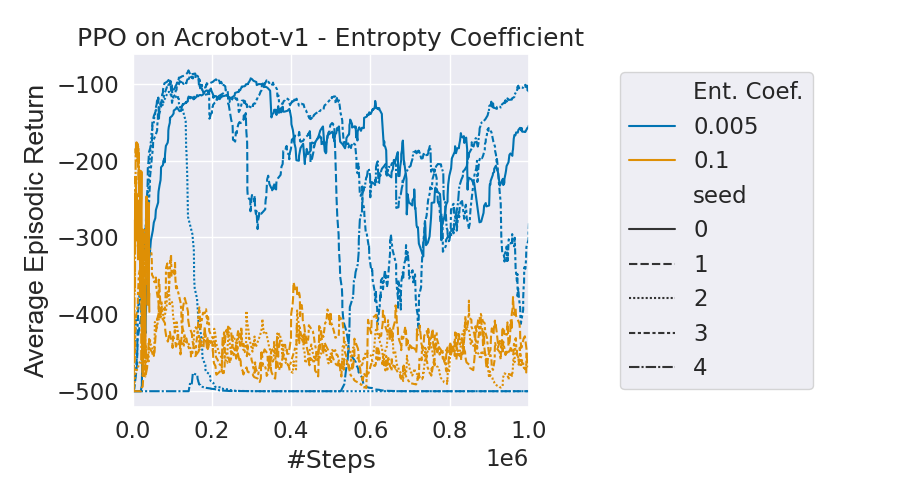}
    \includegraphics[width=0.23\textwidth]{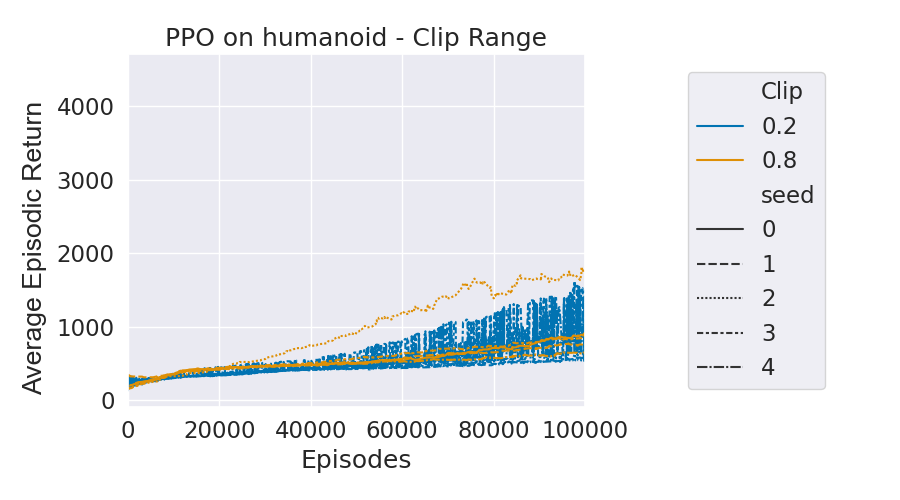}
    \includegraphics[width=0.23\textwidth]{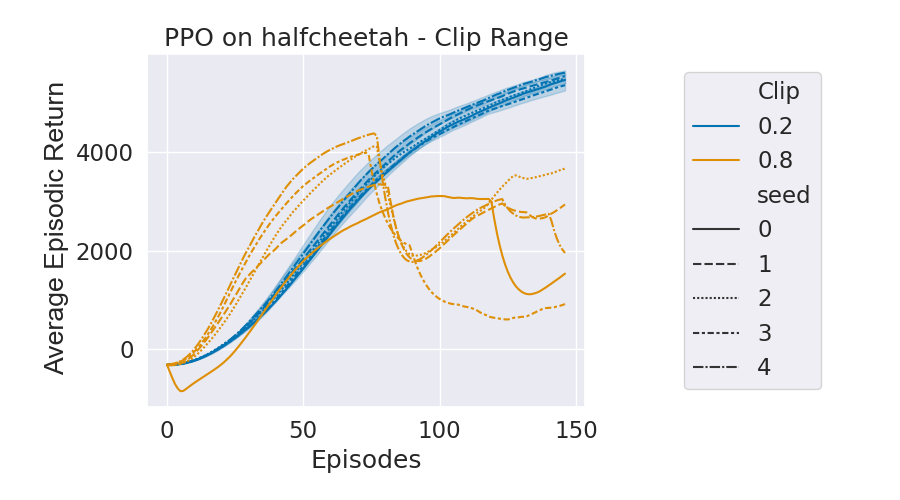}
    \includegraphics[width=0.23\textwidth]{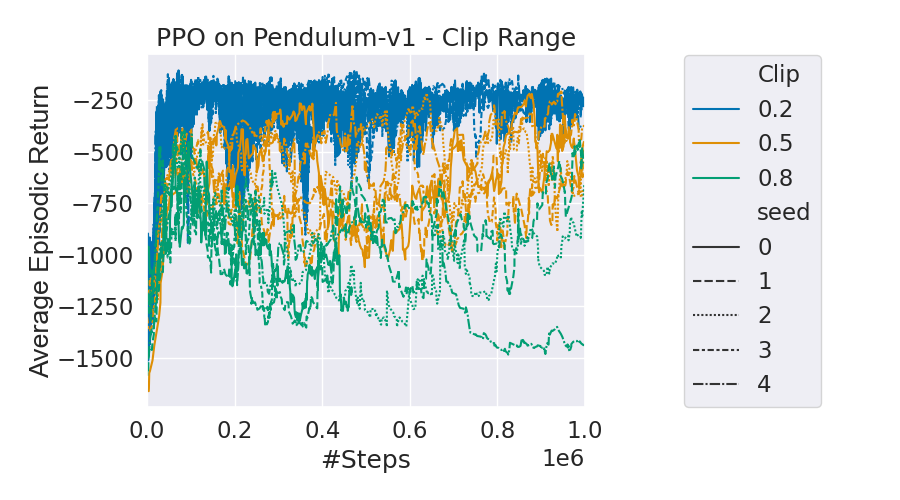}
    \caption{Individual seeds for selected clip range and entropy coefficient values of PPO across various environments.}
    \label{fig:ppo_single_seeds}
\end{figure}

\begin{table*}
\centering
    \caption{Tuning PPO on Acrobot (top) and SAC on Pendulum (bottom) across the full search space and different numbers of seeds. Lower numbers are better, best test performance for each method and values within its standard deviation are highlighted. Test performances are aggregated across 10 separate test seeds using the mean for each tuning run. We report mean and standard deviation of these.}
    \label{tab:tuning_seeds_table}
    \begin{tabular}{ll|cc|cc|cc}
        \toprule
            &  & DEHB Inc. & DEHB Test & PB2 Inc. & PB2 Test & RS Inc. & RS Test \\
        \midrule
        \parbox[t]{4mm}{\multirow{3}{*}{\rotatebox[origin=c]{90}{\small Acrobot}}} &
         $1$ Seed  & $70.6 \pm 3.4$  & $\mathbf{341.3\pm183.1}$  & $305.3 \pm 185.5$ & $\mathbf{353.7\pm134.5}$  & $77.8 \pm 4.9$  & $136.8\pm70.5$ \\
         & $3$ Seeds & $76.2 \pm 0.9$  & $\mathbf{381.1\pm127.6}$  & $301.2 \pm 128.0$ & $\mathbf{411.3\pm117.9}$  & $88.2 \pm 5.7$  & $\mathbf{98.8\pm16.3}$\\
         & $5$ Seeds & $79.3 \pm 1.2$  & $465.1\pm24.6$   & $228.5 \pm 149.5$ & $\mathbf{471.8\pm19.1}$   & $89.2 \pm10.4$ & $116.8\pm43.3$\\
         & $10$ Seeds & $156.0 \pm24.5$& $464.8\pm36.5$ & $404.9 \pm 53.3$ & $\mathbf{474.4\pm23.5}$ & $108.3 \pm 28.2$ & $\mathbf{100.1 \pm 20.0}$\\
         \midrule 
         \parbox[t]{4mm}{\multirow{3}{*}{\rotatebox[origin=c]{90}{\small Pendulum}}}& $1$ Seed & $111.5 \pm 23.6$  & $\mathbf{150.5\pm13.4}$ & $77.8 \pm 19.0$ & $840.7\pm580.1$ & $88.6 \pm 24.9$ & $168.3\pm46.4$\\
         & $3$ Seeds & $125.0 \pm 23.2$  & $\mathbf{144.8\pm9.0}$ & $133.3 \pm 14.7$ & $\mathbf{171.0\pm35.5}$ & $150.7 \pm 13.9$ & $159.0\pm21.6$\\
         & $5$ Seeds & $127.3 \pm 11.5$  & $350.2\pm418.2$  & $134.0 \pm 22.1$ & $661.3\pm586.2$ & $134.8 \pm 9.8$ & $397.8\pm485.5$\\
         & $10$ Seeds & $742.4 \pm 498.8$ & $318.6 \pm 281.3$ & $282.0 \pm 252.9$ & $468.6\pm437.9$ & $144.5 \pm 17.9$ & $\mathbf{150.2\pm4.8}$ \\
         \bottomrule
    \end{tabular}
\end{table*}

While we do see large variability in some configurations, this issue seems to occur largely in medium-well performing configurations, not in the very best or worst ones (see Figure~\ref{fig:ppo_sweeps}).
This supports our claim that hyperparameters are not only useful in increasing performance but have a significant influence on algorithm variability.

During the run itself differences between seeds can become an issue, however, especially for methods using partial runs. 
On many environments, when looking at each seed individually per hyperparameter as in Figure \ref{fig:ppo_single_seeds}, we can see the previously predictable behaviour is replaced with significant differences across seeds.
We observe single seeds with crashing performance, inconsistent learning curves and also exceptionally well performing seeds that end up outperforming the best seeds of configurations which are better on average. 
Given that we believe tuning only a few seeds of the target RL algorithm is still the norm~\cite{schulman-arxiv17a,berner-corr19,raileanu-iclr20,badia-icml20,hambro-neurips22}, such high variability with respect to the seed is likely a bigger difficulty factor for HPO in RL than the optimization landscape itself.

Thus, our conclusion is somewhat surprising: \textit{it should be possible to tune RL hyperparameters just as well as the ones in any other fields without RL-specific additions to the tuning algorithm since RL hyperparameter landscapes appear to be rather smooth.}
The large influence of many different hyperparameters is a potential obstacle, however, as are interaction effects that can ocurr between hyperparameters.
Furthermore, RL's sensitivity to the random seed, can present a challenge in tuning its hyperparameters, both by hand and in an automated manner.

\subsection{How Do We Account for Noise?}
As the variability between random seeds is a potential source of error when tuning and running RL algorithms, 
we investigate how we can account for it in our experiments to generate more reliable performance estimates.

As we have seen high variability both in performance and across seeds for different hyperparameter values, we return to Figure~\ref{fig:ppo_boxplots} to investigate how big the seed's influence on the final performance really is.
The plots show that the standard deviation of the performance for the same hyperparameter configuration can be very large. 
While this performance spread tends to decrease for configurations with better median performance, top-performing seeds can stem from unstable configurations with low median performance (e.g. the learning rate on Humanoid).
In most cases, there is an overlap between adjacent configurations, so \textit{it is certainly possible to select a presumably well-performing hyperparameter configuration on one seed that has low average performance across others}.

As this is a known issue in other fields as well, albeit not to the same degree as in RL, \textit{it is common to evaluate a configuration on multiple seeds in order to achieve a more reliable estimate of the true performance}~\cite{eggensperger-ijcai18a}. 
We verify this for RL by comparing the final performance of agents tuned by DEHB and PB2 on the performance mean across a single, 3 or 5 seeds.
We then test the overall best configuration on $5$ unseen test seeds.

Table~\ref{tab:tuning_seeds_table} shows that RS is able to improve the average test performance on both Acrobot and Pendulum by increasing the number of tuning seeds, as are DEHB and PB2 on Pendulum. 
However, this is only true up to a point as performance estimation across more than $3$ seeds leads to a general decrease in test performance, as well as a sharp increase in variance in some cases (e.g. $5$ seed RS or $10$ seed PB2 on Pendulum).
Especially when tuning across $10$ seeds, we see that the incumbents suffer as well, indicating that evaluating the configurations across multiple seeds increases the difficulty of the HPO problem substantially, even though it can help avoid overfitting. 
The performance difference between tuning and testing is significant in many cases and we can see e.g. on Acrobot that the best incumbent configurations, found by DEHB, perform more than four times worse on test seeds. 
We can find this effect in all tuning methods, especially on Pendulum.
This presents a challenge for reproducibility given that currently it is almost impossible to know what seeds were used for tuning or evaluation.
Simply reporting the performance of tuned seeds for the proposed method and that of testing seeds for the baselines is an unfair comparison which can lead to wrong conclusions.
\begin{figure*}
    \centering
    \includegraphics[width=\textwidth]{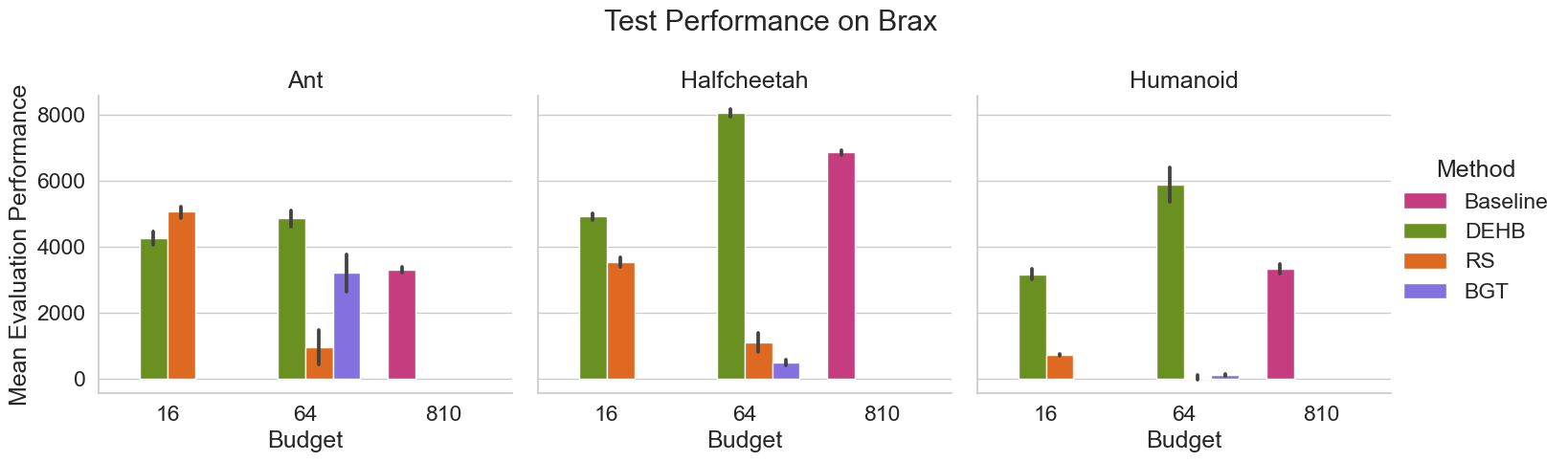}
    \caption{Tuning Results for PPO on Brax. Shown is the mean evaluation reward across $10$ episodes for $3$ tuning runs as well as the $98\%$ confidence interval across tuning runs.}
    \label{fig:brax}
\end{figure*}

To summarize, we have seen that the main challenges are the size of the search space, the variability involved in training RL agents, and the challenging generalization across random seeds.
Since many hyperparameters have a large influence on agent performance, but the optimization landscape is relatively smooth, RL hyperparameters can be efficiently tuned using HPO techniques, as we have shown in our experiments. 
Manual tuning, however, is comparatively costly as its cost scales at least linearly with the size of the search space.
 Dedicated HPO tools, on the other hand are able to find good configurations  on a significantly smaller budget by searching the whole space.
\textit{A major difficulty factor, however, is the high variability of results across seeds, which is an overlooked reproducibility issue that can lead to distorted comparisons of RL algorithms. This problem can be alleviated by tuning the algorithms on multiple seeds and evaluating them on separate test seeds.}

\section{Tradeoffs for Hyperparameter Optimization in Practice}
\label{sec:exp2}

While the experiments in the previous section are meant to highlight what challenges HPO tools face in RL and how well they overcome them, we now turn to more complex use cases of HPO.
To this end, we select three challenging environments each from Brax~\cite{freeman-neurips21} (Ant, Halfcheetah and Humanoid) and three from Procgen~\cite{cobbe-icml20} (Bigfish, Climber and Plunder) 
and automatically tune the state-of-the-art RL algorithms on these domains (PPO for Brax and IDAAC~\cite{raileanu-icml21} for Procgen).
Our goal here is simple:  we want to see if HPO tools can improve upon the state of the art in these domains with respect to both final performance and compute overhead.
As we now want to compare absolute performance values on a more complex problems with a bigger budget, we use BGT~\cite{wan-automl22} as the state-of-the-art population-based approach, and DEHB since it is among the best solver currently available~\cite{eggensperger-neurips21}. 
As before, we use RS as an example of a simple-to-implement tuning algorithm with minimal overhead. In view of the results of \citet{turner-arxiv21a} on HPO for supervised machine learning, we expect that RS should be outperformed by the other approaches.
For each task, we work on the original open-sourced code of each state-of-the-art RL method we test against, using the manually tuned hyperparameter settings as recommended in the corresponding papers as the baseline. 
All tuning algorithms will be given a small budget of up to $16$ full algorithm runs as well as a larger one of $64$ runs.
In comparison, IDAAC's tuning budget is $810$ runs. 
To give an idea of the reliability of both the tuning algorithm and the found configurations, we tune each setting $3$ times across $5$ seeds and test the best-found configuration on $10$ unseen test seeds.

\begin{figure*}
    \centering
    \includegraphics[width=\textwidth]{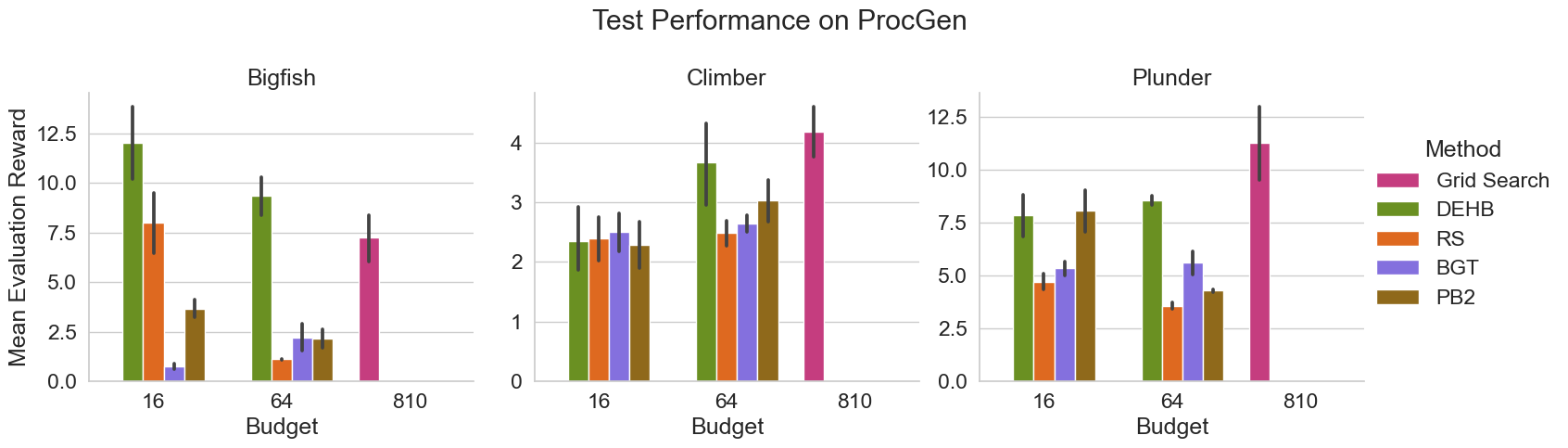}
    \caption{Tuning Results for IDAAC on Procgen. Shown is the mean evaluation reward across $10$ episodes for $3$ tuning runs as well as the $98\%$ confidence interval  across tuning runs.}
    \label{fig:Procgen}
\end{figure*}

As shown in Figures~\ref{fig:brax} and~\ref{fig:Procgen}, these domains are more challenging to tune on our small budgets relative to our previous environments (for tabular results, see Appendix~\ref{app:tuning_tables}).
While we do not know how the Brax baseline agent was tuned as this is not reported in the paper, the IDAAC baseline uses $810$ runs which is $12$ times more than the large tuning budget used by our HPO methods. 
On Brax, DEHB outperforms the baseline with a mean rank of $1.3$ compared to $1.7$ for the $16$ run budget and a rank of $1.0$ compared to the baselines's $1.3$ with $64$ runs. On Procgen the comparison is similar with $1.7$ to $2$ for $16$ runs and $1.0$ to $1.3$ for $64$ runs (see Appendix~\ref{app:brax} for details on how the rank is computed). 
We also see that DEHB's incumbent and test scores improve the most consistently out of all the tuning methods, with the additional run budget being utilized especially well on Brax.
RS, as expected cannot match this performance, ranking $2.3$ and $2.7$ for $16$ runs and $3.3$ and $3$ for $64$ runs respectively. We also see poor scaling behavior in some cases e.g. RS with a larger budget overfits to the tuning seeds on Brax while failing to improve on Procgen. 
As above, we see an instance of PB2 performing around $5$ times worse on the test seeds compared to the incumbent on Bigfish, further suggesting that certain PBT variants may struggle to generalize in such settings. 
On the other environments it does better, however, earning a Procgen rank of $2$ on the $16$ run budget, matching the baseline. With a budget of $64$ runs, it ranks $2.7$, the same as BGT and above RS. 
BGT does not overfit to the same degree as PB2 but performs worse on lower budgets, ranking $3.8$ on Procgen for $16$ runs and $2.7$ for $64$. On Brax, it fails to find good configurations with the exception of a single run on Ant (rank $3$). We do not restart the BGT optimization after a set amount of failures, however, in order to keep within our small maximum budgets. The original paper indicates that it is likely BGT will perform much better given a less restrictive budget.

Overall, HPO tools conceived for the AC setting, as represented by DEHB, are the most consistent and reliable within our experimental setting. 
Random Search, while not a bad choice on smaller budgets, does not scale as well with the number of tuning runs. 
Population-based methods cannot match either; PB2, while finding very well performing incumbent configurations struggles with overfitting, while BGT would likely benefit from larger budgets than used here. Further research into this optimization paradigm that priorities general configurations over incumbent performance could lead to additional improvements.

Across both benchmarks we see large discrepancies between the incumbent and test performance. 
This underlines our earlier point about the importance of using different test and tuning seeds for reporting.
In terms of compute overhead, all tested HPO methods had negligible effects on the total runtime, with BGT, by far the most expensive one, utilising on average under two minutes of time to produce new configurations for the $16$ run budget and less than 2 hours for the $64$ run budget, with all other approaches staying under $5$ minutes in each budget. 
Overall, we see that \textit{even computationally cheap methods with small tuning budgets can generally match or outperform painstakingly hand-tuned configurations that use orders of magnitude more compute.}


\section{Recommendations \& Best Practices}
Our experiments show the benefit of comprehensive hyperparameter tuning in terms of both final performance and compute cost, as well as how common overfitting to the set of tuning seeds is. As a result of our insights, we recommend some good practices for HPO in RL going forward.

\textbf{Complete Reporting} We still find that many RL papers do not state how they obtain their hyperparameter configurations, if they are included at all. As we have seen, however, unbiased comparisons should not take place on the same seeds the hyperparameters are tuned on. Hence, reporting the tuning seeds, the test seeds, and the exact protocol used for hyperparameter selection, should be standard practice to ensure a sound comparison across RL methods. 

\textbf{Adopting AutoML Standards} In many ways, the AutoML community is ahead of the RL community regarding hyperparameter tuning. 
We can leverage this by learning from their best practices, as e.g. stated by \citet{eggensperger-jair19a} and \citet{lindauer-jmlr20a}, and using their HPO tools which can lead to strong performance as shown in this paper. 
One notable good practice is to use separate seeds for tuning and testing hyperparameter configurations. 
Other examples include standardizing the tuning budget for the baselines and proposed methods, as well as tuning on the training and not the test setting.
While HPO in RL provides unique challenges such as the dynamic nature of the training loop or the strong sensitivity to the random seed, we observe significant improvements in both final performance and compute cost by employing state-of-the-art AutoML approaches. 
This can be done by integrating multi-fidelity evaluations into the population-based framework or using optimization tools like DEHB and SMAC.

\textbf{Integrate Tuning Into The Development Pipeline} For fair comparisons and realistic views of RL methods, we have to use competently tuned baselines. More specifically, the proposed method and baselines should use the same tuning budget and be evaluated on test seeds which should be different from the tuning seeds. 
Integrating HPO into RL codebases is a major step towards facilitating such comparisons.
Some RL frameworks have started to include options for automated HPO~\cite{cleanrl,ray} or provide recommended hyperparameters for a set of environments~\cite{sb3} (although usually not how they were obtained).
The choice of tuning tools for each library is still relatively limited, however, while provided hyperparameters are not always well documented and typically do not transfer well to other environments or algorithms.  
Thus, we hope our versatile and easy-to-use \href{https://github.com/facebookresearch/how-to-autorl}{HPO implementations} that can be applied to any RL algorithm and environment will encourage broader use of HPO in RL (see Appendix~\ref{app:implementation} for more information). 
In the future, we hope more RL libraries include AutoRL approaches since in a closed ecosystem, more sophisticated methods that go beyond black-box optimizers (e.g. gradient-based methods, neuro-evolution, or meta-learned hyperparameter agents à la DAC) could be deployed more easily.

\textbf{A Recipe For Efficient RL Research} To summarize, we recommend the following step-by-step process for tuning and selecting hyperparameters in RL: 
\begin{enumerate}
    \itemsep-0.3em 
    \item Define a training and test set which can include:
    \begin{enumerate}
        \itemsep-0em 
        \item environment variations
        \item random seeds for non-deterministic environments
        \item random seeds for initial state distributions
        \item random seeds for the agent (including network initialization)
        \item training random seeds for the HPO tool
    \end{enumerate}
    \item Define a configuration space with all hyperparameters that likely contribute to training success;
    \item Decide which HPO method to use;
    \item Define the limitations of the HPO method, i.e. the budget (or use self-terminating~\cite{makarova-automl22});
    \item Settle on a cost metric -- this should be an evaluation reward across as many episodes as a needed for a reliable performance estimate;
    \item Run this HPO method on the training set across a number of tuning seeds;
    \item Evaluate the resulting incumbent configurations on the test set across a number of separate test seeds and report the results.
\end{enumerate}
To ensure a fair comparison, this procedure should be followed for all RL methods used, including the baselines.
If existing hyperparameters are re-used, their source and tuning protocol should be reported. In addition, their corresponding budget and search space should be the same as those of the other RL methods used for comparison. 
In case the budget is runtime and not e.g. a number of environment steps, it is also important to use comparable hardware for all runs.
Furthermore, it is important to use the same test seeds for all configurations that are separate from all tuning seeds.
If this information is not available, re-tuning the algorithm is preferred.
This procedure, including all information on the search space, cost metric, HPO method settings, seeds and final hyperparameters should be reported.
We provide a checklist containing all of these points in Appendix~\ref{app:checklist} and as a \texttt{LaTeX} template in our \href{https://github.com/facebookresearch/how-to-autorl/blob/main/checklist.tex}{GitHub repository}.

\section{Conclusion}

We showed that hyperparameters in RL deserve more attention from the research community than they currently receive. 
Underreported tuning practices have the potential to distort algorithm evaluations while ignored hyperparameters may lead to suboptimal performance.
With only small budgets, we demonstrate that HPO tools like DEHB can cover large search spaces to produce better performing configurations using fewer computational resources than hyperparameter sweeps or grid searches.
We provide versatile and easy-to-use implementations of these tools which can be applied to any RL algorithm and environment. We hope this will encourage the adoption of AutoML best practices by the RL community, which should enhance the reproducibility of RL results and make solving new domains simpler.

Nevertheless, there is a lot of potential for developing HPO approaches tailored to the key challenges of RL such as the high sensitivty to the random seed for a given hyperparameter configuration.
Frameworks for learnt hyperparameter policies or gradient-based optimization methods could counteract this effect by reacting dynamically to an algorithm's behaviour on a given seed. We believe this is a promising direction for future work since in our experiments, PBT methods yield fairly static configurations instead of flexible schedules.  
Benchmarks like the recent AutoRL-Bench~\cite{shala-metalearn22a} accelerate progress by comparing AutoRL tools without the need for RL algorithm evaluations.
Lastly, higher-level AutoRL approaches that do not aim to find hyperparameter values but replace them entirely by directing the algorithm's behavior could in the long term both simplify and stabilize RL algorithms. Examples include exploration strategies~\cite{tzhang-neurips21}, learnt optimizers~\cite{metz-corr22} or entirely new algorithms~\cite{coreyes-iclr21,lu-corr22}. 

\section*{Acknowledgements}
\begin{wrapfigure}{l}{0.1\textwidth}
\vspace{-2em}
\includegraphics[height=1cm]{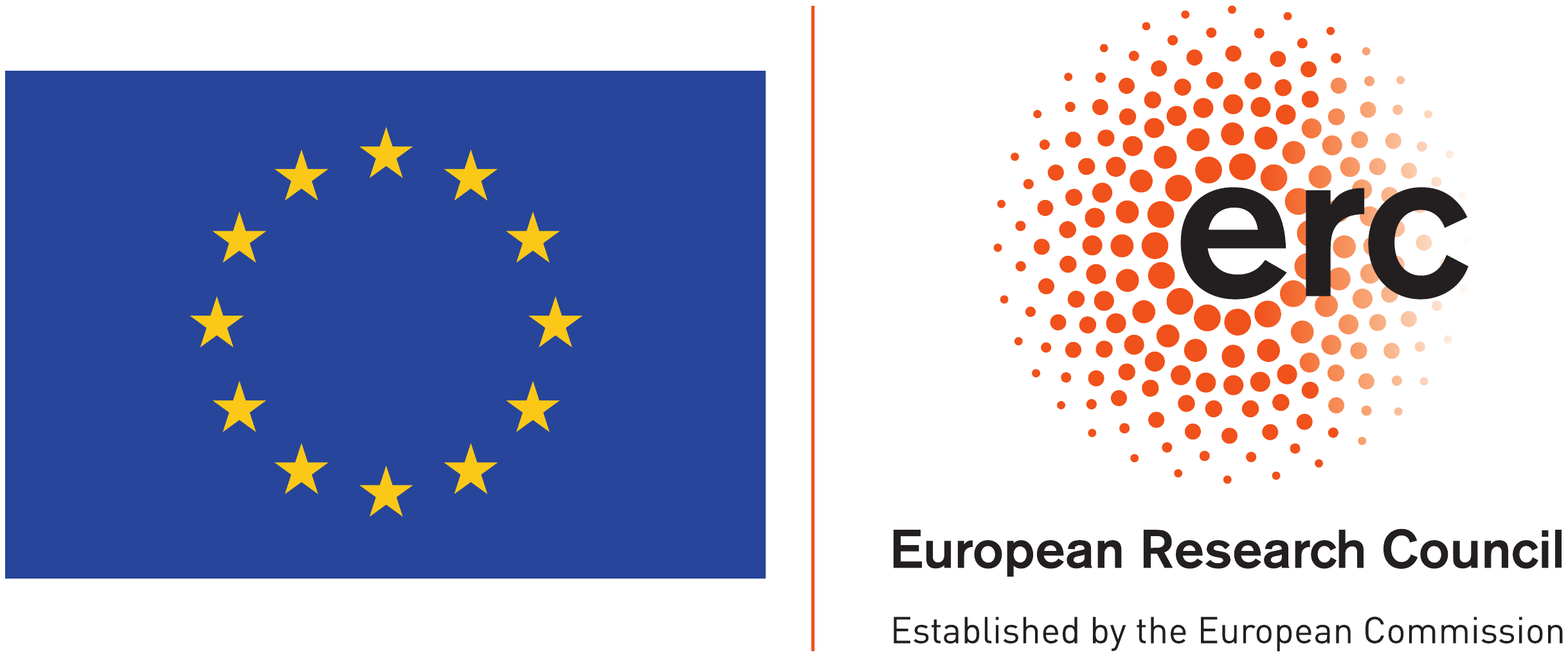}
\vspace{-2em}
\end{wrapfigure}
Marius Lindauer acknowledges funding by the European Union (ERC, ``ixAutoML'', grant no.101041029). Views and opinions expressed are those of the author(s) only and do not necessarily reflect those of the European Union or the European Research Council Executive Agency. Neither the European Union nor the granting authority can be held responsible for them.




\bibliography{bib/lib,bib/local,bib/proc,bib/shortstrings,bib/strings}
\bibliographystyle{icml2023}

\newpage
\appendix
\onecolumn
\section{Reproducibility Checklist for Tuning Hyperparameters in RL}
\label{app:checklist}
Below is a checklist we recommend for conducting experiments and reporting the process in RL. 
It is hard to give general recommendations for all RL settings when it comes to questions of budget, number of seeds or configuration space size.
For guidance on an appropriate number of testing seeds, as well as recommendations on how to report them, see~\citet{agarwal-neurips21}.
The ideal number of tuning seeds will likely depend heavily on the domain, though we recommend to use at least 5 to avoid overtuning on a small number of seeds.
As for configuration space size, we have seen successful tuning across up to $14$ hyperparameters in this paper and only small differences between $3$ and up to $9$ hyperparameters in Section~\ref{sec:exp1}, so we believe there is no reason be too selective for search spaces of around this size unless hyperparameter importances on the algorithm and domain are fairly well known.
Much larger search spaces could benefit from pruning, potentially after an initial analysis of hyperparameter importance.
\begin{enumerate}
    \itemsep0em 
    \item Are there training and test settings available on your chosen domains? \\ If yes:
    \begin{itemize}
        \item Is only the training setting used for training? \cmark \xmark
        \item Is only the training setting used for tuning? \cmark \xmark
        \item Are final results reported on the test setting? \cmark \xmark
    \end{itemize}
    \item Hyperparameters were tuned using \texttt{$<$package-name$>$} which is based on \texttt{$<$an-optimization-method$>$}
    \item The configuration space was:
    \texttt{$<$algorithm-1$>$}:
    \begin{itemize}
        \item \texttt{$<$a-continuous-hyperparameter$>$}: (\texttt{$<$lower$>$}, \texttt{$<$upper$>$})
        \item \texttt{$<$a-logspaced-continuous-hyperparameter$>$}: log((\texttt{$<$lower$>$}, \texttt{$<$upper$>$}))
        \item \texttt{$<$a-discrete-hyperparameter$>$}: [\texttt{$<$lower$>$}, \texttt{$<$upper$>$}]
        \item \texttt{$<$a-categorical-hyperparameter$>$}: {\texttt{$<$choice-a$>$}, \texttt{$<$choice-b$>$}}
        \item ...
    \end{itemize}
    \texttt{$<$algorithm-2$>$}:
    \begin{itemize}
        \item \texttt{$<$an-additional-hyperparameter$>$}: (\texttt{$<$lower$>$}, \texttt{$<$upper$>$})
        \item ...
    \end{itemize}
    \item The search space contains the same hyperparameters and search ranges wherever algorithms share hyperparameters \cmark \xmark \\ If no, why not?
    \item The cost metric(s) optimized was/were \texttt{$<$a-cost-metric$>$}
    \item The tuning budget was \texttt{$<$the-budget$>$}
    \item The tuning budget was the same for all tuned methods \cmark \xmark \\ If no, why not?
    \item If the budget is given in time: the hardware used for all tuning runs is comparable \cmark \xmark
    \item All methods that were reported were tuned with this the methods and settings described above \cmark \xmark \\ If no, why not?
    \item Tuning was done across $<n>$ tuning seeds which were: [$<0>$, $<1>$, $<2>$, $<3>$, $<4>$]
    \item Testing was done across $<m>$ test seeds which were: [$<5>$, $<6>$, $<7>$, $<8>$, $<9>$]
    \item Are all results reported on the test seeds? \cmark \xmark \\ If no, why not?
    \item The final incumbent configurations reported were: \\ \texttt{$<$algorithm-1-env-1$>$}:
    \begin{itemize}
        \item \texttt{$<$a-hyperparameter$>$}: \texttt{$<$value$>$}
        \item ...
    \end{itemize}
    \texttt{$<$algorithm-1-env-2$>$}:
    \begin{itemize}
        \item \texttt{$<$a-hyperparameter$>$}: \texttt{$<$value$>$}
        \item ...
    \end{itemize}
    \texttt{$<$algorithm-2-env-1$>$}:
    \begin{itemize}
        \item \texttt{$<$a-hyperparameter$>$}: \texttt{$<$value$>$}
        \item ...
    \end{itemize}
    \item The code for reproducing these experiments is available at: \texttt{$<$a-link$>$}
    \item The code also includes the tuning process \cmark \xmark
    \item Bundled with the code is an exact version of the original software environment, e.g. a conda environment file with all package versions or a docker image in case some dependencies are not conda installable \cmark \xmark
    \item The following hardware was used in running the experiments:
    \begin{itemize}
        \item \texttt{$<$n$>$} \texttt{$<$gpu-types$>$}
        \item ...
    \end{itemize}
\end{enumerate}

\section{Our AutoRL Hydra Sweepers}
\label{app:implementation}
We provide implementations of DEHB and PBT variations to supplement existing options like Optuna~\cite{optuna} or ray~\cite{ray} with state-of-the-art HPO tools that still have relatively high barriers of use, particularly in RL.
Our goal was to provide a tuning option with as little human overhead as possible but as much flexibility for applying it to different RL codebases as possible.
Hydra allows us to do both by using the tuning algorithms as sweepers that launch different configurations either locally or as parallel cluster jobs.
In practice, this means minimal code changes are necessary to use our sweepers: the return value will need to be a cost metric and for PBT checkpointing and loading is mandatory.
In this way, these plugins are compatible with any RL algorithm and environment.

Once these changes are implemented, a sweeper Hydra configuration that includes a search space definition can be used to run the whole optimization process in one go or resume existing runs (e.g. if the optimization was terminated accidentally or if more tuning budget becomes available after the fact). 
We include the option of using tuning seeds, which is so far uncommon except for CleanRL~\cite{cleanrl} where they are user specified.
Furthermore, we extended the option of initial runs for PBT variations to the original PBT and PB2 instead of just BGT in order to stabilize those methods.
In comparison to existing user-friendly tuners like Optuna, we provide different tuning algorithms that are not BO-based and include the option of using multi-fidelity tuning in hydra directly instead of having to implement a separate script.

\begin{figure}
    \centering
    \includegraphics[width=0.49\textwidth]{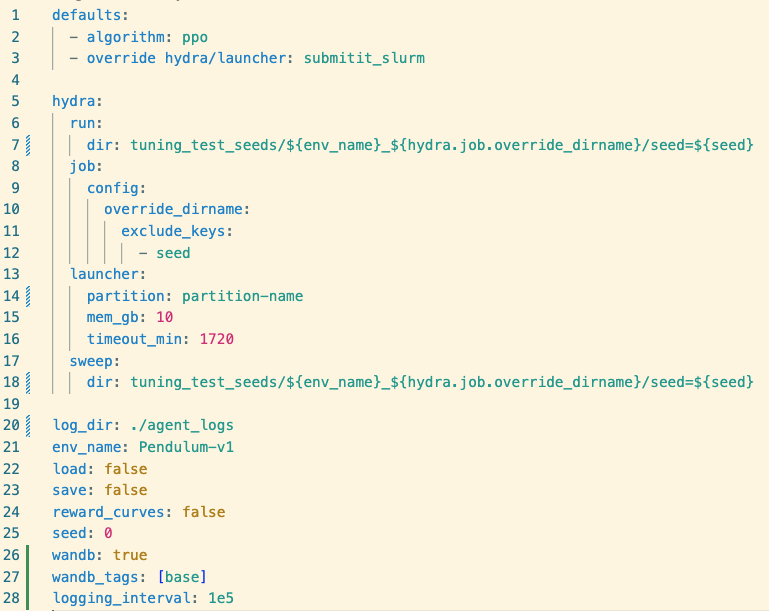}
    \includegraphics[width=0.49\textwidth]{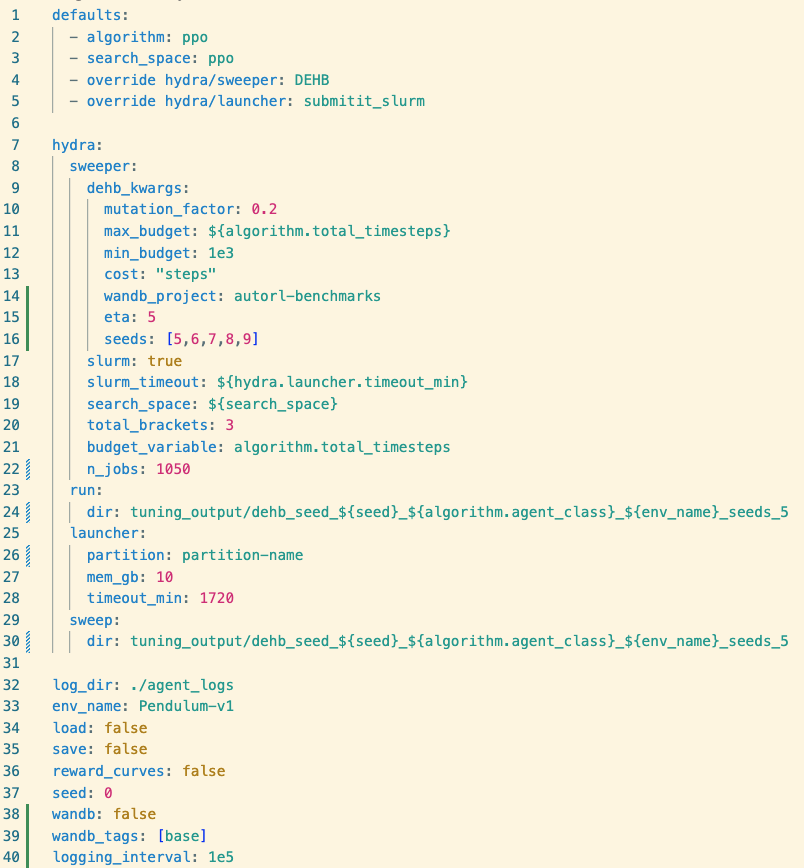}
    \caption{A base Hydra configuration file (left) and the changes necessary to tune this algorithm with DEHB (right).}
    \label{app-fig:hydra}
\end{figure}

\begin{figure}
    \centering
    \includegraphics[width=0.49\textwidth]{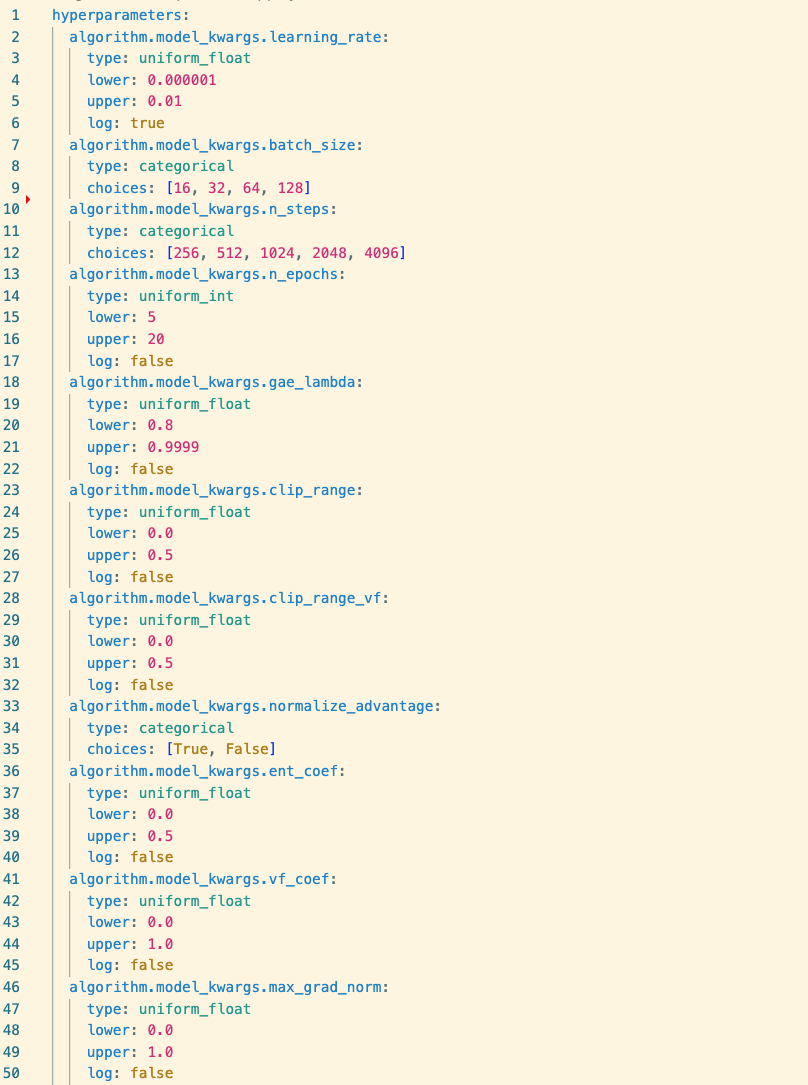}
    \caption{Example definition of a search space for PPO in a separate configuraion file.}
    \label{app-fig:hydra_space}
\end{figure}

Figure~\ref{app-fig:hydra} shows an example Hydra configuration file turned into a ready-to-run tuning configuration file for tuning with DEHB.
The corresponding configuration file, here for the full search space of PPO in StableBaselines3, is shown in Figure~\ref{app-fig:hydra_space}.

\section{Additional Background on Tuning Methods Used}
\label{app:more_background}
Since many in the RL community might be unfamiliar with the state of the art in AutoML and AutoRL, we provide brief descriptions of the RS, DEHB and PBT approaches we use in this paper.

\subsection{Random Search}
Random Search for hyperparameter optimization commonly refers to the method of sampling from a configuration space in a pseudo-random fashion~\cite{bergstra-jmlr12a}. 
The resulting configurations are then evaluated on full algorithm runs and the best performing one selected as incumbent.
While RS is not as reliable with small budgets and larger search spaces as other tuning options, it has proven to be a better alternative to grid search due to its scaling properties.
As Grid Search exhaustively evaluates all combinations of the given hyperparameter value set, it needs $n^m$ algorithm evaluations, with $n$ being the number of hyperparameters per dimension and $m$ the number of dimensions in the search space.
Still we only evaluate $n$ values for each dimension, irrelevant of how important this dimension actually is.
RS implicitly shifts more importance to the more relevant hyperparameters by varying the whole configuration at once, producing good results on smaller budgets.
Furthermore, Grid Search relies entirely on the domain knowledge of the user since they provide all configurations. 
This is of course an issue with new methods, domains or if the optimal hyperparameter configuration falls outside of the norm.

\subsection{DEHB}
DEHB is the combination of the evolutionary algorithm Differential Evolution (DE)~\cite{storn-glob97} and the multi-fidelity method HyperBand~\cite{li-jmlr18a}.
HyperBand as a multi-fidelity method is based on the idea of running many configurations with a small budget, i.e. only a fraction of training steps, and progressing promising ones to the next higher budget level. 
In this way we see many datapoints, but avoid spending time on bad configurations.
DEHB starts with a full set of HyperBand budgets, from very low to full budget, and runs it in its first iteration. 
For each budget, DEHB runs the equivalent of one full algorithm run in steps, e.g. if the current budget is $\frac{1}{10}$ of the full run budget, $10$ configurations will be evaluated.
For the second one, the lowest budget is left out and the second lowest is initialised with a population of configurations evolved by DEHB from the previous iteration's results. 
This procedure continues until either a maximum number of iterations is reached or only the full run budget budget is left.
The number of budgets is decided by a hyperparameter $\eta$.

In our experiments in Section~\ref{sec:exp1} we run $3$ iterations with $\eta = 5$ so only $3$ budgets, and in our larger DEHB experiments in Section~\ref{sec:exp2} we use $2$ iterations with $\eta=1.9$ so $8$ budgets.
We set the minimum budget as $\frac{1}{100}$ of the full run training steps in each case.

\subsection{PBT Variants}
PBT is based on the idea of maintaining a population of agents in parallel, each with its own hyperparameter configuration.
These agents are then trained for $n$ steps, after which their performance is evaluated and a checkpoint of their training state is created.
Now a portion of the worst agents are replaced by the best ones and the rest of the configurations are refined for the next iteration.
This will result in a hyperparameter schedule utilizing the best performing configurations at each iteration.

The original PBT~\cite{li-kdd19} randomly samples the initial configurations and then subsequently perturbes them by either randomly increasing or decreasing each hyperparameter by a constant factor.
Categorical values are randomly resampled with a fixed probability.

This undirected sampling proved successful, but only with large population sizes upwards of $64$ agents, therefore newer iterations of PBT often use a model to select new hyperparameter configurations, as e.g. PB2 uses Bayesian Optimization with a Gaussian Process~\cite{parkerholder-neurips20}.
This enables optimization with a significantly smaller population size of as little as $4$ agents.

As we have seen, however, the result can be volatile, therefore ~\citet{wan-automl22} suggested two main extensions on top of PB2, in addition to the ability to tune the architecture with the PBT framwork, forming the current state of the art across PBT methods. The extensions are (1) the use of periodic kernel restarts in case no improvements are visible and (2) the use of full budget initial runs to warmstart the Gaussian Process with high quality datapoints.

In our experiments we use $20$ configuration changes for each method with a population size of $8$ for PB2 in Section~\ref{sec:exp1} and $16$/$64$ for PB2 in Section~\ref{sec:exp2}. 
For BGT, we use $8$ initial runs and a population size of $8$ for the smaller budget and $48$ and $16$, respectively for the larger one. In both PB2 and BGT, we always replace the worst $12.5\%$ of agents with the best $12.5\%$.

\section{An Overview of Hyperparameter Configurations \& Search Spaces}
\label{app:search_spaces}
\subsection{Stable Baselines Default Configurations}
Table~\ref{app-tab:baseline_defaults} shows the default hyperparameters we use throughout Section~\ref{sec:exp1}.
\begin{table}
\centering
    \begin{tabular}{ll|ccc}
        \toprule
            & & Acrobot \& Pendulum & Brax & MiniGrid \\
         \midrule
         \parbox[t]{4mm}{\multirow{15}{*}{\rotatebox[origin=c]{90}{\small PPO}}}& Policy Class & \texttt{MlpPolicy} & \texttt{MlpPolicy} & \texttt{CnnPolicy}\\
         & leaning\_rate & 1e-3 & 1e-4 & 7e-4\\
         & batch\_size & 64 & 512 & 64 \\
         & gamma & 0.9 & 0.99 & 0.999 \\
         & n\_steps & 1024 & 1024 & 256 \\
         & n\_epochs & 10 & 16 & 4 \\
         & gae\_lambda & 0.95 & 0.96 & 0.95 \\
         & clip\_range & 0.2 & 0.2 & 0.2 \\
         & clip\_range\_vf & null & null & null \\
         & normalize\_advantage & \texttt{True} & \texttt{True} & \texttt{True} \\
         & ent\_coef & 0.01 & 0.01 & 0.01 \\
         & vf\_coef & 0.5 & 0.5 & 0.5 \\
         & max\_grad\_norm & 0.5 & 0.5 & 0.5 \\
         & use\_sde & \texttt{False} & \texttt{False} & \texttt{False} \\
         & sde\_sample\_freq & 4 & 4 & 4 \\
         \midrule
         \parbox[t]{4mm}{\multirow{11}{*}{\rotatebox[origin=c]{90}{\small SAC}}}& Policy Class & \texttt{MlpPolicy} & \texttt{MlpPolicy} & \\
         & leaning\_rate & 1e-4 & 1e-4 & \\
         & batch\_size & 256 & 512 &  \\
         & gamma & 0.99 & 0.99 &  \\
         & tau & 1.0 & 1.0 &  \\
         & learning\_starts & 100 & 100 & \\
         & buffer\_size & 1000000 & 1000000 & \\
         & train\_freq & 1 & 1 & \\
         & gradient\_steps & 1 & 1 & \\
         & use\_sde & \texttt{False} & \texttt{False} & \\
         & sde\_sample\_freq & -1 & -1 & \\
         \midrule
         \parbox[t]{4mm}{\multirow{12}{*}{\rotatebox[origin=c]{90}{\small DQN}}}& Policy Class & \texttt{MlpPolicy} &  & \texttt{CnnPolicy}\\
         & learning\_rate & 1e-3 & & 5e-7 \\
         & batch\_size & 64 & & 64\\
         & tau & 1.0 & & 1.0 \\
         & gamma & 0.9 & & 0.999\\
         & learning\_starts & 50000 & & 100 \\
         & train\_freq & 4 & & 4\\
         & gradient\_steps & 1 & & 1\\
         & exploration\_fraction & 0.1 & & 0.1\\
         & exploration\_initial\_eps & 1.0 & & 1.0\\
         & exploration\_final\_eps & 0.05 & & 0.05\\
         & buffer\_size & 1000000 & & 1000000\\
         \bottomrule
    \end{tabular}
    \caption{StableBaselines hyperparameter defaults for different environments.}
    \label{app-tab:baseline_defaults}
\end{table}

\subsection{Stable Baseline Sweep Values}
We sweep over the same hyperparameter values for each environment one dimension at a time. 
For PPO, these are \texttt{learning rate} $\in \{1e-2,5e-3,1e-3,5e-4,1e-4,5e-5,1e-5,5e-6,1e-6,5e-7\}$, \texttt{entropy coefficient} $\in \{0.1,0.05,0.01,0.005,0.001\}$ and \texttt{clip range} $\in \{0.0,0.1,0.2,0.3,0.4,0.5,0.6,0.7,0.8,0.9\}$.

For SAC, \texttt{learning rate} $\in \{1e-2,5e-3,1e-3,5e-4,1e-4,5e-5,1e-5,5e-6,1e-6,5e-7\}$, \texttt{tau} $\in \{1.0,0.9,0.8,0.7,0.6,0.5,0.4,0.3,0.2,0.1\}$ and \texttt{training frequency} $\in \{1,2,4,8,16\}$.

For DQN, \texttt{learning rate} $\in \{1e-2,5e-3,1e-3,5e-4,1e-4,5e-5,1e-5,5e-6,1e-6,5e-7\}$, \texttt{epsilon} $\in \{1.0,0.9,0.8,0.7,0.6,0.5,0.4,0.3,0.2,0.1\}$ and \texttt{training frequency} $\in \{1,2,4,8,16\}$.

\subsection{Stable Baselines Search Spaces}
Table~\ref{app-tab:baseline_spaces} shows the search spaces we used for the experiments in Section~\ref{sec:exp1}. The search spaces are the same for all tuning methods and across environments.
We denote floating point intervals as \texttt{interval(lower,upper)}, integer ranges as \texttt{range[lower,upper]}, categorical choices as \texttt{\{choice 1, choice 2\}} and add \texttt{log} if the search space for this hyperparameter is traversed logarithmically.
\begin{table}
\centering
    \begin{tabular}{ll|ccc}
        \toprule
            & Hyperparameter & Full Space & Small Space & LR Only\\
         \midrule
         \parbox[t]{4mm}{\multirow{11}{*}{\rotatebox[origin=c]{90}{\small PPO}}} & leaning\_rate & log(interval(1e-6, 0.1)) & log(interval(1e-6, 0.1)) & log(interval(1e-6, 0.1))\\
         & ent\_coef & interval(0.0, 0.5) & interval(0.0, 0.5) & \\
         & n\_epochs & range[5,20] & range[5,20] &  \\
         & batch\_size & \{16, 32, 64, 128\} & & \\
         & n\_steps & \{256, 512, 1024, 2048, 4096\} &  &  \\
         & gae\_lambda & interval(0.8, 0.9999) & & \\
         & clip\_range & interval(0.0, 0.5) &  &  \\
         & clip\_range\_vf & interval(0.0, 0.5) &  & \\
         & normalize\_advantage & \{True, False\} &  &  \\
         & vf\_coef & interval(0.0, 1.0) &  &  \\
         & max\_grad\_norm & interval(0.0, 1.0) &  &  \\
         \midrule
         \parbox[t]{4mm}{\multirow{6}{*}{\rotatebox[origin=c]{90}{\small SAC}}}& leaning\_rate & log(interval(1e-6, 0.1)) & log(interval(1e-6, 0.1)) & log(interval(1e-6, 0.1))\\
         & train\_freq & range[1,1e3] & range[1,1e3] & \\
         & tau & interval(0.01, 1.0) & interval(0.01, 1.0) &  \\
         & batch\_size & \{64, 128, 256, 512\} & &  \\
         & learning\_starts & range[0,1e4] & & \\
         & buffer\_size & range[5e3,5e7] & & \\
         & gradient\_steps & range[1,10] & & \\
         \midrule
         \parbox[t]{4mm}{\multirow{9}{*}{\rotatebox[origin=c]{90}{\small DQN}}}& learning\_rate & log(interval(1e-6, 0.1)) & log(interval(1e-6, 0.1)) & log(interval(1e-6, 0.1)) \\
         & batch\_size & \{4, 8, 16, 32\} & \{4, 8, 16, 32\} & \\
         & exploration\_fraction & interval(0.005, 0.5) & interval(0.005, 0.5) & \\
         & learning\_starts & range[0,1e4] & &  \\
         & train\_freq & range[1,1e3] & & \\
         & gradient\_steps & range[1,10] & & \\
         & exploration\_initial\_eps & interval(0.5, 1.0) & & \\
         & exploration\_final\_eps & interval(0.001, 0.2) & & \\
         & buffer\_size & range[5e3,5e7] & & \\
         \bottomrule
    \end{tabular}
    \caption{StableBaselines search spaces.}
    \label{app-tab:baseline_spaces}
\end{table}

\subsection{Brax Experiment Settings}
\label{app:brax}
We base our implementations the training code provided with Brax with minor additions like only a single final evaluation and agent loading. 
The GitHub commit ID for the \href{https://github.com/google/brax}{code} version we use is \texttt{3843d433050a08cb492c301e039e04409b3557fc}.
The cost metric we optimize is the evaluation reward across one episode of the environment batch.
We tune on seeds $0-4$ and evaluate on seeds $5-14$.
The baseline hyperparameters are taken from this commit as well and are shown in \ref{app-tab:brax_defaults} together with our search space.

For both Procgen and Brax, we compute the rank of a method as follows: the best performing method on the test seeds and all other methods within its standard deviation receive rank one.
The method with the next best mean (and all methods in its standard deviation) receive the next free rank -- 2 in case there was a single best method, 3 if there were two and so on. 
These ranks are determined for each environment from which we can compute a mean across the whole domain.

\begin{table}
\centering
    \begin{tabular}{l|cc}
        \toprule
         Hyperparameter & Search Space & Defaults \\
         \midrule
         leaning\_rate & log(interval(1e-6, 0.1)) & 3e-4\\
         num\_update\_epochs & range[1,15] & 4\\
         batch\_size & \{128,256,512,1024,2048\} & 1024 \\
         num\_minibatches & range[0,7] & 6 \\
         entropy\_cost & interval(0.0001, 0.5) & 1e-2 \\
         gae\_lambda & interval(0.5, 0.9999) & 0.95 \\
         epsilon & interval(0.01, 0.9) & 0.3 \\
         vf\_coef & interval(0.01, 0.9) & 0.5 \\
         reward\_scaling & interval(0.01, 1.0) & 0.1\\
         \bottomrule
    \end{tabular}
    \caption{Search Space and baseline hyperparameters for Brax. Actual number of minibatches is $2^{num\_minibatches}$. Epsilon refers to the clip range in this implementation.}
    \label{app-tab:brax_defaults}
\end{table}

\subsection{Procgen Experiment Settings}
We use the open-source code provided by \citep{raileanu-icml21} with minor additions like loading agents.
The GitHub commit ID for the \href{https://github.com/rraileanu/idaac}{code} version we use is \texttt{2fe30202942898b1b09d76e5d8c71d5a7db3686b}.
The cost metric we optimize is the evaluation reward across ten episodes of the training environment.
We tune on seeds $0-4$ and evaluate on seeds $5-14$.
Our baseline are the provided best hyperparameters per environment (see \ref{app-tab:Procgen_defaults} for the configuration and our search space).

\begin{table}
\centering
    \begin{tabular}{l|cccc}
        \toprule
         Hyperparameter & Search Space & Bigfish & Climber & Plunder \\
         \midrule
         lr & log(interval(1e-6, 0.1)) & 5e-4 & 5e-4 & 5e-4\\
         eps & log(interval(1e-6, 0.1)) & 1e-5 & 1e-5 & 1e-5 \\
         hidden\_size & - & 256 & 256 & 256\\
         clip\_param & interval(0.0, 0.5) & 0.2 & 0.2 & 0.2\\
         num\_mini\_batch & - & 8 & 8 & 8\\
         ppo\_epoch & range[1, 5] & 3 & 3 & 3\\
         num\_steps & - & 256 & 256 & 256\\
         max\_grad\_norm & interval(0.0, 1.0) & 0.5 & 0.5 & 0.5\\
         value\_loss\_coef & interval(0.0, 1.0) & 0.5 & 0.5 & 0.5\\
         entropy\_coef & interval(0.0, 0.5) & 0.01 & 0.01 & 0.01\\
         gae\_lambda & interval(0.8, 0.9999) & 0.95 & 0.95 & 0.95\\
         gamma & - & 0.999 & 0.999 & 0.999\\
         alpha & interval(0.8, 0.9999) & 0.99 & 0.99 & 0.99\\
        clf\_hidden\_size & - & 4 & 64 & 4\\
        order\_loss\_coef & interval(0.0, 0.1) & 0.01 & 0.001 & 0.1 \\
        use\_nonlinear\_clf & \{\texttt{True}, \texttt{False}\}& \texttt{True} & \texttt{True} & \texttt{False} \\
        adv\_loss\_coef & interval(0.0, 1.0) & 0.05 & 0.25 & 0.3 \\
        value\_freq & range[1, 5] & 32 & 1 & 8 \\
        value\_epoch & range[1, 10] & 9 & 9 & 1\\
         \bottomrule
    \end{tabular}
    \caption{Search Space and baselines hyperparameters for IDAAC on Procgen.}
    \label{app-tab:Procgen_defaults}
\end{table}

\subsection{Hardware}
All of our experiments were run on a compute cluster with two Intel CPUs per node (these were used for the experiments in Section~\ref{sec:exp1}) and four different node configurations for GPU (used for the experiments in Section~\ref{sec:exp2}).
These configurations are: $2$ Pascal $130$ GPUs, $2$ Pascal $144$ GPUs, $8$ Volta $10$ GPUs or $8$ Volta $332$.
We ran the CPU experiments with $10$GB of memory on single nodes and the GPU experiments with $10$GB for Procgen and $40$GB for Brax on a single GPU each.

\section{Details on the Tuned Configurations}
\label{app:configurations}
We want to give some insights into how much the incumbents of our HPO methods differ from the baselines and one another.
We show an example comparisons between different incumbent configurations on the full search space of PPO on Acrobot in Table~\ref{app-tab:config_comparison}.
This result is consistent with what we find across other algorithms and environments: the differences between incumbents as well as between incumbents and baseline are fairly large.
The result of HPO in our experiment has not meant small changes to only a subset of the search space, but usually significant deviations from the baseline in most of them.
Still, we can see some similarities at times, in this case the batch size stays consistently at $64$ across all configurations (with the exception of the final training phase of PB2).
We can also see common trends among the incumbents at times, e.g. in the value of the GAE $\lambda$ which is between $0.81$ and $0.85$ for the incumbents, but at $0.95$ in the default configuration. 
Unfortunately, the other hyperparameter values do not seem to share any trends and often have significantly different values as e.g. in the entropy coefficient which varies between $0.01$ and $0.42$.

Why do we then see such similar performance from all of these configurations? We believe three main factors are at play: hyperparameter importance, the algorithm's sensitivity to a hyperparameter value and interaction effects between hyperparameters.
It is likely that not all hyperparameters are crucial to optimize in this setting, so seeing very different values for unimportant hyperparameters can make the configurations appear more different than they are. 
We know from our experiments, however, that a mistake in the entropy coefficient can be highly damaging to the algorithm's performance in Acrobot (see Figure~\ref{app-fig:ppo_boxplots} below).
Comparing the entropy coefficient curve of Acrobot and Ant in this figure, however, reveals that the median performance across different entropy coefficients degrades much less quickly for Acrobot than for Ant -- on Acrobot, PPO is less sensitive to changes in hyperparameter values.
To put it another way: hyperparameter values may look different between configurations but result in the same algorithm behaviour as long as they are within a similar range.
Lastly, since we optimize many hyperparameters and the agent's behaviour depends on these hyperparameters, it is possible that hyperparameters interact with each other to produce a similar outcome as long as their relation stays similar.
This could be an explanation for combining lower learning rates with more update epochs as DEHB does.
Analysing hyperparameter configurations on their own, however, will not provide enough information to determine each of these factors for each hyperparameter.
They have to be explored through separate experiments first before we can draw conclusions on how similar the configurations we found in HPO tools are and what that means for optimal hyperparameter values in our settings.

\begin{table}
\centering
    \begin{tabular}{ll|cccc}
        \toprule
            & Hyperparameter & Default & DEHB & PB2 & RS \\
         \midrule
         \parbox[t]{4mm}{\multirow{15}{*}{\rotatebox[origin=c]{90}{\small PPO Acrobot}}}& leaning\_rate & 1e-3 & 5e-05 & 5e-3 for 90e5 steps, then 3e-6 & 1e-6\\
         & batch\_size & 64 & 64 & 64 for 90e5 steps, then 128 & 64\\
         & n\_steps & 1024 & 1024 & 256 & 256 \\
         & n\_epochs & 10 & 14 & 8 for 90e5 steps, then 20 & 6 \\
         & gae\_lambda & 0.95 & 0.82 & 0.85 for 90e5 steps, then 0.82 & 0.81 \\
         & clip\_range & 0.2 & 0.14 & 0.06 for $90$e5 steps, then 0.27 & 0.05 \\
         & clip\_range\_vf & null & 0.43 & 0.06 & 0.11 \\
         & normalize\_advantage & \texttt{True} & \texttt{True} & \texttt{True} \\
         & ent\_coef & 0.01 & 0.21 & 0.42 for $90$e5 steps, then 0.29 & 0.01 \\
         & vf\_coef & 0.5 & 0.02 & 0.5 for $90$e5 steps, then 0.7 & 0.07 \\
         & max\_grad\_norm & 0.5 & 0.75 & 0.24 for $90$e5 steps, then 0.5 & 0.97 \\
         \bottomrule
    \end{tabular}
    \caption{StableBaselines hyperparameter defaults for different environments.}
    \label{app-tab:config_comparison}
\end{table}

\section{Tuning Results on Brax \& Procgen in Tabular Form}
\label{app:tuning_tables}
For ease of comparison, we provide the results of Section~\ref{sec:exp2} in tabular form.
\begin{table}
    \centering
    \caption{Tuning PPO on Brax's Ant, Halfcheetah and Humanoid environments. Shown are tuning results across $3$ runs across $5$ seeds each, tested on $10$ different test seeds.}
    \begin{tabular}{c|ccc}
        \toprule
              & Ant & Halfcheetah & Humanoid  \\
         \midrule
         Baseline & $3448 \pm 343$ & $6904 \pm 377$ & $\mathbf{3235 \pm 758}$ \\
         \midrule
         DEHB Inc. & $5745 \pm 878$ & $3993\pm 871$& $1788 \pm 718$\\
         DEHB Test & $\mathbf{4288 \pm 1017}$ & $4928\pm 500$ & $\mathbf{3167 \pm 874}$ \\
         RS Inc. & $2515 \pm 1750$ & $2978 \pm 1007$ & $763\pm 317$ \\
         RS Test & $\mathbf{5165 \pm 896}$& $3646 \pm 699$ &$ 753 \pm 209$ \\
         \midrule
         DEHB Inc. (64) & $7170 \pm 1045$ & $8202\pm 445$& $4338 \pm 1655$\\
         DEHB Test (64 ) & $\mathbf{4696 \pm 1252}$& $\mathbf{8039 \pm 636}$& $\mathbf{5205 \pm 2781}$\\
         BGT Inc. (64) & $1119\pm 1321$ & $1051\pm 752$ & $434\pm 15$\\
         BGT Test (64) & $3196 \pm 3307$ & $456 \pm 461$ & $132\pm0$\\
         RS Inc. (64) & $6344\pm 654$ & $7891\pm 386$ & $2932\pm 798$  \\
         RS Test (64) & $669 \pm 2447$ &$950 \pm 1461$ & $325 \pm 162$\\
         \bottomrule
    \end{tabular}
    \label{tab:brax}
\end{table}

\begin{table}
    \centering
    \caption{Tuning IDAAC on Procgen's Bigfish, Climber and Plunder. Results are across $3$ runs using $5$ seeds each, and tested on $10$ different test seeds.}
    \begin{tabular}{c|ccc}
        \toprule
              & Bigfish & Climber & Plunder  \\
         \midrule
         Baseline & $6.8 \pm 3.2$ & $\mathbf{4.1 \pm 1.4}$ & $\mathbf{11.8 \pm 5.5}$\\
         \midrule
         DEHB Inc. & $7.3\pm 2.0$ & $3.7\pm0.2$ & $5.8 \pm 0.2$ \\
         DEHB Test & $\mathbf{11.9 \pm 4.3}$ & $\mathbf{2.7 \pm 1.5}$ & $\mathbf{8.6 \pm 2.6}$ \\
         BGT Inc. & $1.3 \pm 0.2$ & $2.5 \pm 0.4$ & $4.5 \pm 0.3$\\
         BGT Test & $0.9 \pm 0.4$ & $2.4 \pm 1.1$ & $5.3 \pm 1.0$\\
         PB2 Inc. & $26.1 \pm 2.2$ & $3.2 \pm 0.3$ & $4.7 \pm 0.3$\\
         PB2 Test & $3.4 \pm 1.9$ & $2.6 \pm 1.1$ & $\mathbf{8.3 \pm 2.7}$\\
         RS Inc. & $4.4 \pm 2.1$ & $5.5\pm 0.6$ & $6.2\pm 1.3$ \\
         RS Test & $7.4\pm 5.0$ & $2.6 \pm 1.0$ & $4.7 \pm 1.0$\\
         \midrule
         DEHB Inc. (64 runs) & $11.5\pm 0.7$ & $6.0\pm 1.0$ & $7.3 \pm 0.9$ \\
         DEHB Test (64 runs) & $\mathbf{9.4 \pm 2.5}$& $\mathbf{3.9 \pm 1.9}$& $\mathbf{8.7 \pm 0.7}$ \\
         BGT Inc. (64 runs) & $4.7\pm 5.1$  & $3.2\pm 1.5$ & $5.3\pm 0.1$\\
         BGT Test (64 runs) & $2.1 \pm 1.9$& $2.6 \pm 0.4$ & $5.9 \pm 1.6$\\
         PB2 Inc. (64 runs) & $10.5\pm 10.1$ & $3.4\pm 0.3$ & $5.2\pm 0.2$\\
         PB2 Test (64 runs) & $2.1 \pm 1.1$ & $3.0 \pm 0.9$ & $4.3 \pm 0.2$\\
         RS Inc. (64 runs) & $1.9 \pm 0.3$ &$5.9\pm 0.8 $  &  $6.7\pm0.4$\\
         RS Test (64 runs) & $1.1\pm0.1$ & $2.3 \pm 0.7$ & $3.6 \pm 0.5$ \\
         \bottomrule
    \end{tabular}
    \label{tab:Procgen}
\end{table}

\clearpage
\section{Hyperparameter Sweeps for PPO, DQN and SAC}
\label{app:more_sweeps}
\begin{figure}[h]
    \centering
    \includegraphics[width=0.2\textwidth]{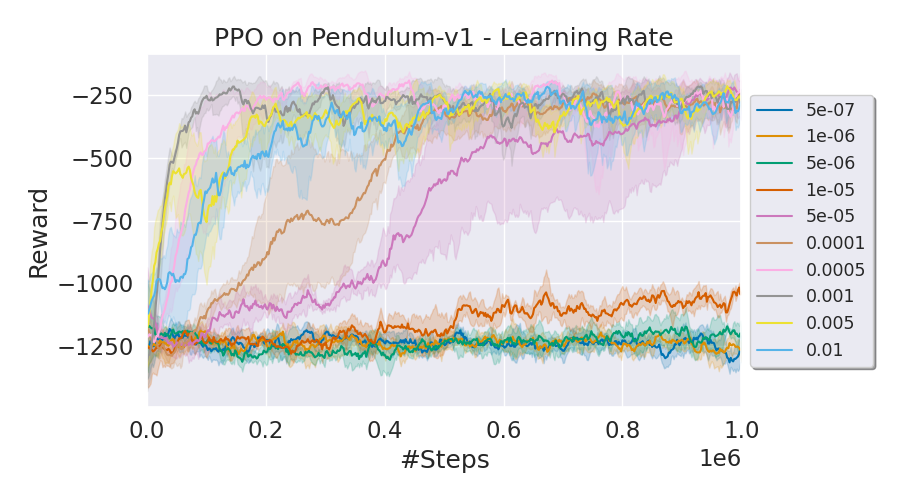}
    \includegraphics[width=0.2\textwidth]{sweep_Pendulum-v1_PPO_algorithm.model_kwargs.ent_coef}
    \includegraphics[width=0.2\textwidth]{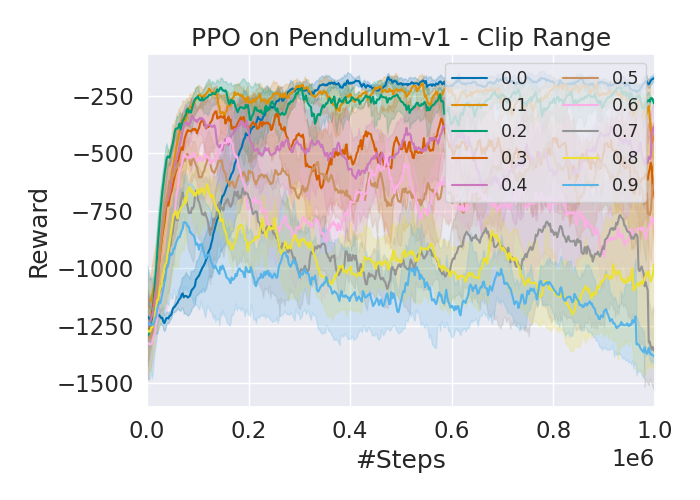}
    \includegraphics[width=0.2\textwidth]{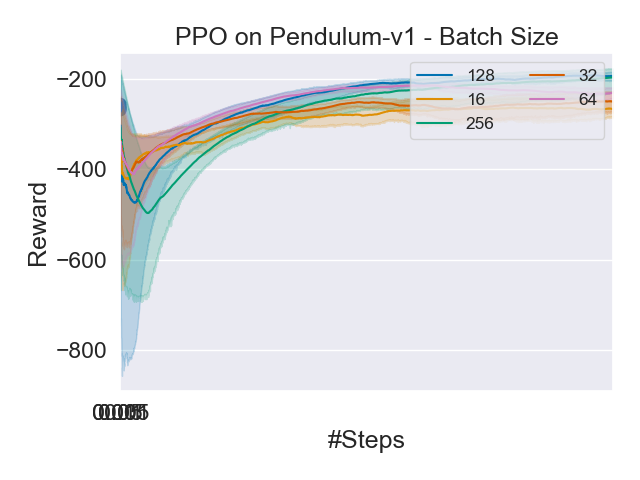}
    \includegraphics[width=0.2\textwidth]{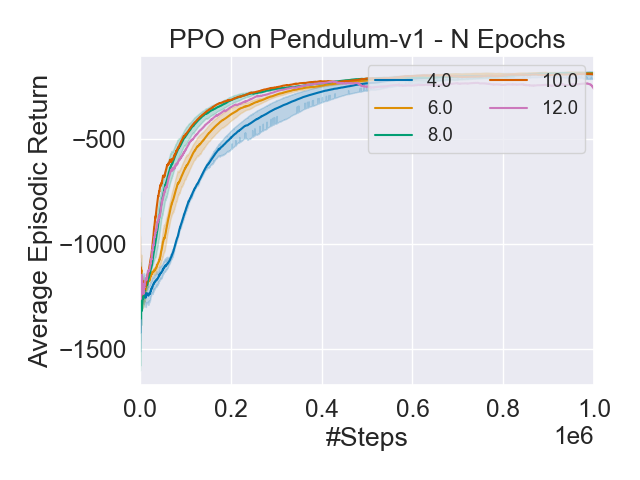}
    \includegraphics[width=0.2\textwidth]{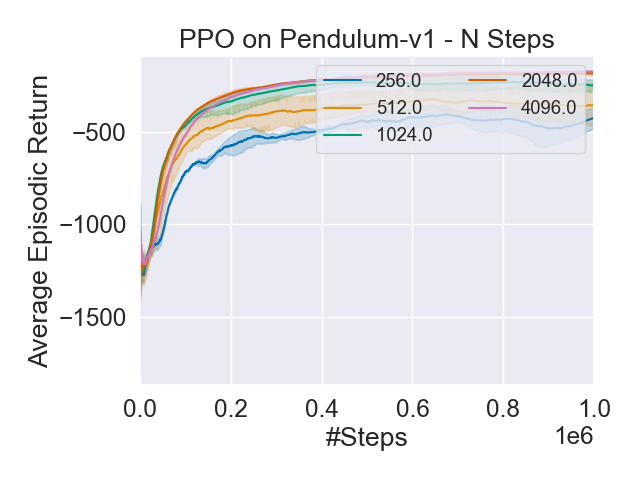}
    \includegraphics[width=0.2\textwidth]{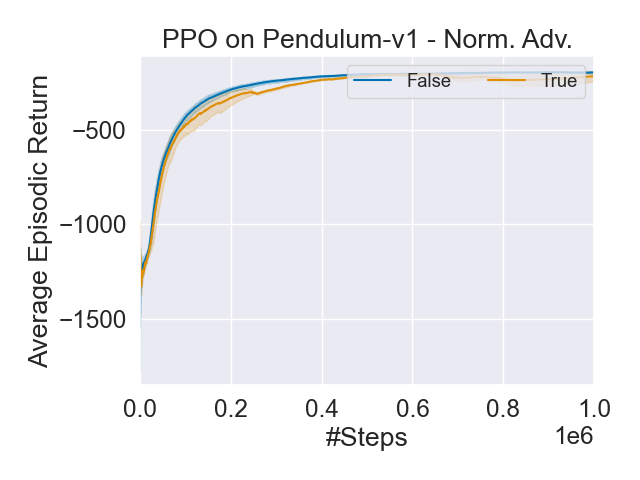}
    \includegraphics[width=0.2\textwidth]{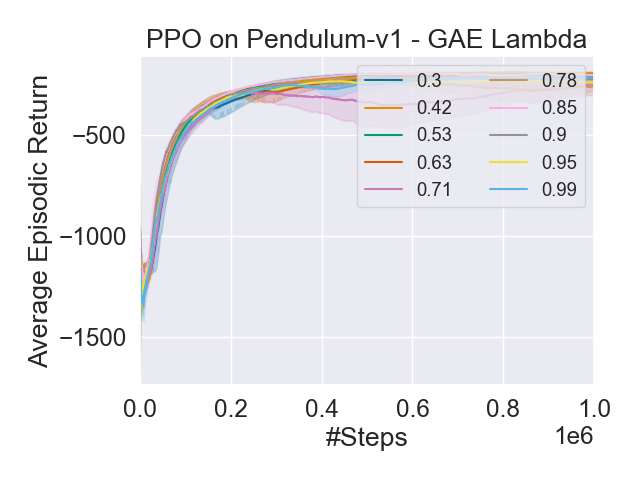}
    \includegraphics[width=0.2\textwidth]{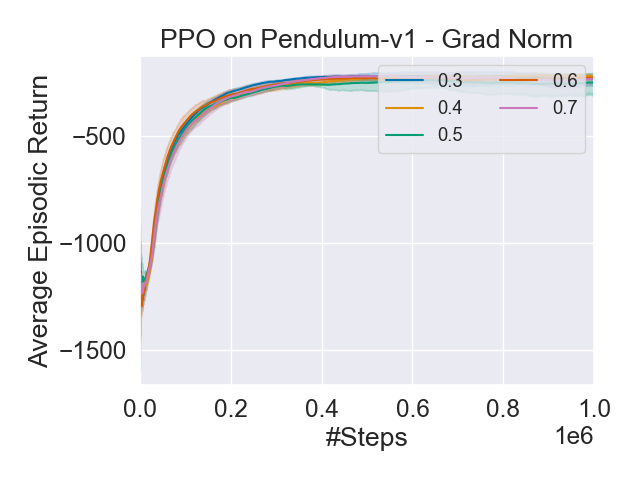}
    \includegraphics[width=0.2\textwidth]{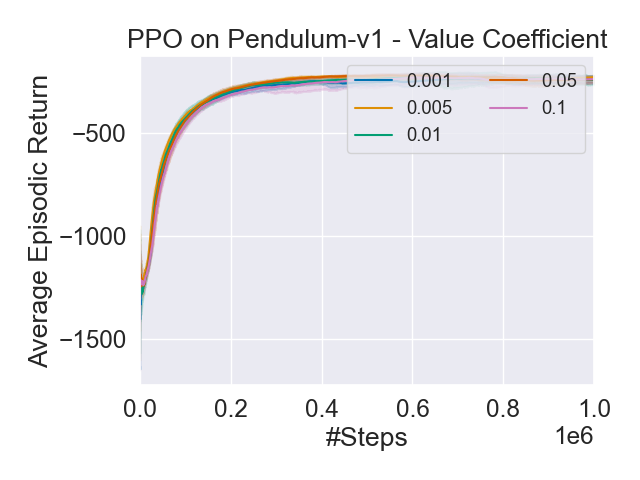}
    \includegraphics[width=0.2\textwidth]{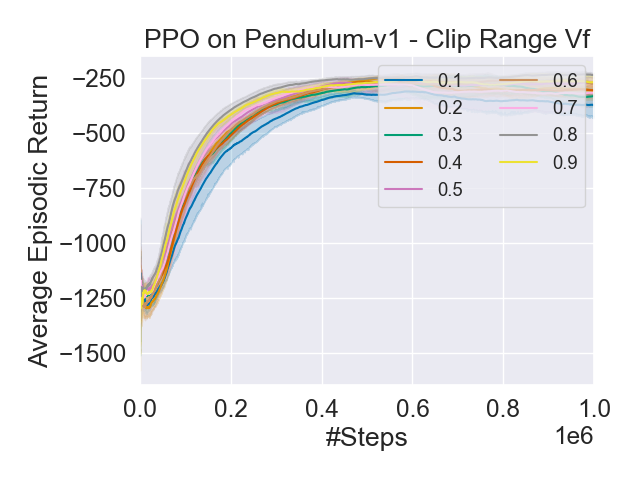}
    \includegraphics[width=0.2\textwidth]{sweep_Acrobot-v1_PPO_algorithm.model_kwargs.learning_rate}
    \includegraphics[width=0.2\textwidth]{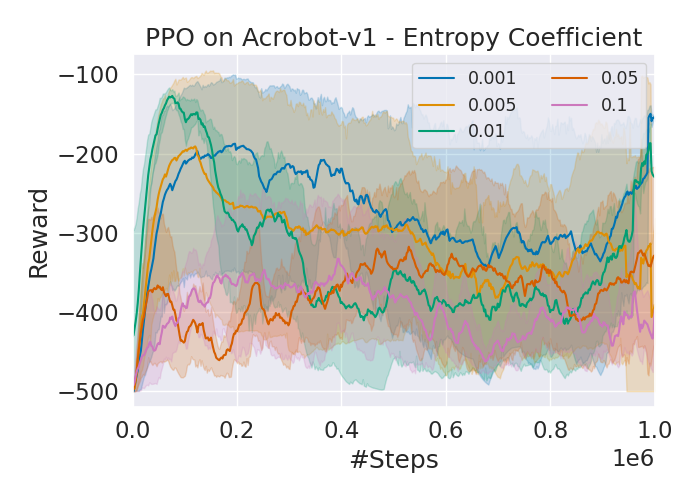}
    \includegraphics[width=0.2\textwidth]{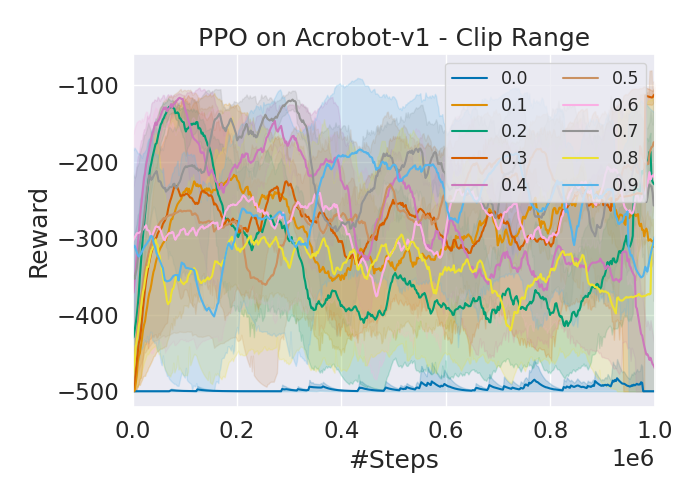}
    \includegraphics[width=0.2\textwidth]{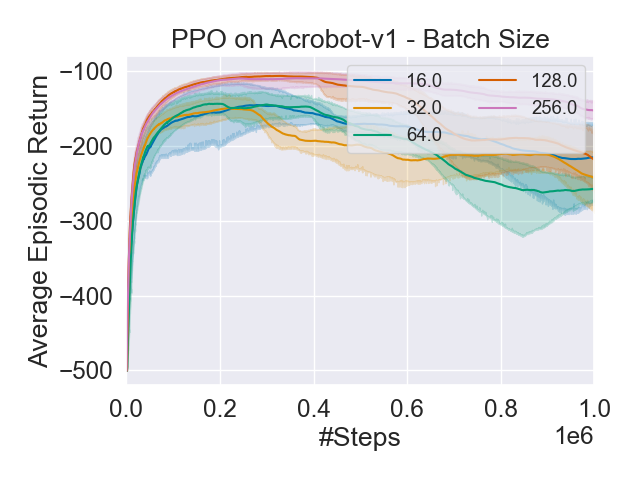}
    \includegraphics[width=0.2\textwidth]{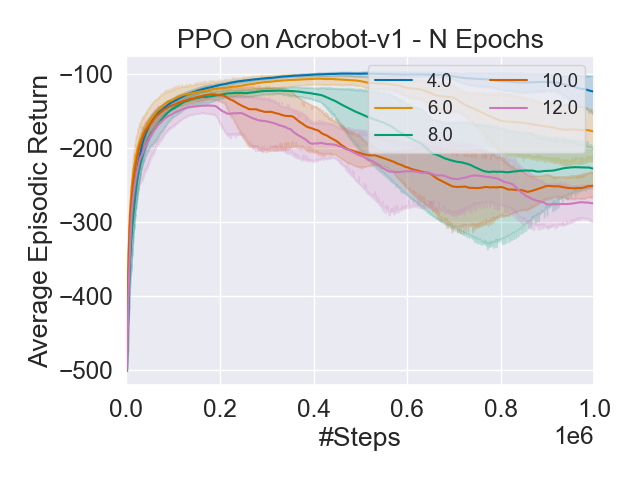}
    \includegraphics[width=0.2\textwidth]{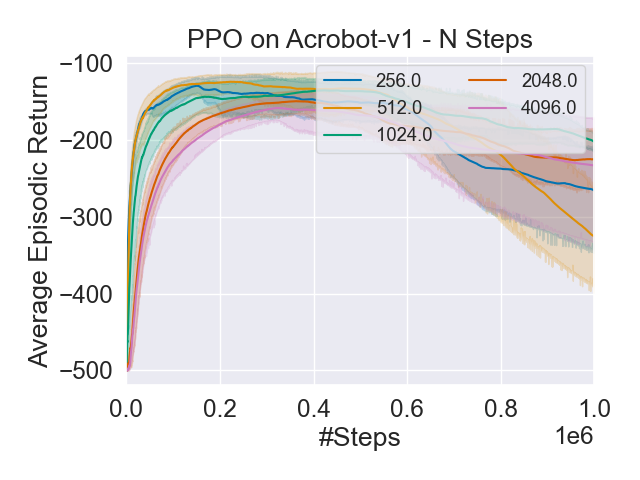}
    \includegraphics[width=0.2\textwidth]{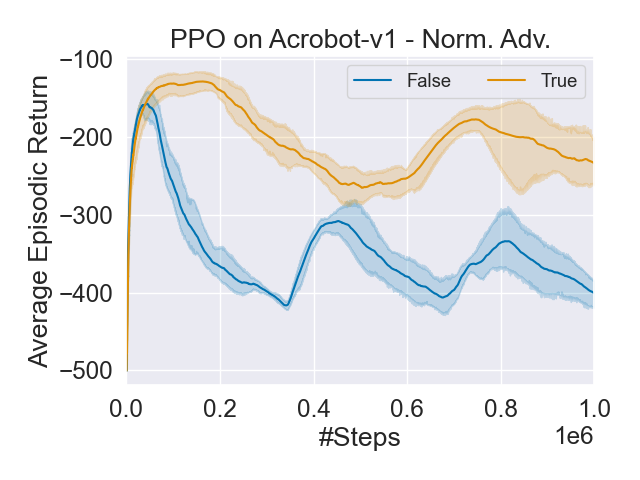}
    \includegraphics[width=0.2\textwidth]{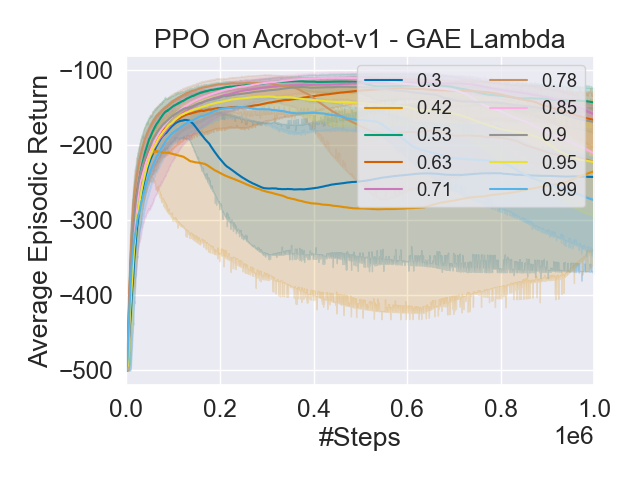}
    \includegraphics[width=0.2\textwidth]{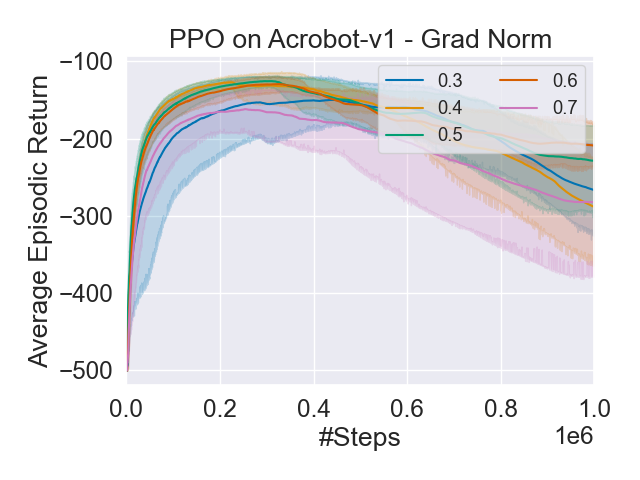}
    \includegraphics[width=0.2\textwidth]{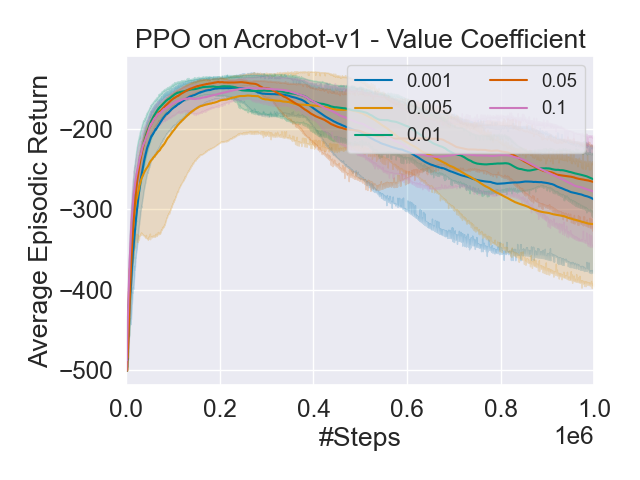}
    \includegraphics[width=0.2\textwidth]{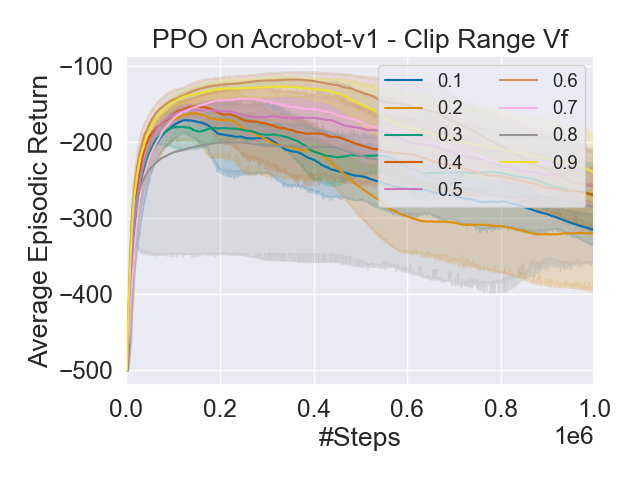}
    \caption{Hyperparameter Sweeps for PPO on Pendulum and Acrobot.}
    \label{app-fig:ppo_sweeps}
\end{figure}

\begin{figure}
    \centering
    \includegraphics[width=0.2\textwidth]{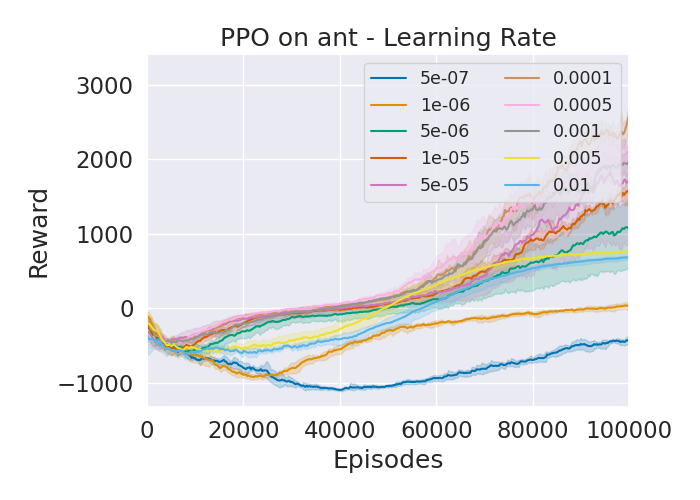}
    \includegraphics[width=0.2\textwidth]{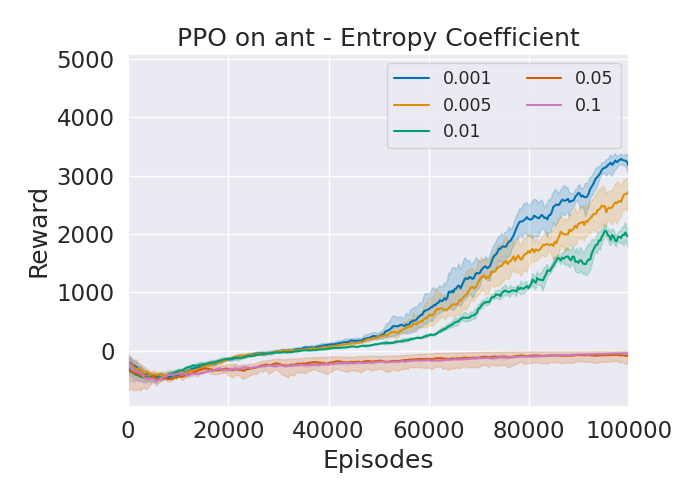}
    \includegraphics[width=0.2\textwidth]{sweep_ant_PPO_algorithm.model_kwargs.clip_range}
    \includegraphics[width=0.2\textwidth]{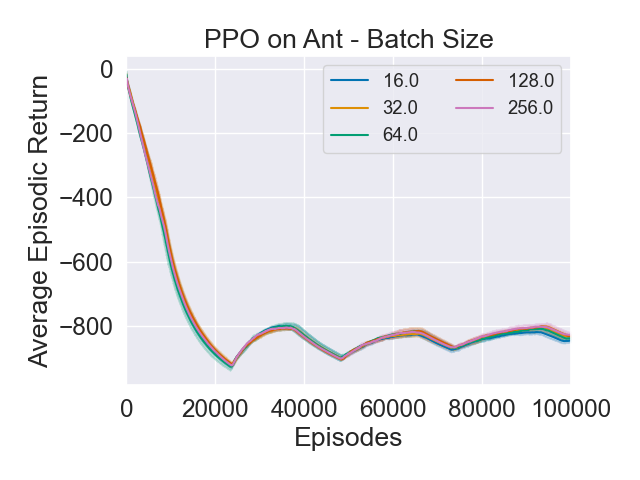}
    \includegraphics[width=0.2\textwidth]{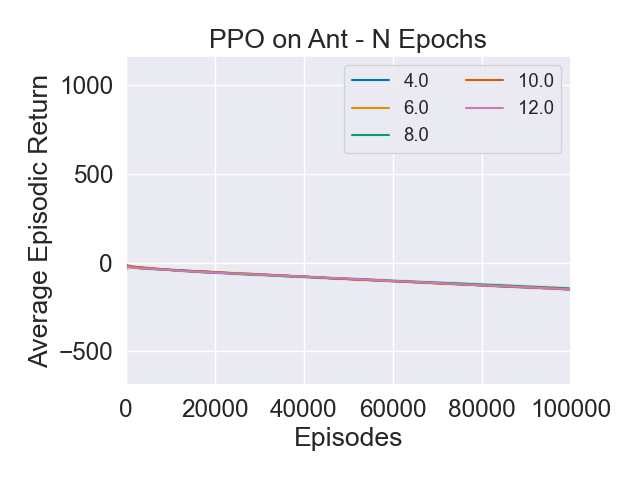}
    \includegraphics[width=0.2\textwidth]{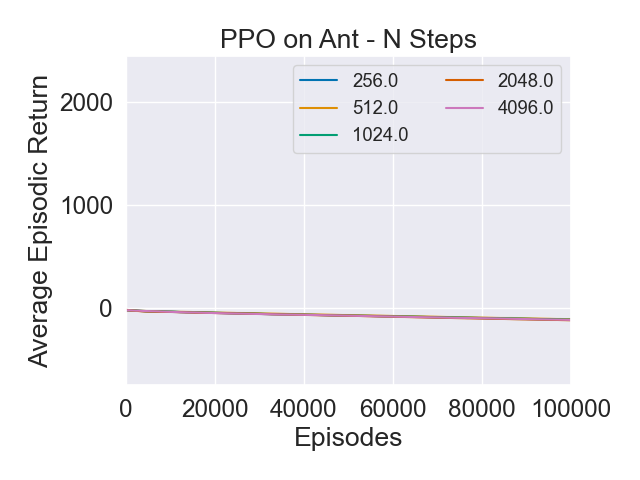}
    \includegraphics[width=0.2\textwidth]{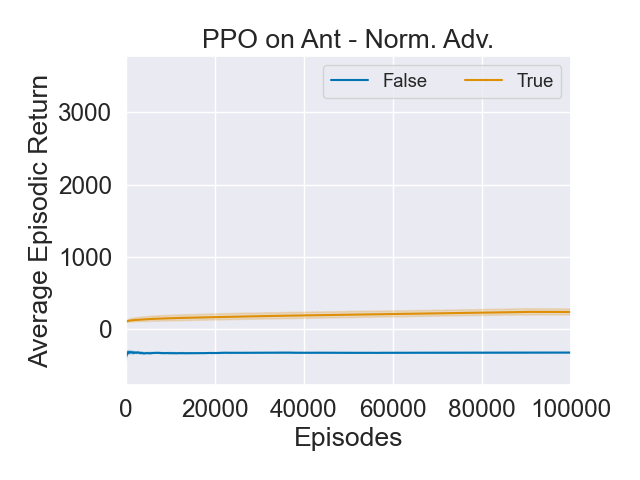}
    \includegraphics[width=0.2\textwidth]{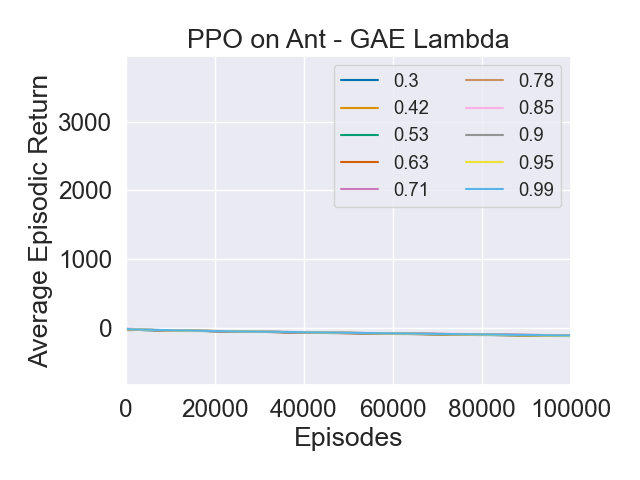}
    \includegraphics[width=0.2\textwidth]{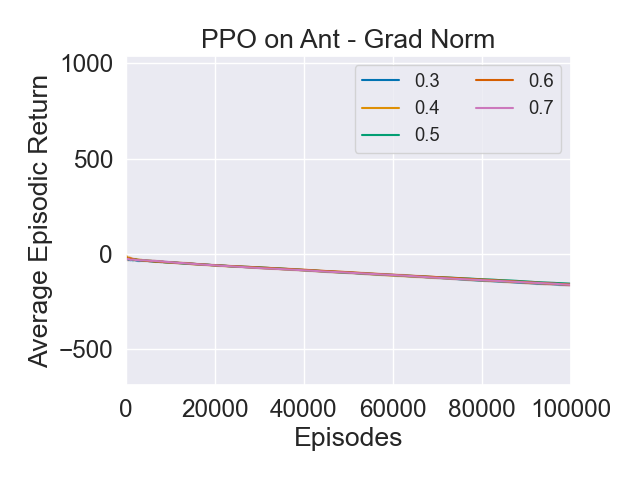}
    \includegraphics[width=0.2\textwidth]{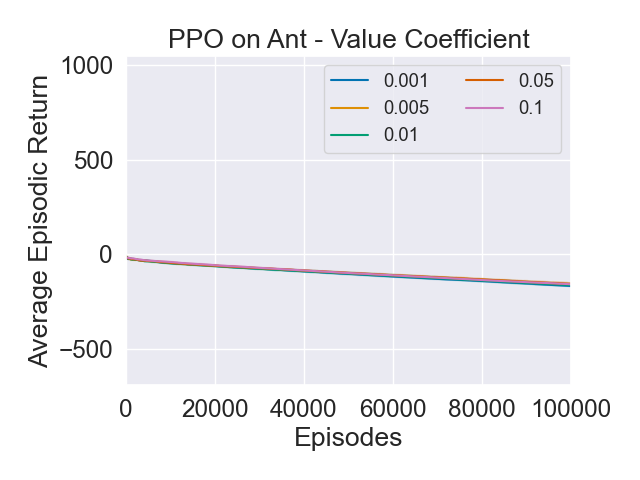}
    \includegraphics[width=0.2\textwidth]{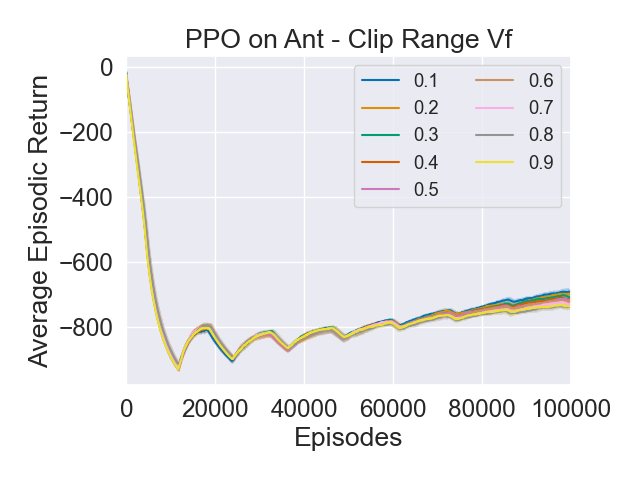}
    \includegraphics[width=0.2\textwidth]{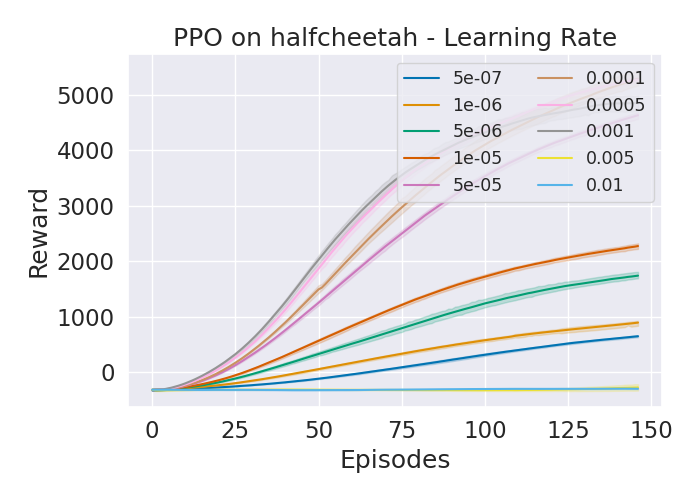}
    \includegraphics[width=0.2\textwidth]{sweep_halfcheetah_PPO_algorithm.model_kwargs.ent_coef}
    \includegraphics[width=0.2\textwidth]{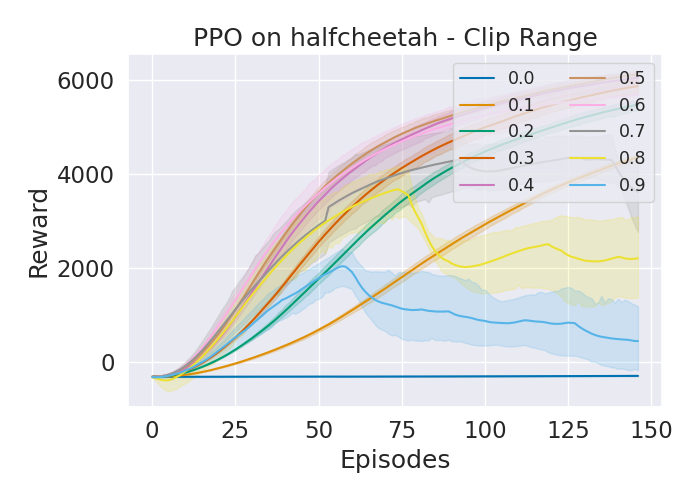}
    \includegraphics[width=0.2\textwidth]{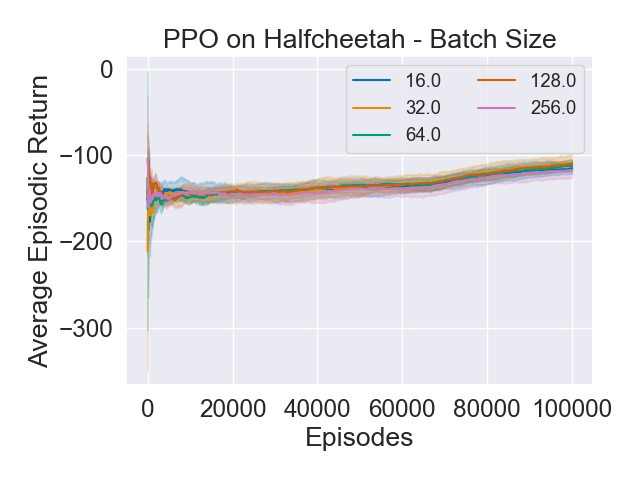}
    \includegraphics[width=0.2\textwidth]{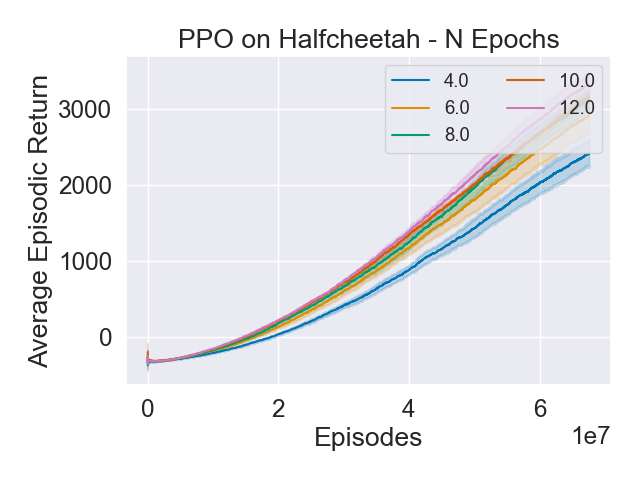}
    \includegraphics[width=0.2\textwidth]{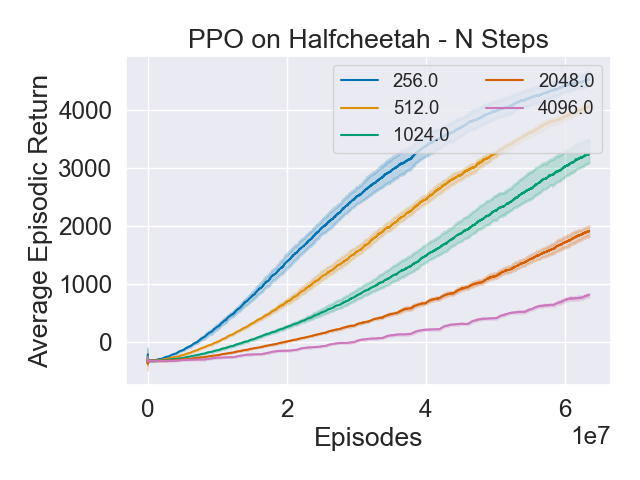}
    \includegraphics[width=0.2\textwidth]{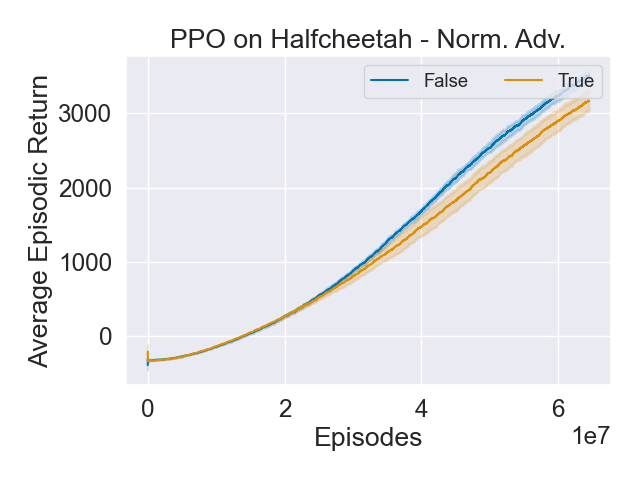}
    \includegraphics[width=0.2\textwidth]{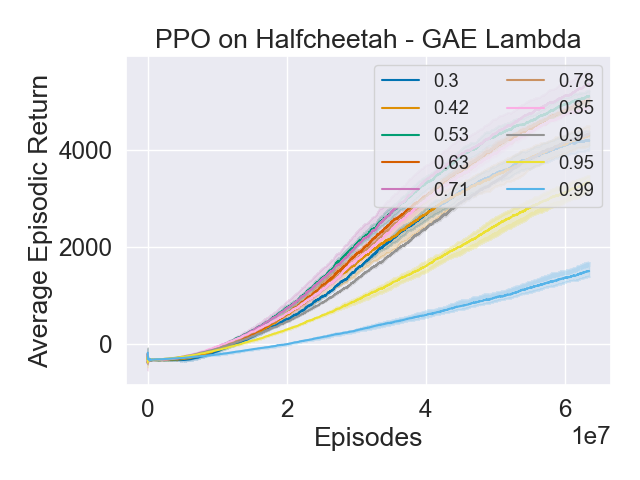}
    \includegraphics[width=0.2\textwidth]{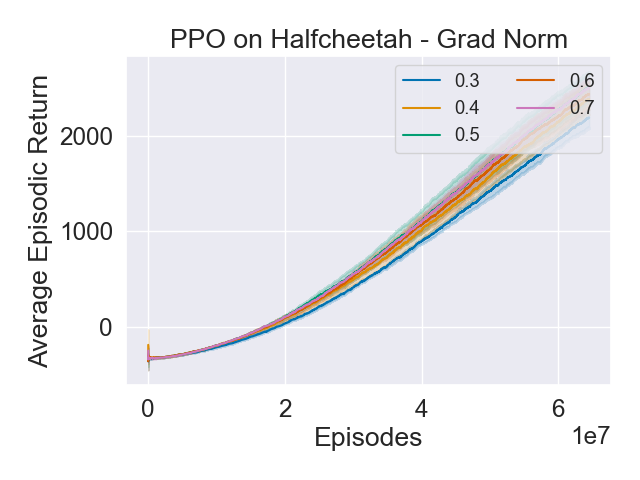}
    \includegraphics[width=0.2\textwidth]{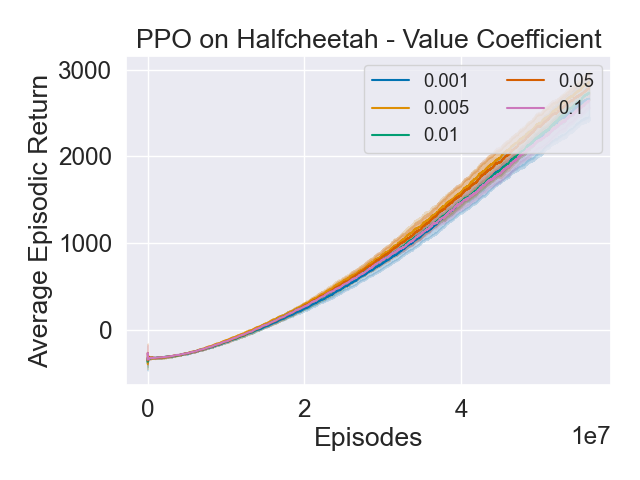}
    \includegraphics[width=0.2\textwidth]{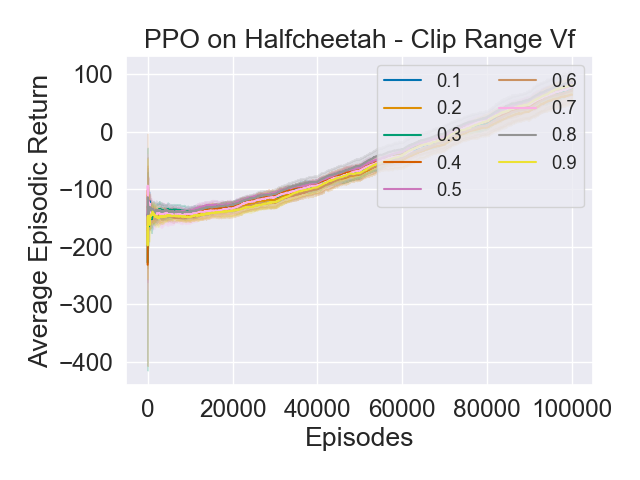}
    \caption{Hyperparameter Sweeps for PPO on Ant and Halfcheetah.}
    \label{app-fig:ppo_sweeps2}
\end{figure}

\begin{figure}
    \centering
    \includegraphics[width=0.2\textwidth]{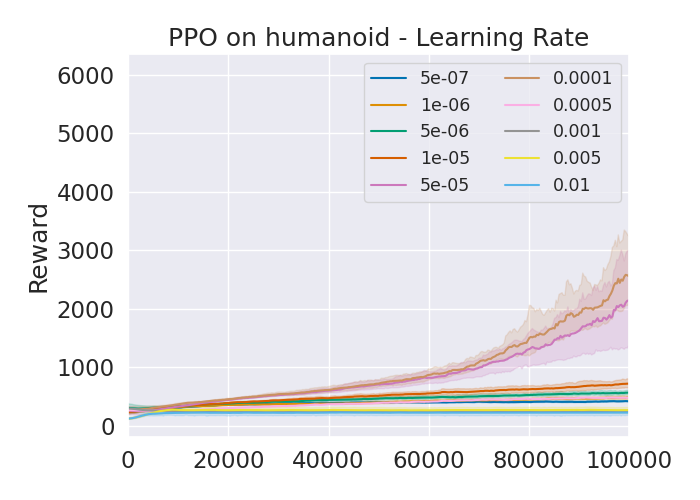}
    \includegraphics[width=0.2\textwidth]{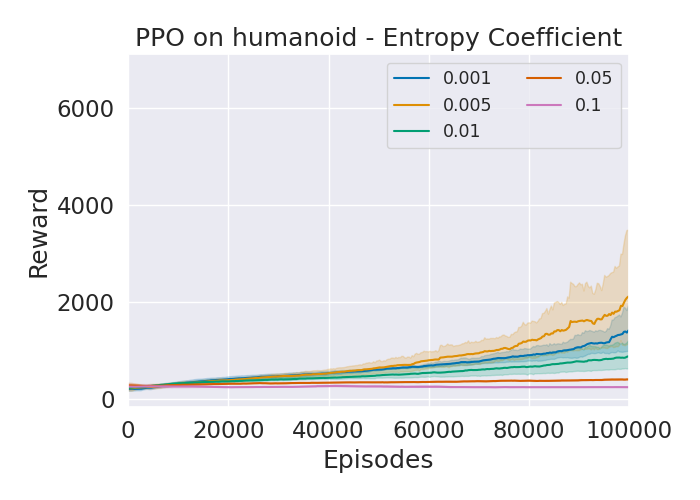}
    \includegraphics[width=0.2\textwidth]{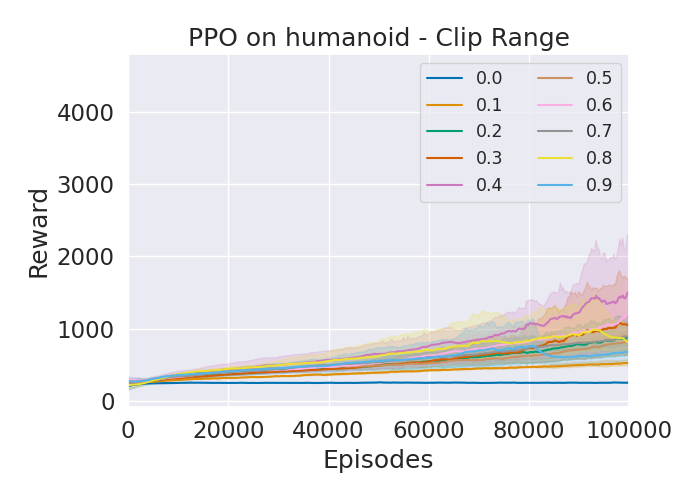}
    \includegraphics[width=0.2\textwidth]{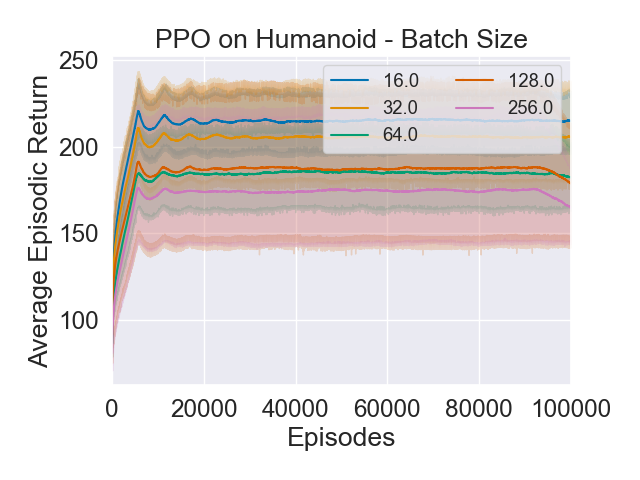}
    \includegraphics[width=0.2\textwidth]{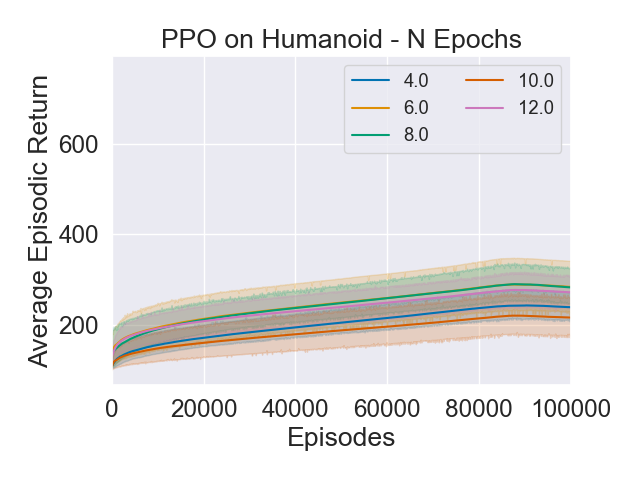}
    \includegraphics[width=0.2\textwidth]{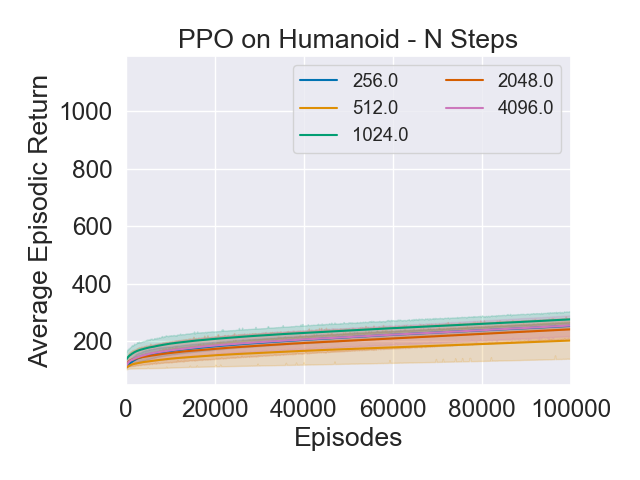}
    \includegraphics[width=0.2\textwidth]{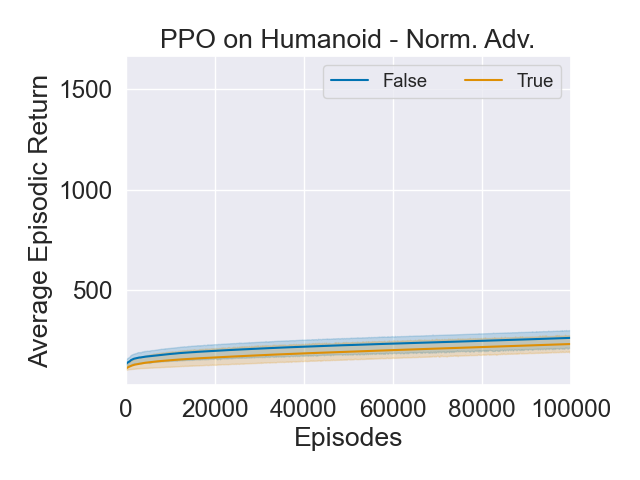}
    \includegraphics[width=0.2\textwidth]{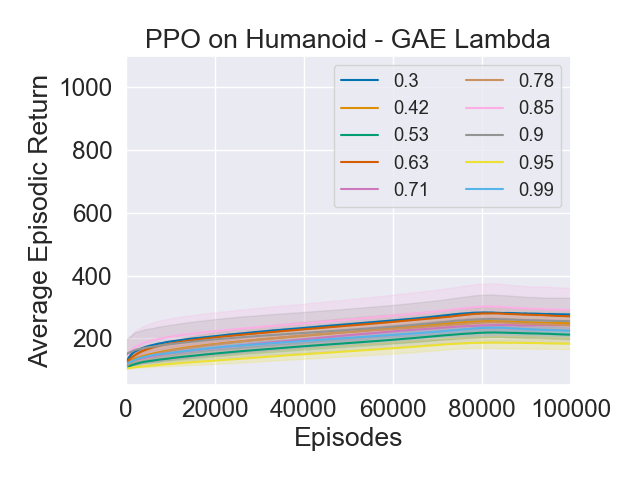}
    \includegraphics[width=0.2\textwidth]{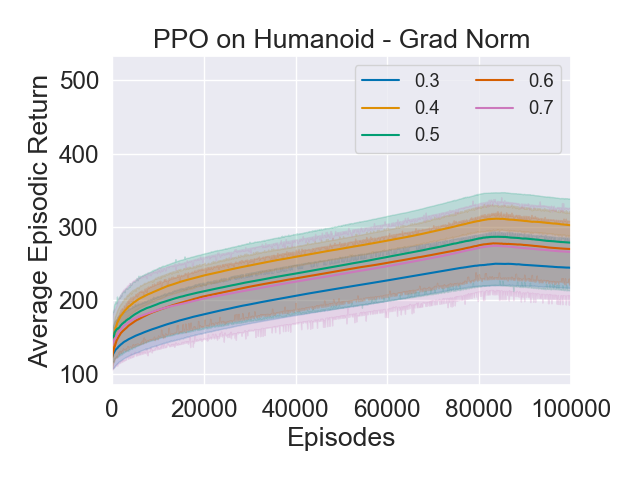}
    \includegraphics[width=0.2\textwidth]{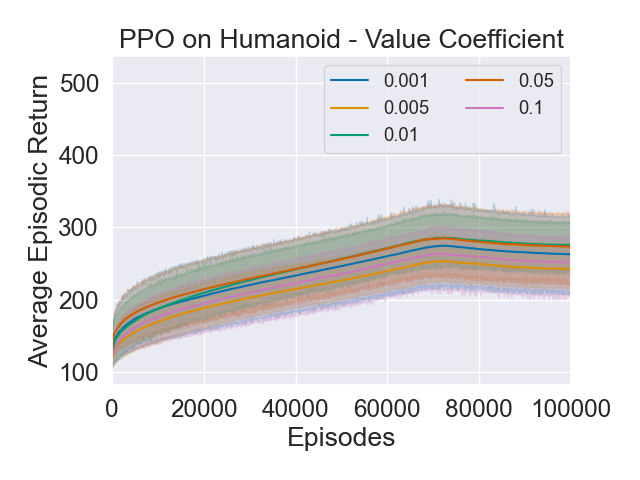}
    \includegraphics[width=0.2\textwidth]{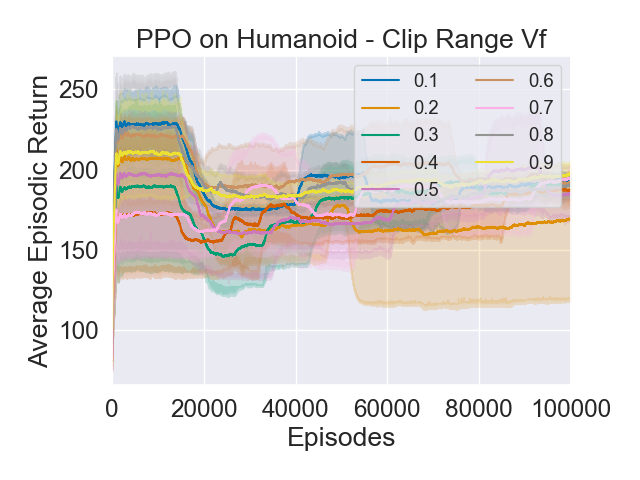}
    \caption{Hyperparameter Sweeps for PPO on Humanoid.}
    \label{app-fig:ppo_sweeps3}
\end{figure}

\begin{figure}
    \centering
    \includegraphics[width=0.2\textwidth]{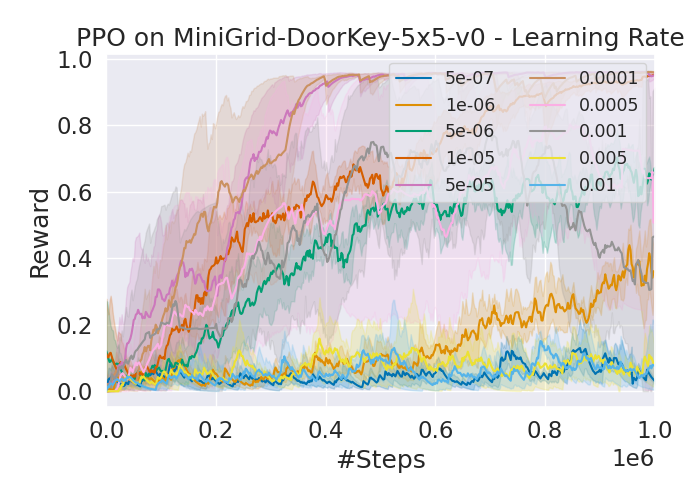}
    \includegraphics[width=0.2\textwidth]{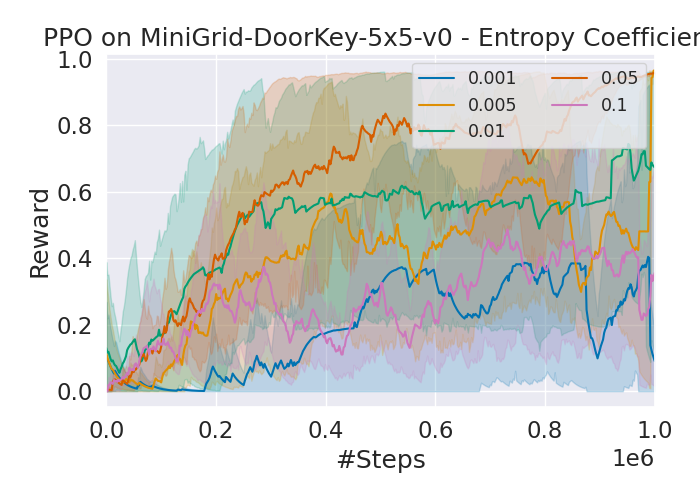}
    \includegraphics[width=0.2\textwidth]{sweep_MiniGrid-DoorKey-5x5-v0_PPO_algorithm.model_kwargs.clip_range}
    \includegraphics[width=0.2\textwidth]{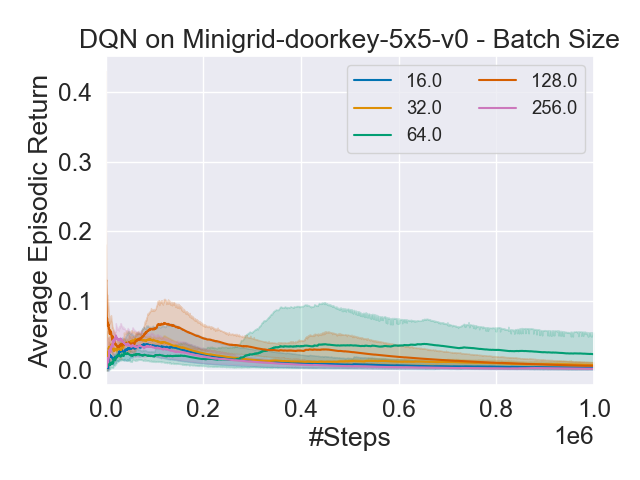}
    \includegraphics[width=0.2\textwidth]{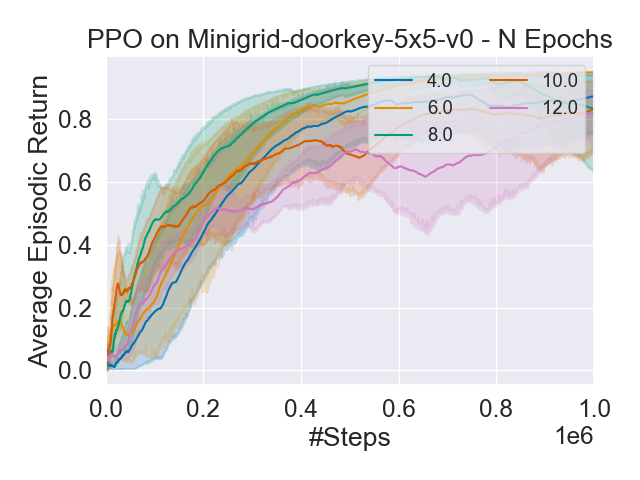}
    \includegraphics[width=0.2\textwidth]{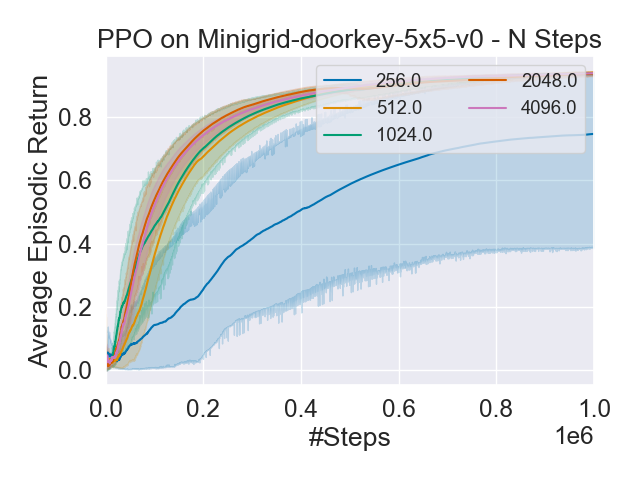}
    \includegraphics[width=0.2\textwidth]{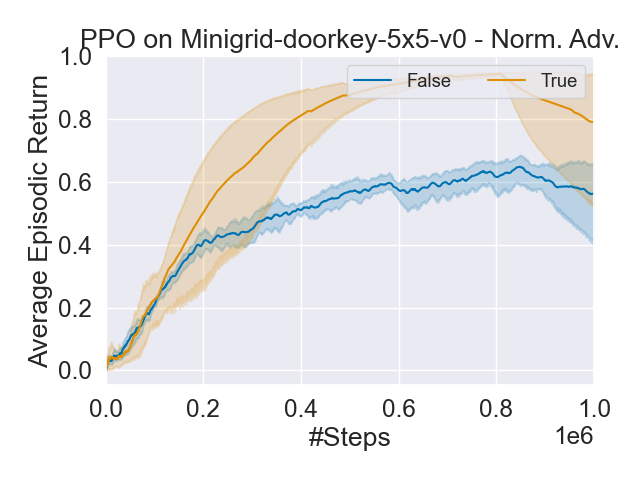}
    \includegraphics[width=0.2\textwidth]{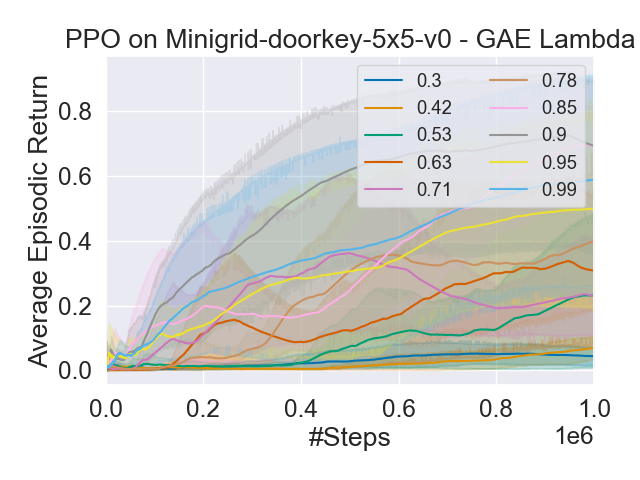}
    \includegraphics[width=0.2\textwidth]{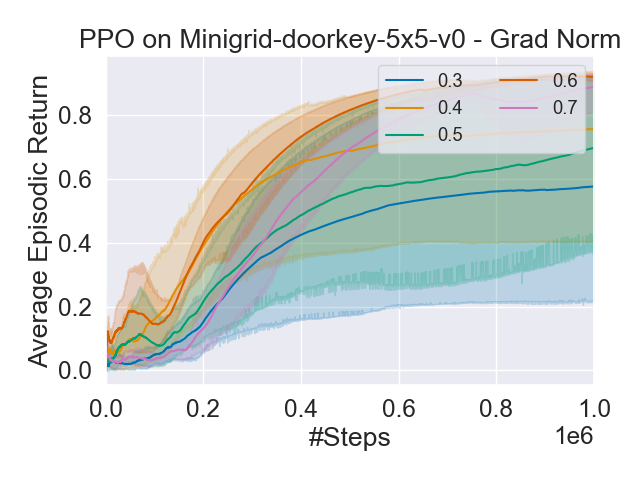}
    \includegraphics[width=0.2\textwidth]{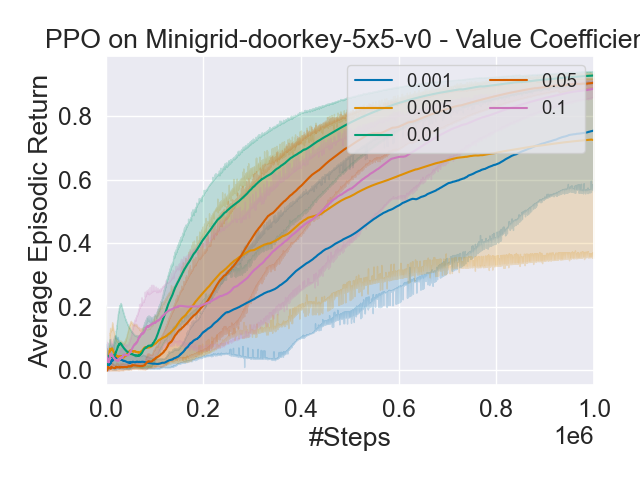}
    \includegraphics[width=0.2\textwidth]{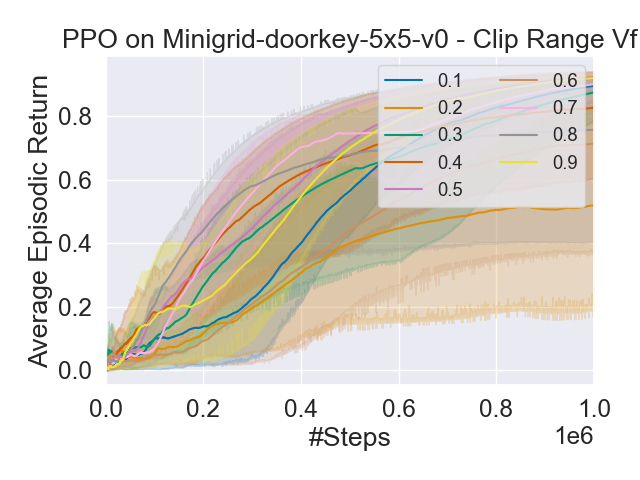}
    \includegraphics[width=0.2\textwidth]{sweep_MiniGrid-Empty-5x5-v0_PPO_algorithm.model_kwargs.learning_rate}
    \includegraphics[width=0.2\textwidth]{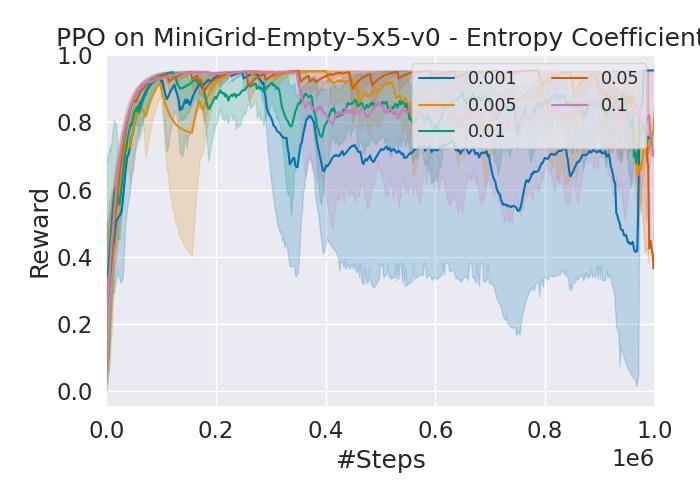}
    \includegraphics[width=0.2\textwidth]{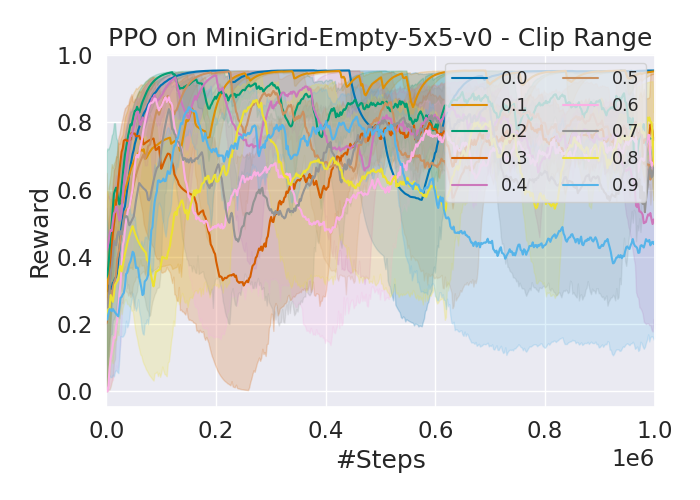}
    \includegraphics[width=0.2\textwidth]{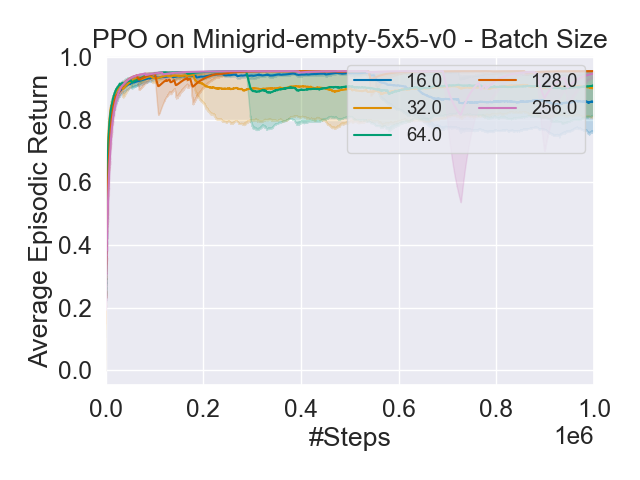}
    \includegraphics[width=0.2\textwidth]{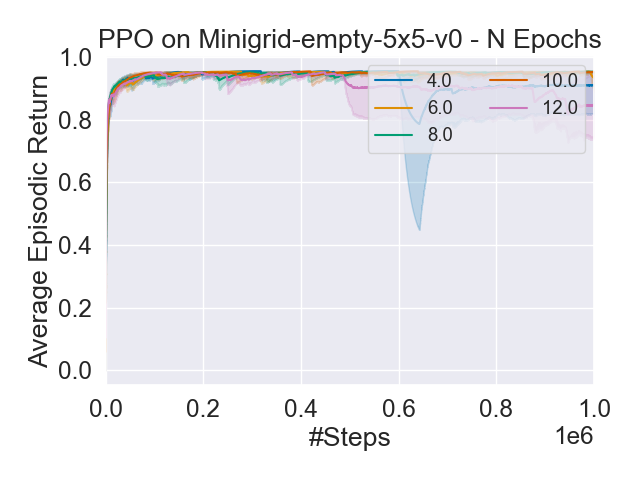}
    \includegraphics[width=0.2\textwidth]{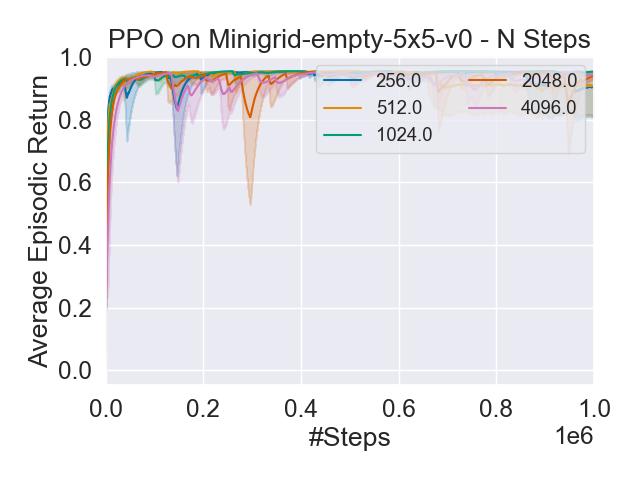}
    \includegraphics[width=0.2\textwidth]{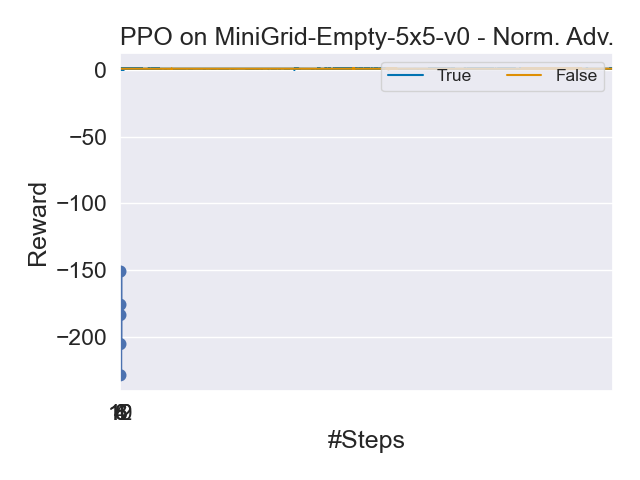}
    \includegraphics[width=0.2\textwidth]{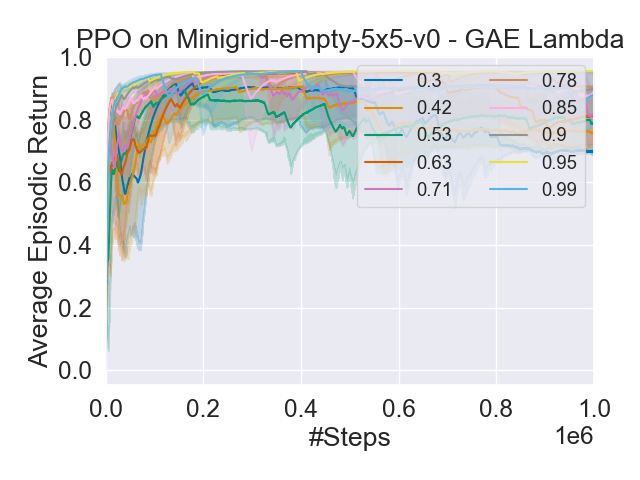}
    \includegraphics[width=0.2\textwidth]{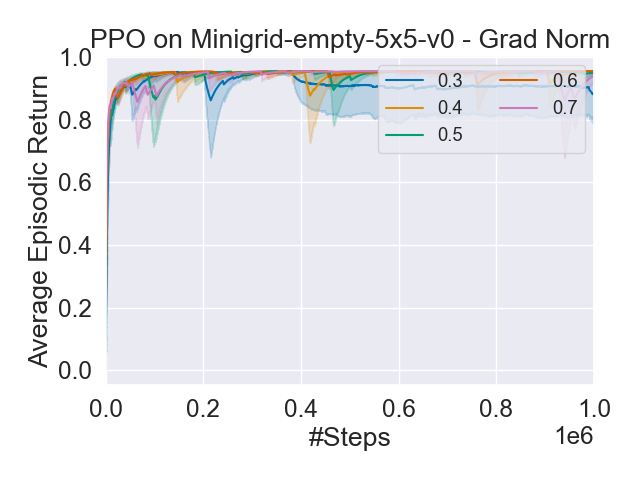}
    \includegraphics[width=0.2\textwidth]{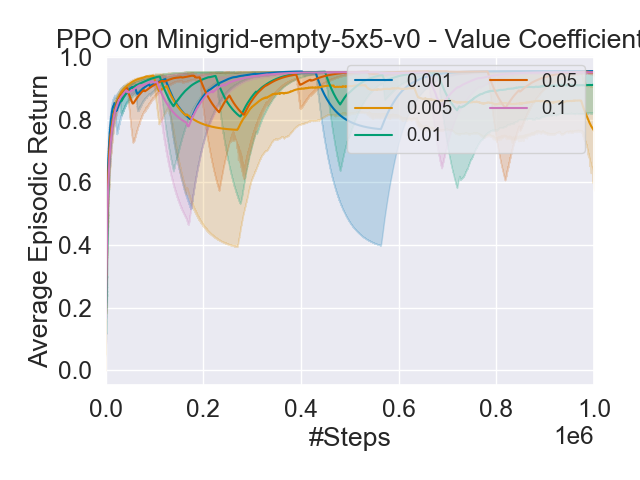}
    \includegraphics[width=0.2\textwidth]{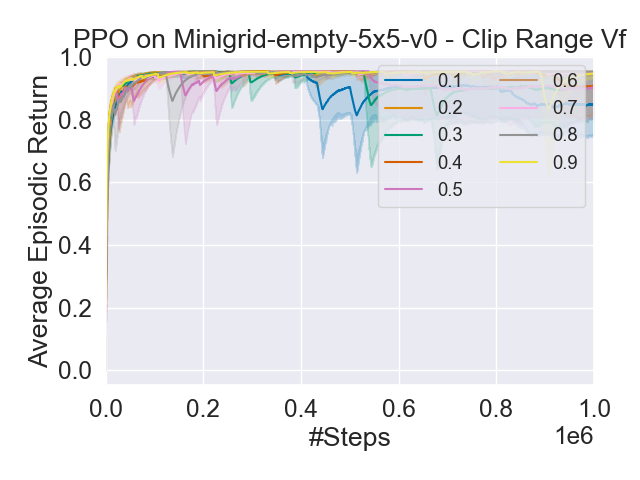}
    \caption{Hyperparameter Sweeps for PPO on MiniGrid.}
    \label{app-fig:ppo_sweeps4}
\end{figure}

\begin{figure}[h]
    \centering
    \includegraphics[width=0.2\textwidth]{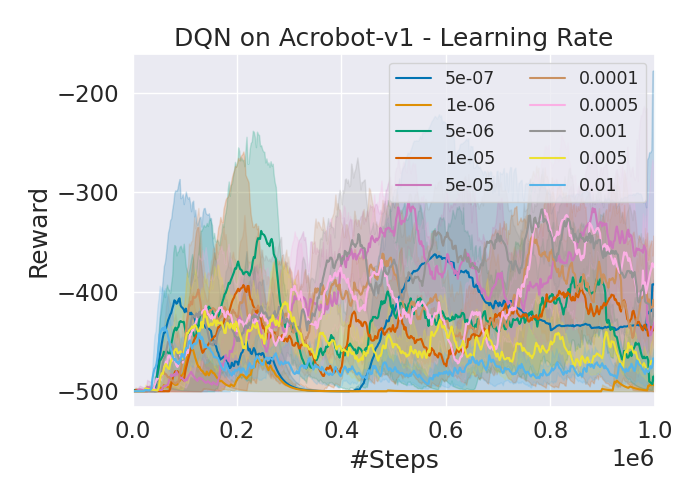}
    \includegraphics[width=0.2\textwidth]{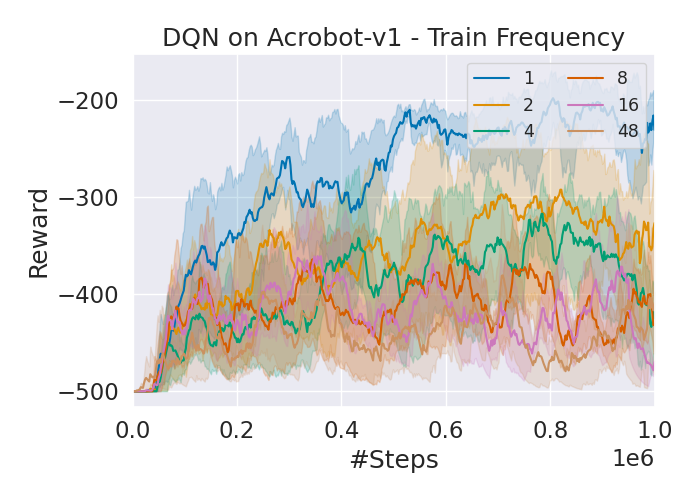}
    \includegraphics[width=0.2\textwidth]{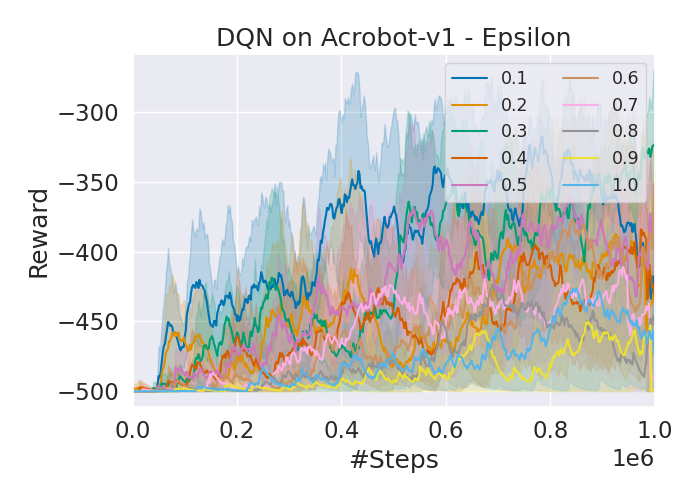}
    \includegraphics[width=0.2\textwidth]{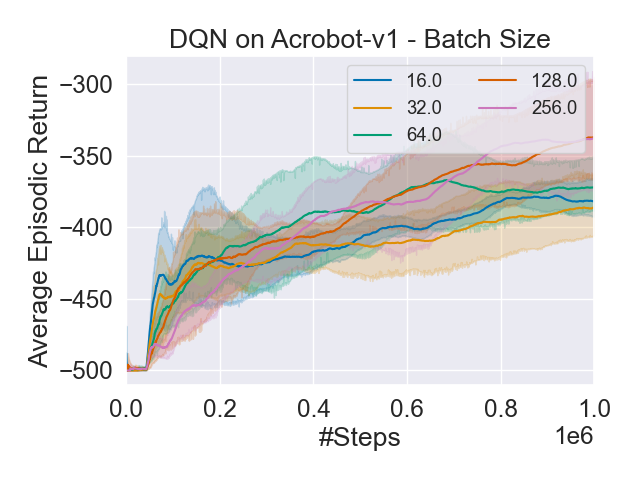}
    \includegraphics[width=0.2\textwidth]{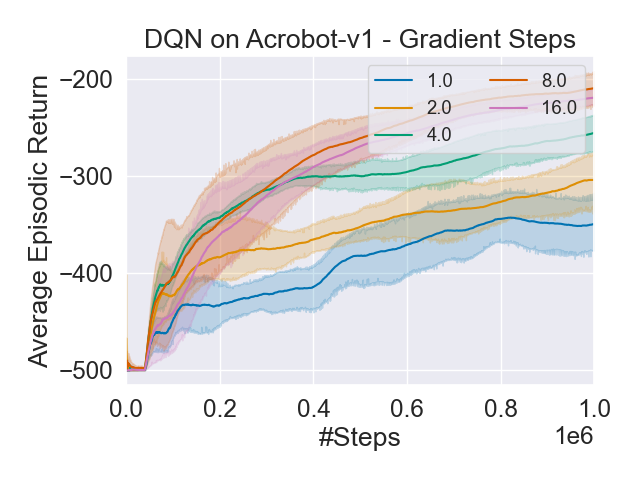}
    \includegraphics[width=0.2\textwidth]{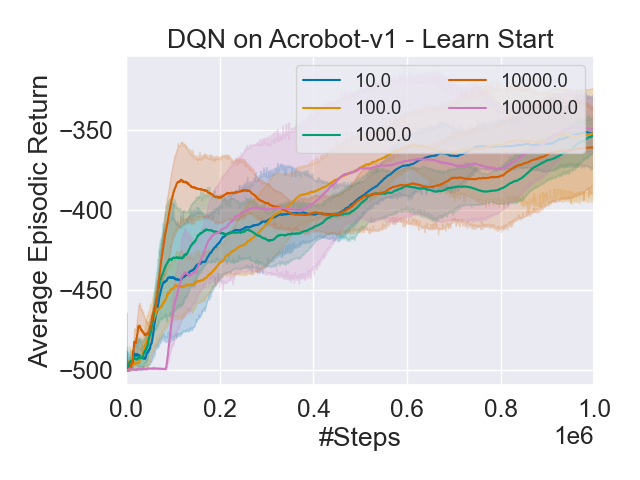}
    \includegraphics[width=0.2\textwidth]{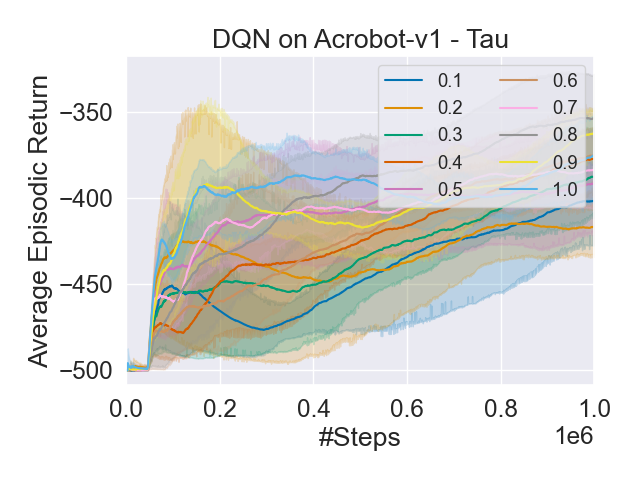}
    \includegraphics[width=0.2
    \textwidth]{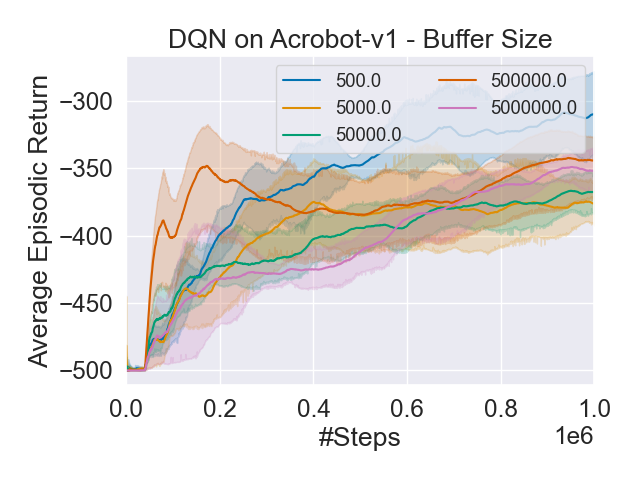}
    \caption{Hyperparameter Sweeps for DQN on Acrobot.}
    \label{app-fig:dqn_sweeps}
\end{figure}

\begin{figure}[h]
    \centering
    \includegraphics[width=0.2\textwidth]{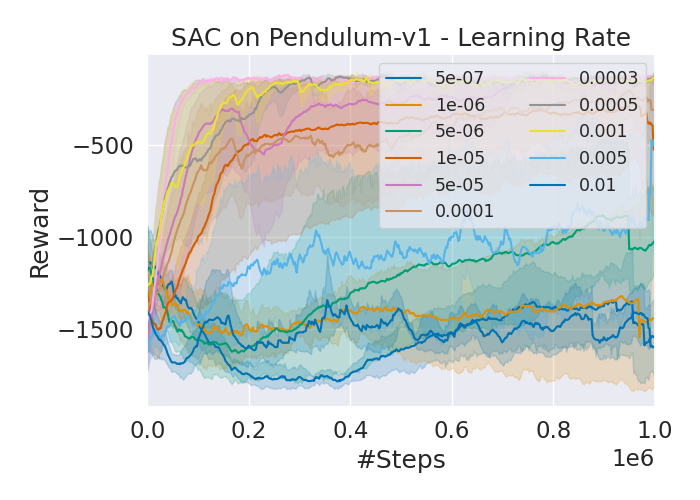}
    \includegraphics[width=0.2\textwidth]{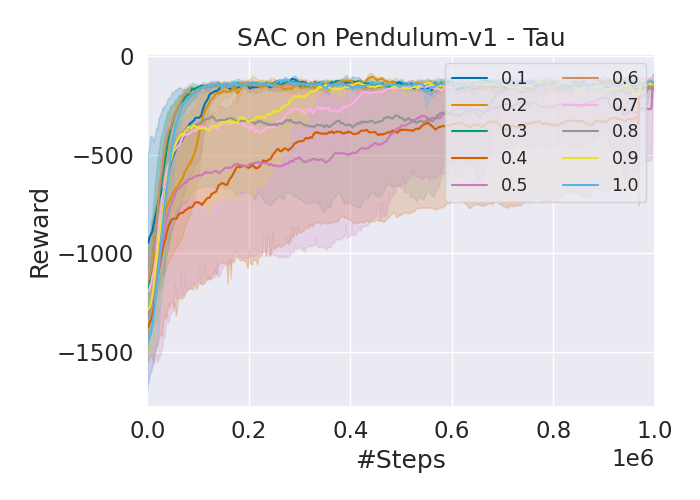}
    \includegraphics[width=0.2\textwidth]{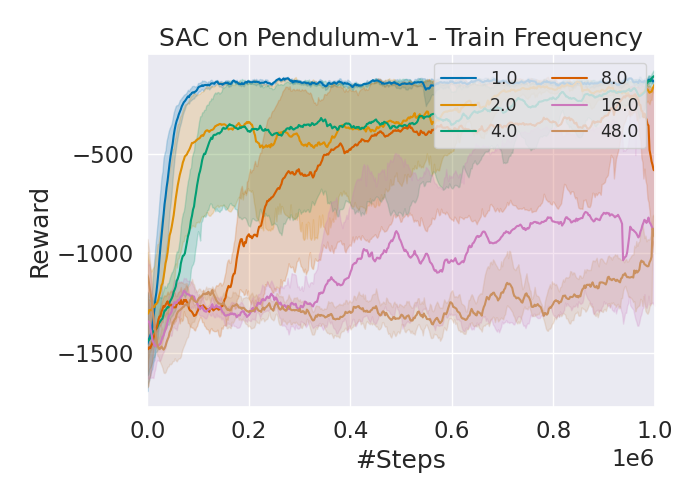}
    \includegraphics[width=0.2\textwidth]{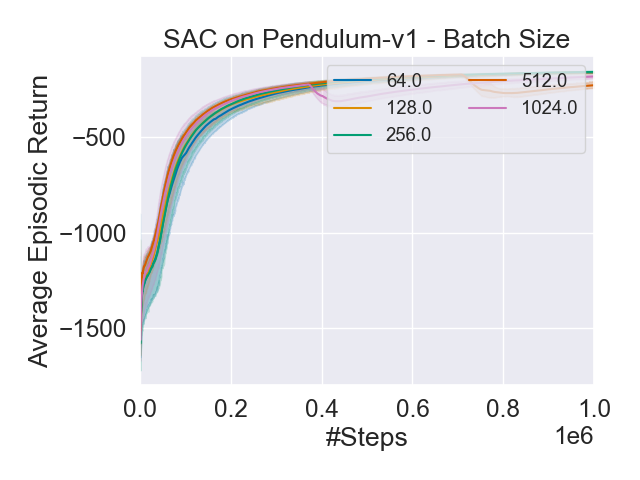}
    \includegraphics[width=0.2\textwidth]{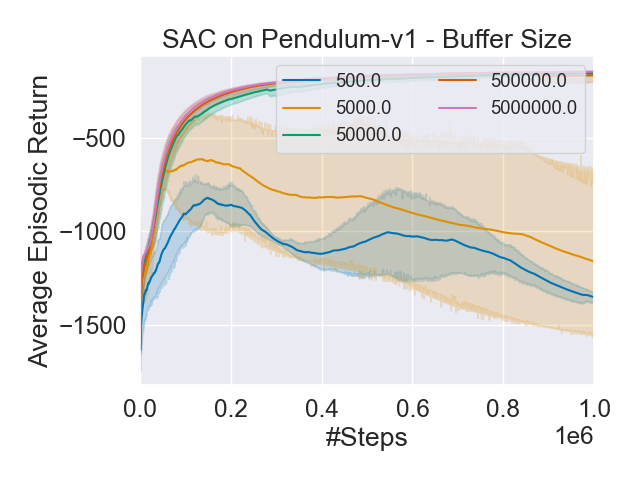}
    \includegraphics[width=0.2\textwidth]{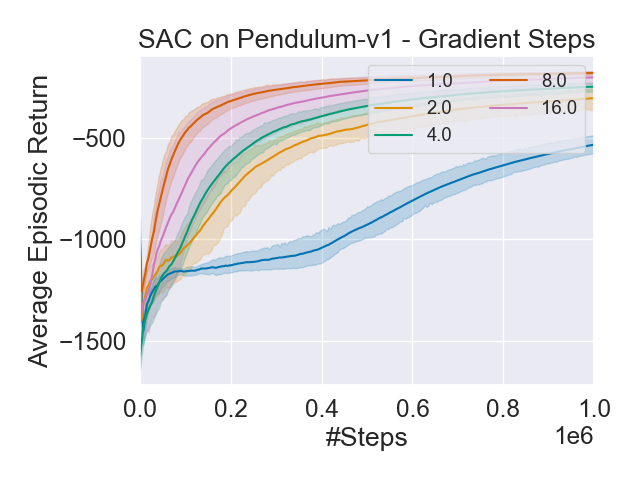}
    \includegraphics[width=0.2\textwidth]{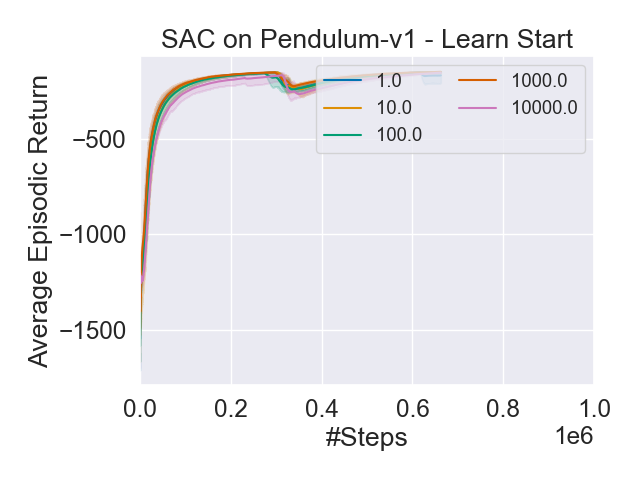}
    \includegraphics[width=0.2\textwidth]{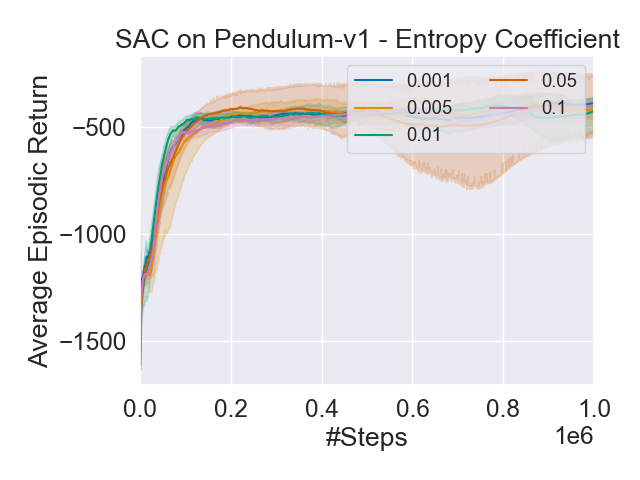}
    \includegraphics[width=0.2\textwidth]{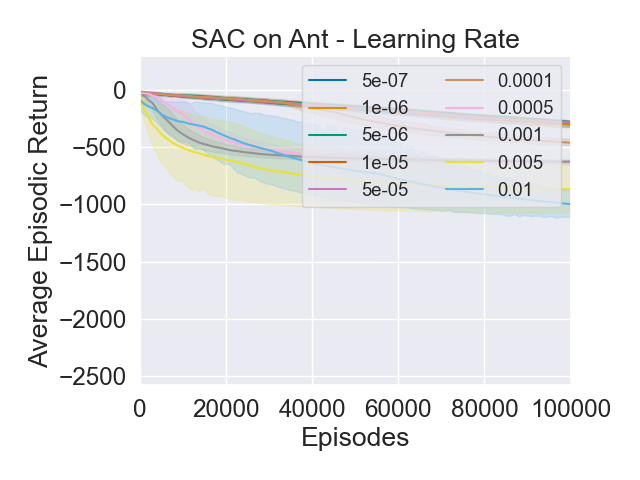}
    \includegraphics[width=0.2\textwidth]{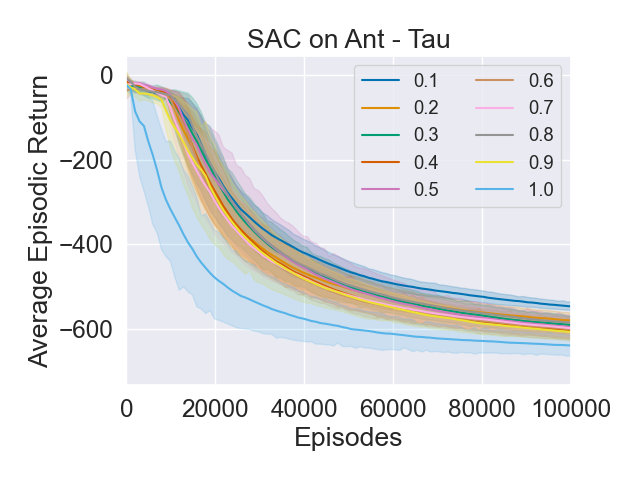}
    \includegraphics[width=0.2\textwidth]{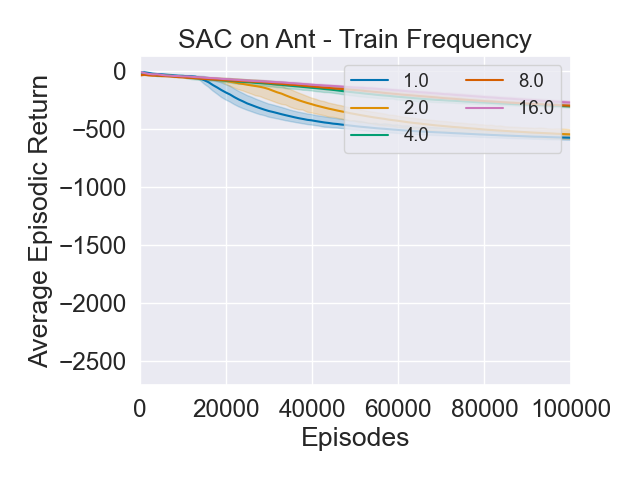}
    \includegraphics[width=0.2\textwidth]{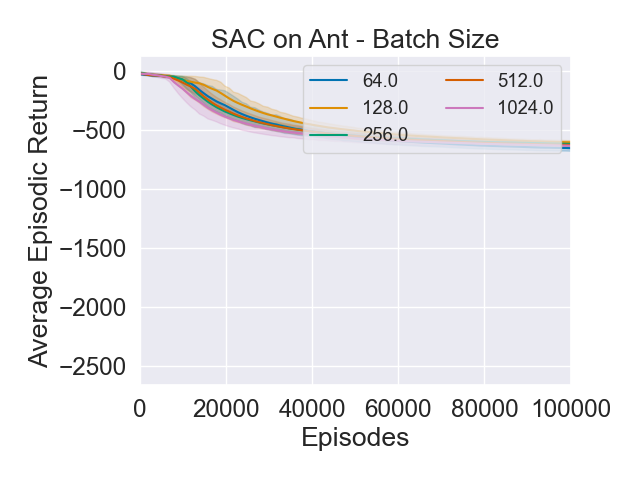}
    \includegraphics[width=0.2\textwidth]{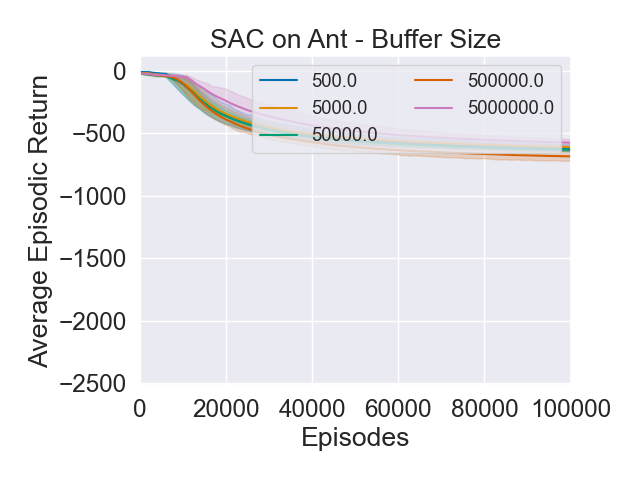}
    \includegraphics[width=0.2\textwidth]{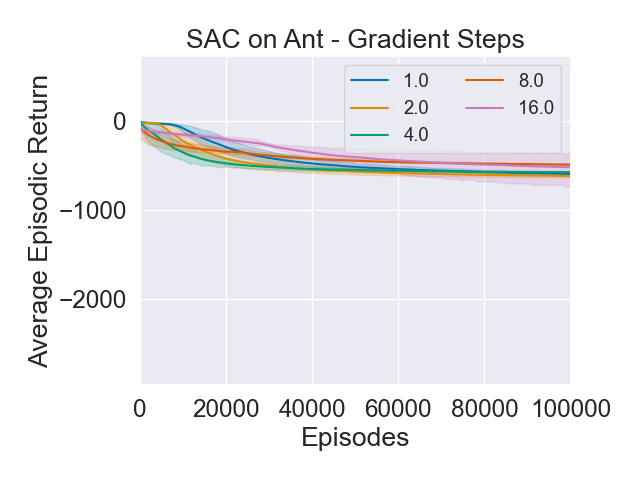}
    \includegraphics[width=0.2\textwidth]{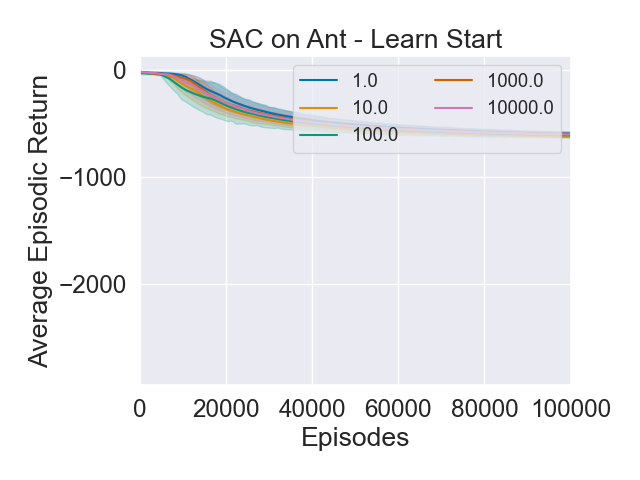}
    \caption{Hyperparameter Sweeps for SAC on Pendulum and Ant.}
    \label{app-fig:sac_sweeps}
\end{figure}

\begin{figure}[h]
    \centering
    \includegraphics[width=0.2\textwidth]{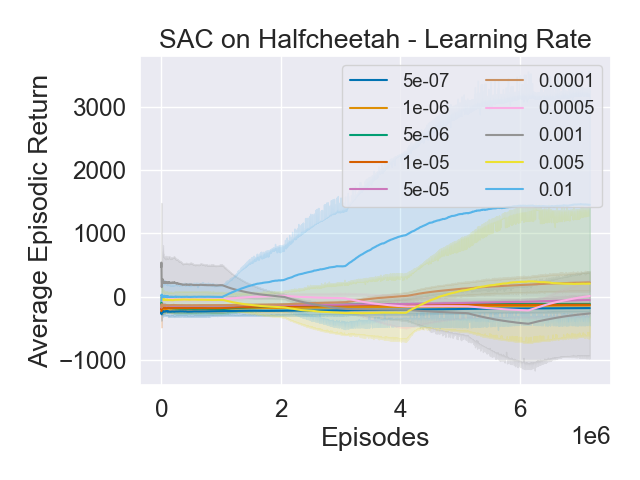}
    \includegraphics[width=0.2\textwidth]{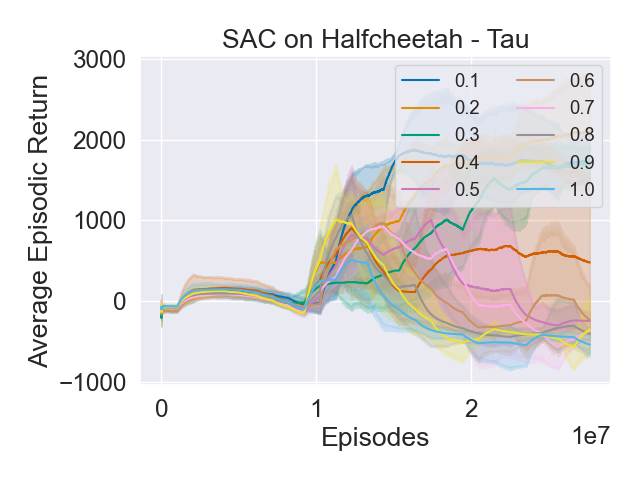}
    \includegraphics[width=0.2\textwidth]{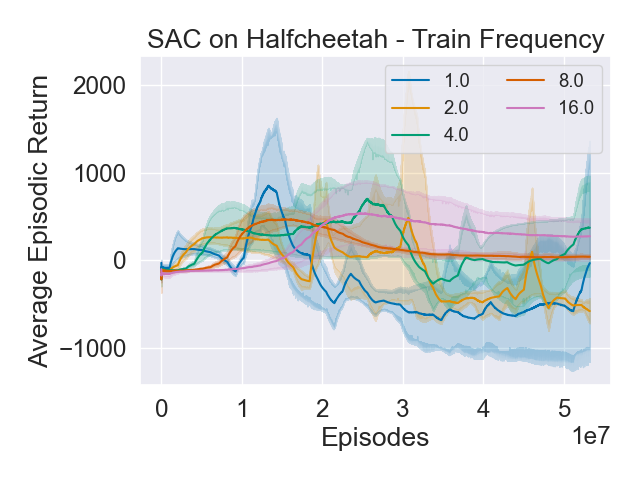}
    \includegraphics[width=0.2\textwidth]{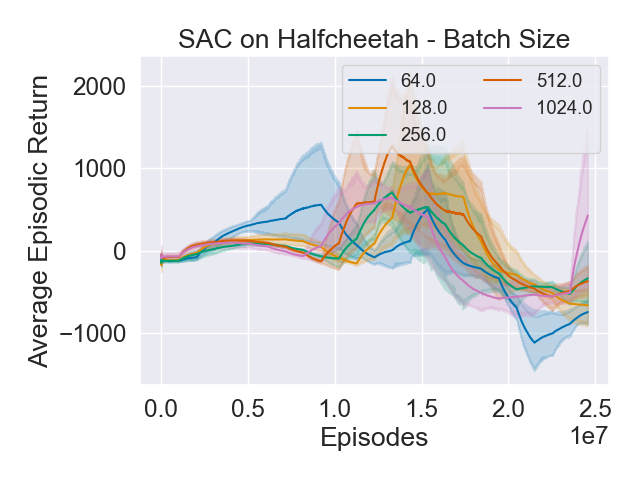}
    \includegraphics[width=0.2\textwidth]{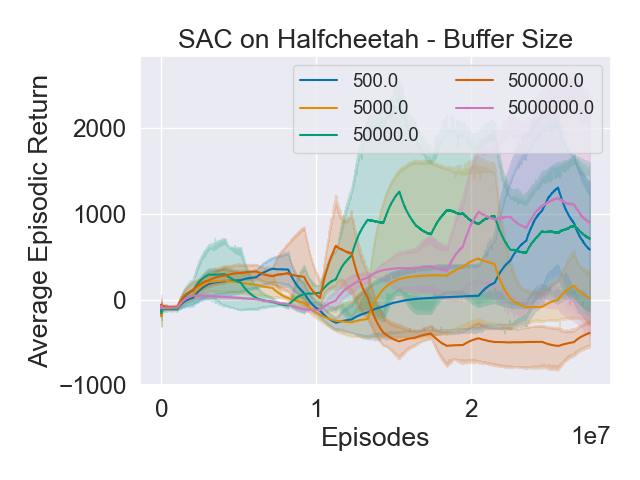}
    \includegraphics[width=0.2\textwidth]{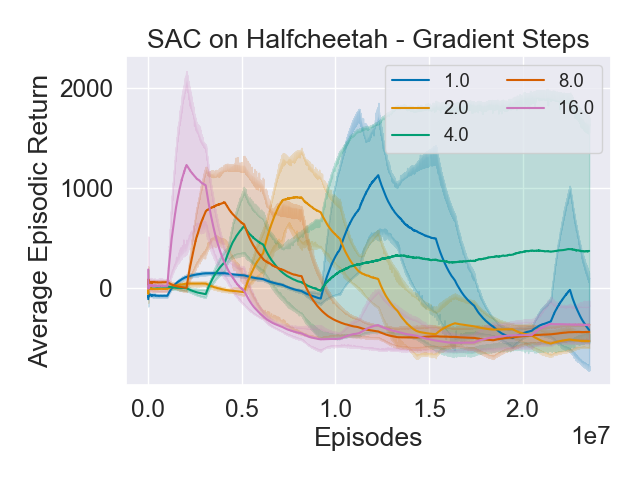}
    \includegraphics[width=0.2\textwidth]{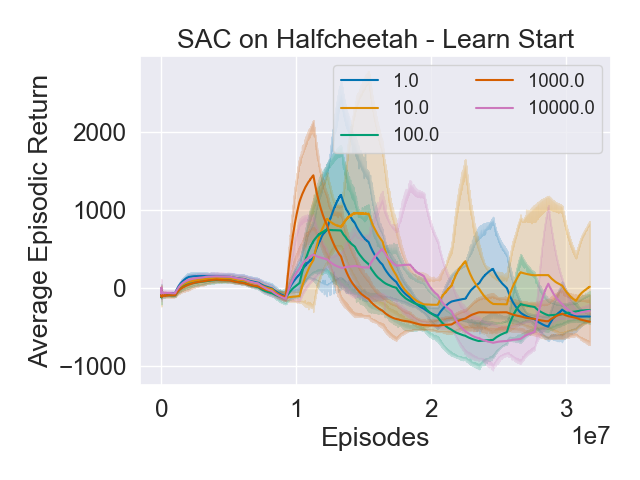}
    \includegraphics[width=0.2\textwidth]{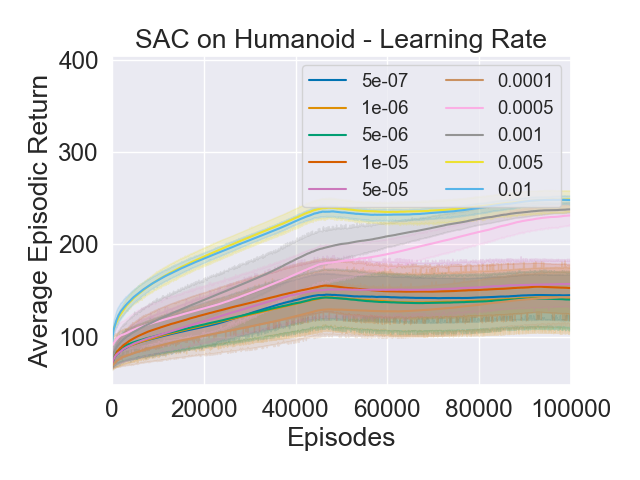}
    \includegraphics[width=0.2\textwidth]{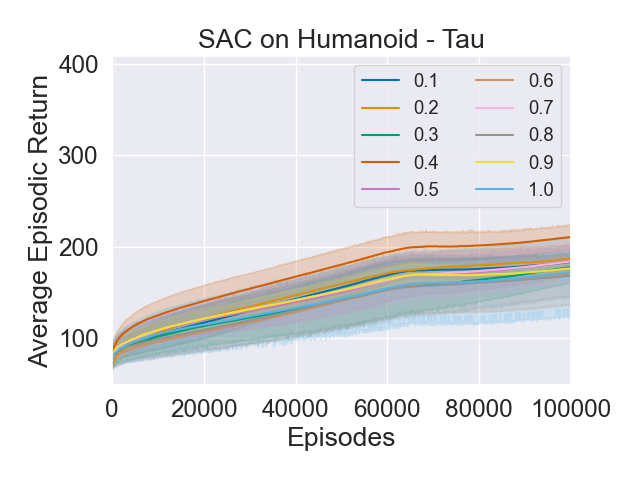}
    \includegraphics[width=0.2\textwidth]{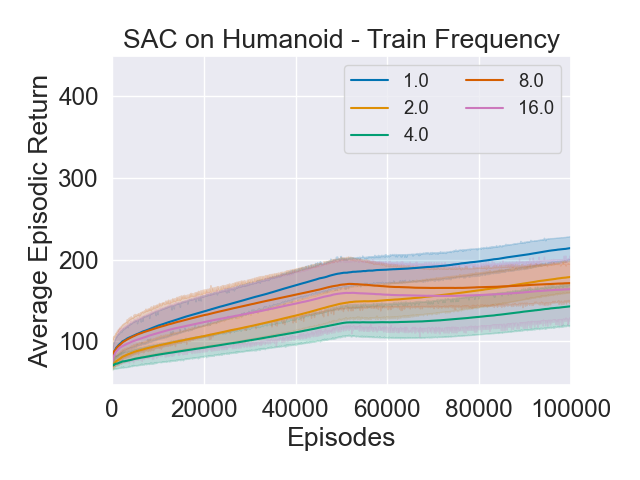}
    \includegraphics[width=0.2\textwidth]{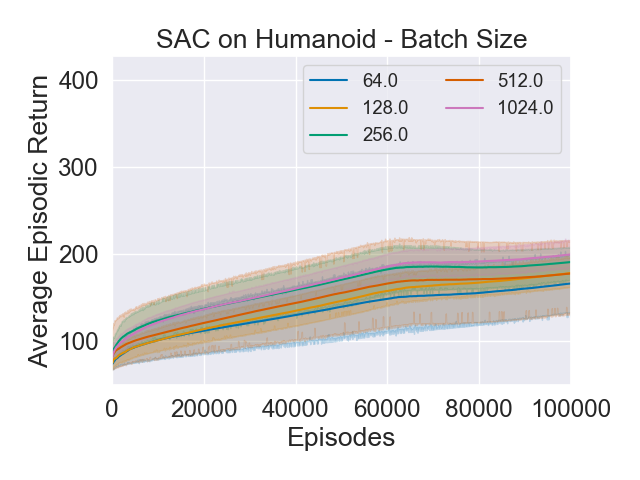}
    \includegraphics[width=0.2\textwidth]{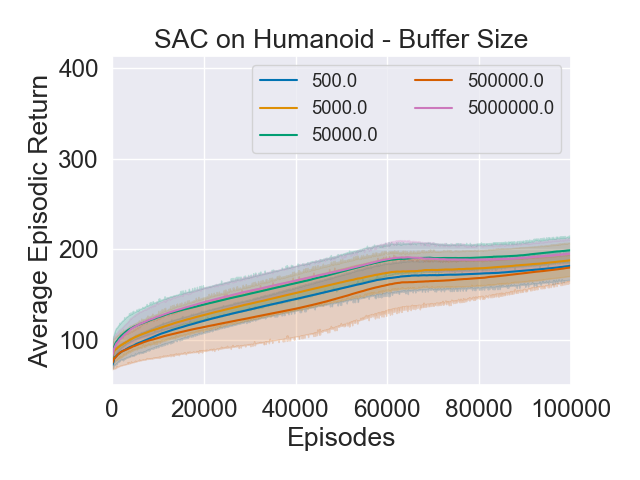}
    \includegraphics[width=0.2\textwidth]{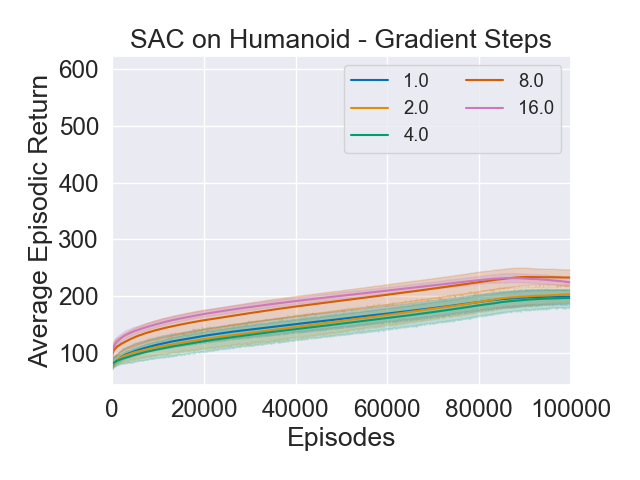}
    \includegraphics[width=0.2\textwidth]{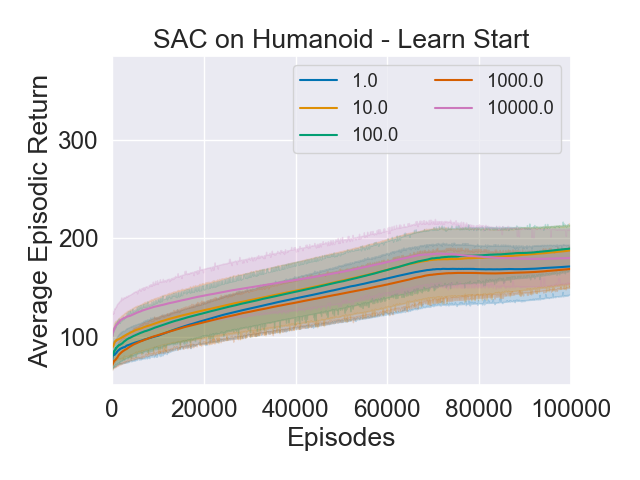}
    \caption{Hyperparameter Sweeps for SAC on Halfcheetah and Humanoid.}
    \label{app-fig:sac_sweeps2}
\end{figure}

The full set of PPO sweeps can be found in Figures~\ref{app-fig:ppo_sweeps}, \ref{app-fig:ppo_sweeps2}, \ref{app-fig:ppo_sweeps3} and \ref{app-fig:ppo_sweeps4}, the SAC sweeps in Figures~\ref{app-fig:sac_sweeps} and \ref{app-fig:sac_sweeps2} and the DQN sweeps in Figures~\ref{app-fig:dqn_sweeps} and \ref{app-fig:ppo_sweeps2}.

\subsection{PPO Sweeps}

\clearpage
\subsection{DQN Sweeps}
\begin{figure}
    \centering
    \includegraphics[width=0.2\textwidth]{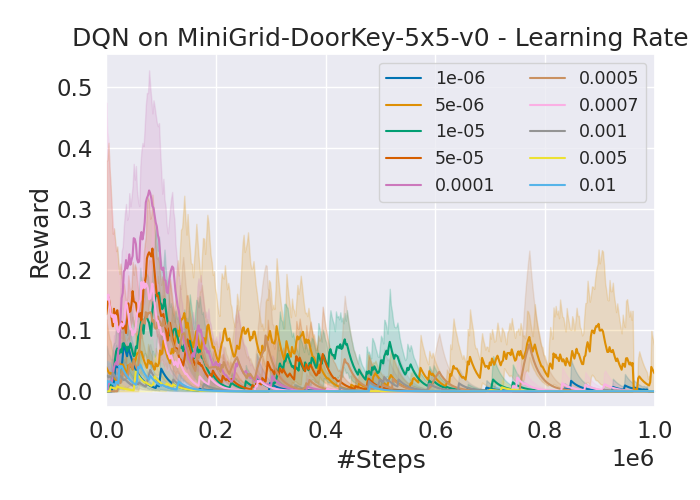}
    \includegraphics[width=0.2\textwidth]{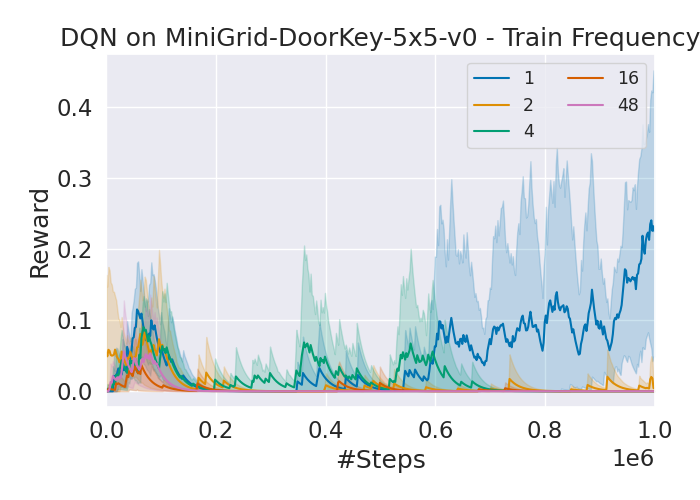}
    \includegraphics[width=0.2\textwidth]{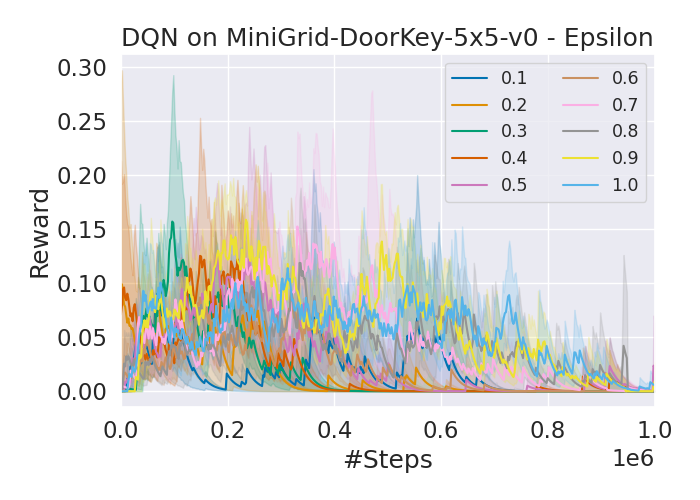}
    \includegraphics[width=0.2\textwidth]{figs/appendix_sweeps/sweep_MiniGrid-DoorKey-5x5-v0_DQN_batch_size.png}
    \includegraphics[width=0.2\textwidth]{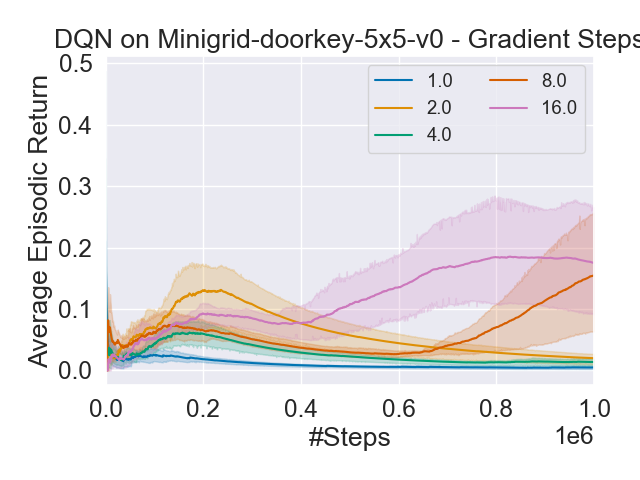}
    \includegraphics[width=0.2\textwidth]{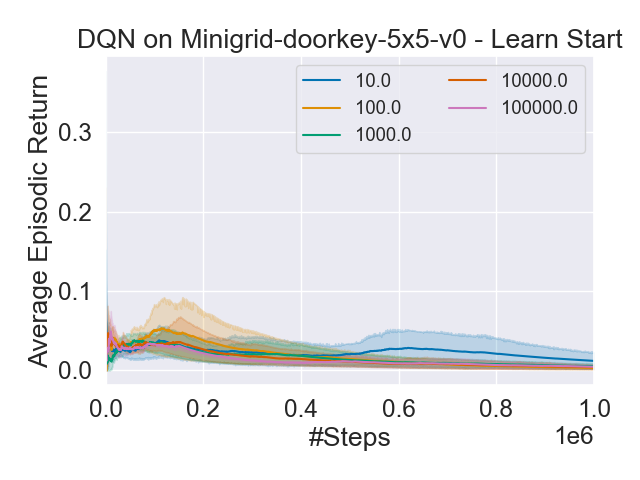}
    \includegraphics[width=0.2\textwidth]{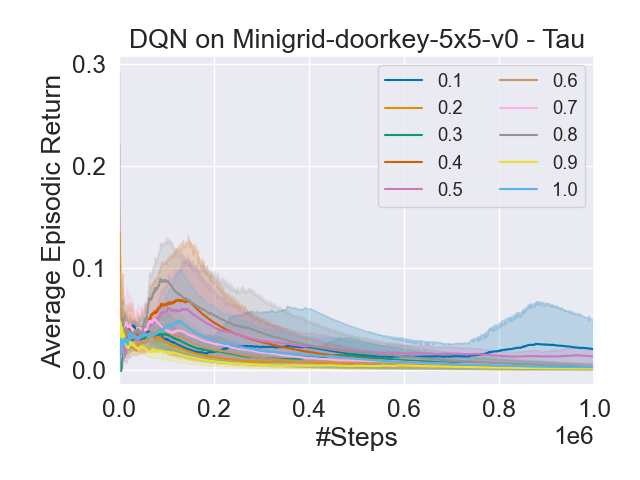}
    \includegraphics[width=0.2\textwidth]{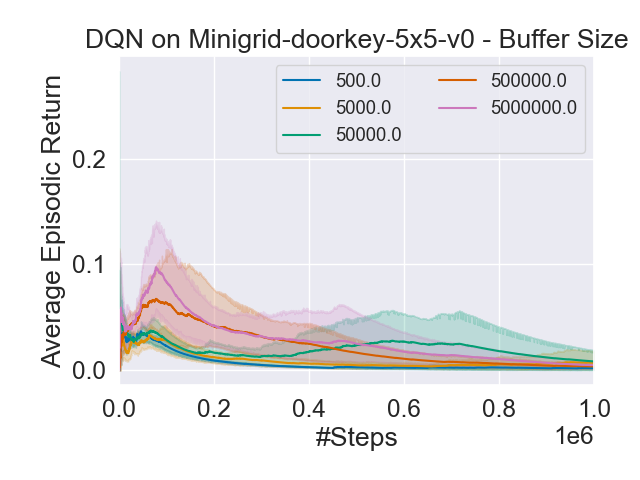}\includegraphics[width=0.2\textwidth]{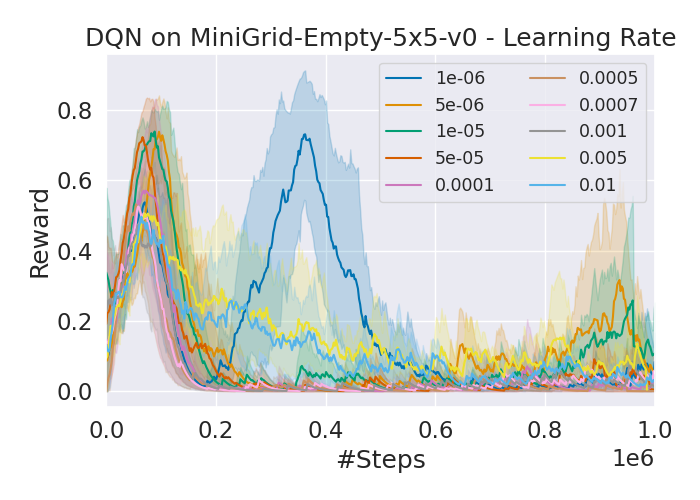}
    \includegraphics[width=0.2\textwidth]{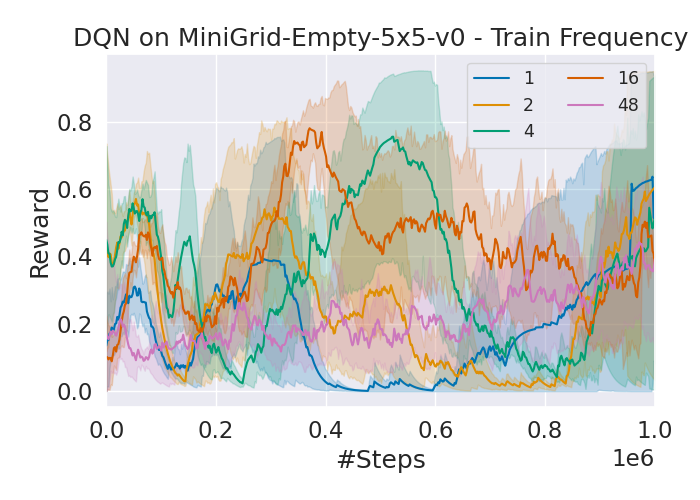}
    \includegraphics[width=0.2\textwidth]{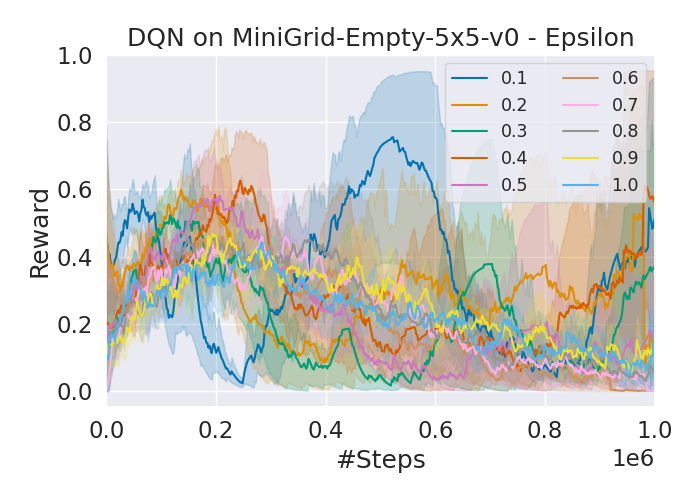}
    \includegraphics[width=0.2\textwidth]{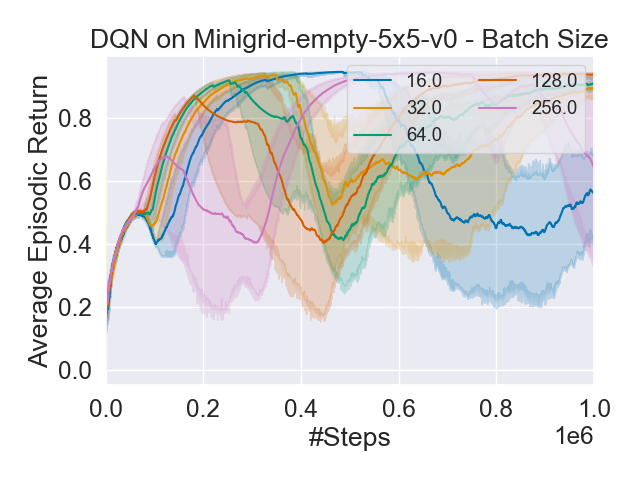}
    \includegraphics[width=0.2\textwidth]{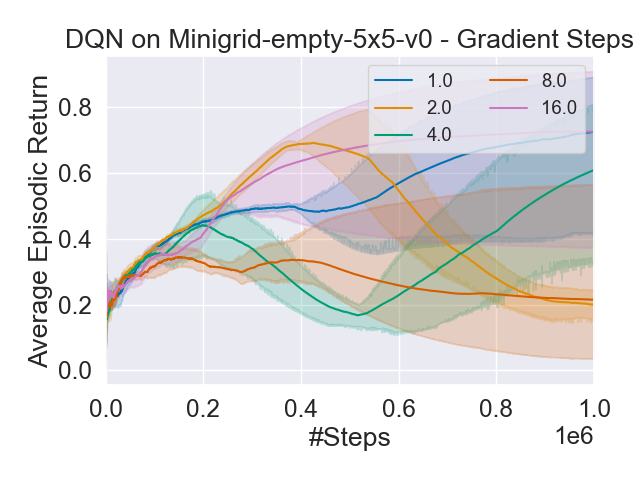}
    \includegraphics[width=0.2\textwidth]{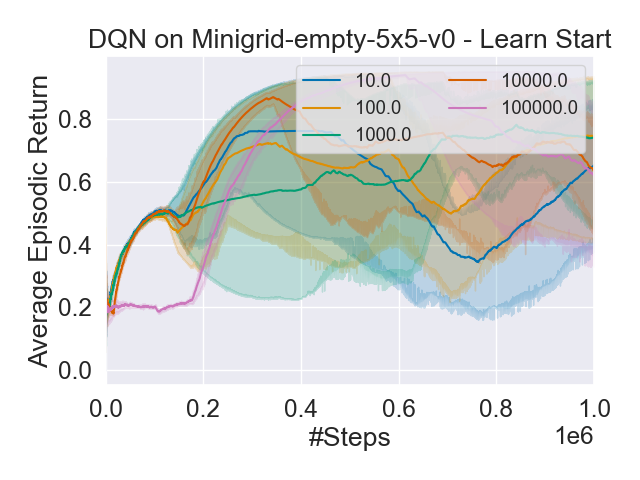}
    \includegraphics[width=0.2\textwidth]{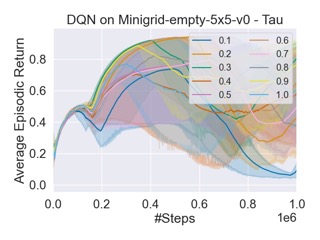}
    \includegraphics[width=0.2\textwidth]{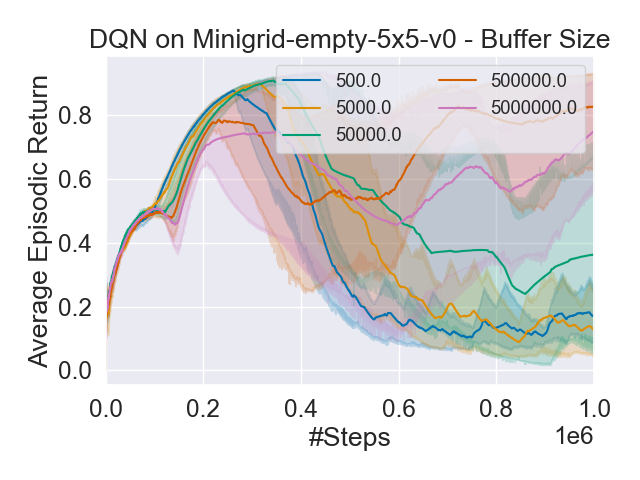}
    \caption{Hyperparameter Sweeps for DQN on MiniGrid.}
    \label{app-fig:dqn_sweeps2}
\end{figure} 

\clearpage
\subsection{SAC Sweeps}

\clearpage
\section{Full Performance Pointplots}
\label{app:more_pointplots}
\begin{figure}[h]
    \centering
    \includegraphics[width=0.2\textwidth]{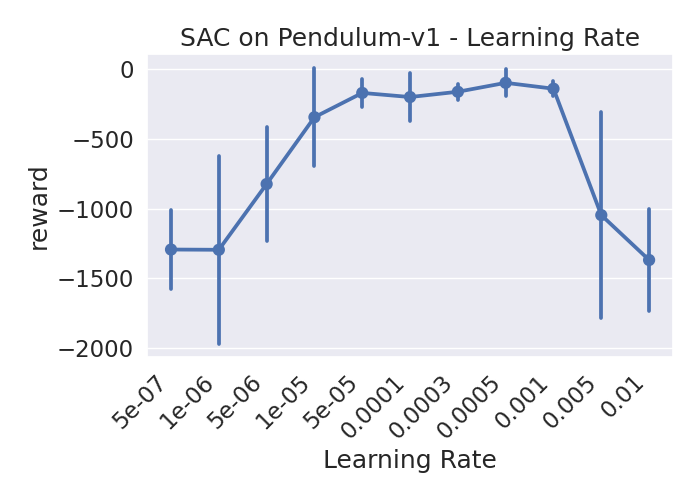}
    \includegraphics[width=0.2\textwidth]{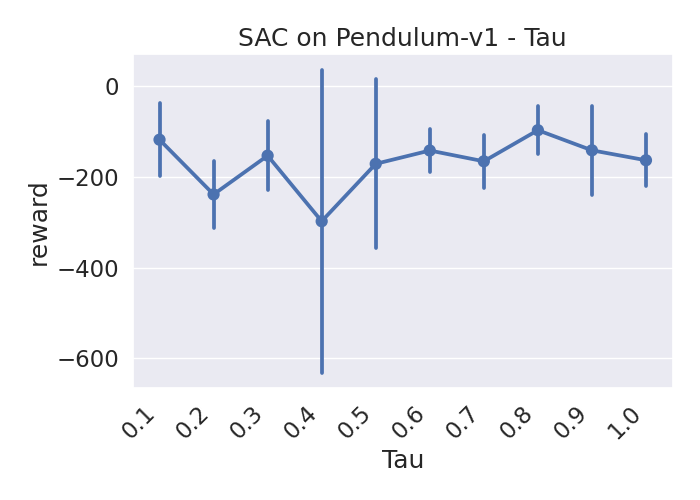}
    \includegraphics[width=0.2\textwidth]{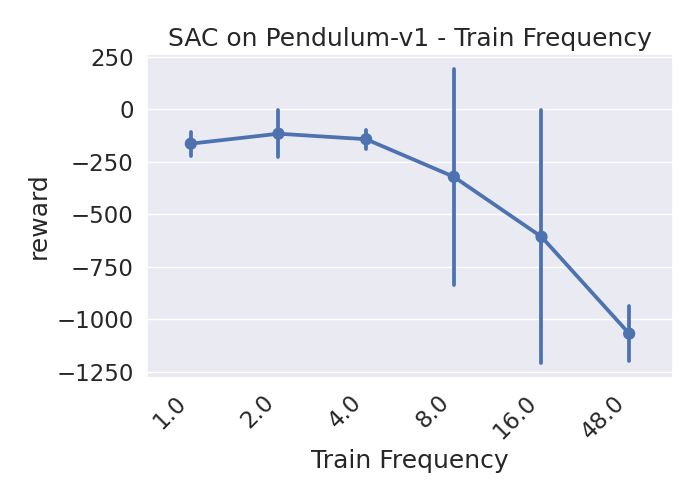}
    \includegraphics[width=0.2\textwidth]{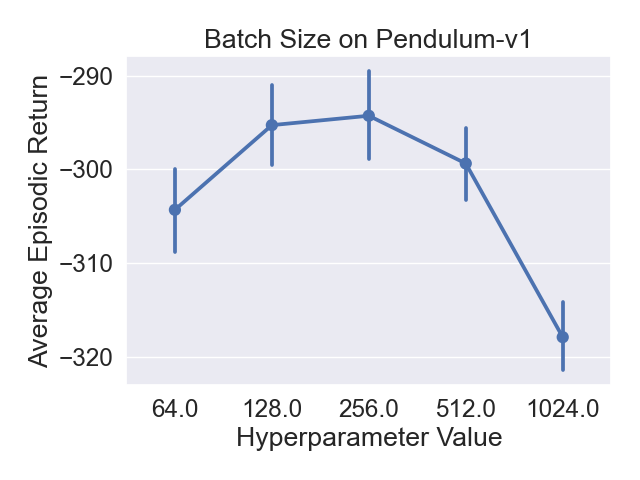}
    \includegraphics[width=0.2\textwidth]{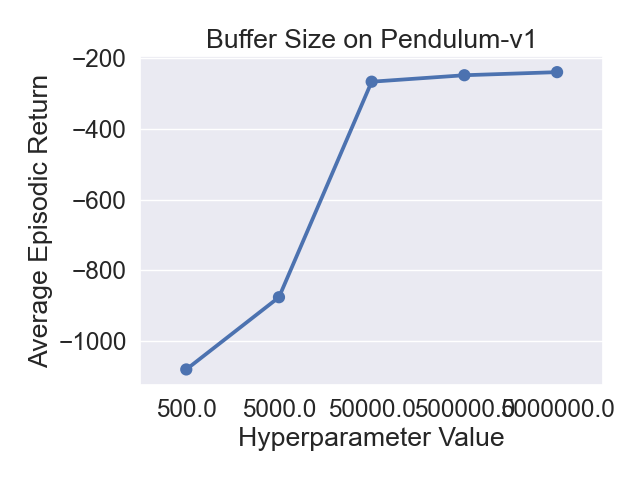}
    \includegraphics[width=0.2\textwidth]{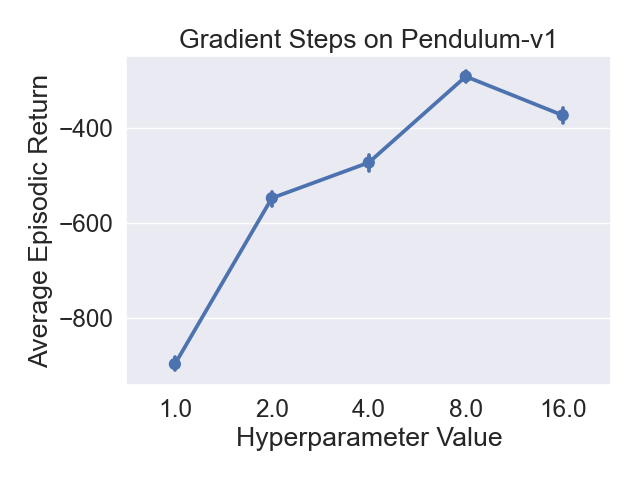}
    \includegraphics[width=0.2\textwidth]{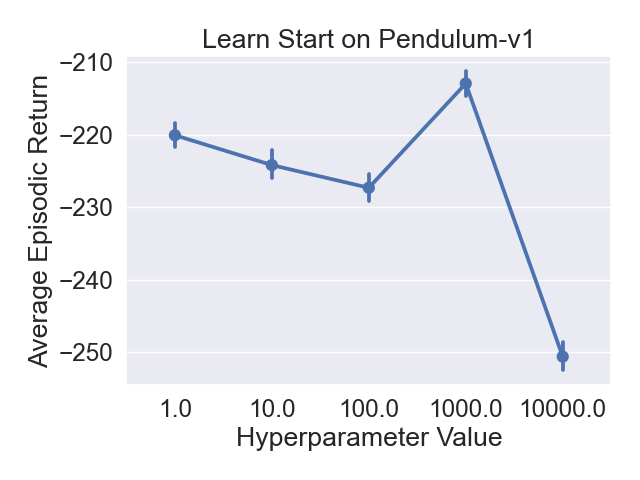}
    \includegraphics[width=0.2\textwidth]{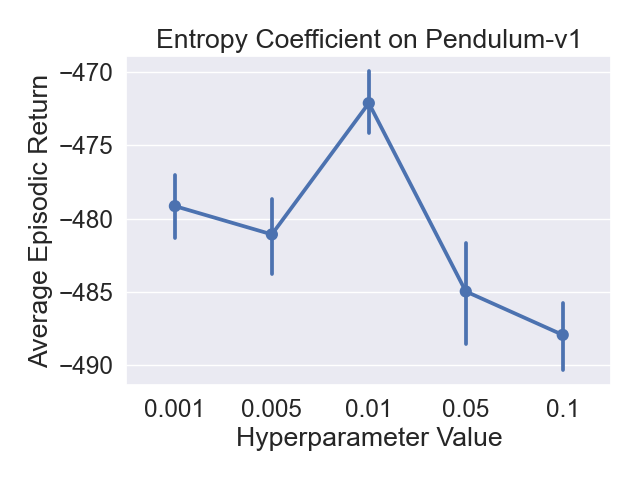}
    \includegraphics[width=0.2\textwidth]{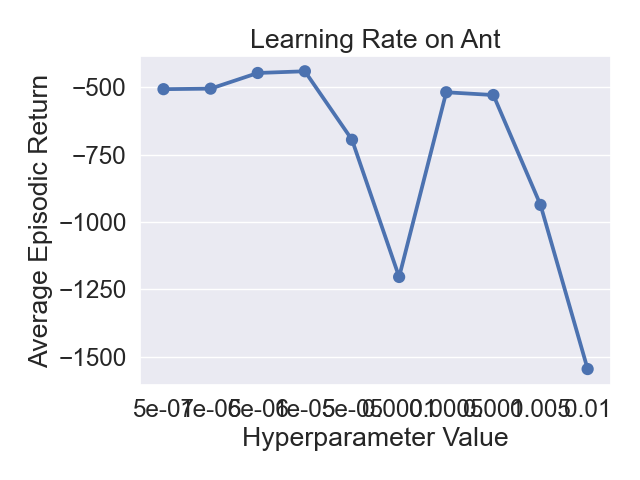}
    \includegraphics[width=0.2\textwidth]{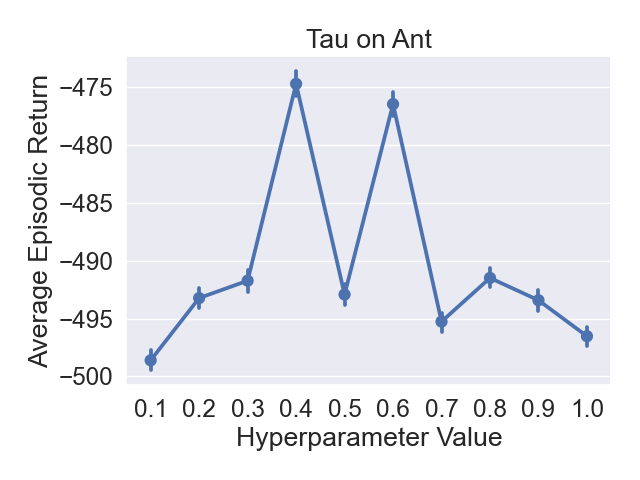}
    \includegraphics[width=0.2\textwidth]{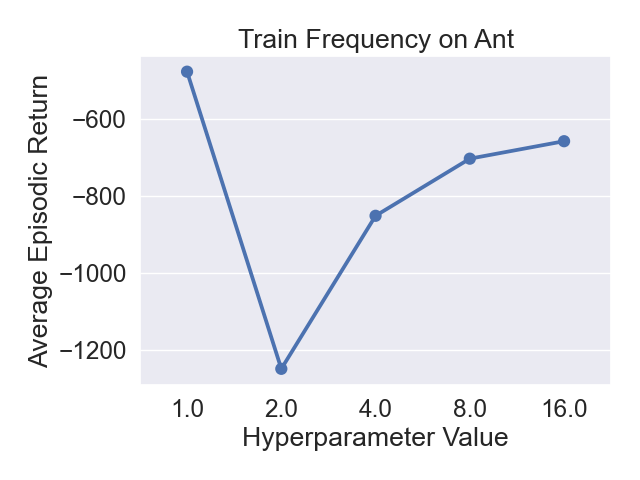}
    \includegraphics[width=0.2\textwidth]{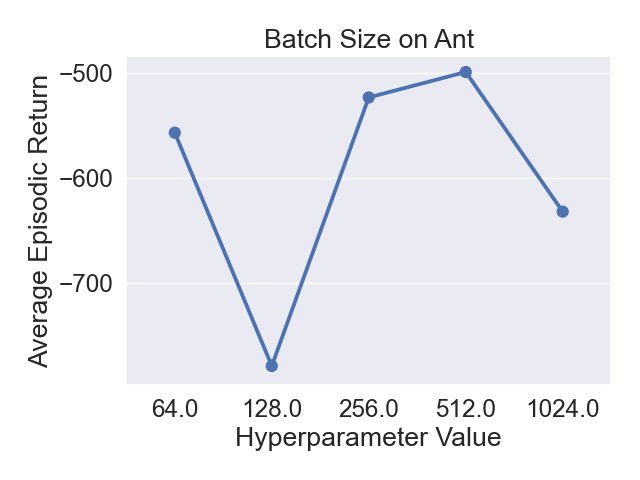}
    \includegraphics[width=0.2\textwidth]{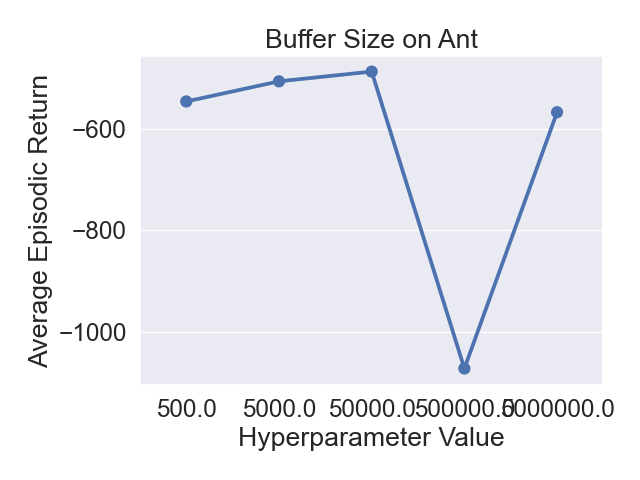}
    \includegraphics[width=0.2\textwidth]{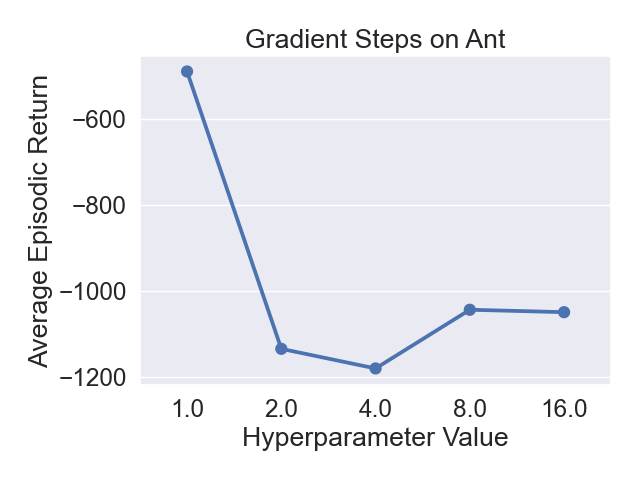}
    \includegraphics[width=0.2\textwidth]{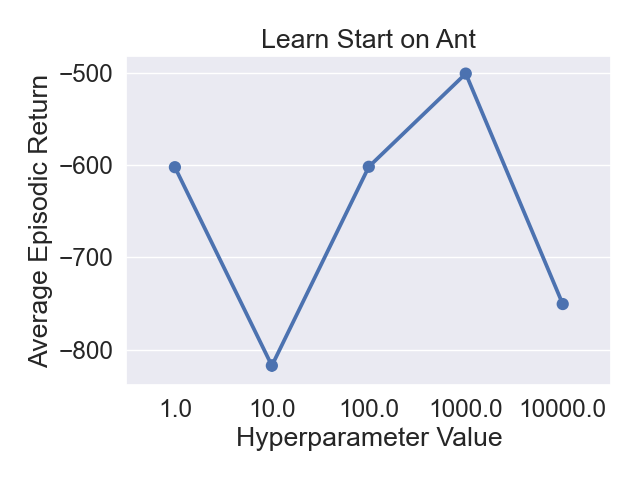}
    \caption{Final returns across 5 seeds for different hp variations of SAC on Pendulum and Ant.}
    \label{app-fig:sac_boxplots}
\end{figure}

\begin{figure}[h]
    \centering
    \includegraphics[width=0.2\textwidth]{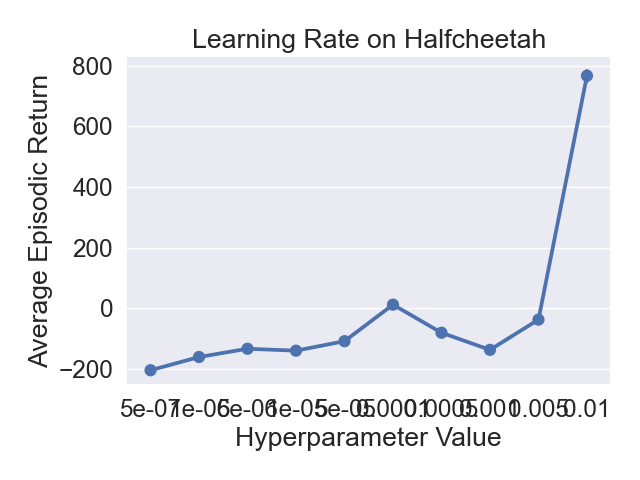}
    \includegraphics[width=0.2\textwidth]{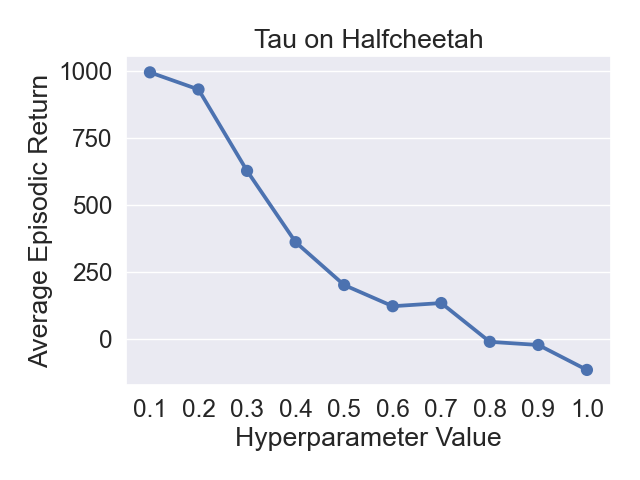}
    \includegraphics[width=0.2\textwidth]{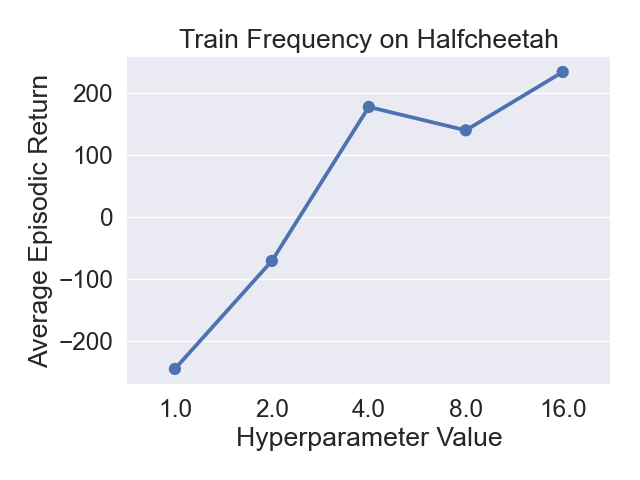}
    \includegraphics[width=0.2\textwidth]{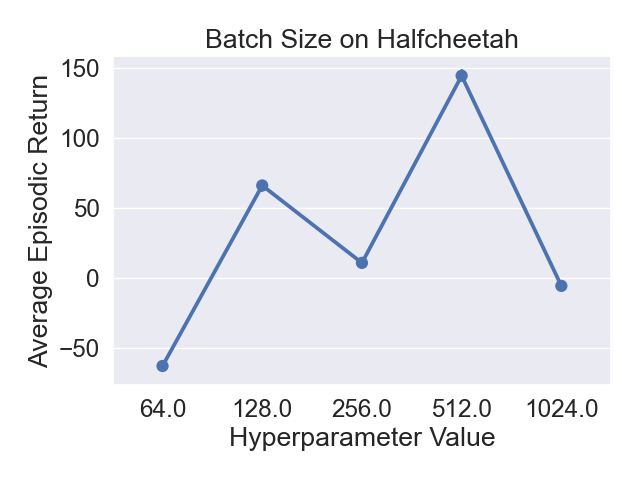}
    \includegraphics[width=0.2\textwidth]{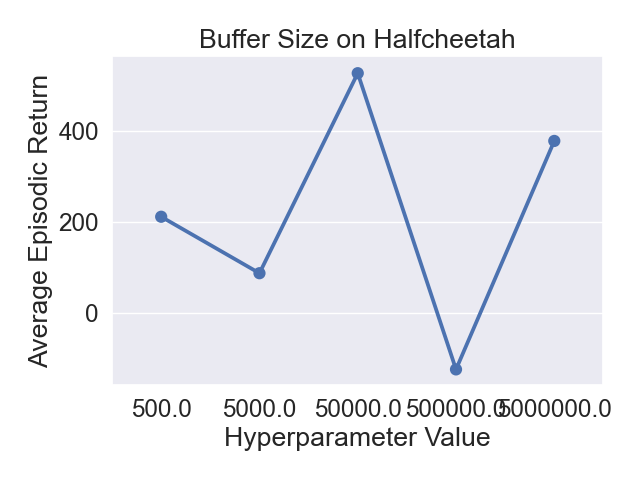}
    \includegraphics[width=0.2\textwidth]{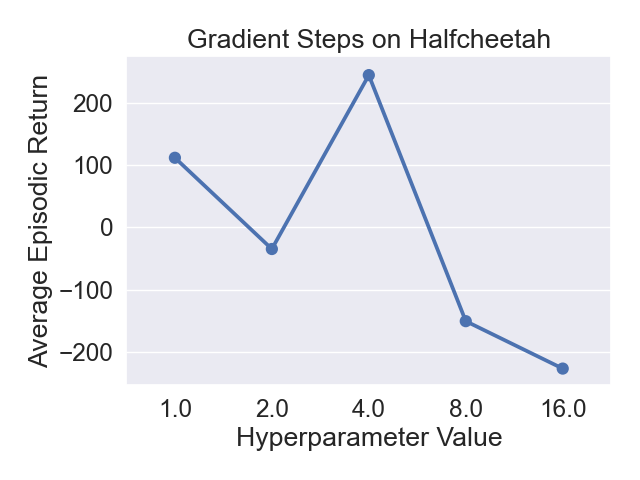}
    \includegraphics[width=0.2\textwidth]{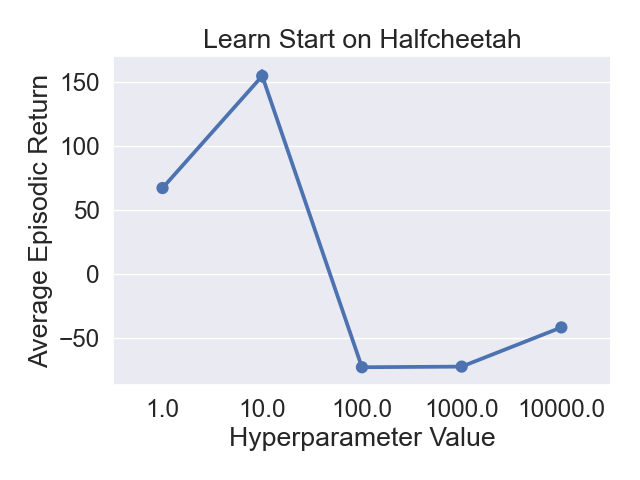}
    \includegraphics[width=0.2\textwidth]{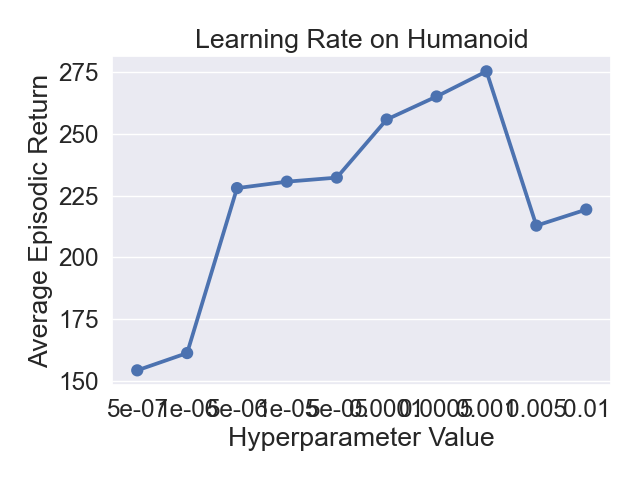}
    \includegraphics[width=0.2\textwidth]{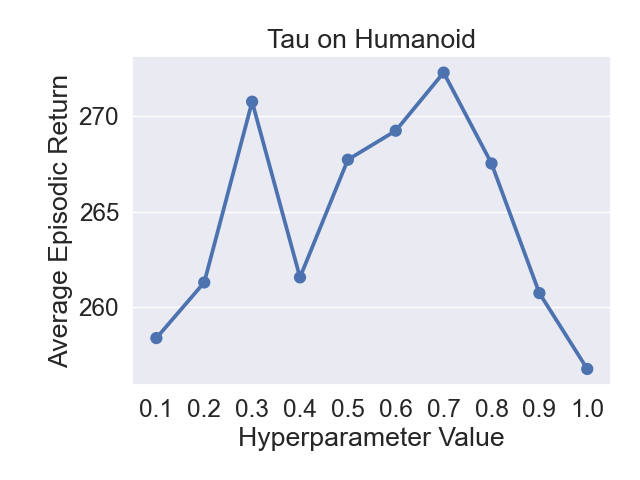}
    \includegraphics[width=0.2\textwidth]{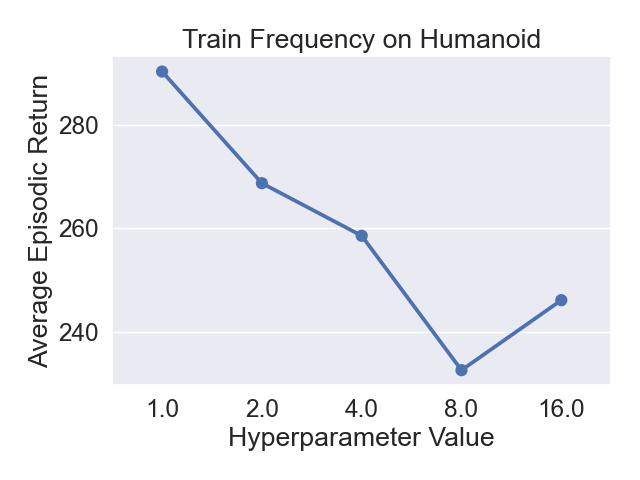}
    \includegraphics[width=0.2\textwidth]{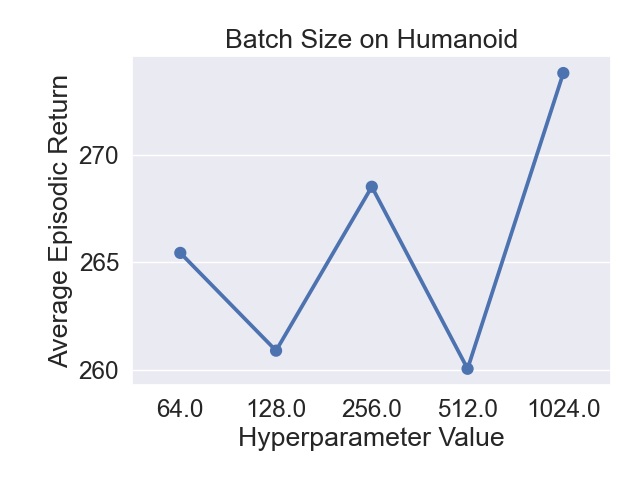}
    \includegraphics[width=0.2\textwidth]{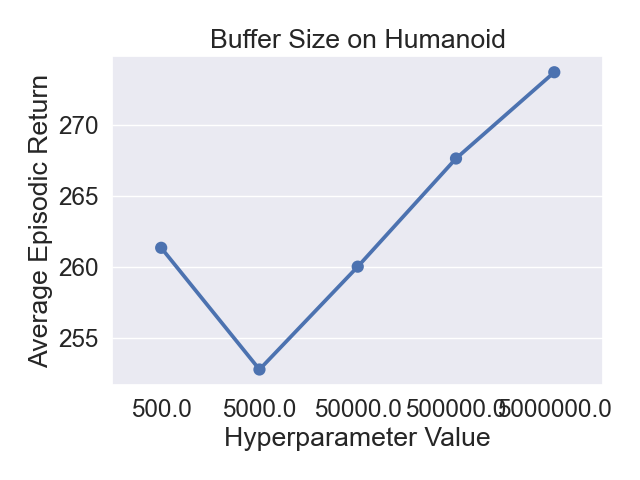}
    \includegraphics[width=0.2\textwidth]{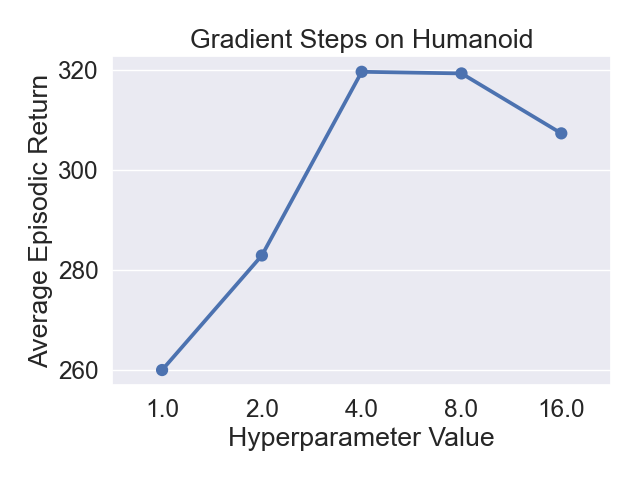}
    \includegraphics[width=0.2\textwidth]{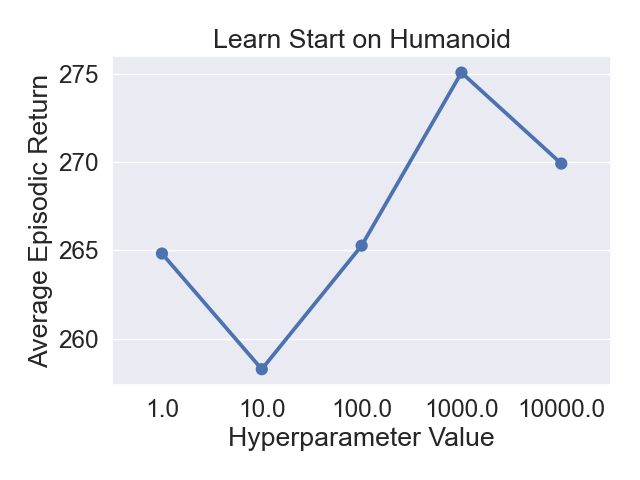}
    \caption{Final returns across 5 seeds for different hp variations of SAC of Halfcheetah and Humanoid.}
    \label{app-fig:sac_boxplots2}
\end{figure}

\begin{figure}[h]
    \centering
    \includegraphics[width=0.2\textwidth]{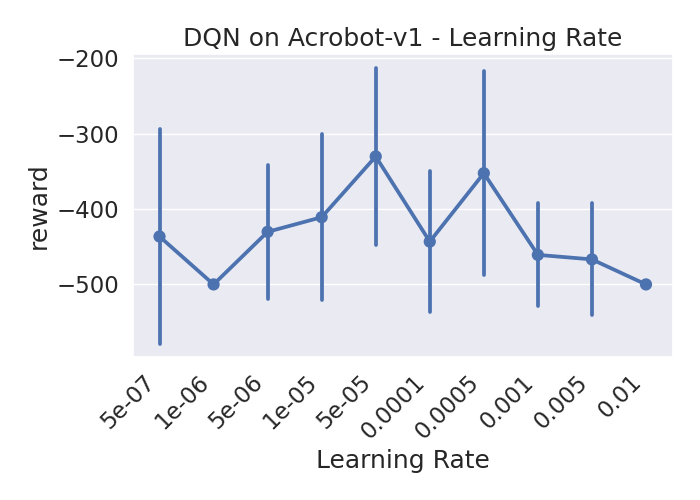}
    \includegraphics[width=0.2\textwidth]{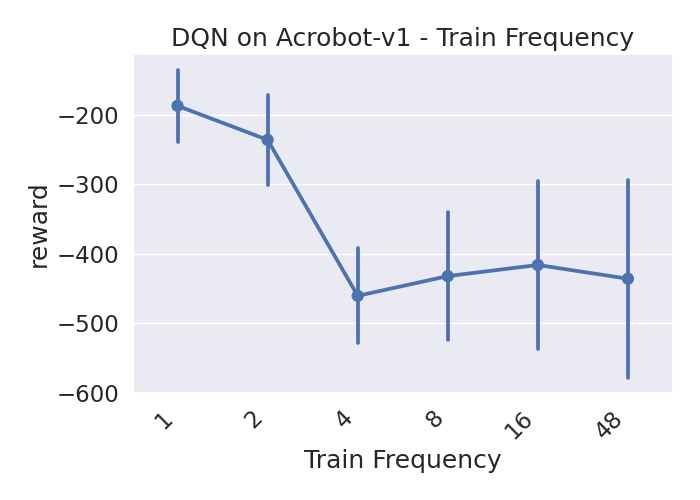}
    \includegraphics[width=0.2\textwidth]{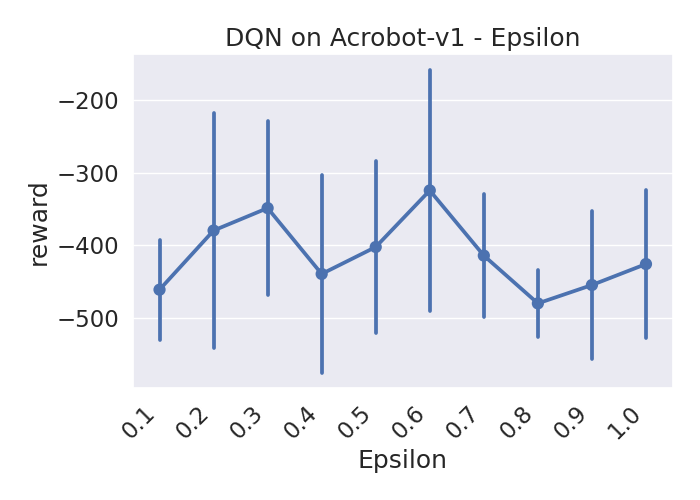}
    \includegraphics[width=0.2\textwidth]{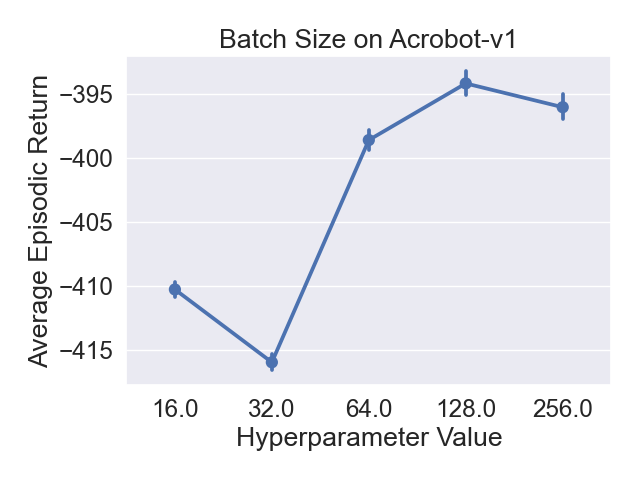}
    \includegraphics[width=0.2\textwidth]{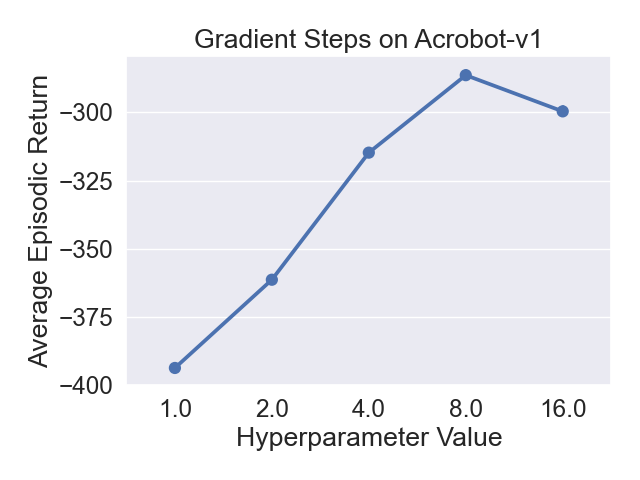}
    \includegraphics[width=0.2\textwidth]{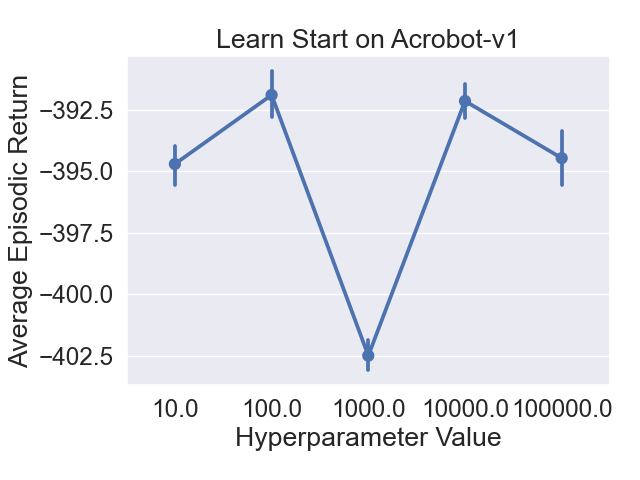}
    \includegraphics[width=0.2\textwidth]{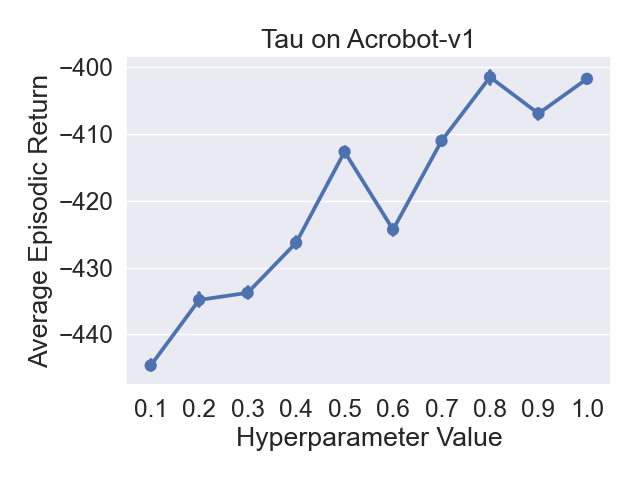}
    \includegraphics[width=0.2\textwidth]{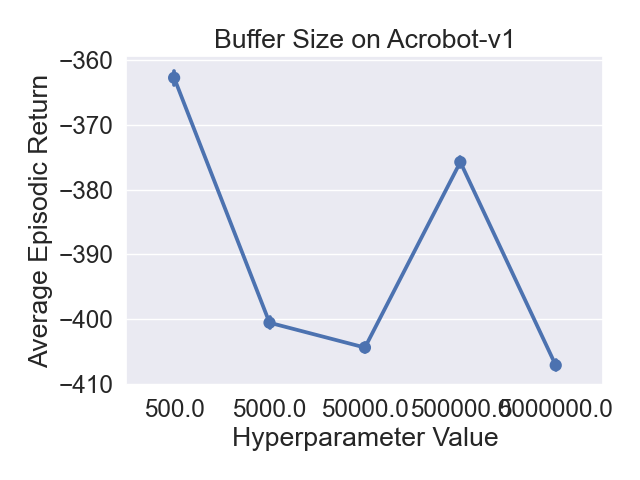}
    \caption{Final returns across 5 seeds for different hp variations of DQN on Acrobot.}
    \label{app-fig:dqn_boxplots}
\end{figure}

\begin{figure}
    \centering
    \includegraphics[width=0.2\textwidth]{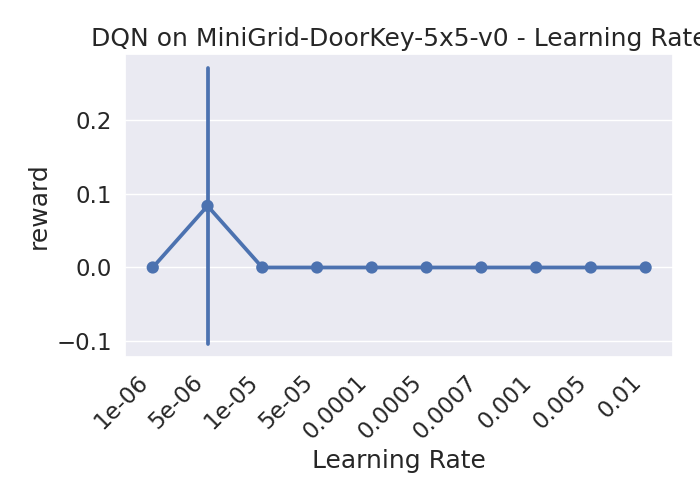}
    \includegraphics[width=0.2\textwidth]{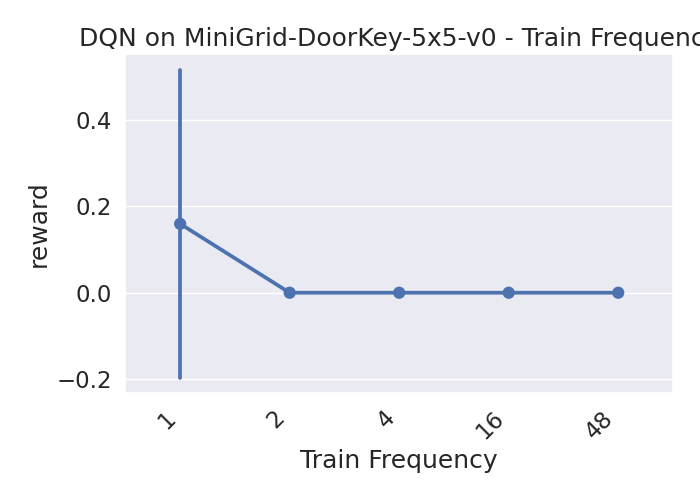}
    \includegraphics[width=0.2\textwidth]{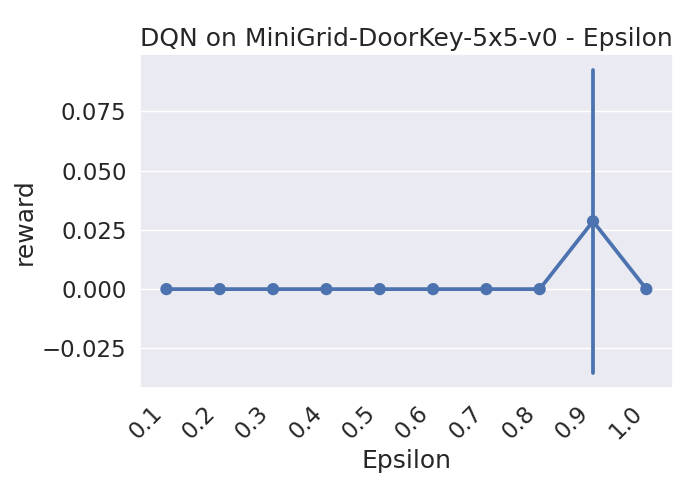}
    \includegraphics[width=0.2\textwidth]{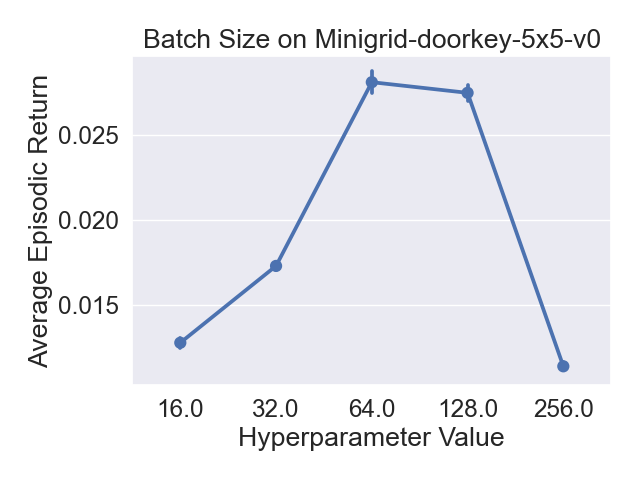}
    \includegraphics[width=0.2\textwidth]{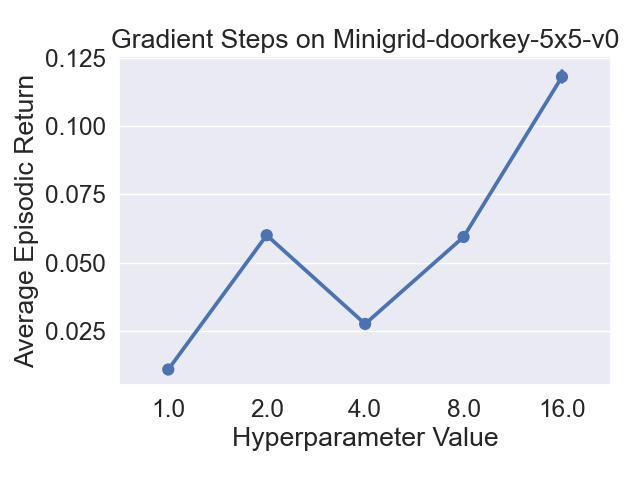}
    \includegraphics[width=0.2\textwidth]{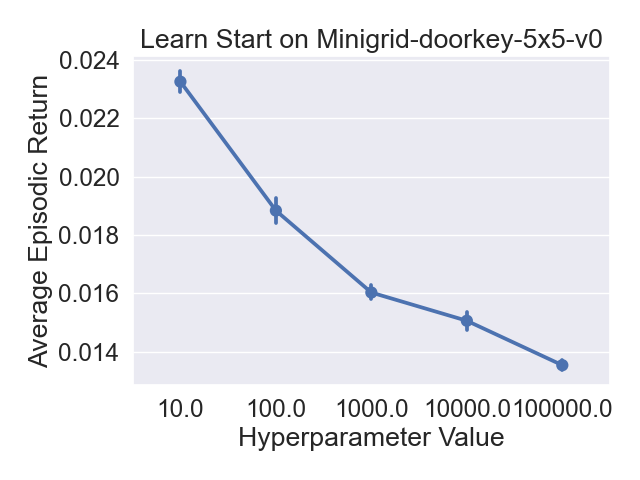}
    \includegraphics[width=0.2\textwidth]{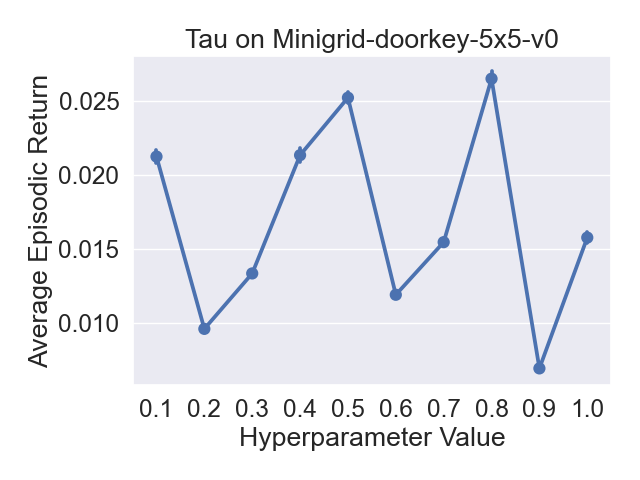}
    \includegraphics[width=0.2\textwidth]{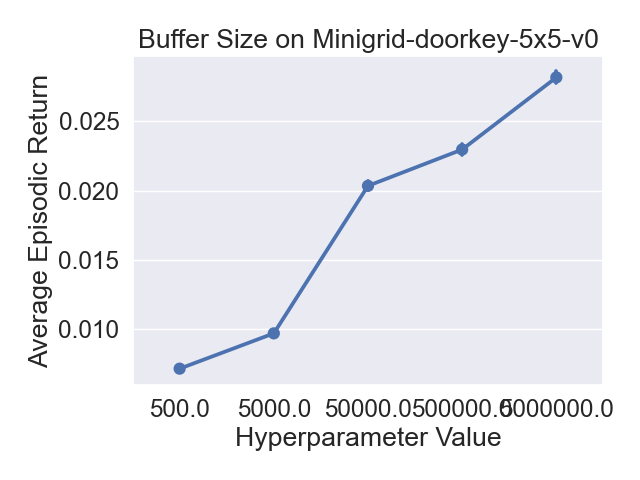}
    \includegraphics[width=0.2\textwidth]{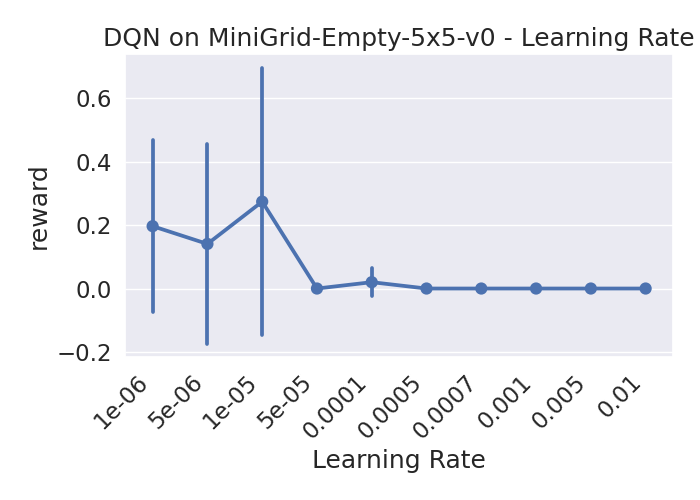}
    \includegraphics[width=0.2\textwidth]{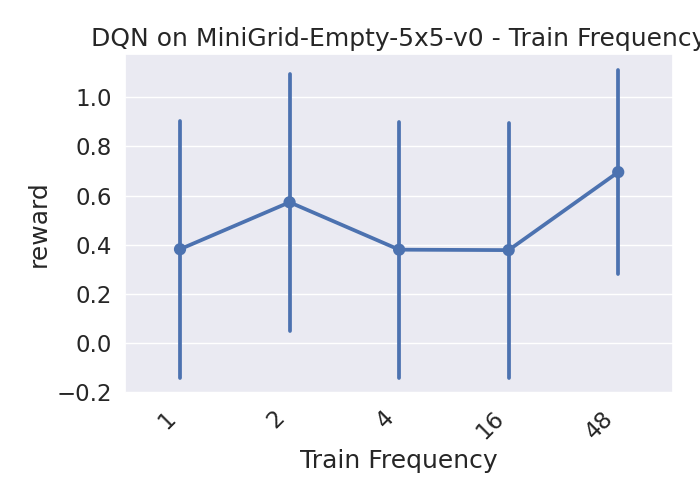}
    \includegraphics[width=0.2\textwidth]{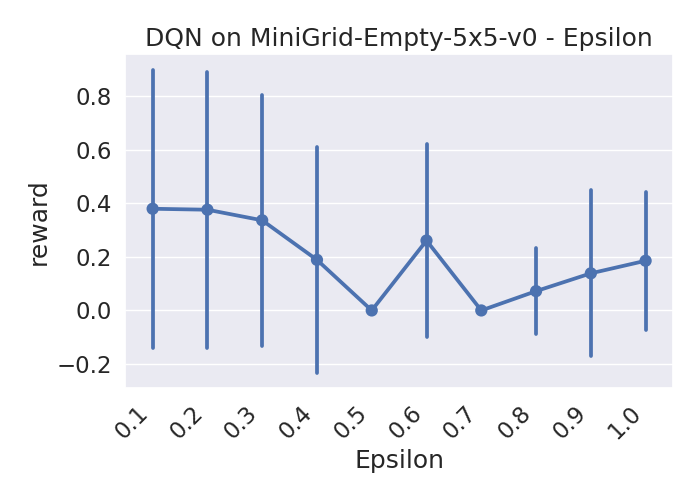}
    \includegraphics[width=0.2\textwidth]{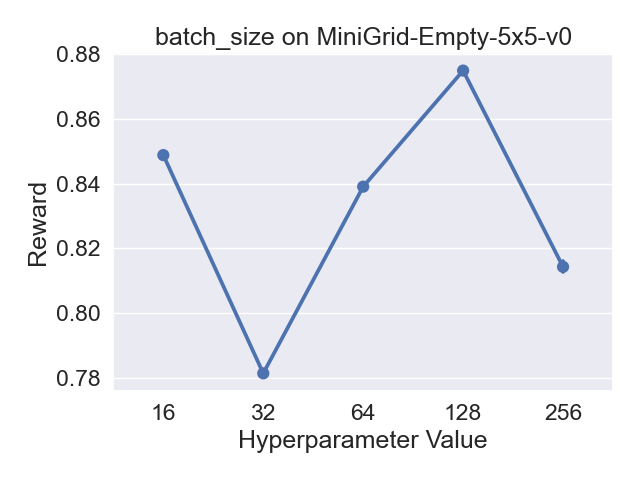}
    \includegraphics[width=0.2\textwidth]{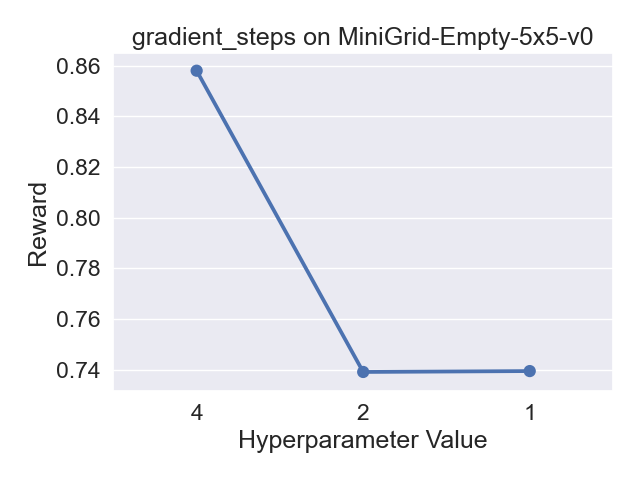}
    \includegraphics[width=0.2\textwidth]{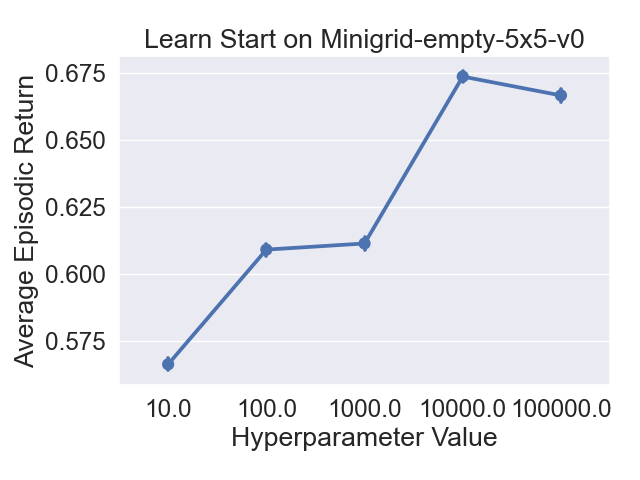}
    \includegraphics[width=0.2\textwidth]{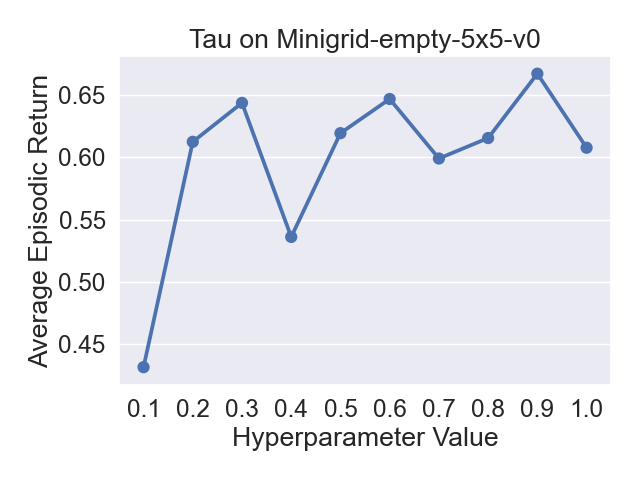}
    \includegraphics[width=0.2\textwidth]{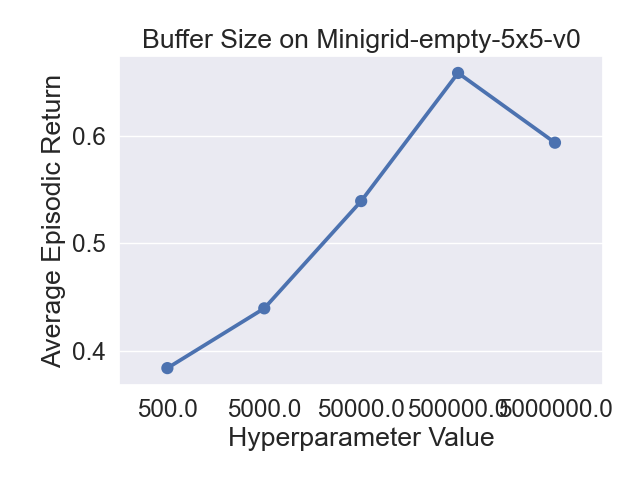}
    \caption{Final returns across 5 seeds for different hp variations of DQN on MiniGrid.}
    \label{app-fig:dqn_boxplots2}
\end{figure}

\begin{figure}[h]
    \centering
    \includegraphics[width=0.2\textwidth]{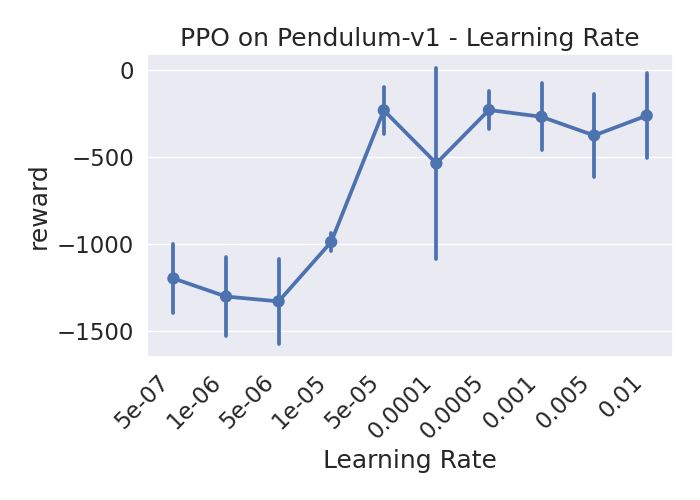}
    \includegraphics[width=0.2\textwidth]{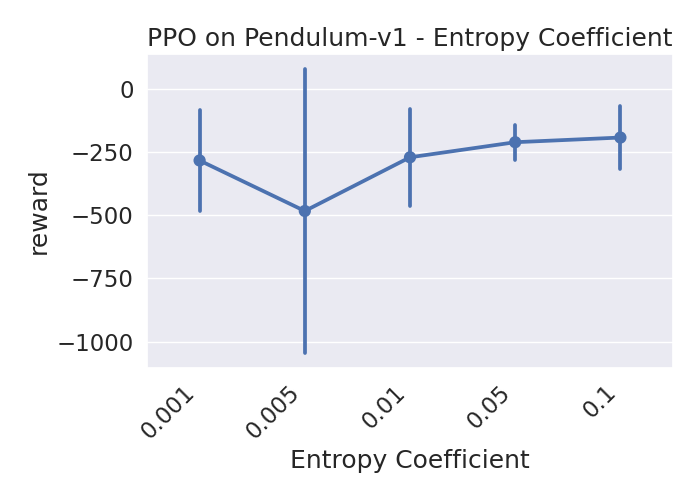}
    \includegraphics[width=0.2\textwidth]{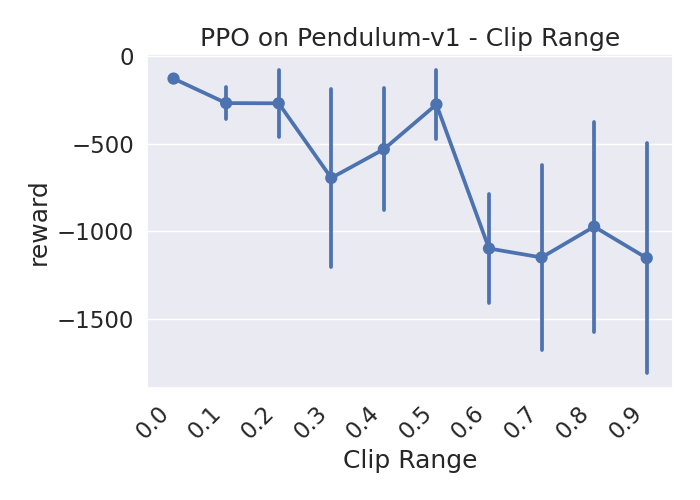}
    \includegraphics[width=0.2\textwidth]{"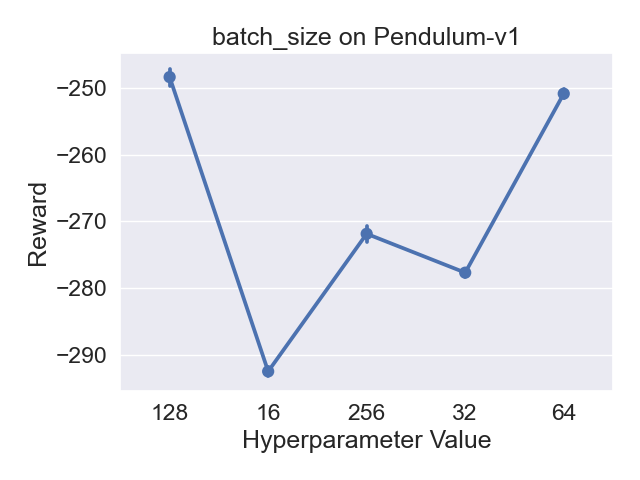"}
    \includegraphics[width=0.2\textwidth]{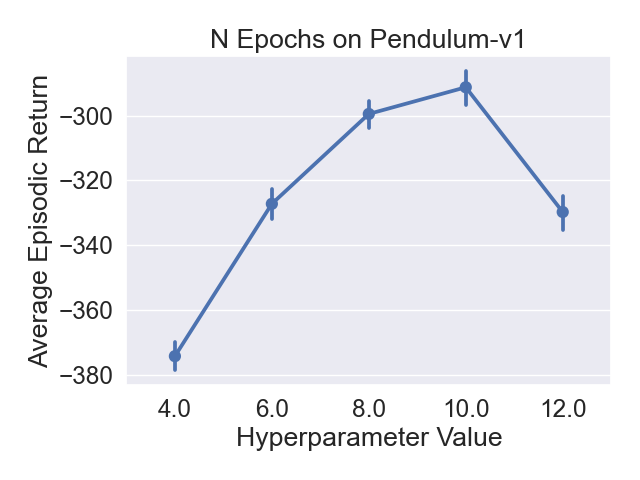}
    \includegraphics[width=0.2\textwidth]{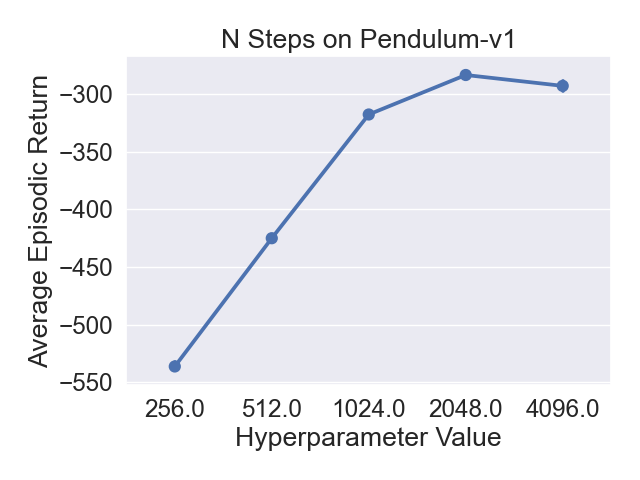}
    \includegraphics[width=0.2\textwidth]{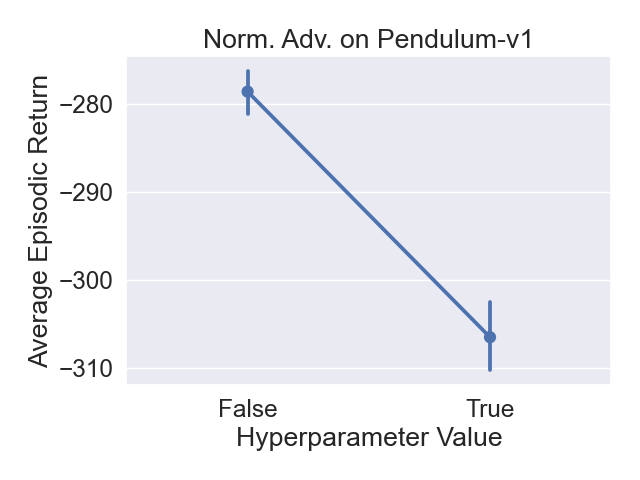}
    \includegraphics[width=0.2\textwidth]{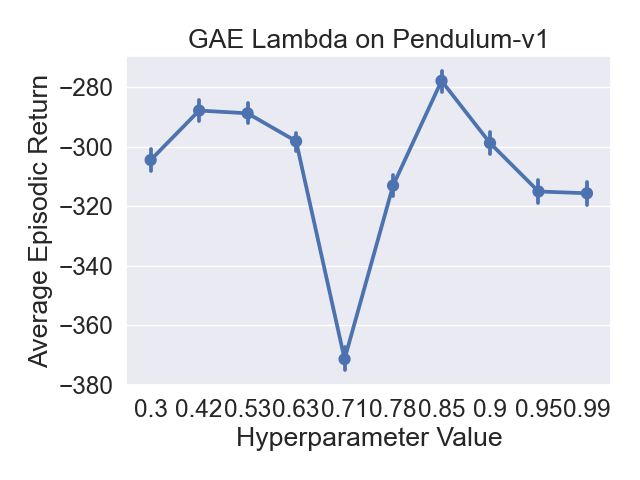}
    \includegraphics[width=0.2\textwidth]{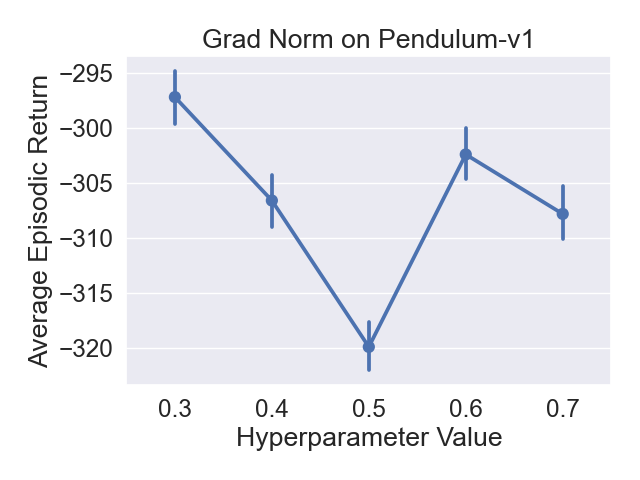}
    \includegraphics[width=0.2\textwidth]{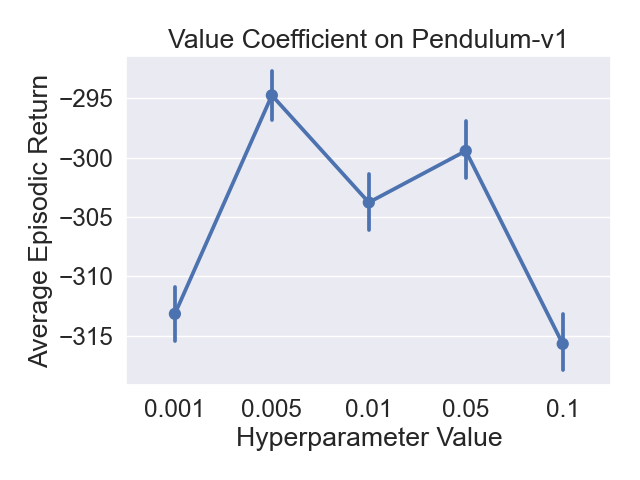}
    \includegraphics[width=0.2\textwidth]{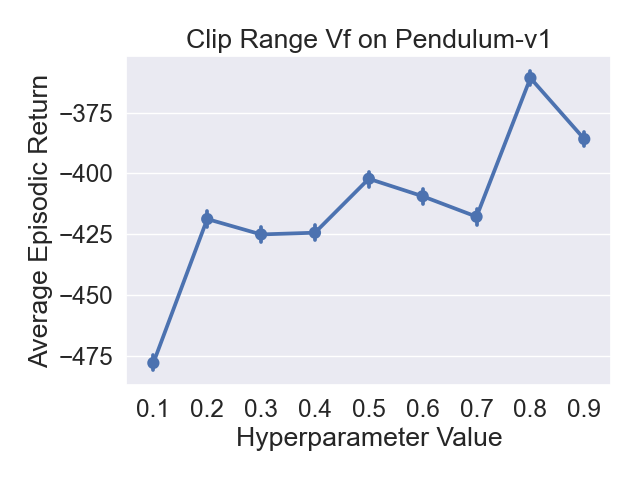}
    \includegraphics[width=0.2\textwidth]{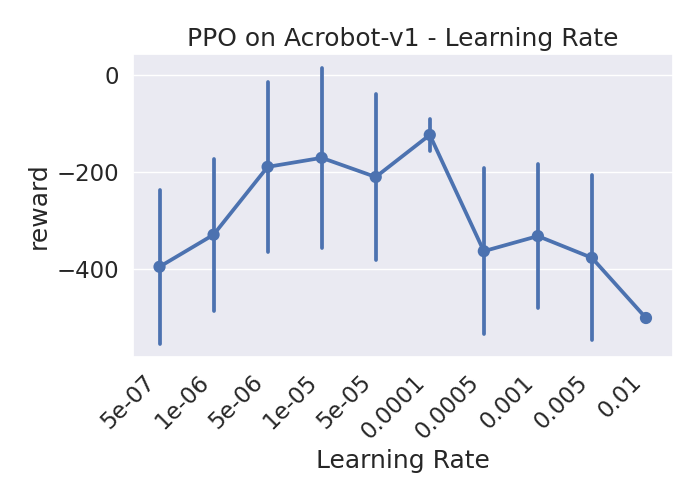}
    \includegraphics[width=0.2\textwidth]{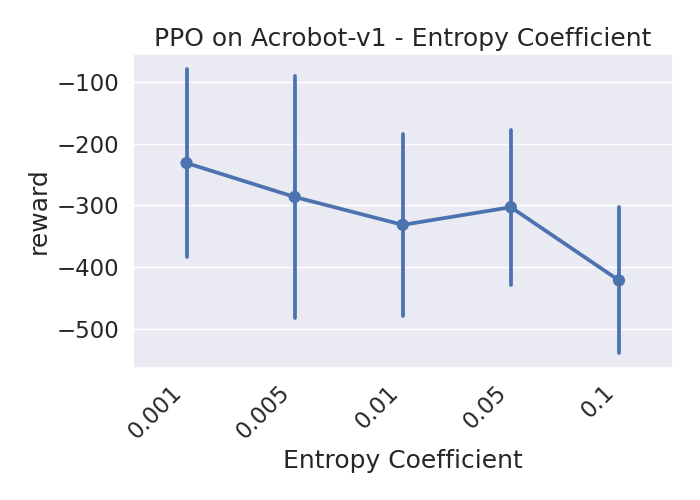}
    \includegraphics[width=0.2\textwidth]{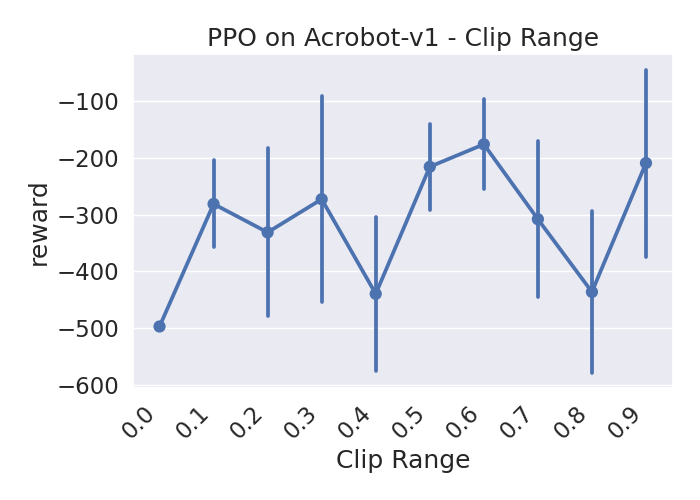}
    \includegraphics[width=0.2\textwidth]{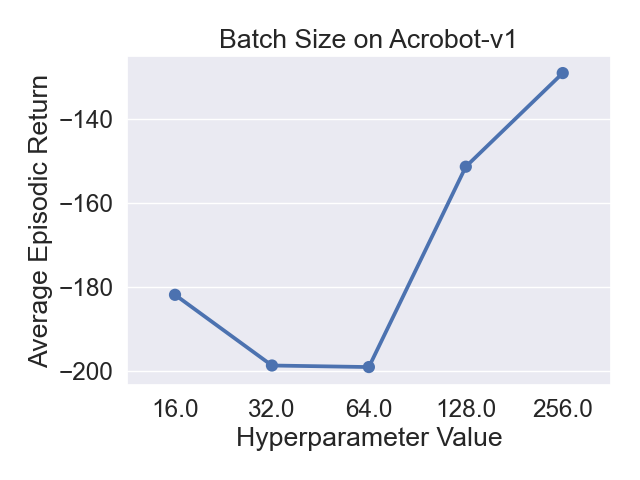}
    \includegraphics[width=0.2\textwidth]{"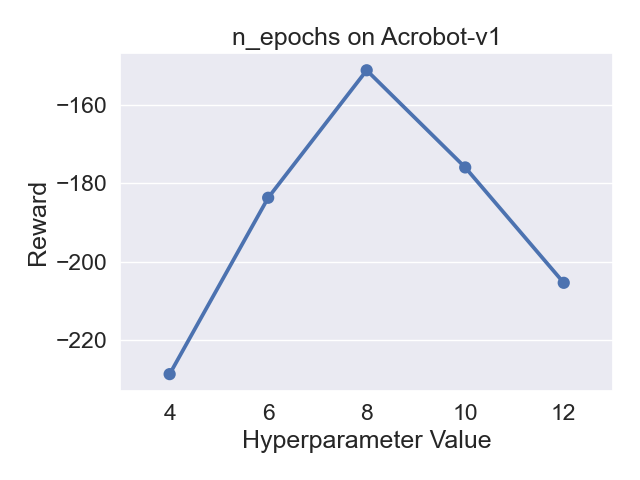"}
    \includegraphics[width=0.2\textwidth]{"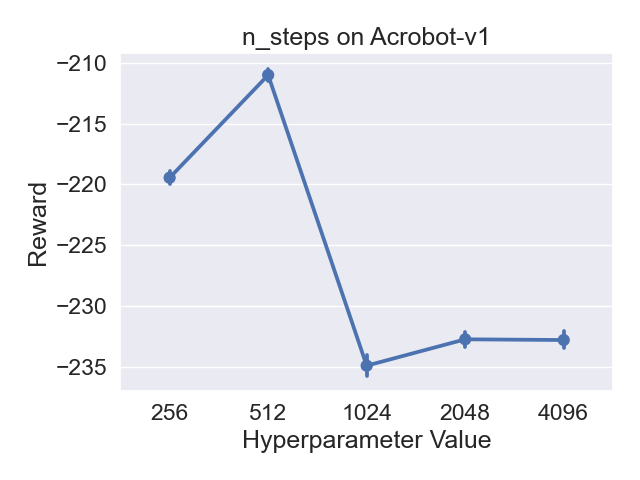"}
    \includegraphics[width=0.2\textwidth]{"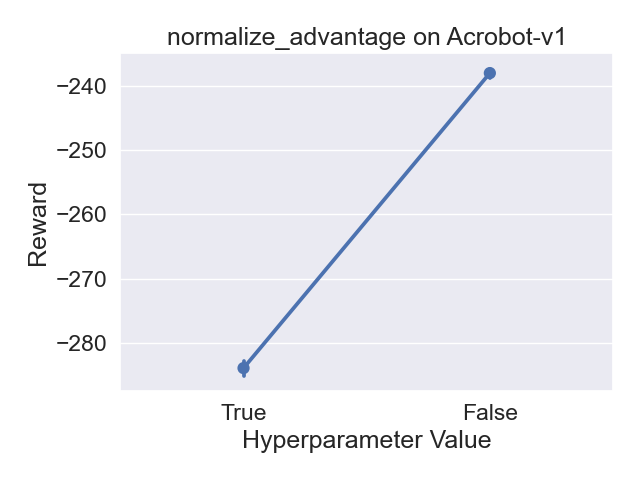"}
    \includegraphics[width=0.2\textwidth]{"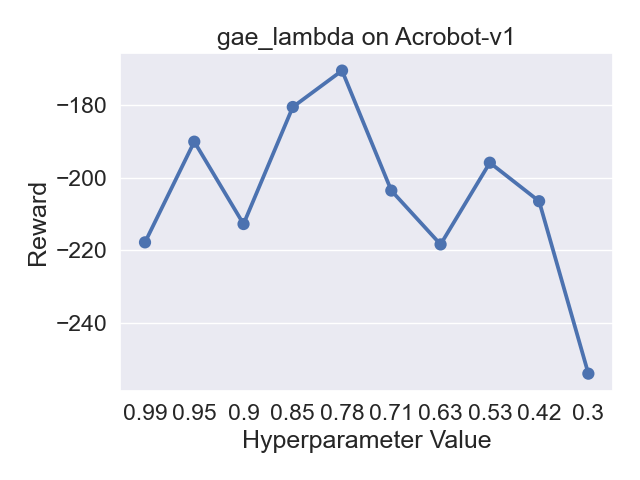"}
    \includegraphics[width=0.2\textwidth]{"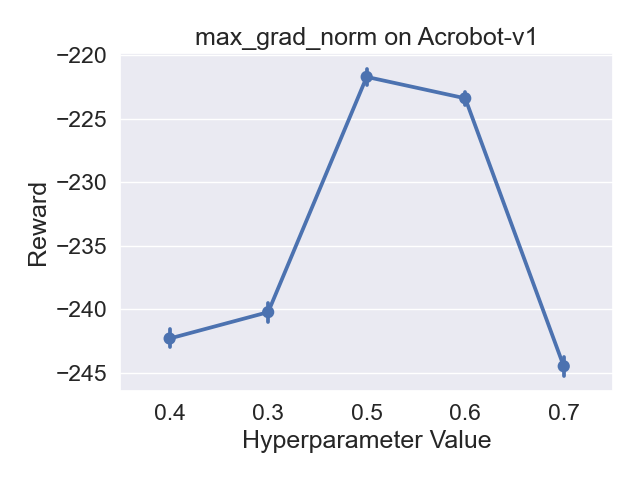"}
    \includegraphics[width=0.2\textwidth]{"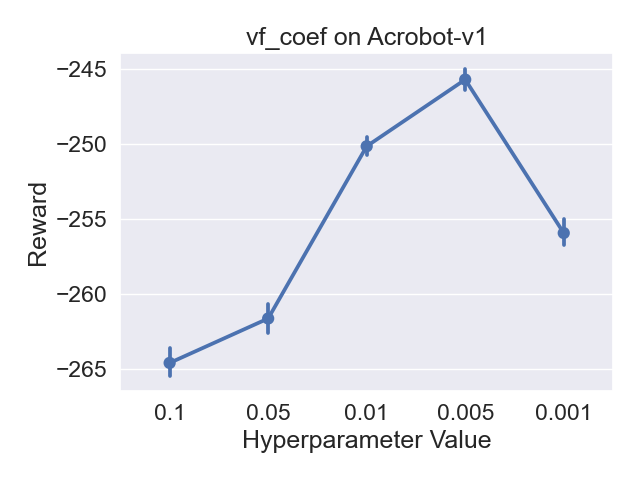"}
    \includegraphics[width=0.2\textwidth]{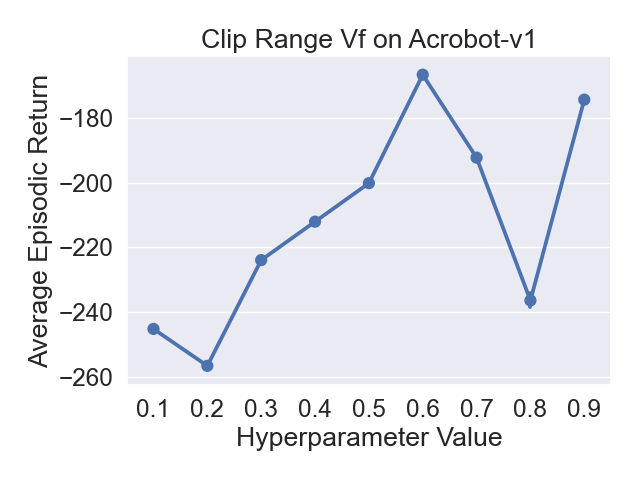}
    \caption{Final returns across 5 seeds for different hp variations of PPO on Pendulum and Acrobot.}
    \label{app-fig:ppo_boxplots}
\end{figure}

\begin{figure}
    \centering
    \includegraphics[width=0.2\textwidth]{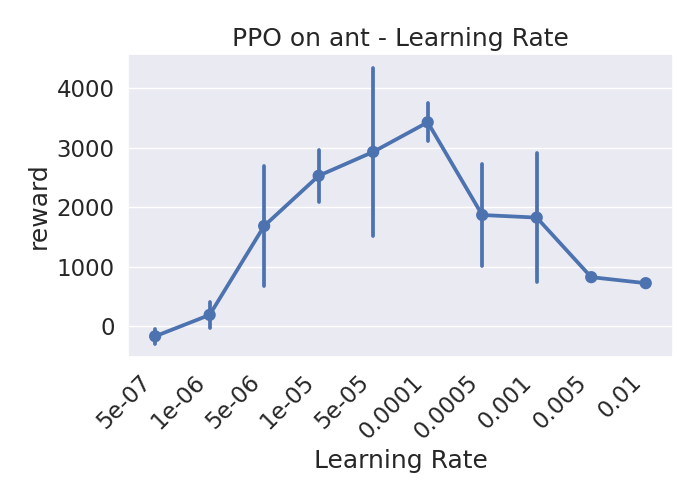}
    \includegraphics[width=0.2\textwidth]{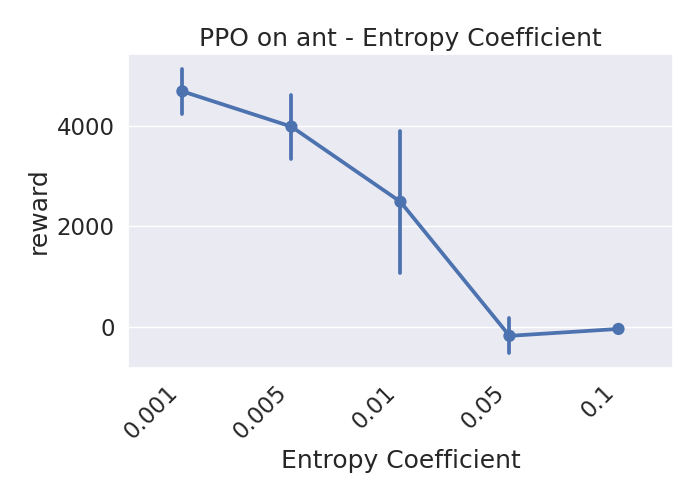}
    \includegraphics[width=0.2\textwidth]{boxplot_ant_PPO_algorithm.model_kwargs.clip_range}
    \includegraphics[width=0.2\textwidth]{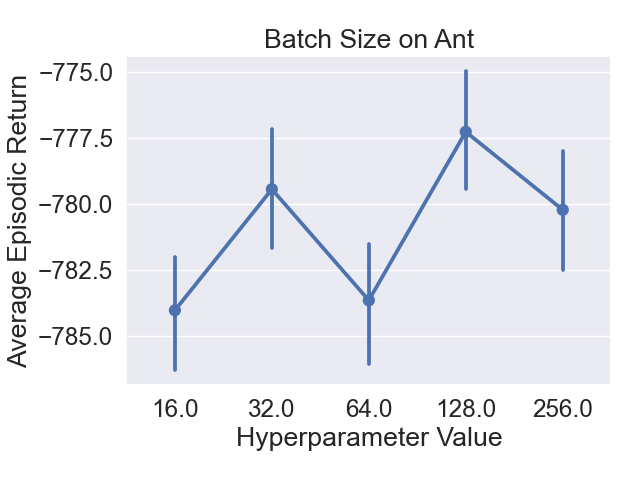}
    \includegraphics[width=0.2\textwidth]{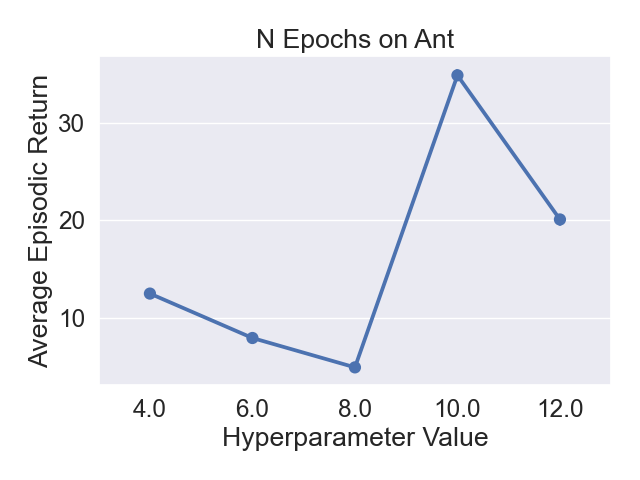}
    \includegraphics[width=0.2\textwidth]{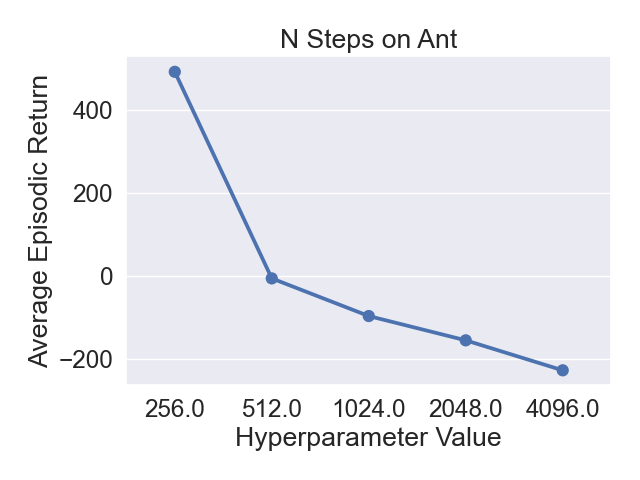}
    \includegraphics[width=0.2\textwidth]{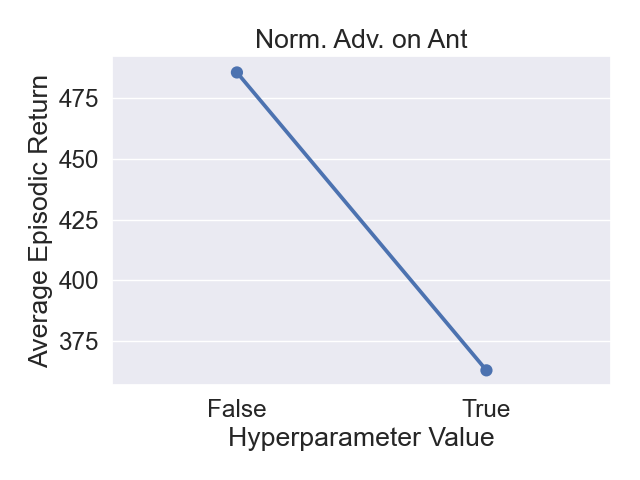}
    \includegraphics[width=0.2\textwidth]{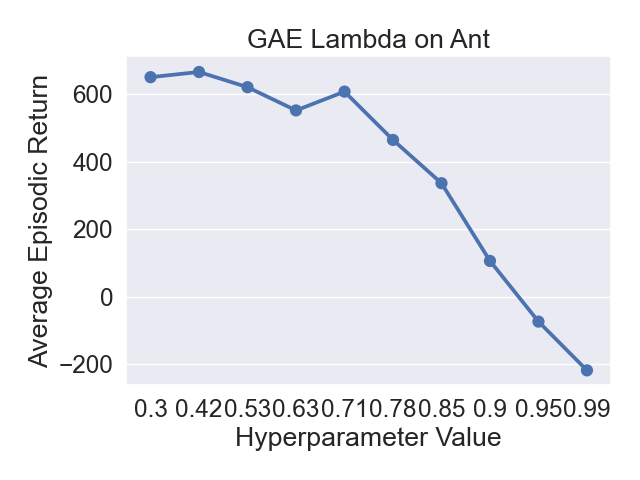}
    \includegraphics[width=0.2\textwidth]{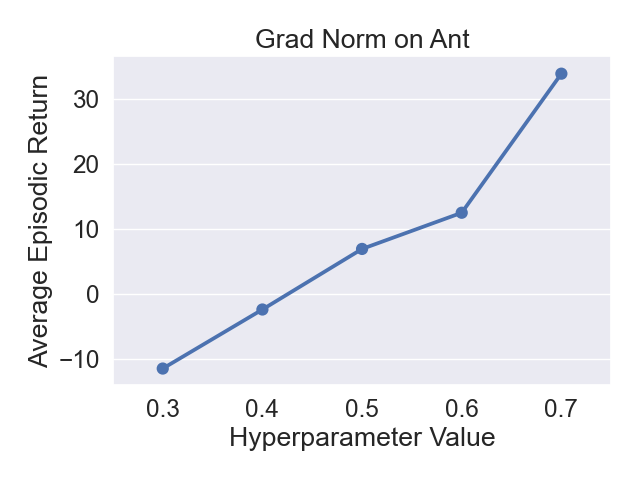}
    \includegraphics[width=0.2\textwidth]{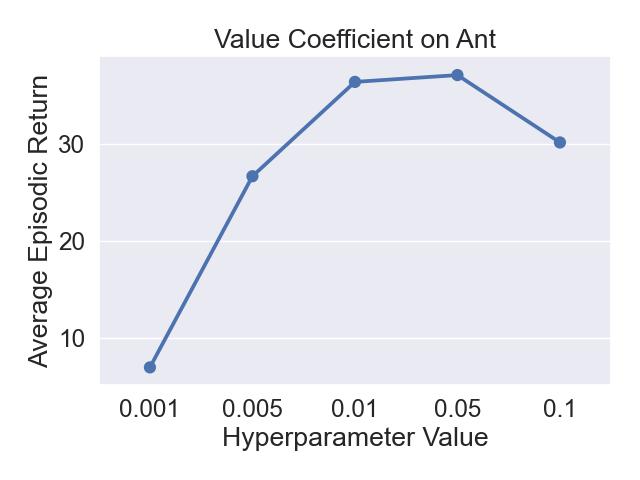}
    \includegraphics[width=0.2\textwidth]{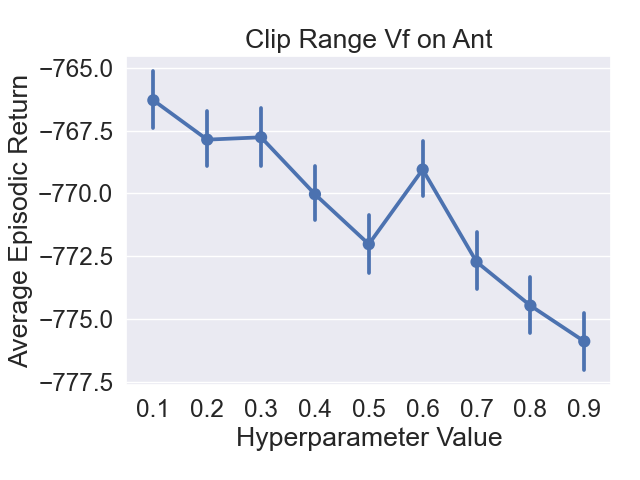}
    \includegraphics[width=0.2\textwidth]{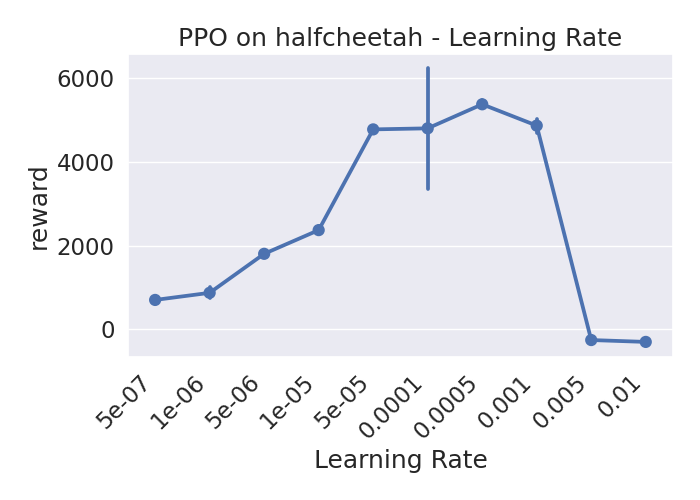}
    \includegraphics[width=0.2\textwidth]{boxplot_halfcheetah_PPO_algorithm.model_kwargs.ent_coef}
    \includegraphics[width=0.2\textwidth]{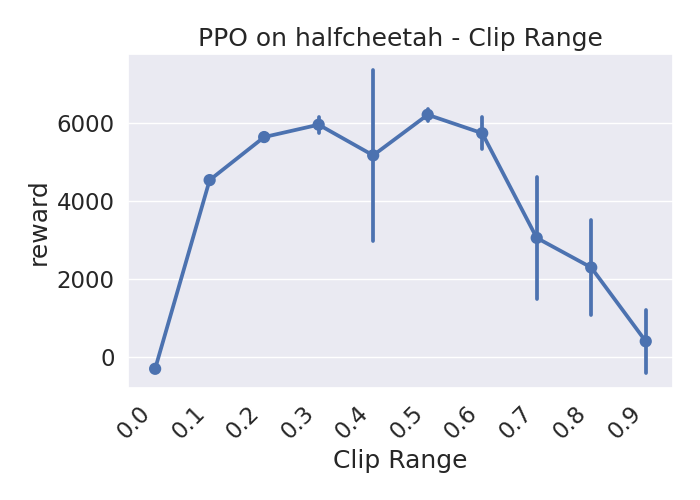}
    \includegraphics[width=0.2\textwidth]{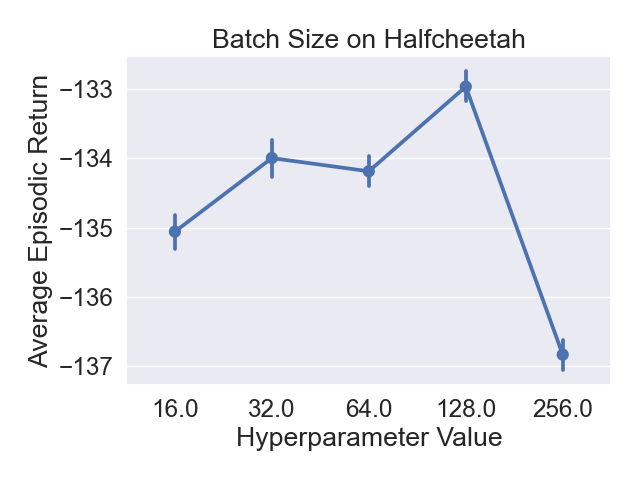}
    \includegraphics[width=0.2\textwidth]{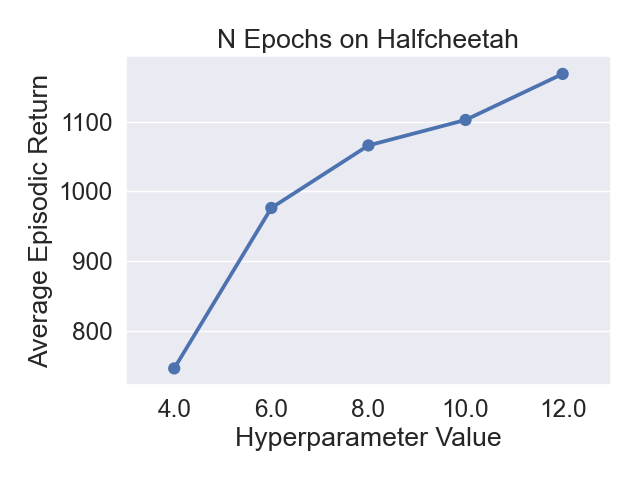}
    \includegraphics[width=0.2\textwidth]{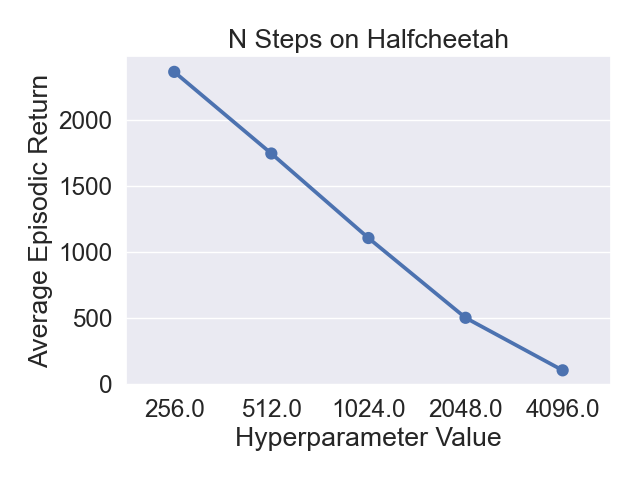}
    \includegraphics[width=0.2\textwidth]{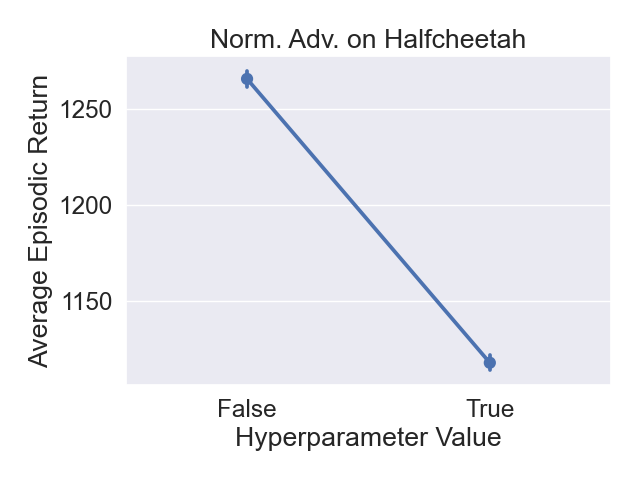}
    \includegraphics[width=0.2\textwidth]{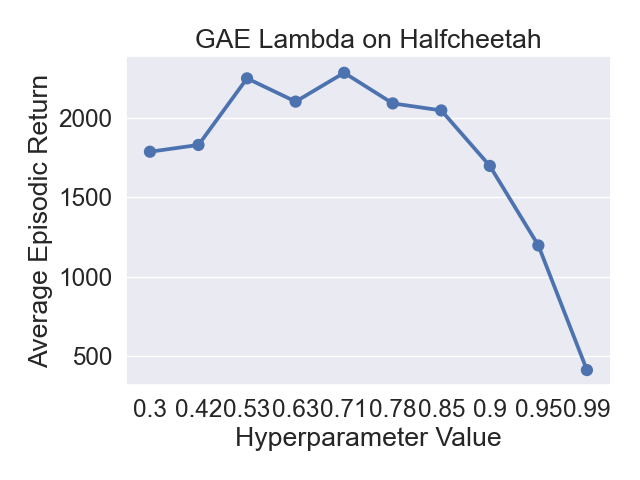}
    \includegraphics[width=0.2\textwidth]{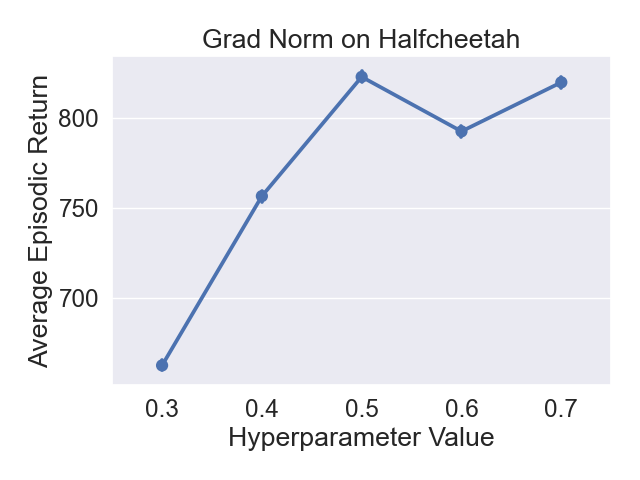}
    \includegraphics[width=0.2\textwidth]{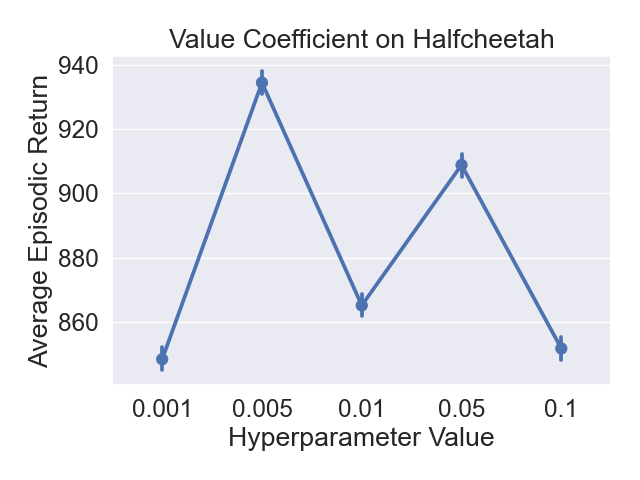}
    \includegraphics[width=0.2\textwidth]{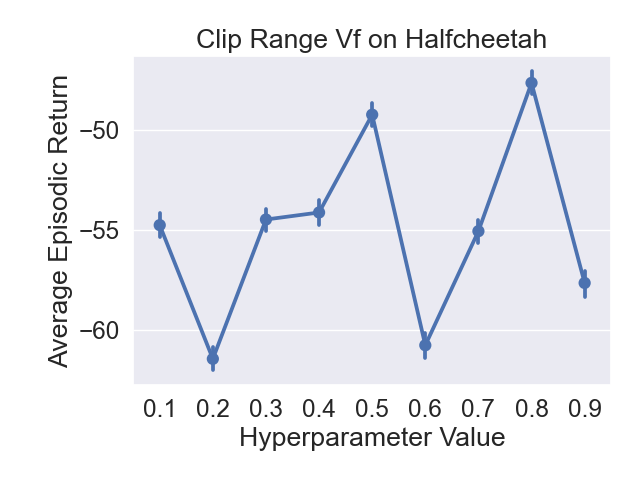}
    \caption{Final returns across 5 seeds for different hp variations of PPO on Ant and Halfcheetah.}
    \label{app-fig:ppo_boxplots2}
\end{figure}

\begin{figure}
    \centering
    \includegraphics[width=0.2\textwidth]{boxplot_humanoid_PPO_algorithm.model_kwargs.learning_rate}
    \includegraphics[width=0.2\textwidth]{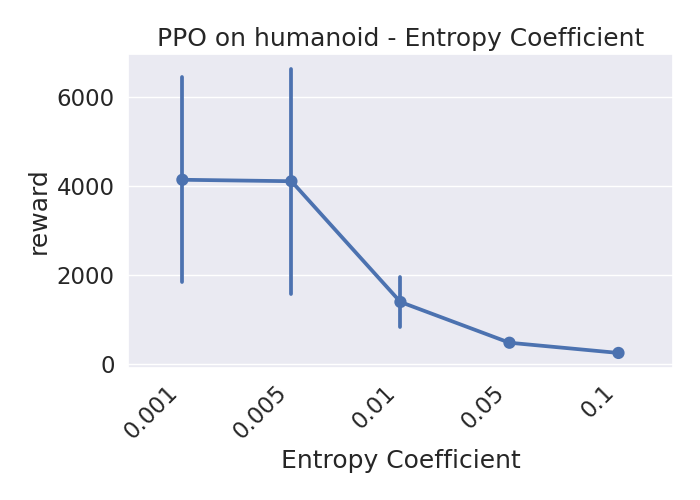}
    \includegraphics[width=0.2\textwidth]{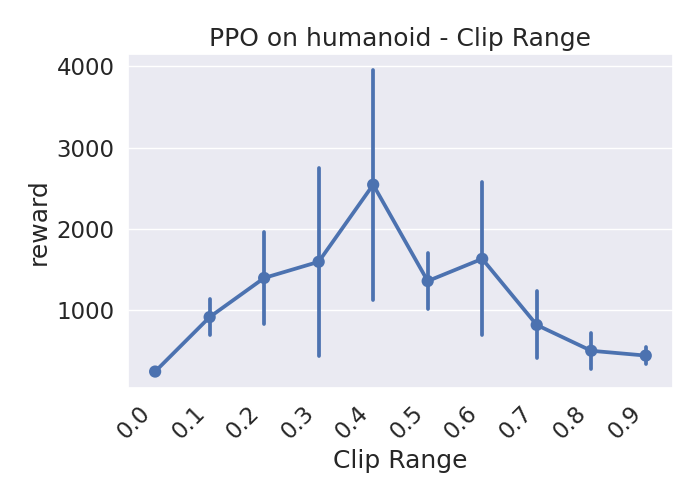}
    \includegraphics[width=0.2\textwidth]{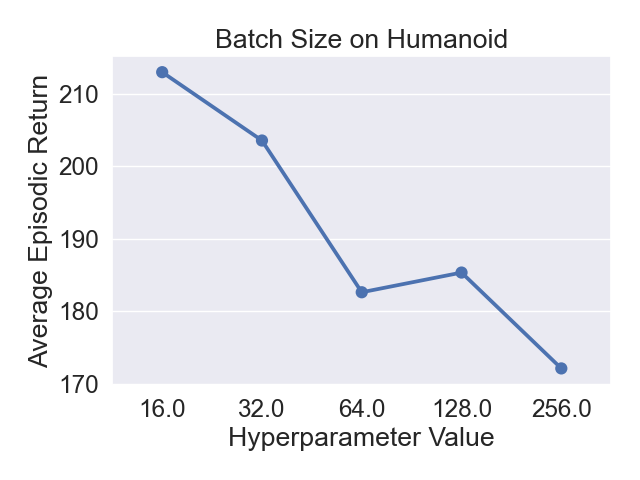}
    \includegraphics[width=0.2\textwidth]{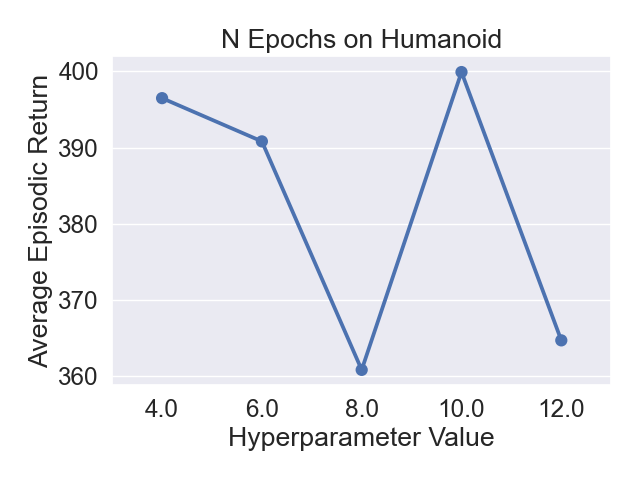}
    \includegraphics[width=0.2\textwidth]{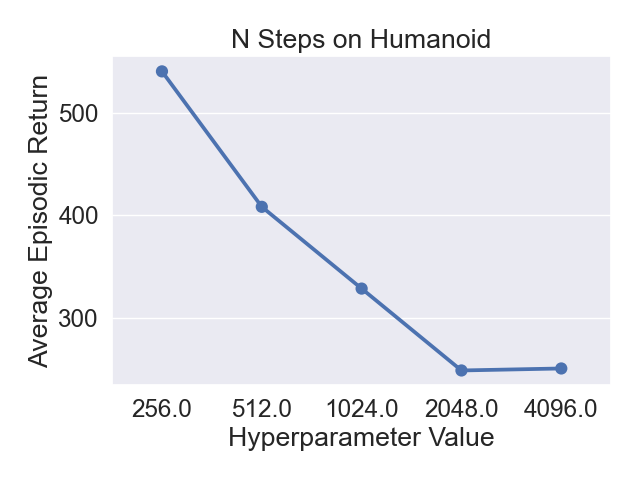}
    \includegraphics[width=0.2\textwidth]{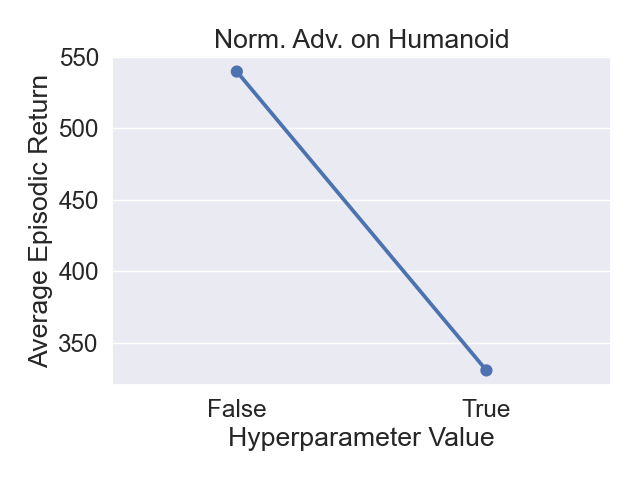}
    \includegraphics[width=0.2\textwidth]{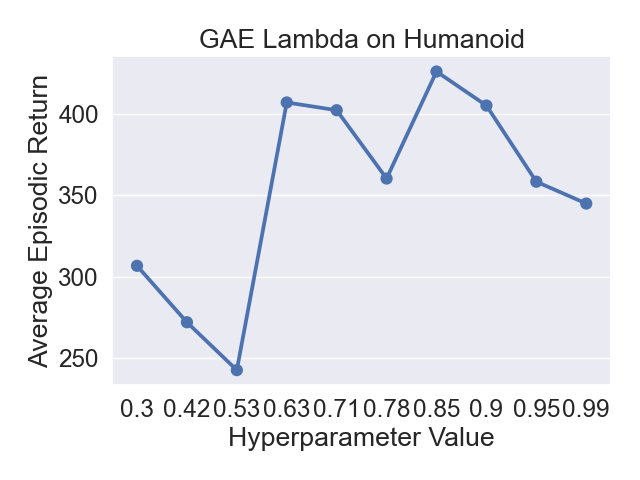}
    \includegraphics[width=0.2\textwidth]{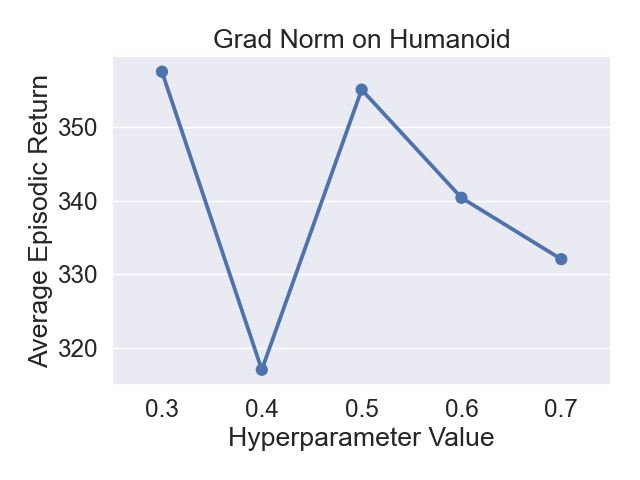}
    \includegraphics[width=0.2\textwidth]{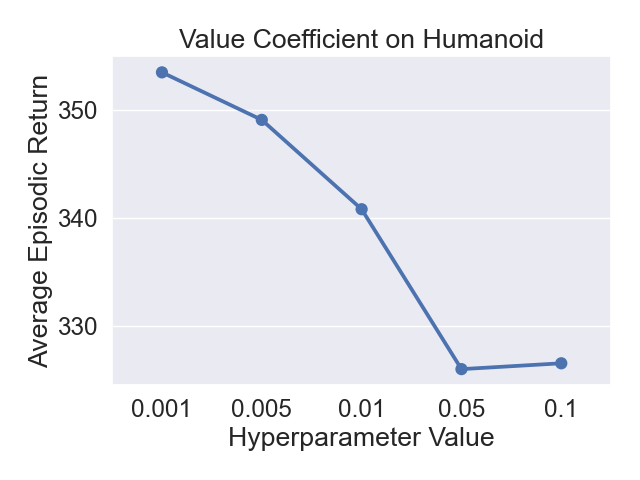}
    \includegraphics[width=0.2\textwidth]{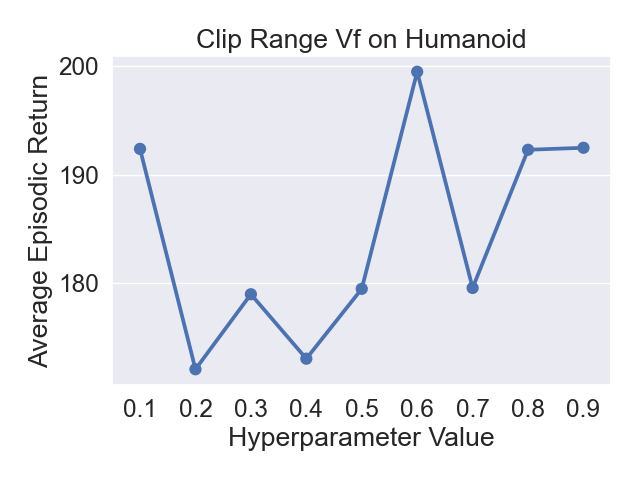}
    \caption{Final returns across 5 seeds for different hp variations of PPO on Humanoid.}
    \label{app-fig:ppo_boxplots3}
\end{figure}

\begin{figure}
    \centering
    \includegraphics[width=0.2\textwidth]{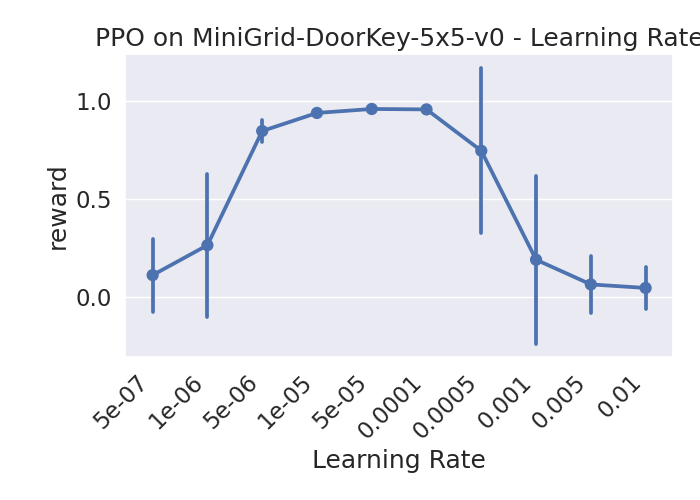}
    \includegraphics[width=0.2\textwidth]{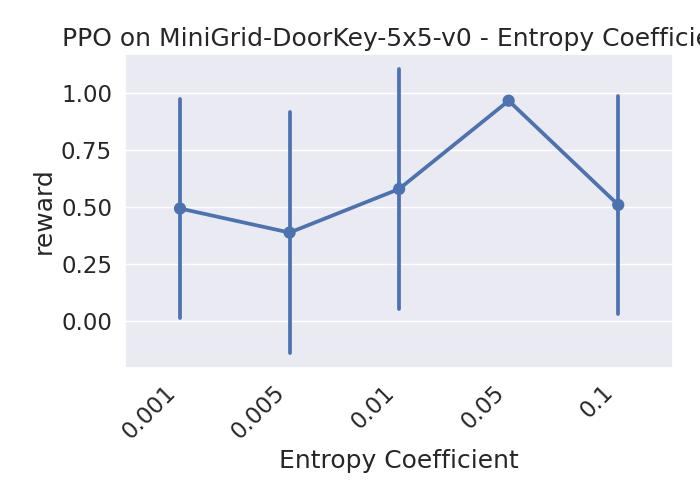}
    \includegraphics[width=0.2\textwidth]{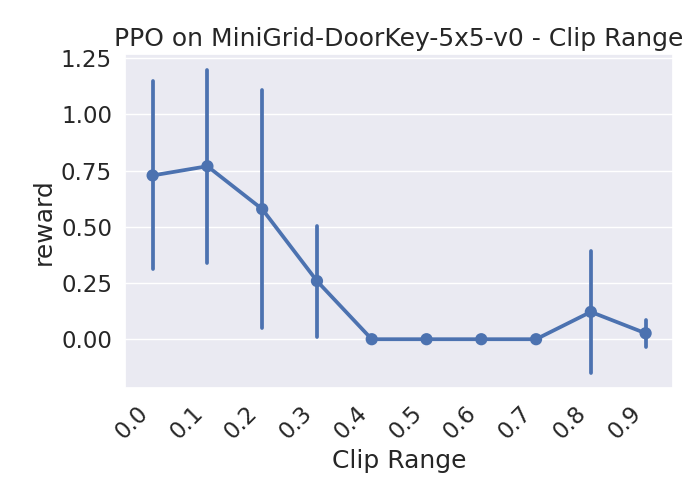}
    \includegraphics[width=0.2\textwidth]{figs/appendix_sweeps/pointplot_MiniGrid-DoorKey-5x5-v0_DQN_batch_size.png}
    \includegraphics[width=0.2\textwidth]{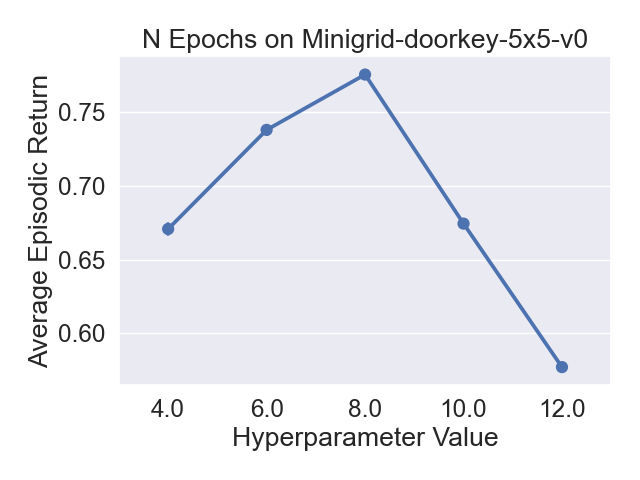}
    \includegraphics[width=0.2\textwidth]{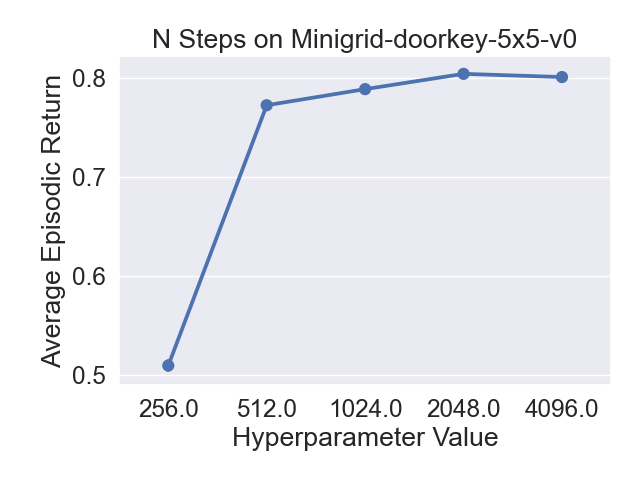}
    \includegraphics[width=0.2\textwidth]{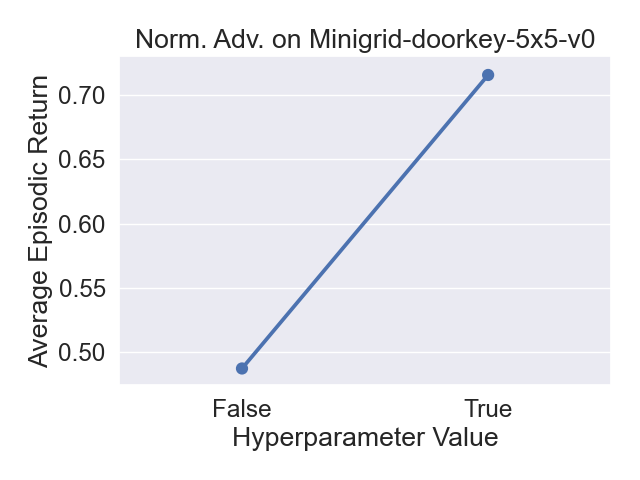}
    \includegraphics[width=0.2\textwidth]{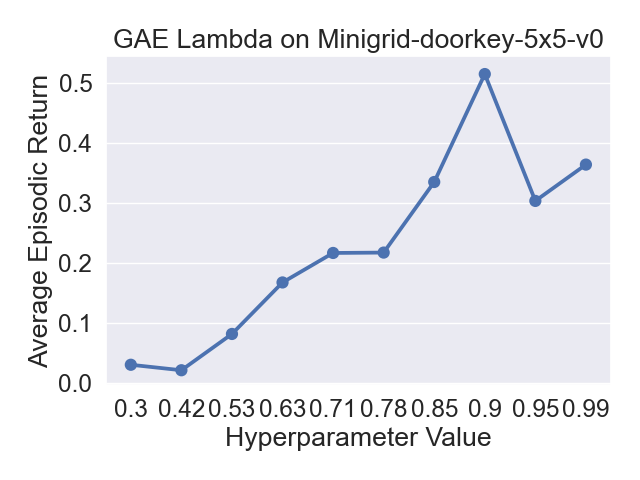}
    \includegraphics[width=0.2\textwidth]{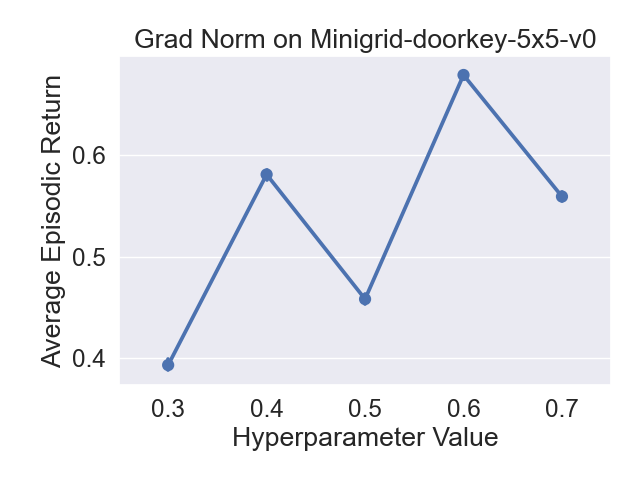}
    \includegraphics[width=0.2\textwidth]{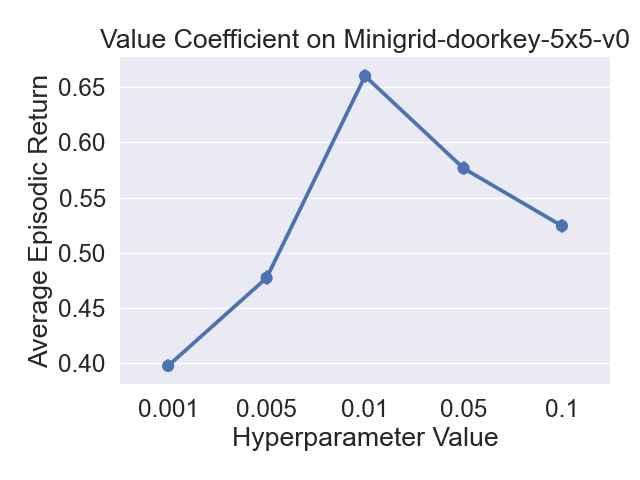}
    \includegraphics[width=0.2\textwidth]{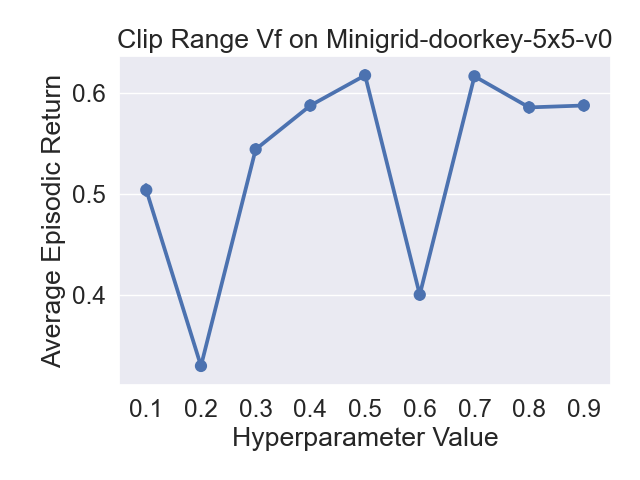}
    \includegraphics[width=0.2\textwidth]{boxplot_MiniGrid-Empty-5x5-v0_PPO_algorithm.model_kwargs.learning_rate}
    \includegraphics[width=0.2\textwidth]{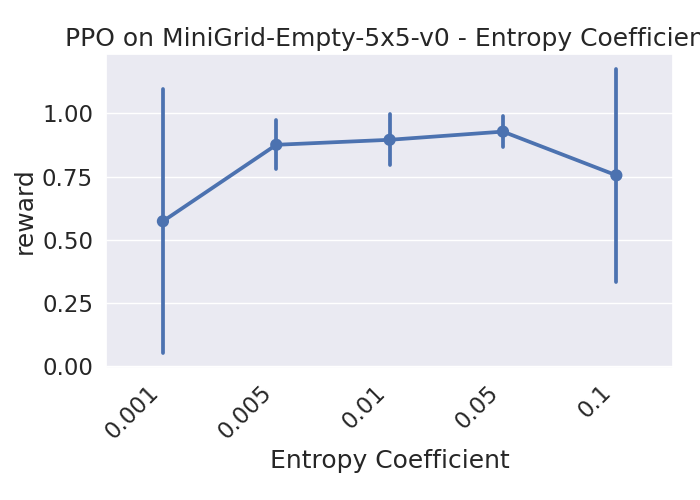}
    \includegraphics[width=0.2\textwidth]{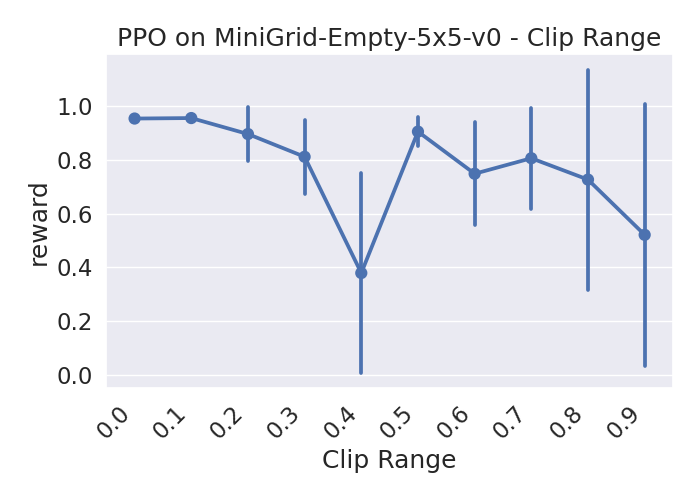}
    \includegraphics[width=0.2\textwidth]{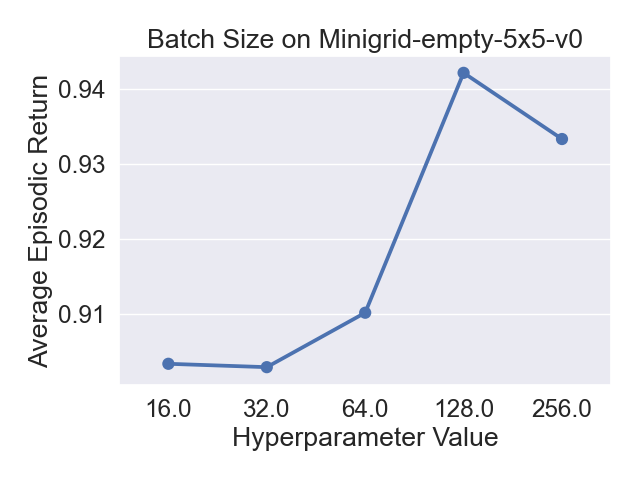}
    \includegraphics[width=0.2\textwidth]{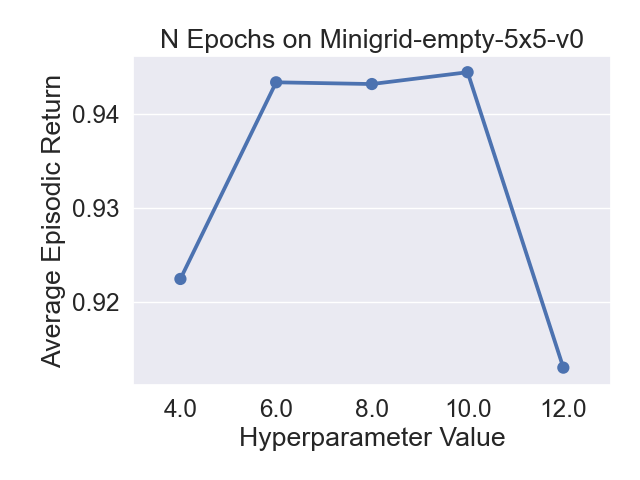}
    \includegraphics[width=0.2\textwidth]{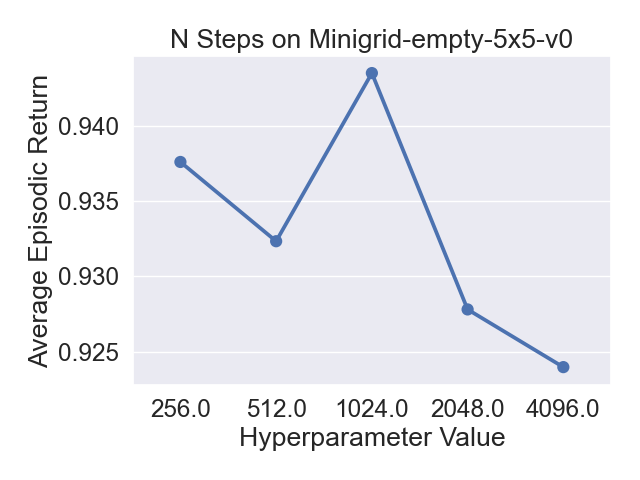}
    \includegraphics[width=0.2\textwidth]{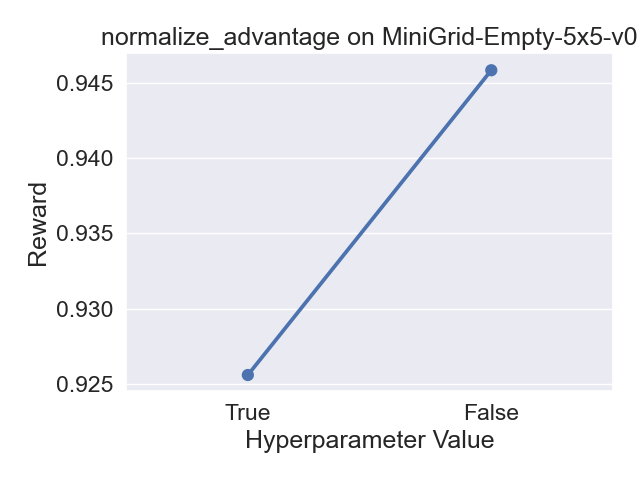}
    \includegraphics[width=0.2\textwidth]{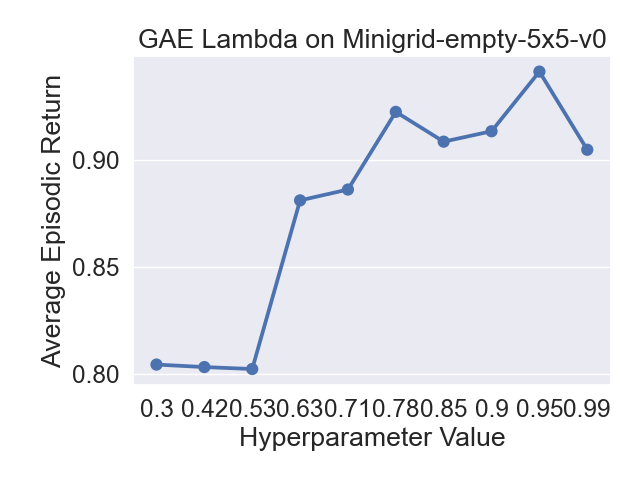}
    \includegraphics[width=0.2\textwidth]{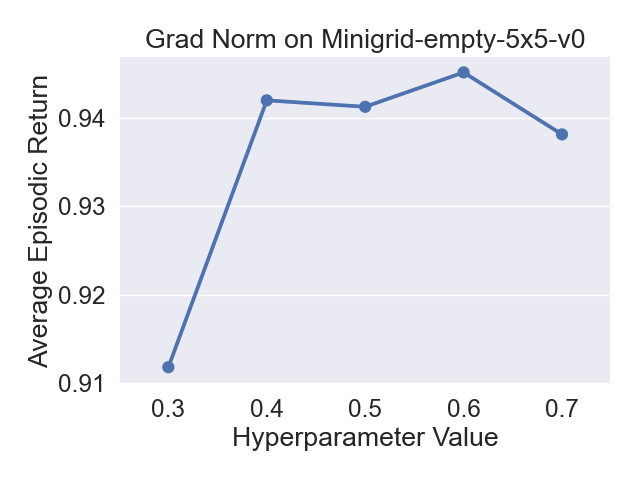}
    \includegraphics[width=0.2\textwidth]{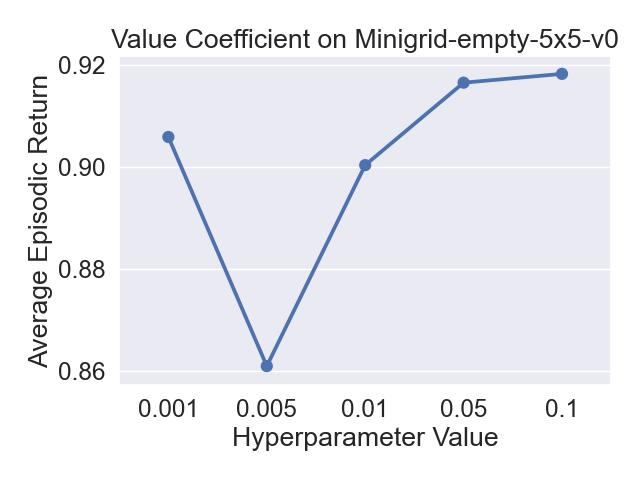}
    \includegraphics[width=0.2\textwidth]{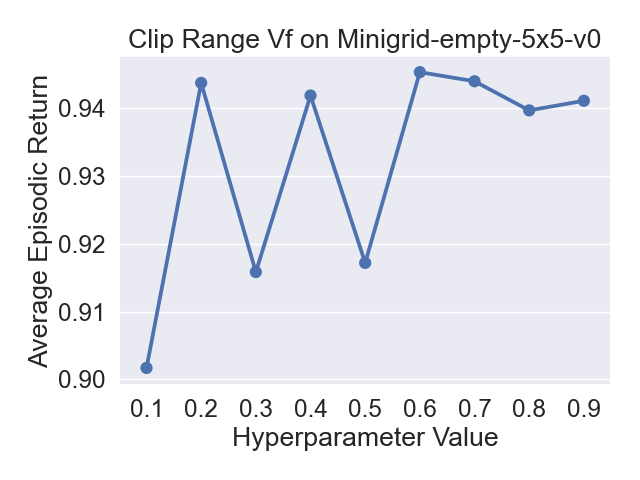}
    \caption{Final returns across 5 seeds for different hp variations of PPO on MiniGrid.}
    \label{app-fig:ppo_boxplots4}
\end{figure}

The full set of SAC pointplots can be found in Figures~\ref{app-fig:sac_boxplots} and \ref{app-fig:sac_boxplots2}, the DQN pointplots in Figures~\ref{app-fig:dqn_boxplots} and \ref{app-fig:dqn_boxplots2} and the PPO pointplots in Figures~\ref{app-fig:ppo_boxplots}, \ref{app-fig:ppo_boxplots2}, \ref{app-fig:ppo_boxplots3} and \ref{app-fig:ppo_boxplots4}.

\subsection{SAC Pointplots}
\clearpage
\subsection{DQN Pointplots}
\clearpage

\subsection{PPO Pointplots}


\clearpage
\section{Hyperparameter Importances using fANOVA}
\label{app:importance}
These hyperparameter importance plots were made using the fANOVA~\cite{hutter-icml14a} plugin of DeepCAVE~\cite{sass-arxiv22}.

\begin{figure}[h]
    \centering
    \includegraphics[width=0.47\textwidth]{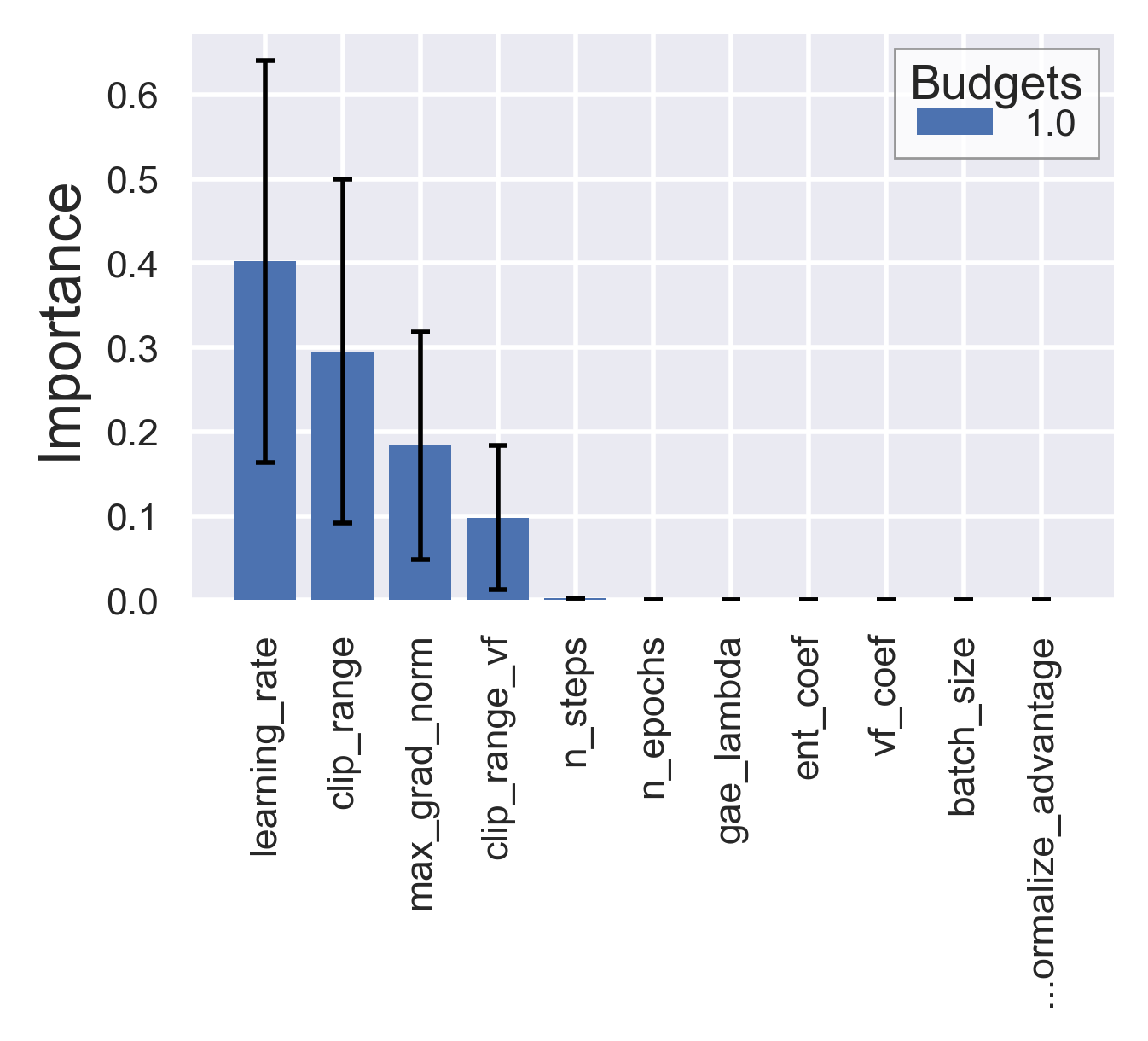}
    \includegraphics[width=0.47\textwidth]{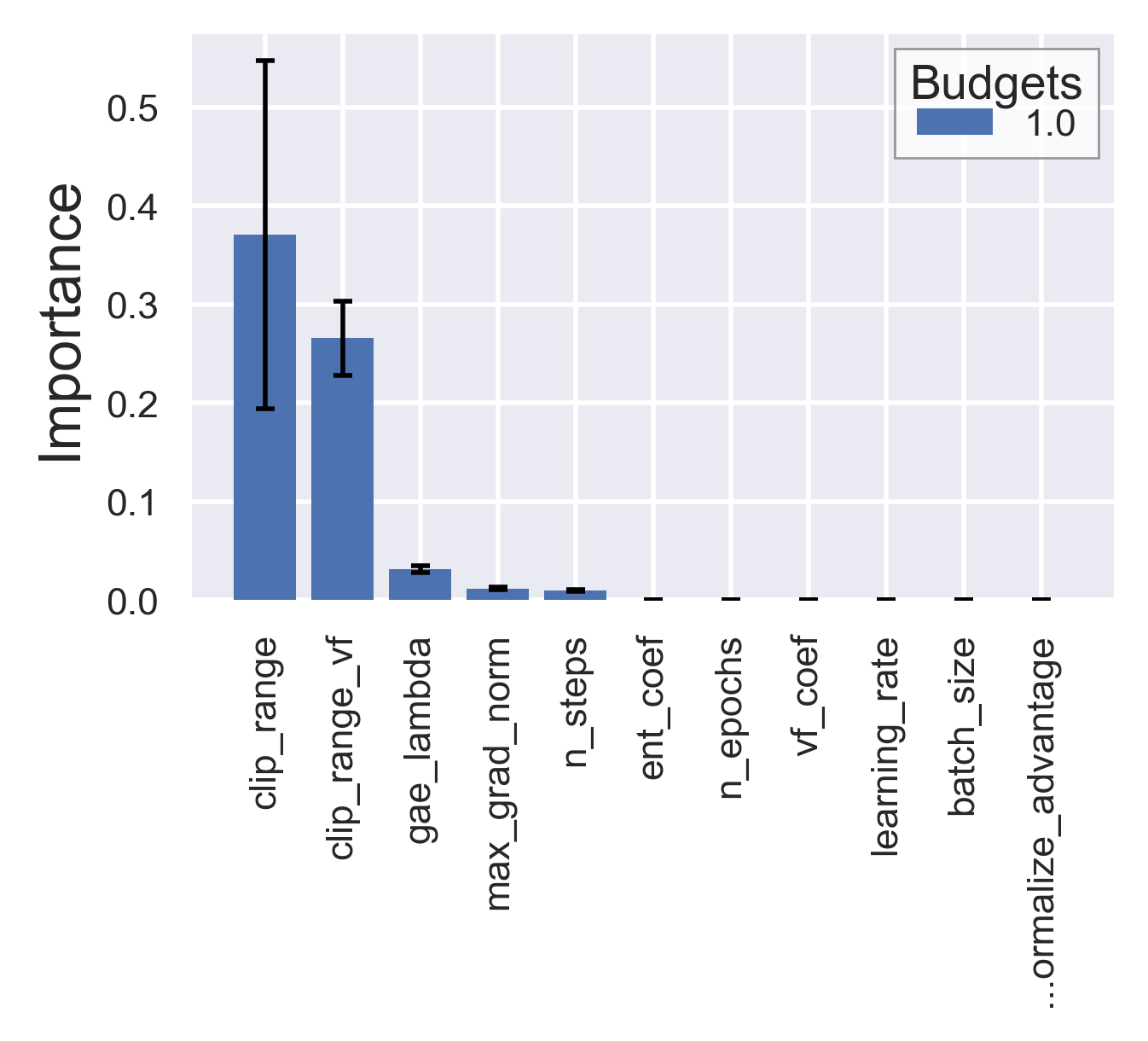}
    \caption{PPO Hyperparameter Importances on Acrobot (left) and Pendulum (right).}
\end{figure}

\begin{figure}[h]
    \centering
    \includegraphics[width=0.47\textwidth]{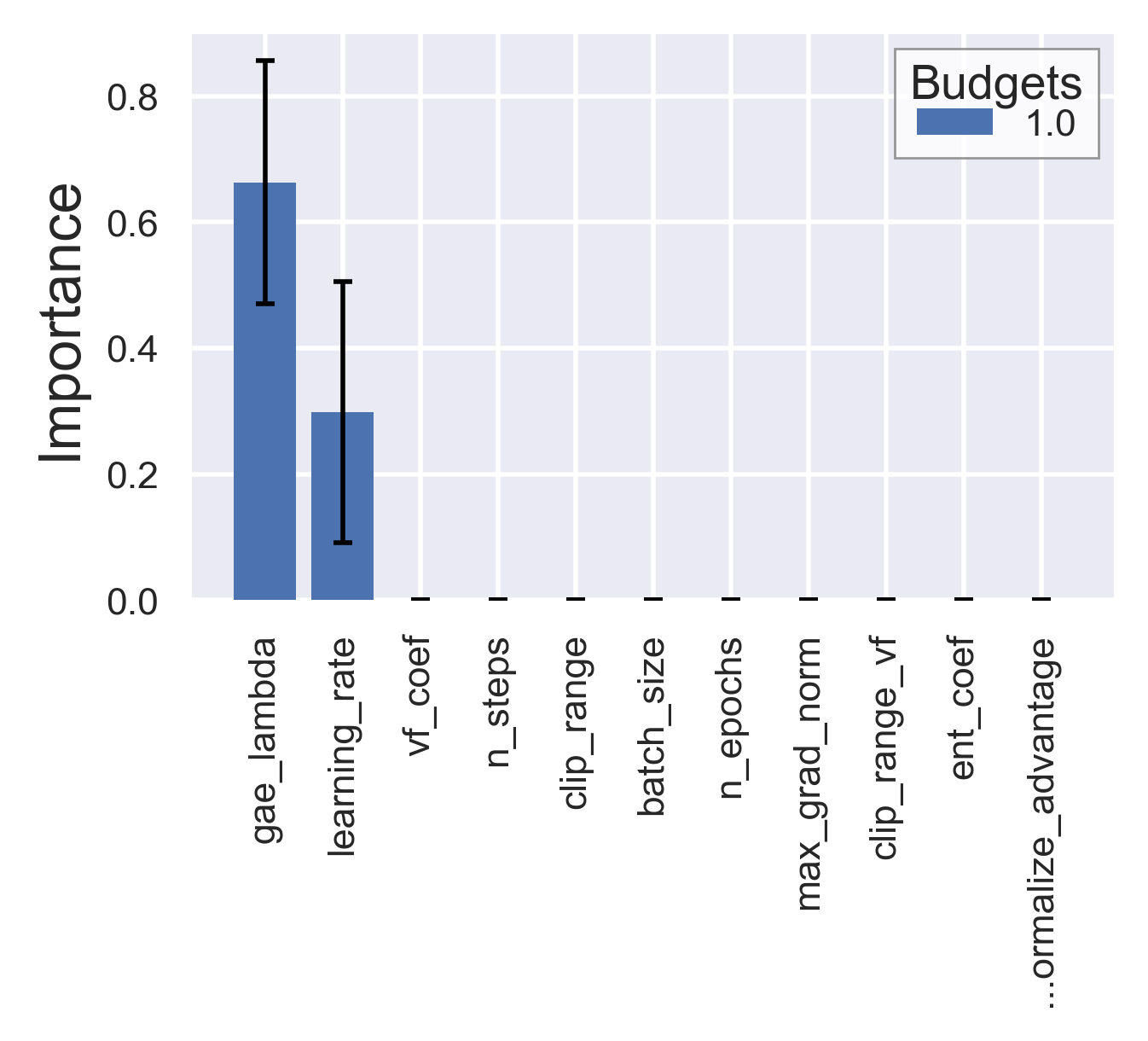}
    \includegraphics[width=0.47\textwidth]{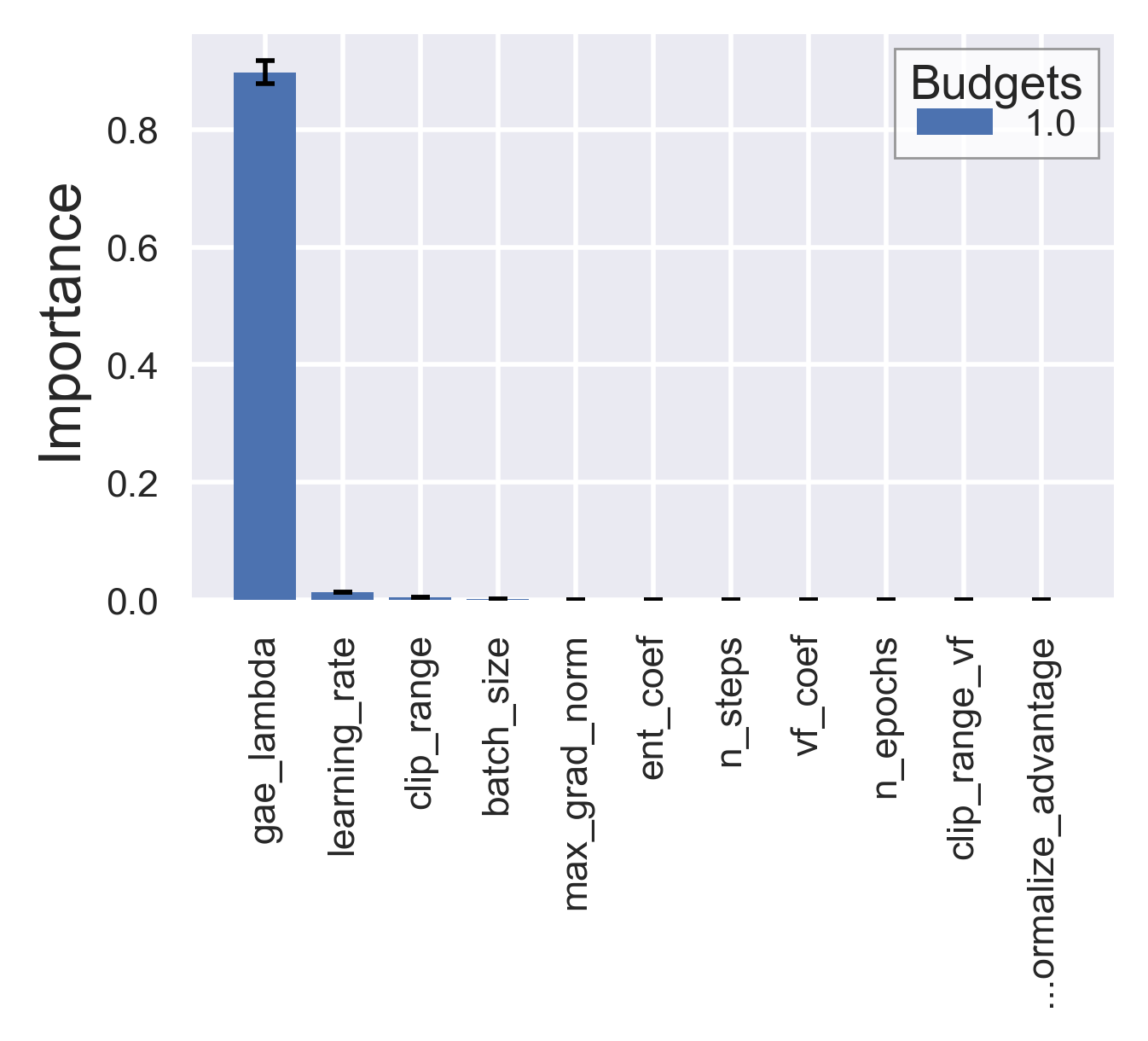}
    \caption{PPO Hyperparameter Importances on MiniGrid Empty (left) and Minigrid DoorKey(right).}
\end{figure}

\begin{figure}
    \centering
    \includegraphics[width=0.3\textwidth]{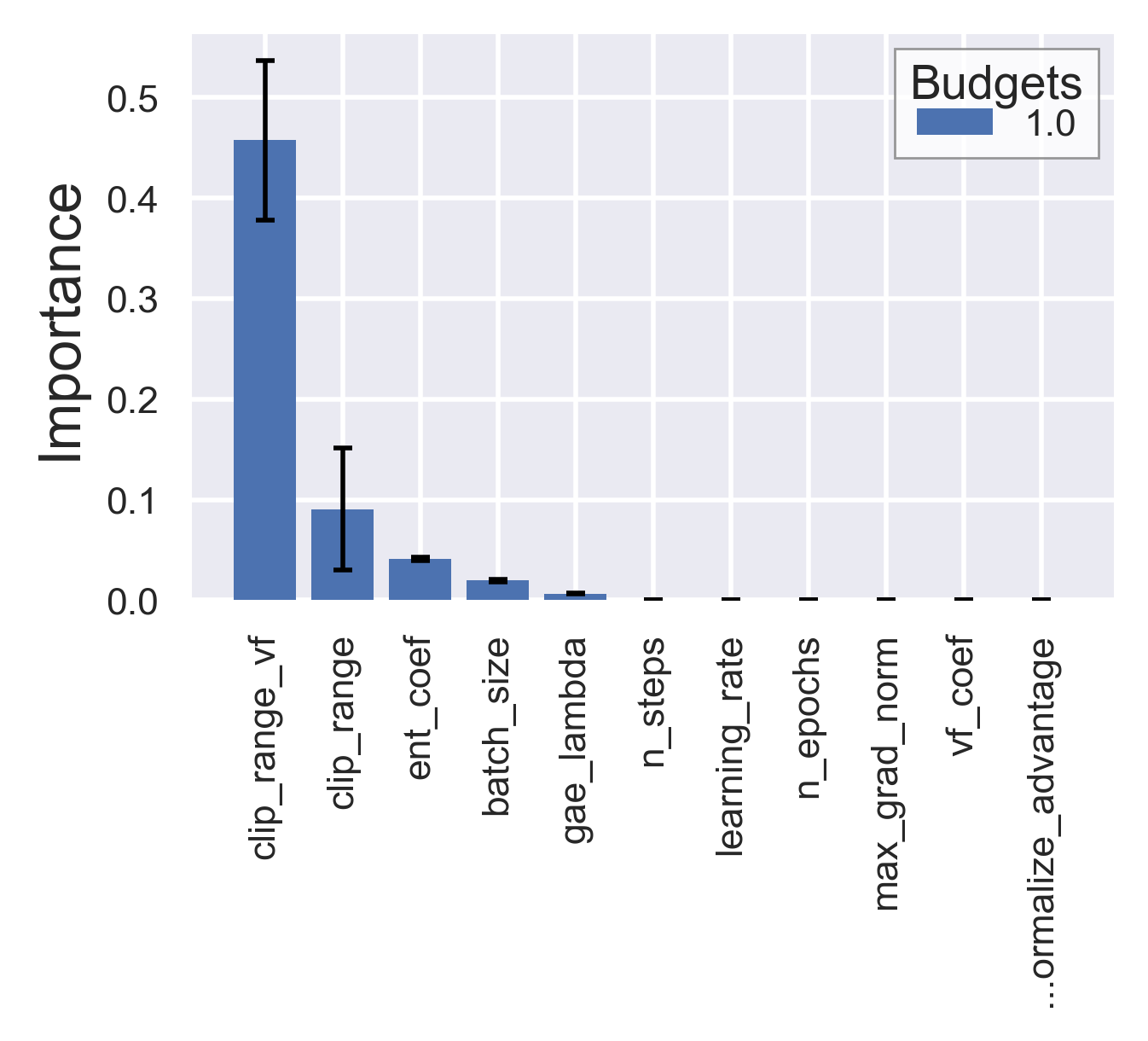}
    \includegraphics[width=0.3\textwidth]{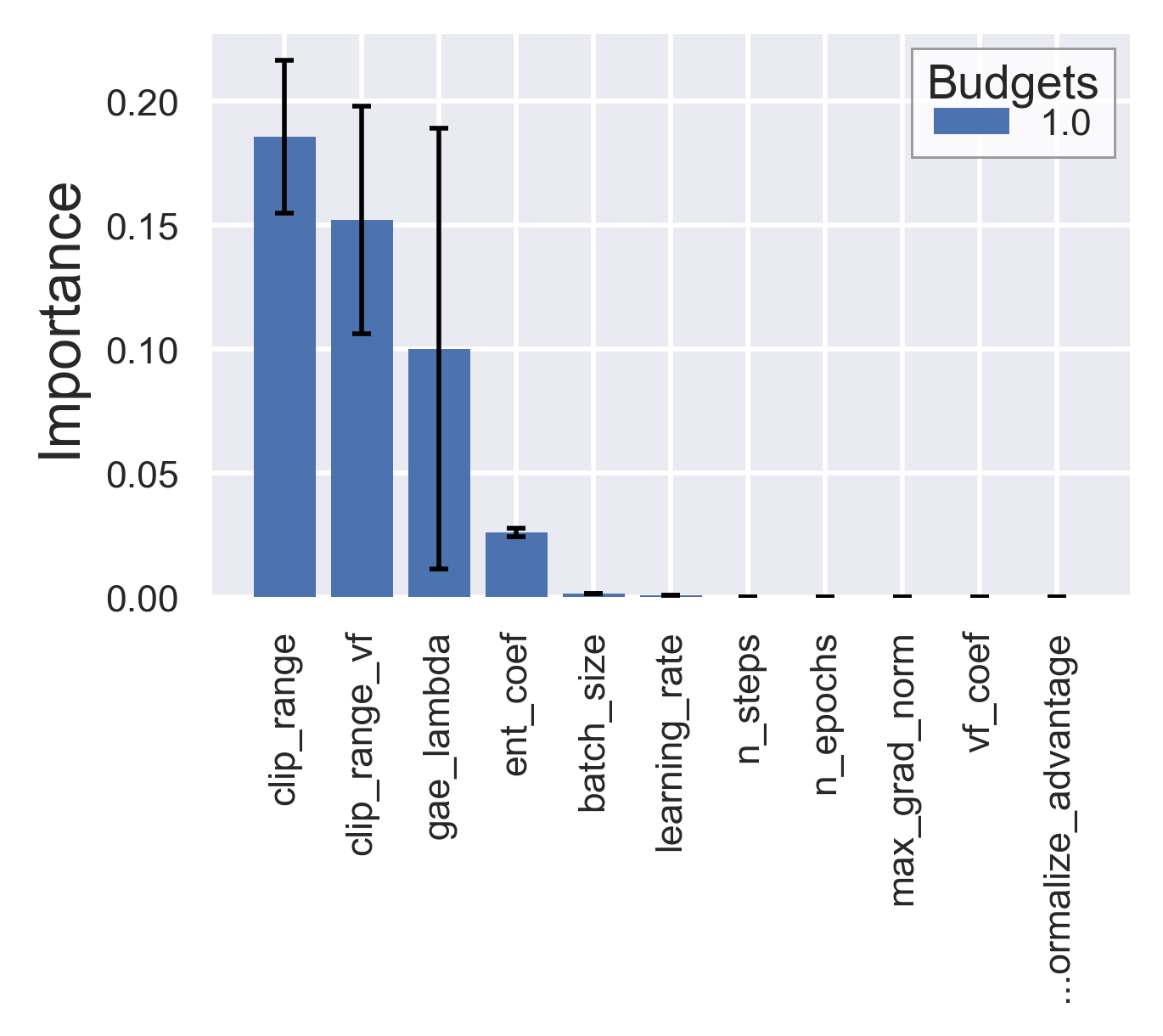}
    \includegraphics[width=0.3\textwidth]{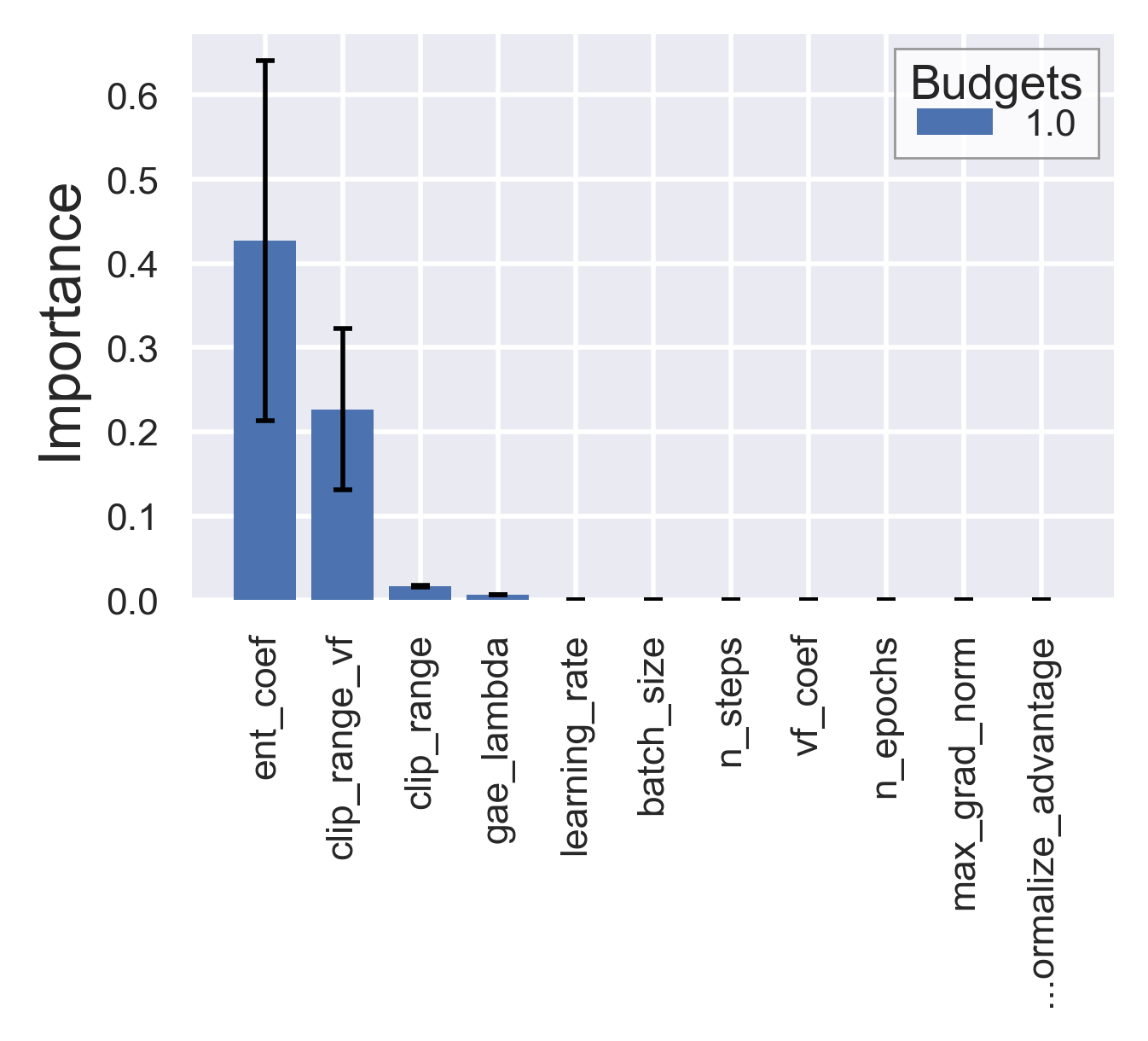}
    \caption{PPO Hyperparameter Importances on Brax Ant (left), Halfcheetah (middle) and Humanoid (right).}
\end{figure}

\begin{figure}
    \centering
    \includegraphics[width=0.3\textwidth]{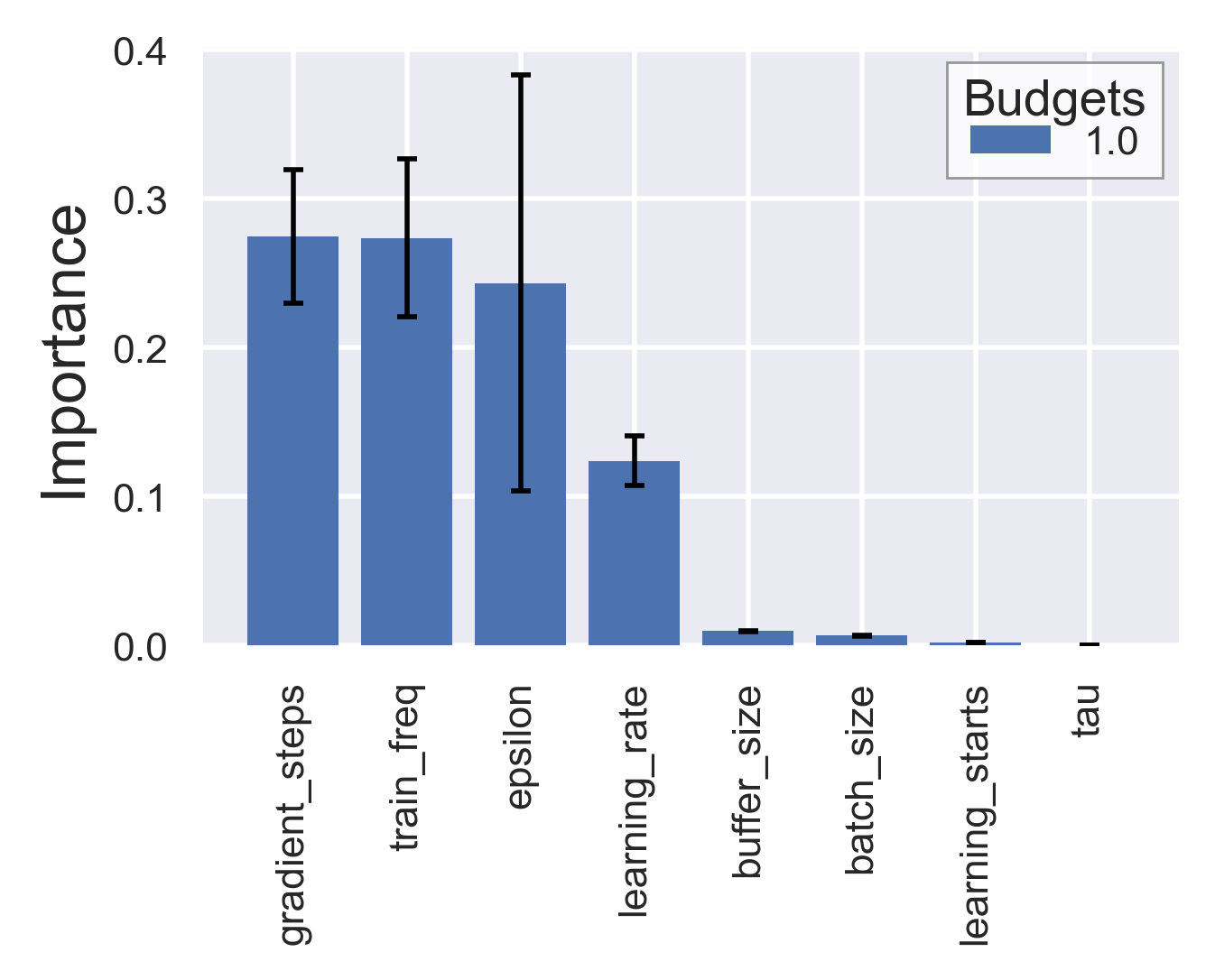}
    \includegraphics[width=0.3\textwidth]{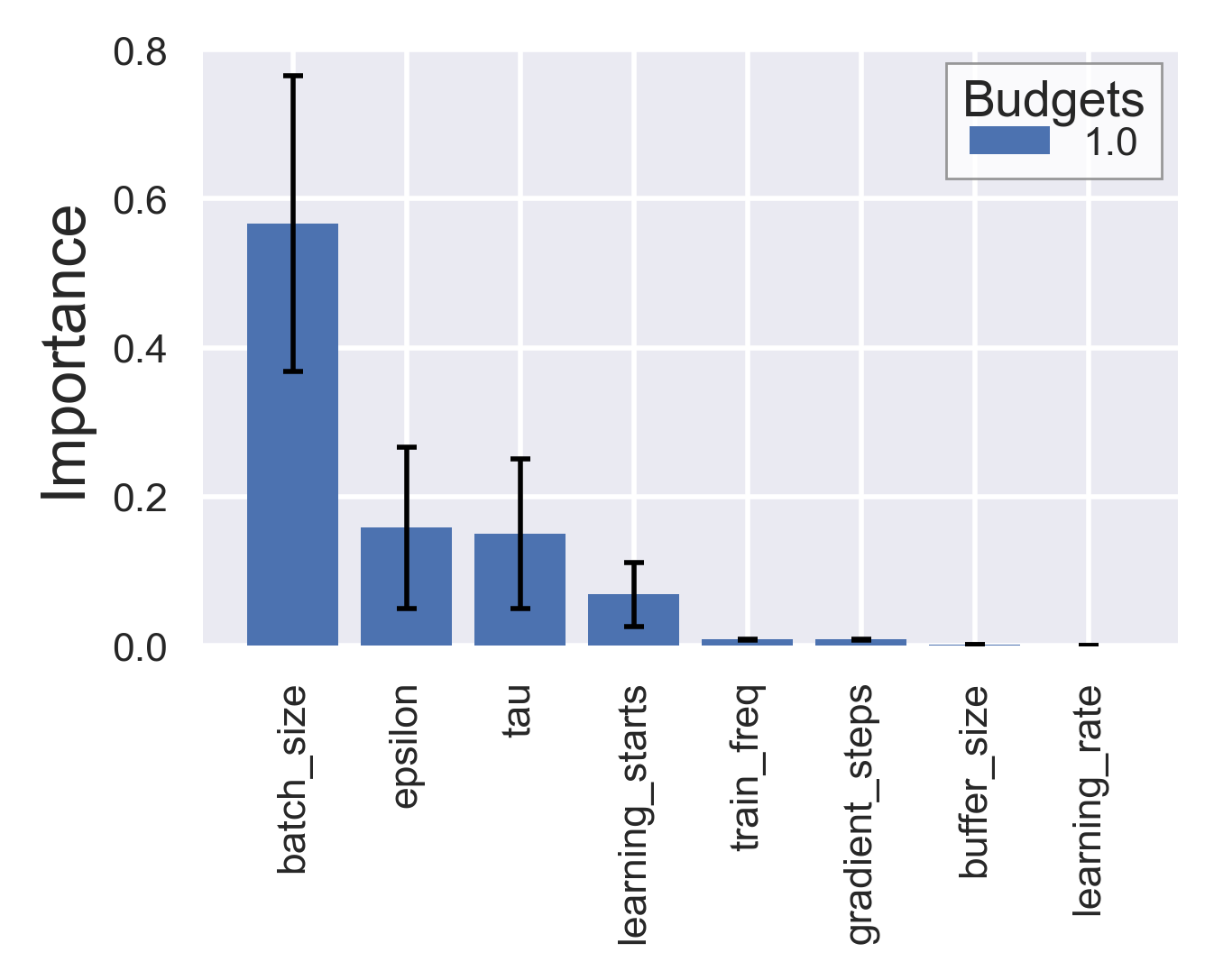}
    \includegraphics[width=0.3\textwidth]{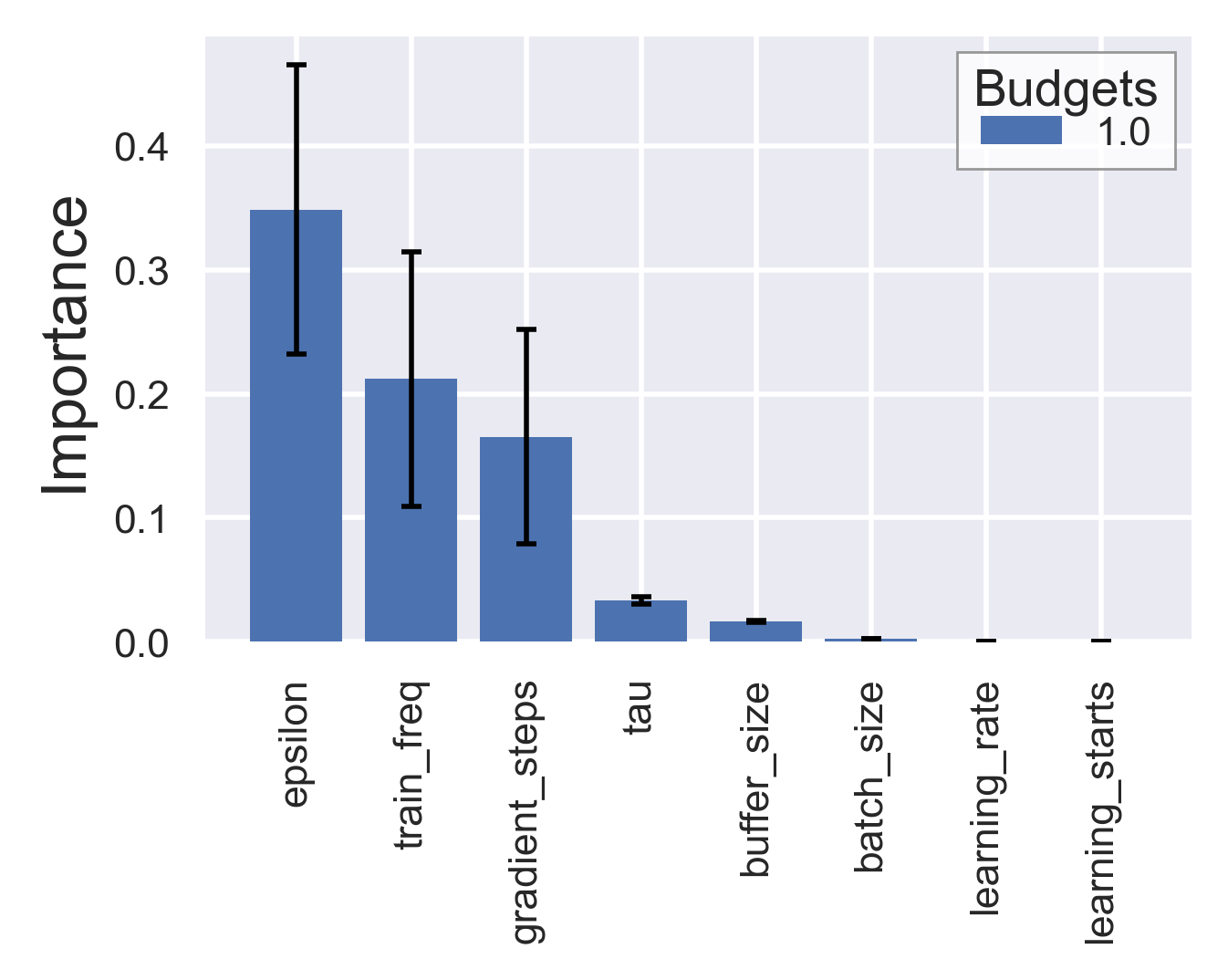}
    \caption{DQN Hyperparameter Importances on Acrobot (left), MiniGrid Empty (middle) and MiniGrid DoorKey (right).}
\end{figure}

\begin{figure}
    \centering
    \includegraphics[width=0.3\textwidth]{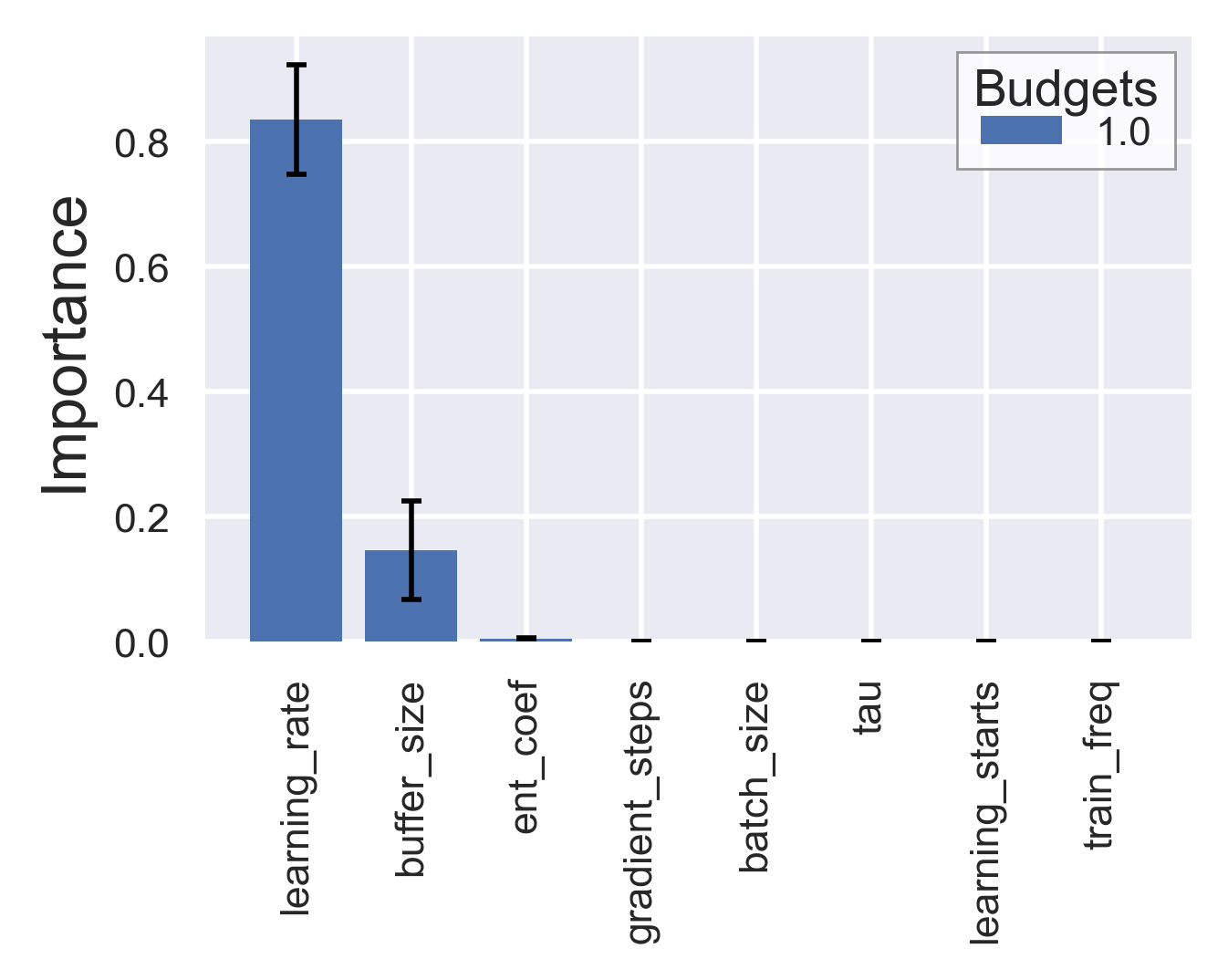}
    \includegraphics[width=0.3\textwidth]{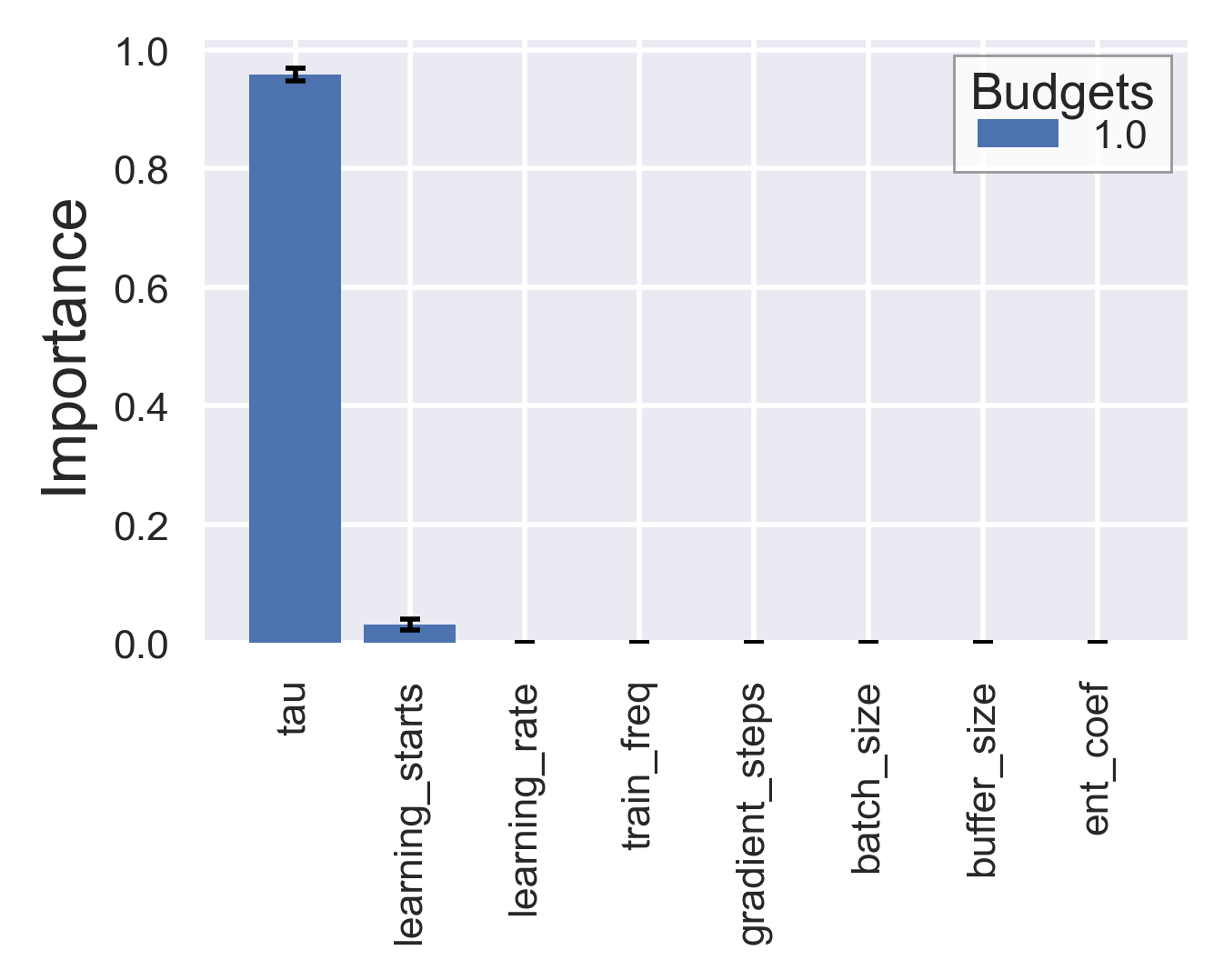}
    \caption{SAC Hyperparameter Importances on Pendulum (left) and Brax Ant(right).}
\end{figure}

\begin{figure}
    \centering
    \includegraphics[width=0.3\textwidth]{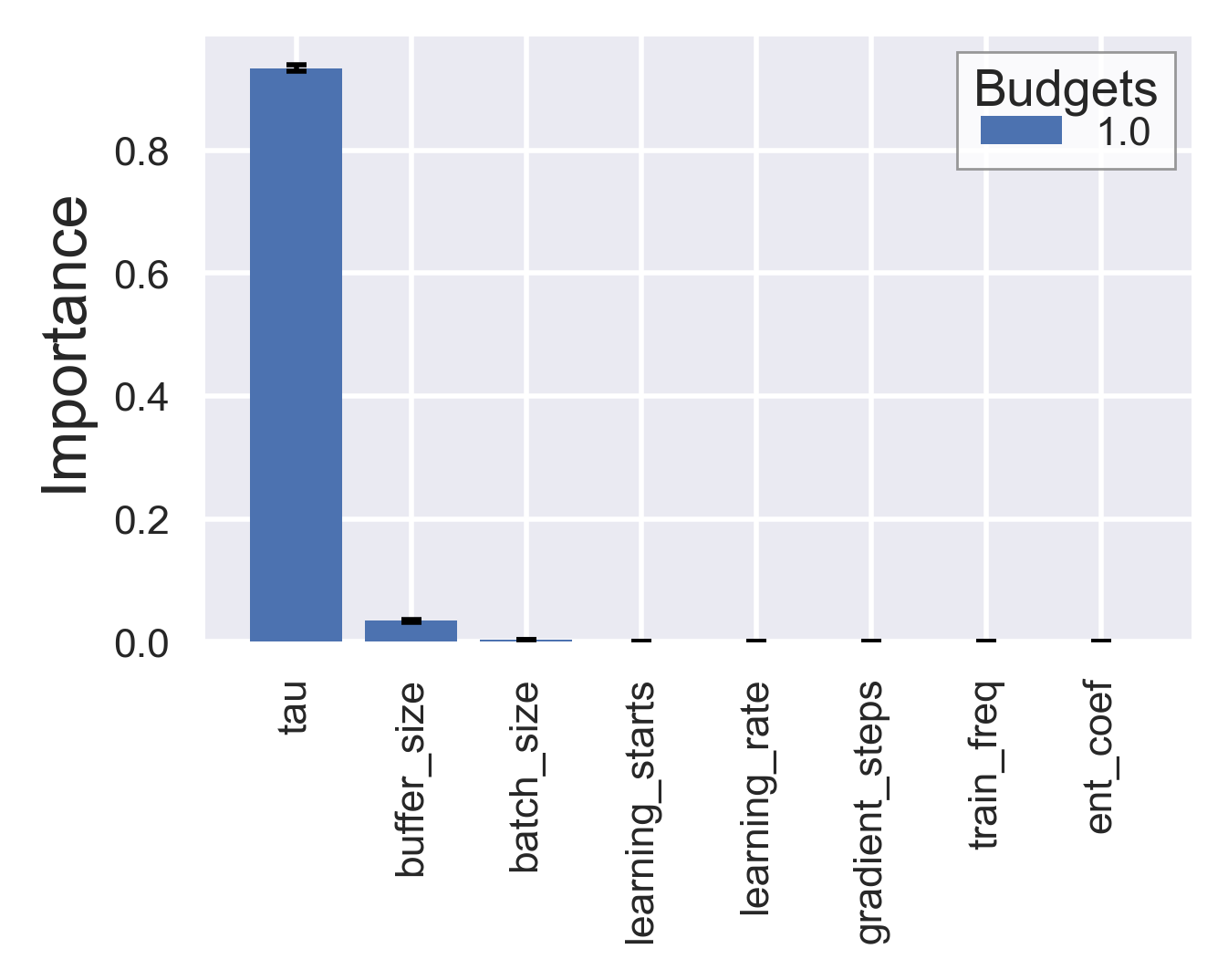}
    \includegraphics[width=0.3\textwidth]{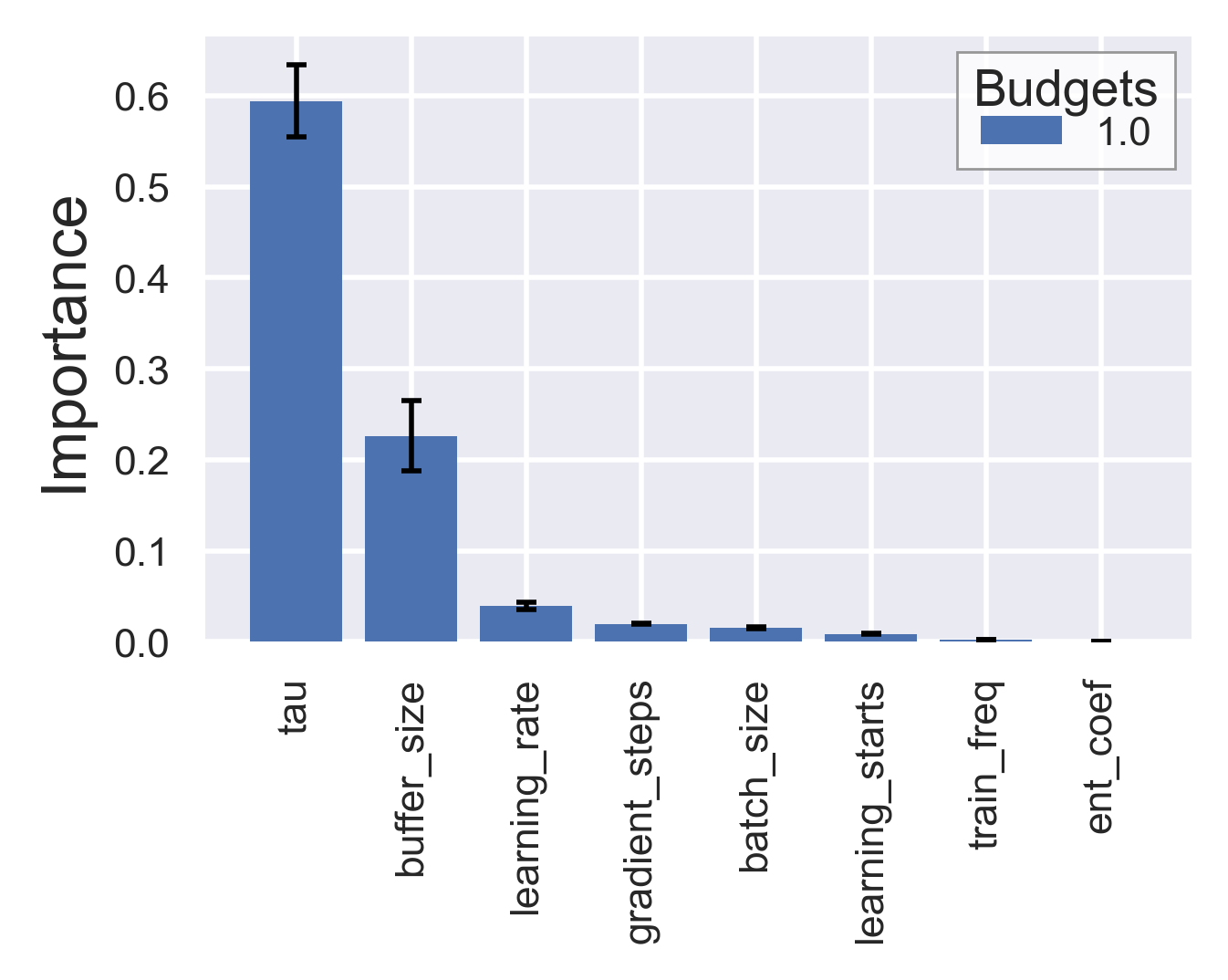}
    \caption{SAC Hyperparameter Importances on Brax Halfcheetah (left) and Humanoid (right).}
\end{figure}

\clearpage
\section{Partial Dependency Plots}
\label{app:pdps}
These plots show performance (lighter is better) across the value ranges of two hyperparameters. 
\subsection{SAC on Pendulum}
\begin{figure}[h]
    \centering
    \includegraphics[width=0.47\textwidth]{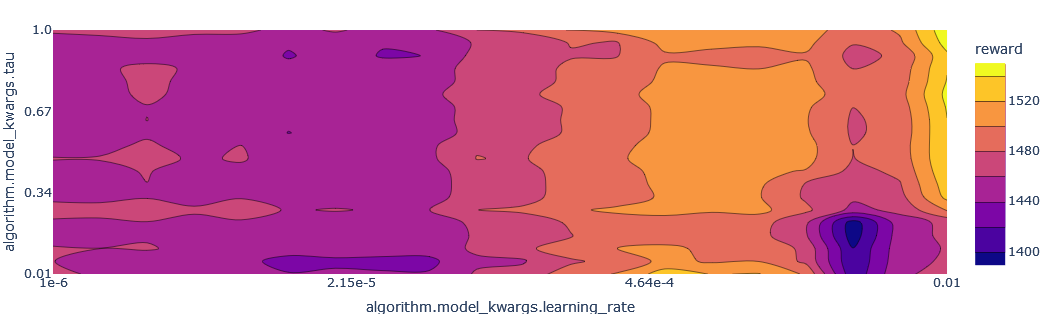}
    \includegraphics[width=0.47\textwidth]{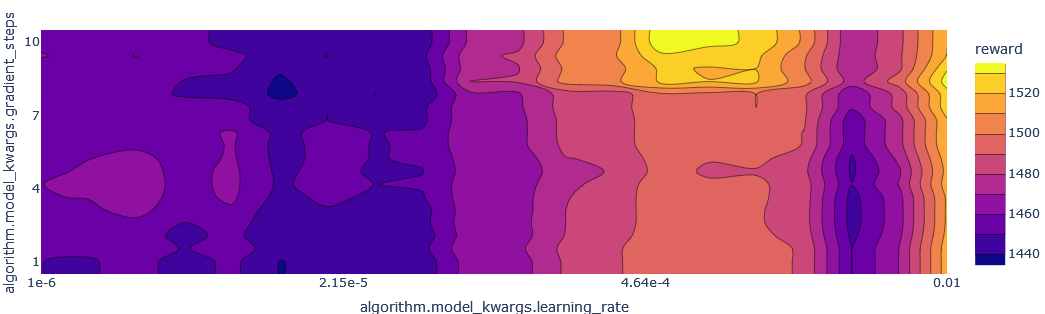}
    \includegraphics[width=0.47\textwidth]{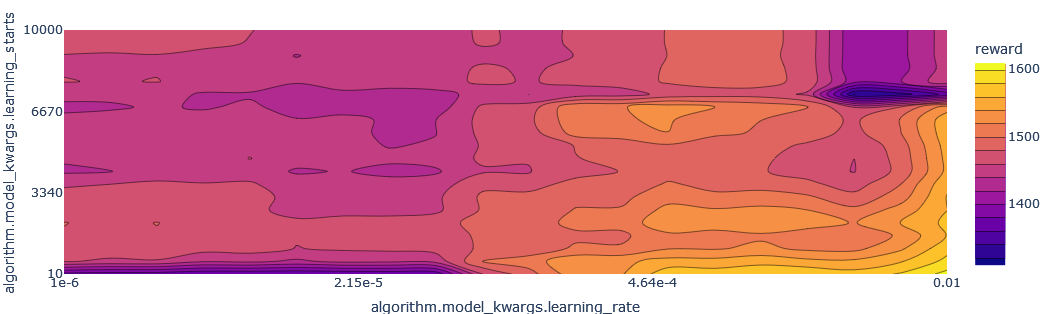}
    \includegraphics[width=0.47\textwidth]{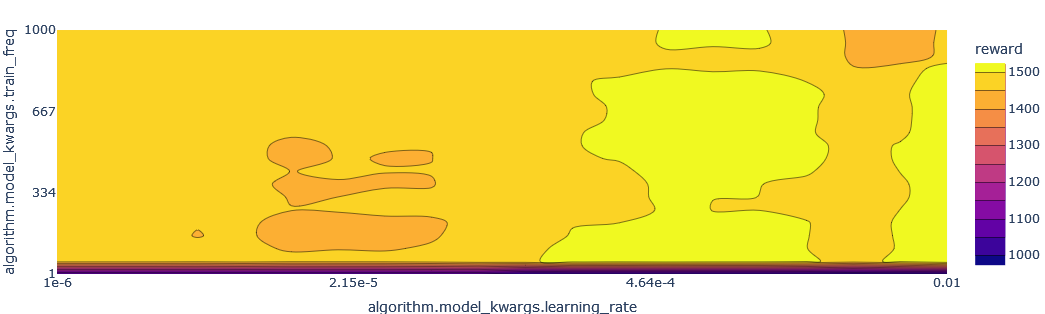}
    \includegraphics[width=0.47\textwidth]{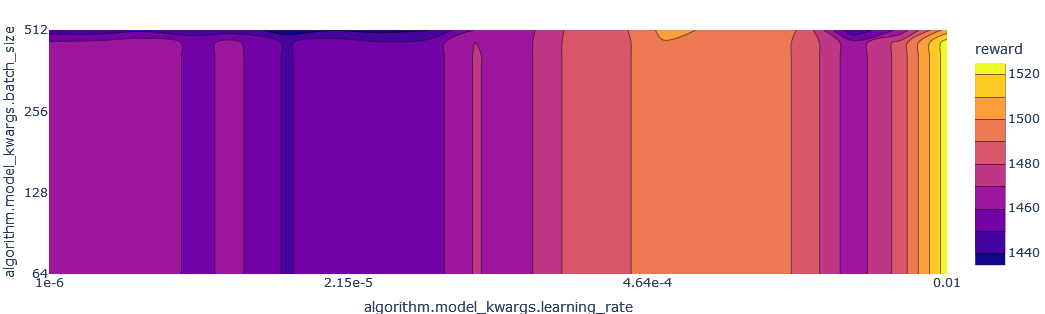}
    \includegraphics[width=0.47\textwidth]{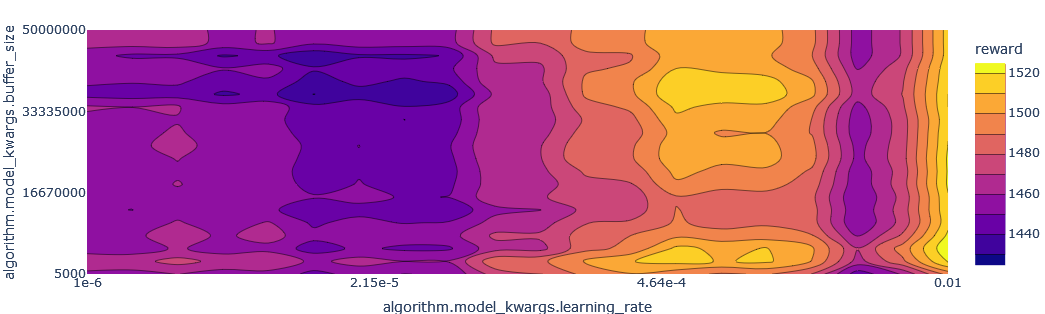}
    \includegraphics[width=0.47\textwidth]{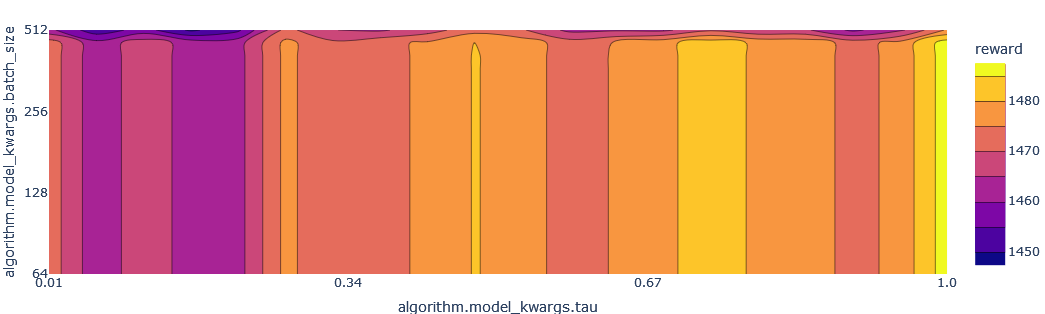}
    \includegraphics[width=0.47\textwidth]{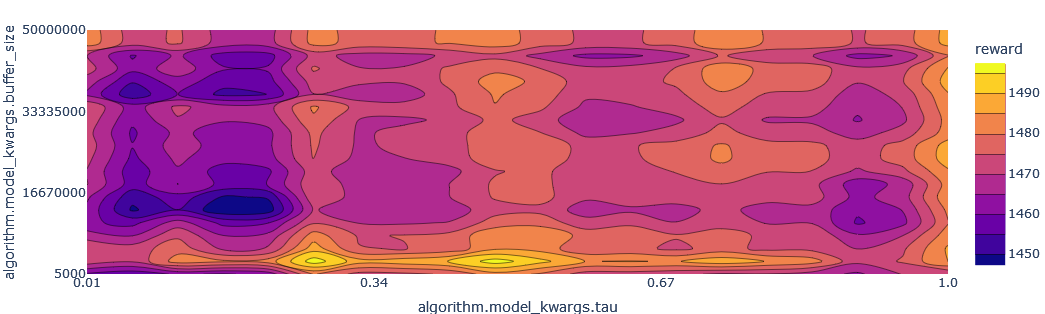}
    \includegraphics[width=0.47\textwidth]{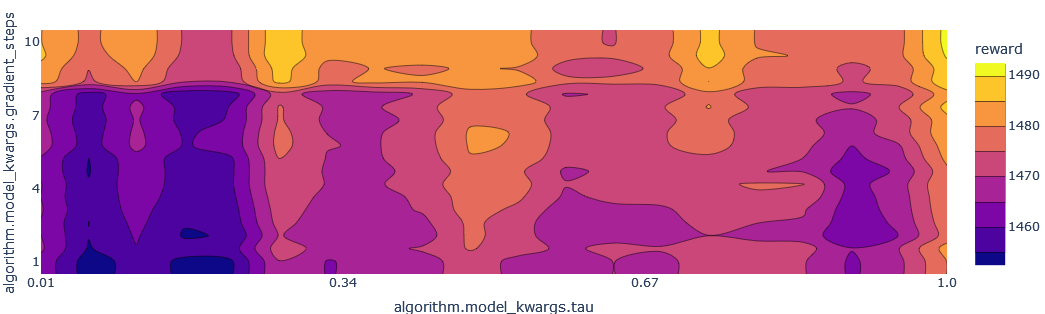}
    \includegraphics[width=0.47\textwidth]{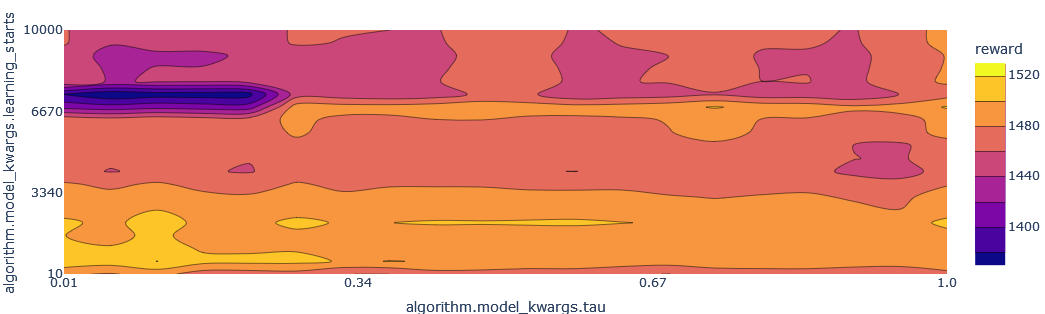}
    \includegraphics[width=0.47\textwidth]{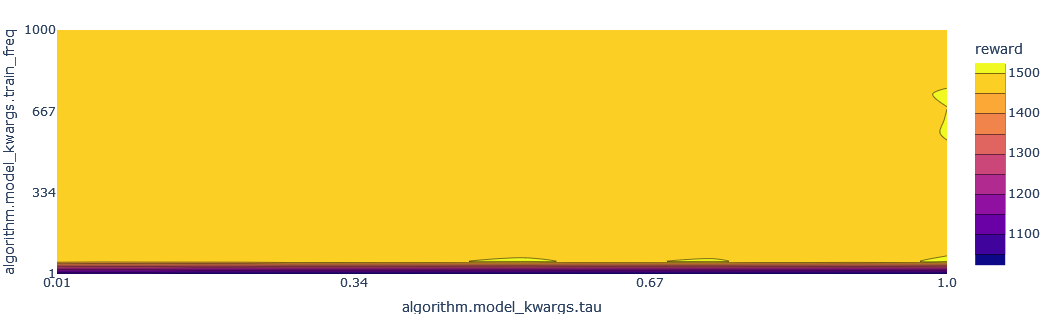}
    \includegraphics[width=0.47\textwidth]{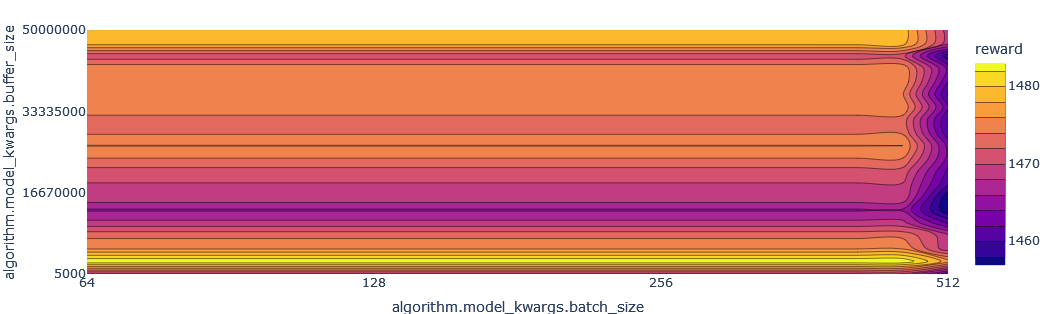}
\end{figure}

\begin{figure}
    \centering
    \includegraphics[width=0.47\textwidth]{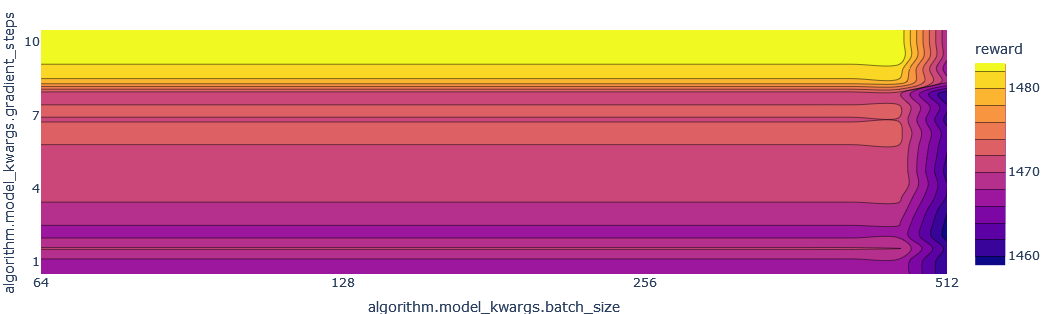}
    \includegraphics[width=0.47\textwidth]{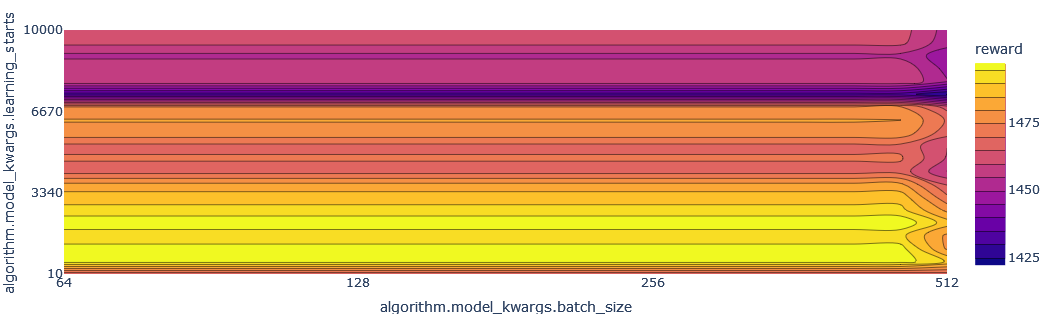}
    \includegraphics[width=0.47\textwidth]{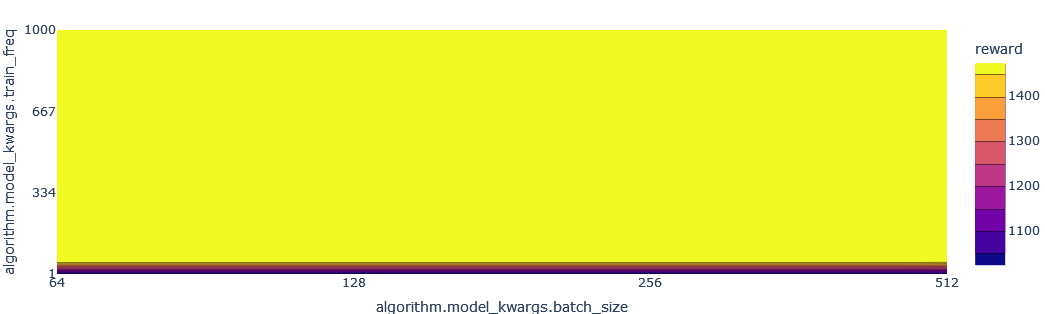}
    \includegraphics[width=0.47\textwidth]{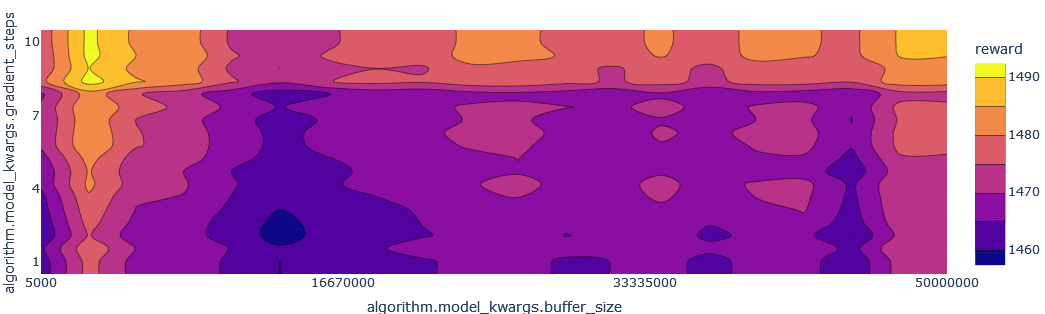}
    \includegraphics[width=0.47\textwidth]{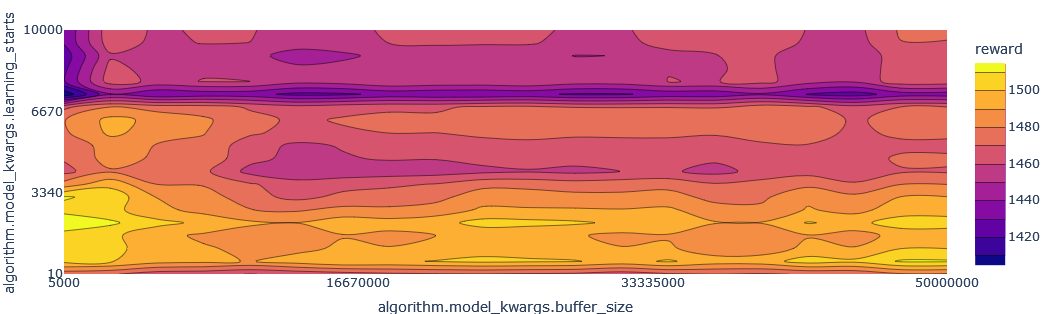}
    \includegraphics[width=0.47\textwidth]{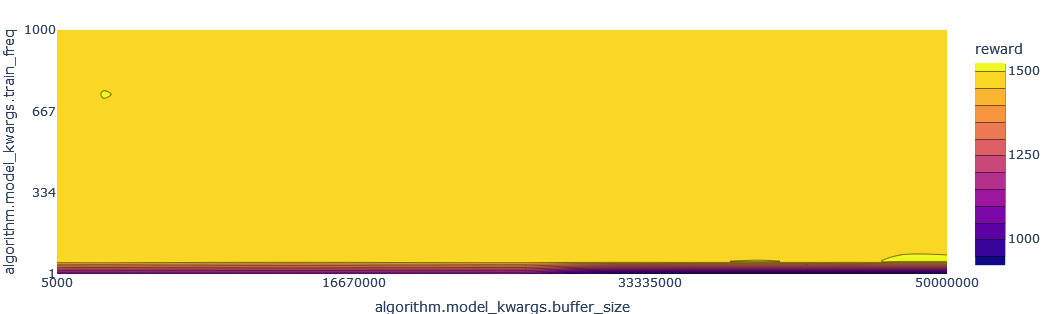}
    \includegraphics[width=0.47\textwidth]{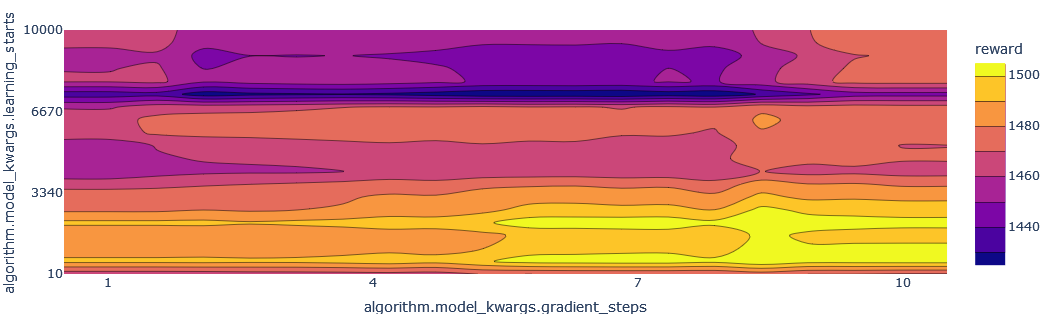}
    \includegraphics[width=0.47\textwidth]{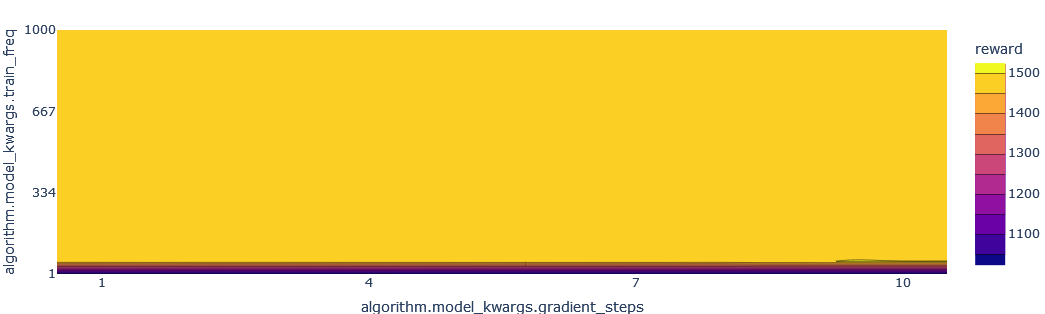}
    \includegraphics[width=0.47\textwidth]{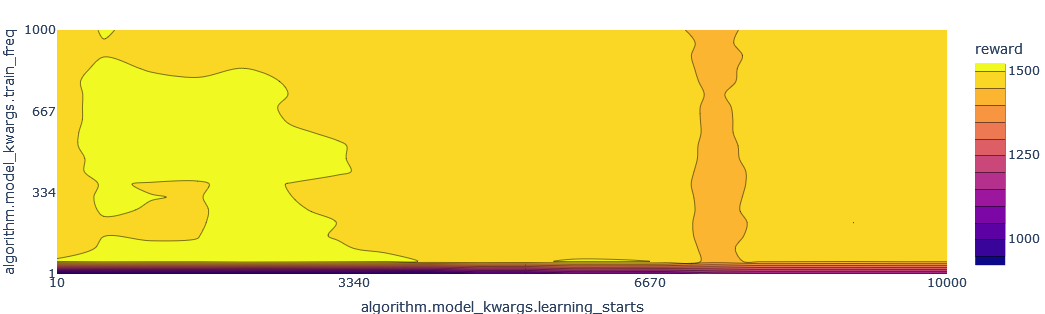}
\end{figure}
\newpage
\subsection{PPO on Acrobot}
\begin{figure}[h]
    \centering
    \includegraphics[width=0.47\textwidth]{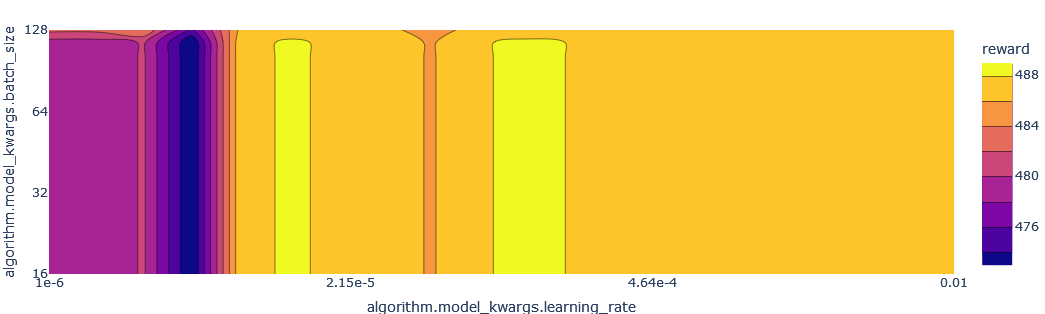}
    \includegraphics[width=0.47\textwidth]{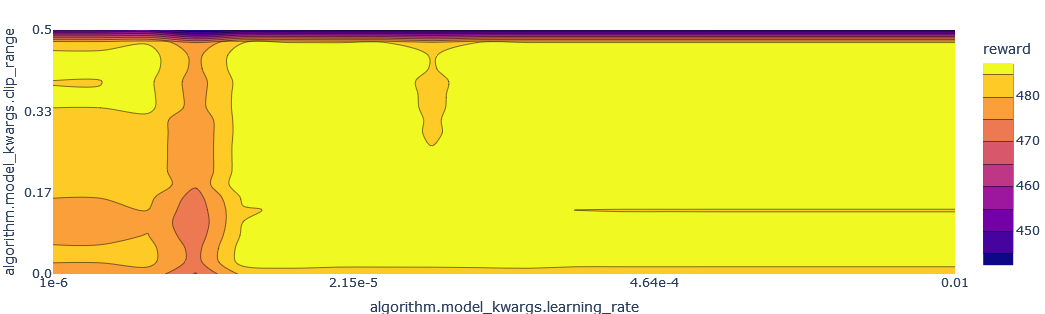}
    \includegraphics[width=0.47\textwidth]{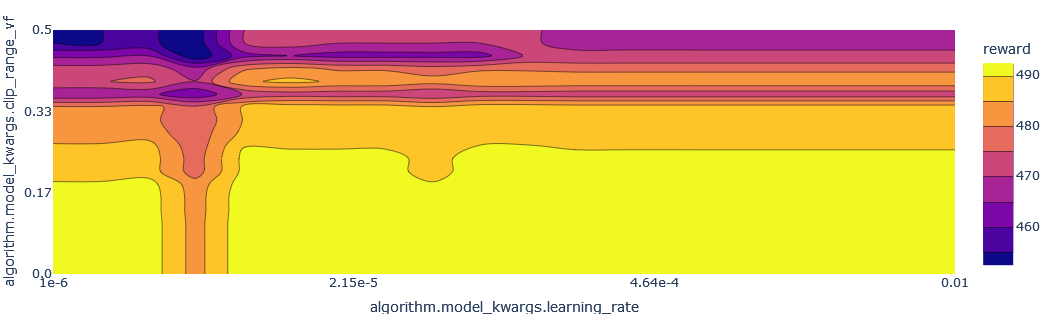}
    \includegraphics[width=0.47\textwidth]{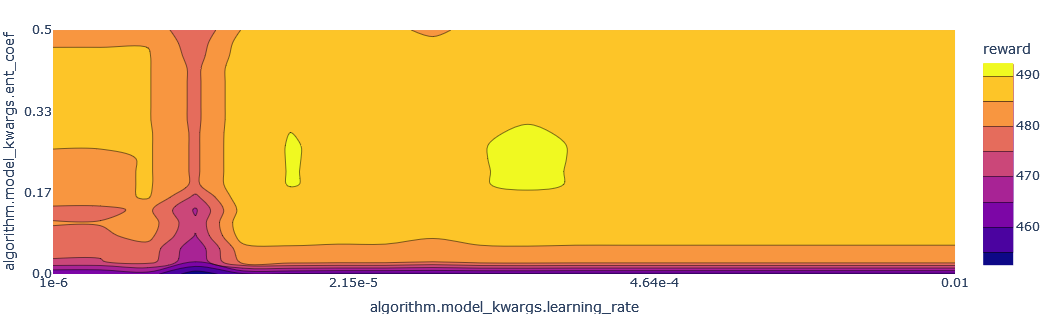}
    \includegraphics[width=0.47\textwidth]{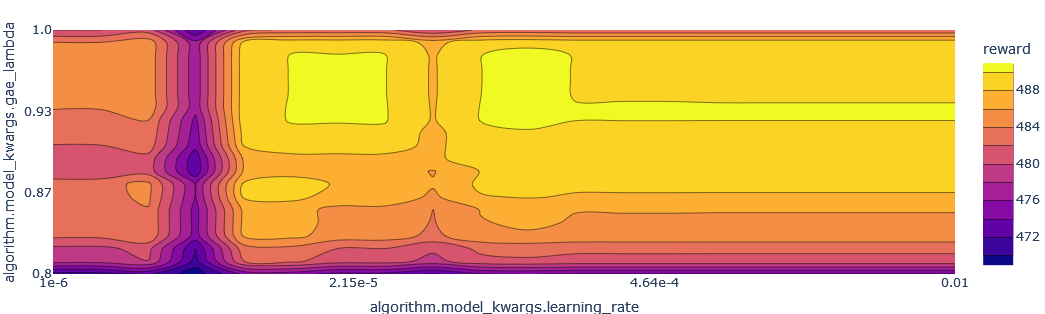}
    \includegraphics[width=0.47\textwidth]{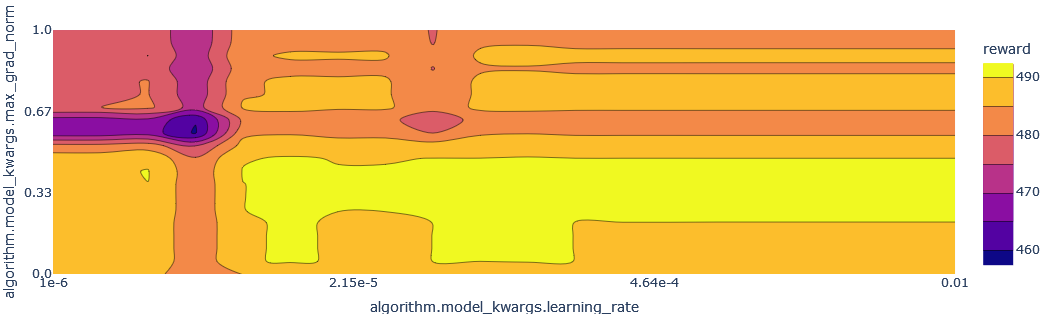}
    \includegraphics[width=0.47\textwidth]{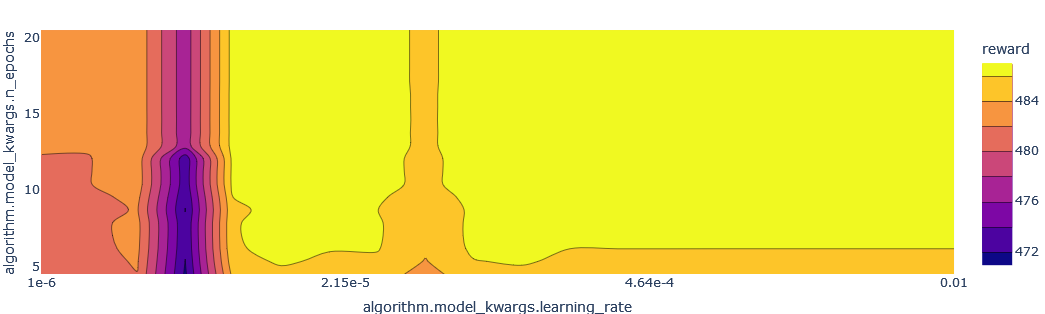}
    \includegraphics[width=0.47\textwidth]{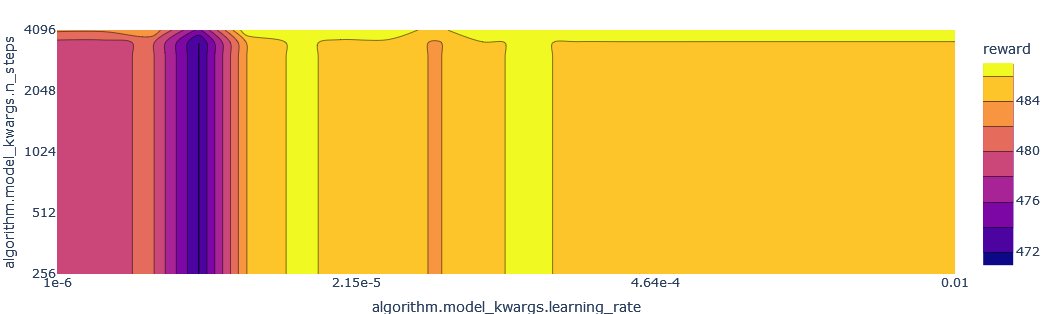}
    \includegraphics[width=0.47\textwidth]{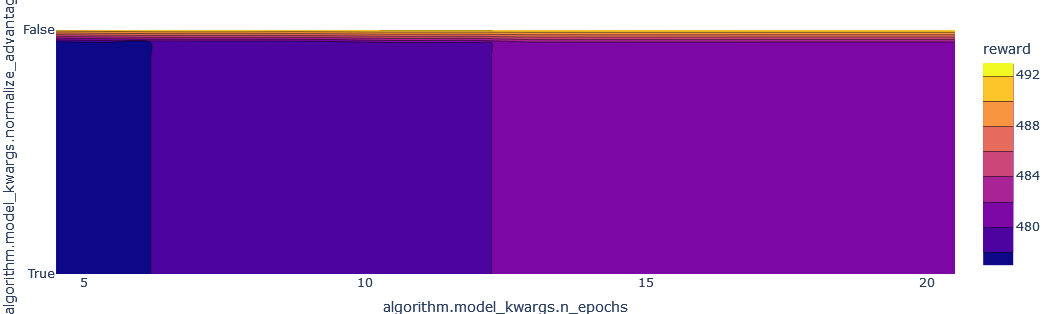}
    \includegraphics[width=0.47\textwidth]{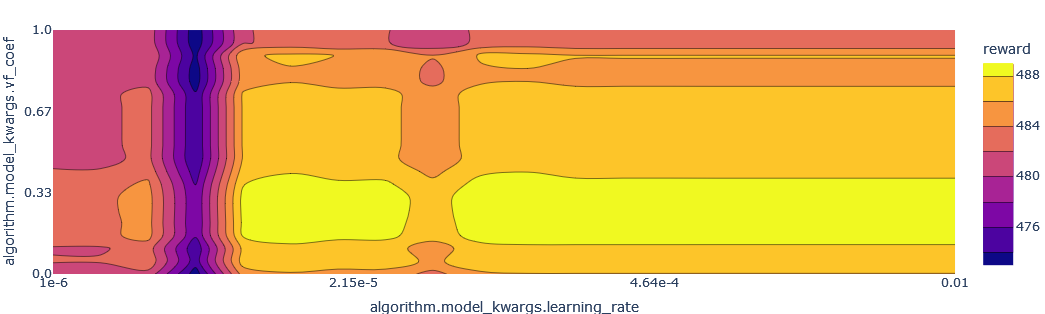}
    \includegraphics[width=0.47\textwidth]{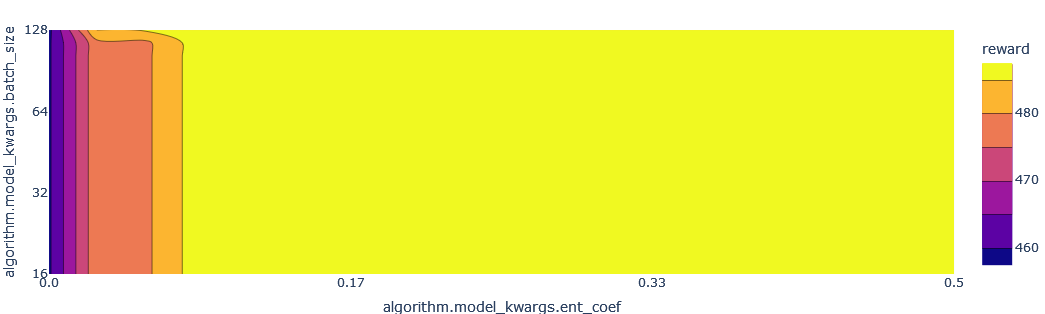}
    \includegraphics[width=0.47\textwidth]{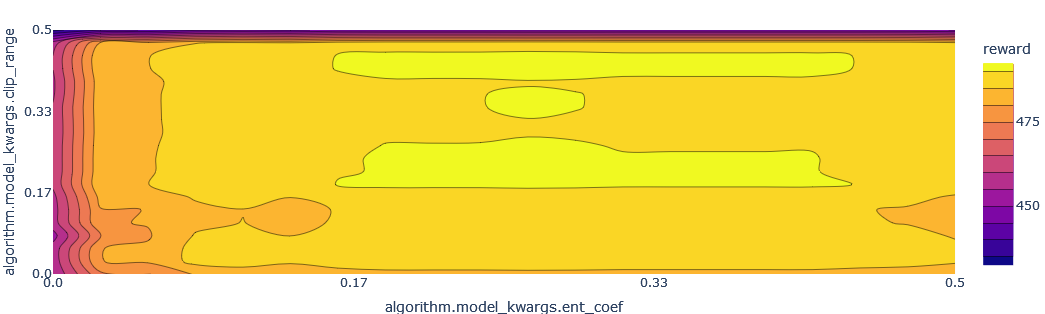}
    \includegraphics[width=0.47\textwidth]{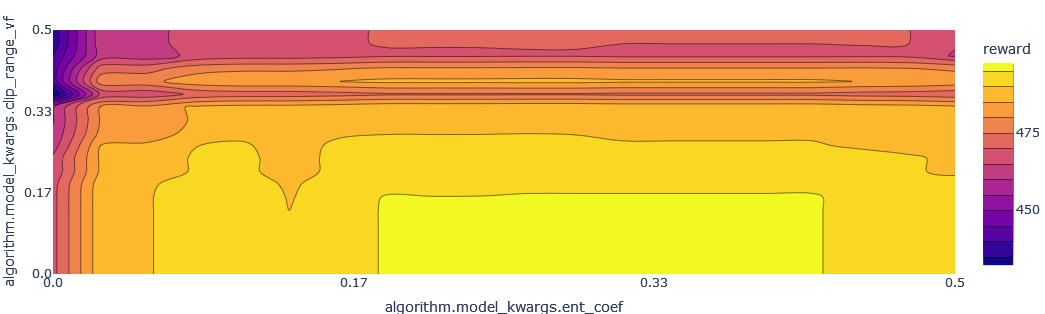}
    \includegraphics[width=0.47\textwidth]{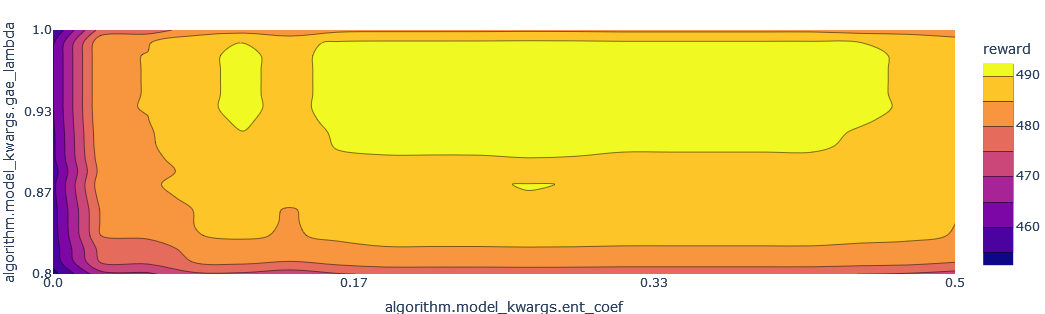}
\end{figure}

\begin{figure}
    \centering
    \includegraphics[width=0.47\textwidth]{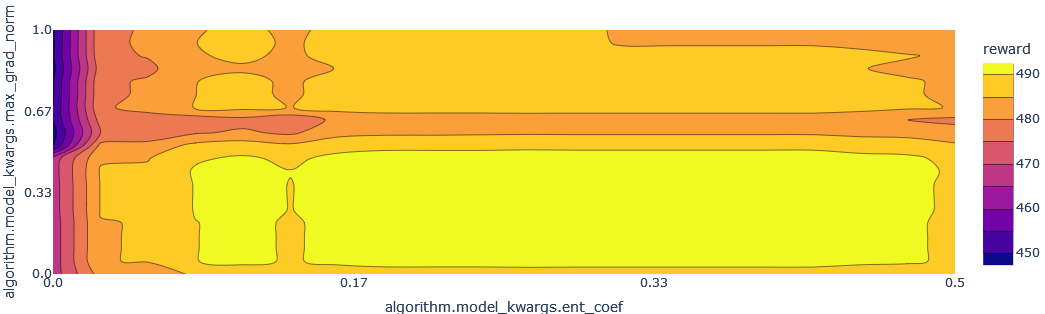}
    \includegraphics[width=0.47\textwidth]{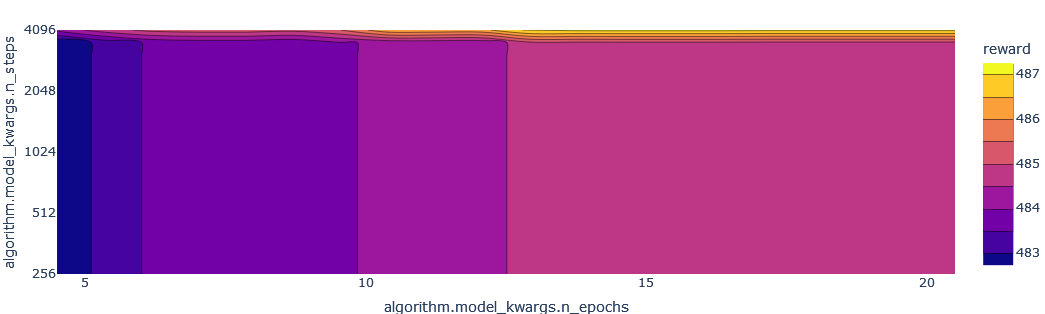}
    \includegraphics[width=0.47\textwidth]{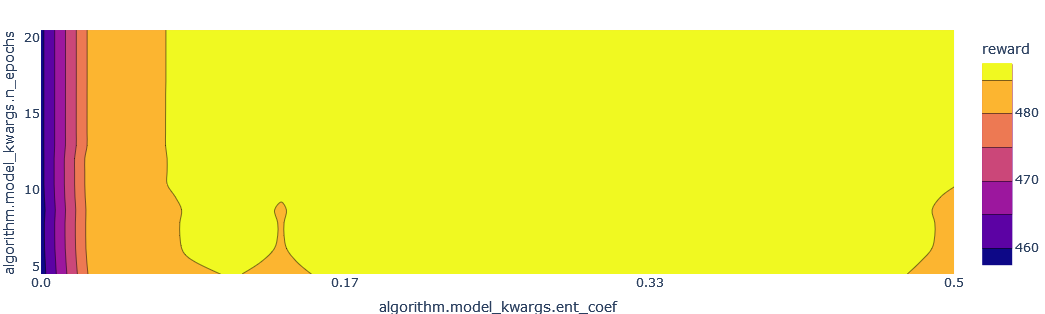}
    \includegraphics[width=0.47\textwidth]{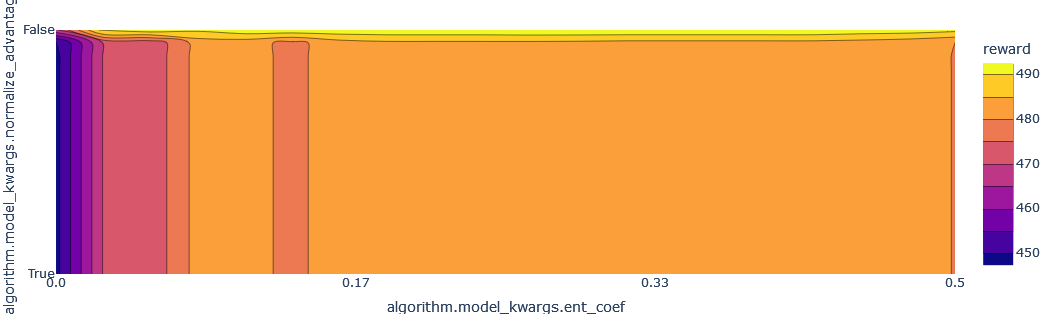}
    \includegraphics[width=0.47\textwidth]{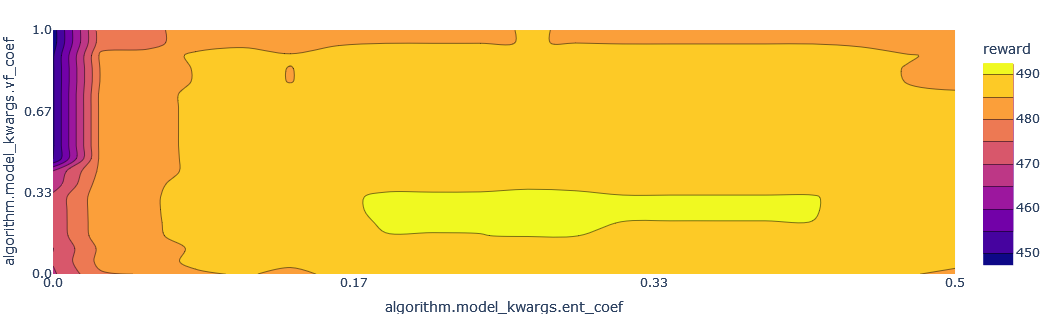}
    \includegraphics[width=0.47\textwidth]{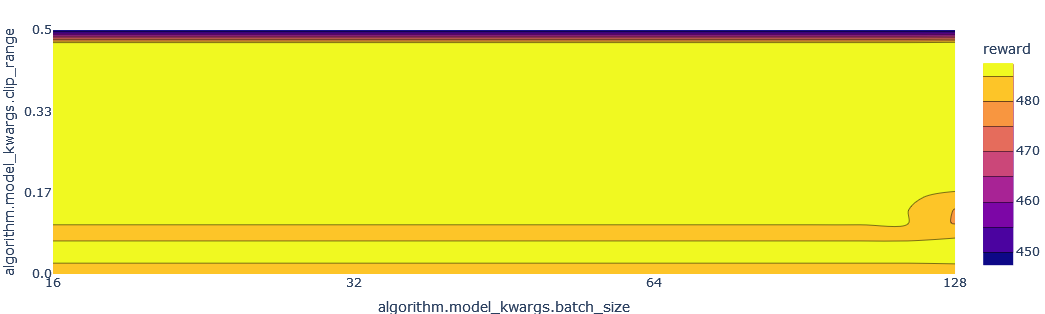}
    \includegraphics[width=0.47\textwidth]{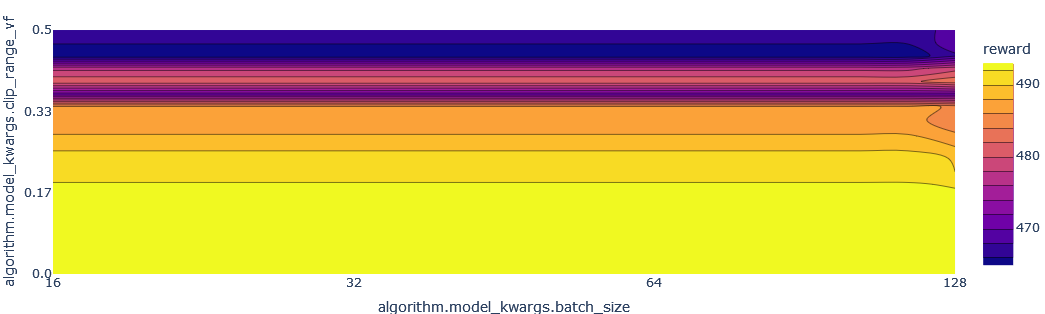}
    \includegraphics[width=0.47\textwidth]{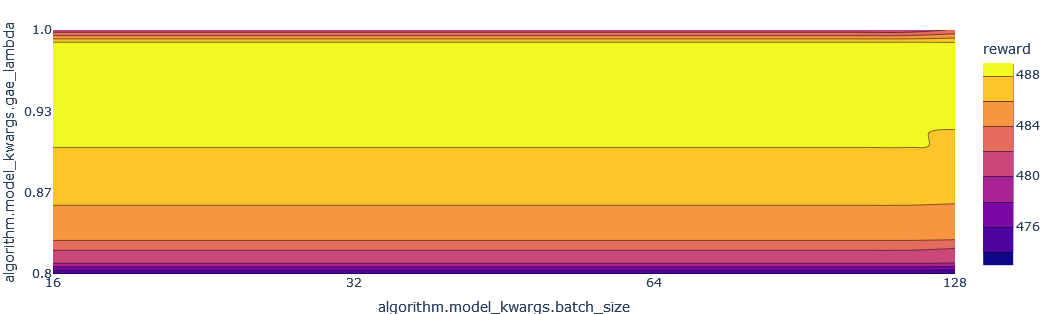}
    \includegraphics[width=0.47\textwidth]{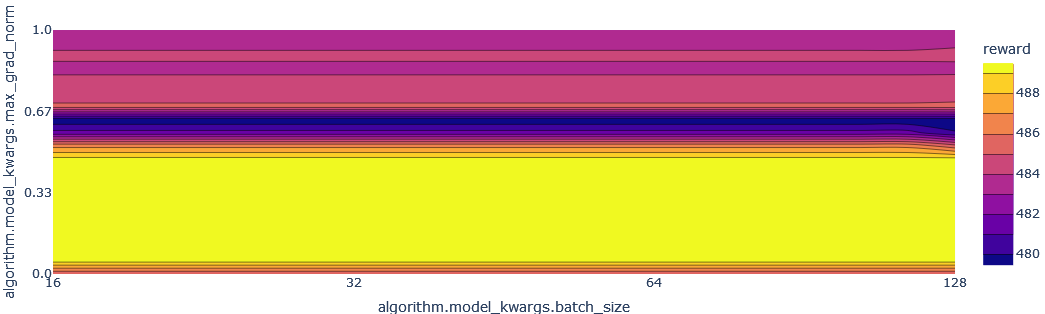}
    \includegraphics[width=0.47\textwidth]{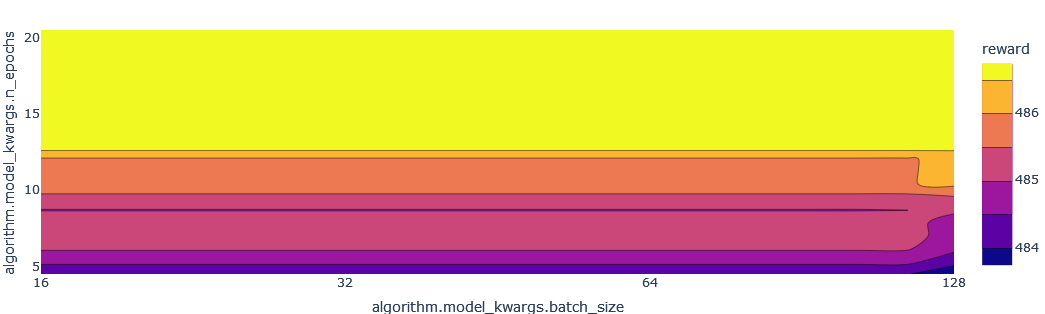}
    \includegraphics[width=0.47\textwidth]{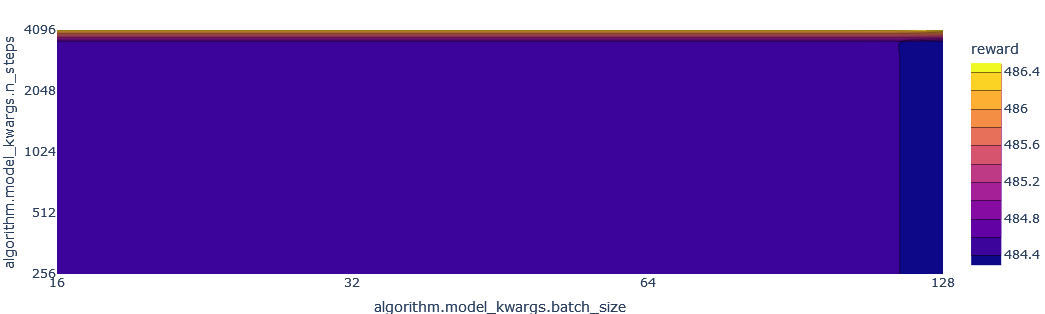}
    \includegraphics[width=0.47\textwidth]{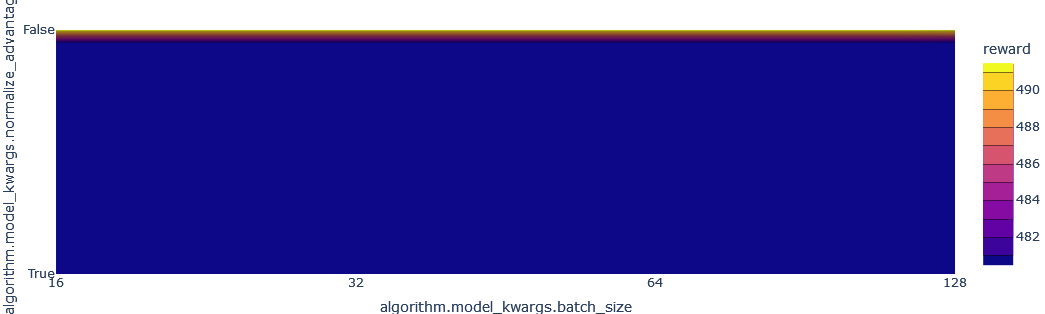}
    \includegraphics[width=0.47\textwidth]{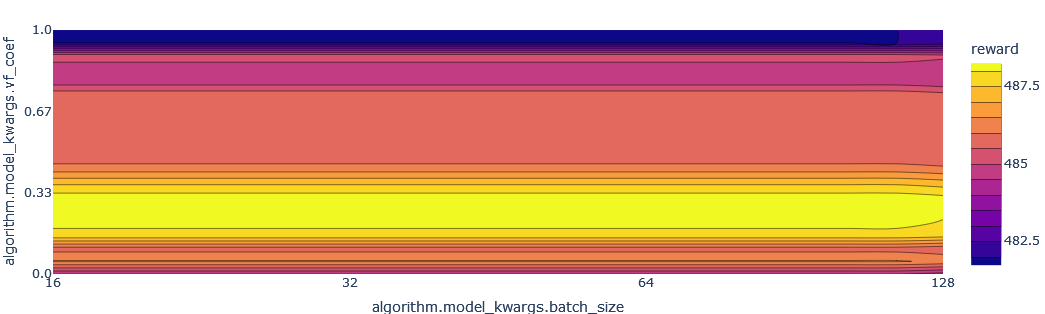}
    \includegraphics[width=0.47\textwidth]{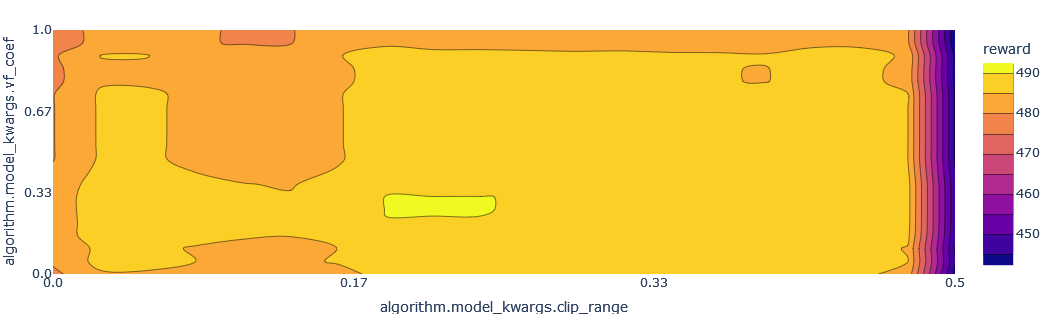}
\end{figure}

\begin{figure}
    \centering
    \includegraphics[width=0.47\textwidth]{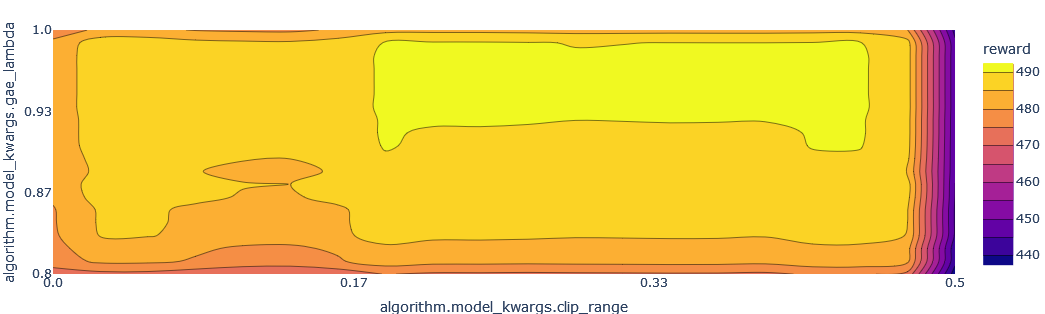}
    \includegraphics[width=0.47\textwidth]{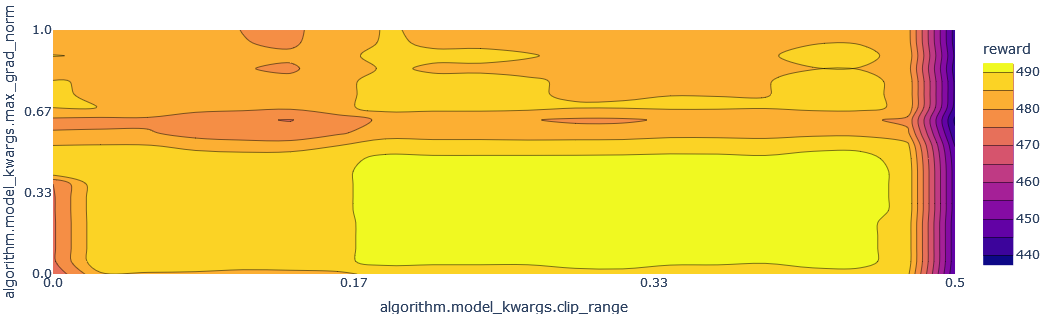}
    \includegraphics[width=0.47\textwidth]{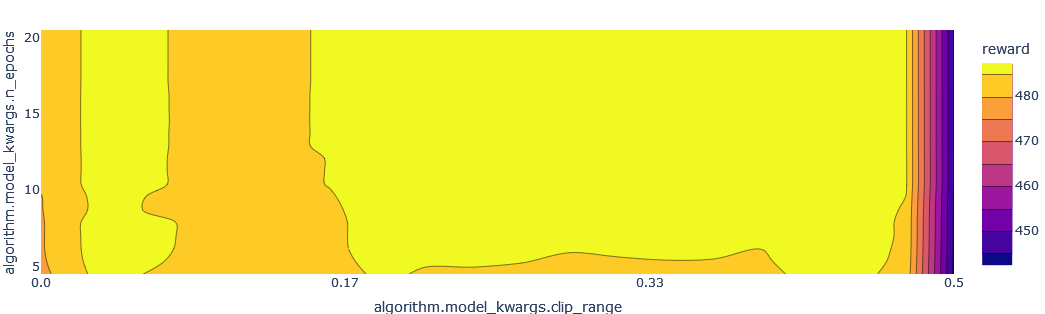}
    \includegraphics[width=0.47\textwidth]{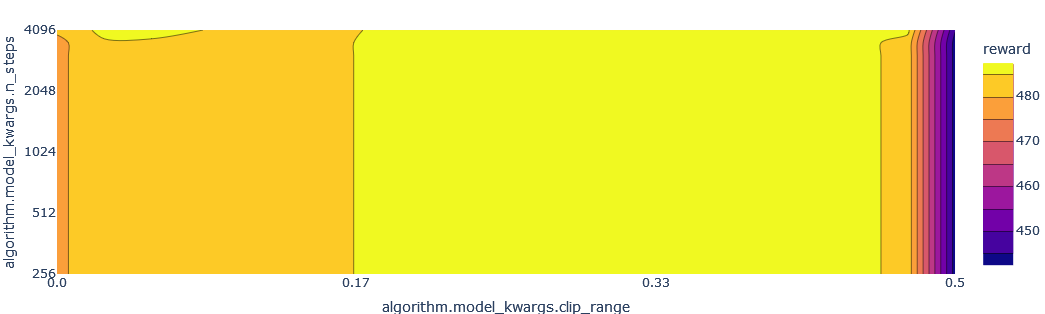}
    \includegraphics[width=0.47\textwidth]{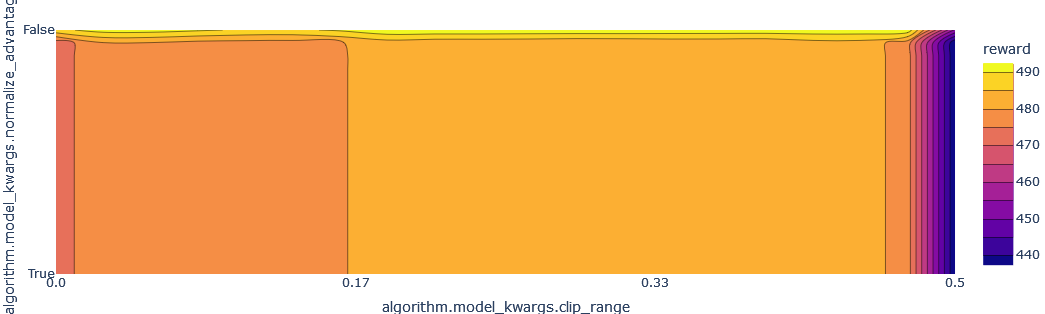}
    \includegraphics[width=0.47\textwidth]{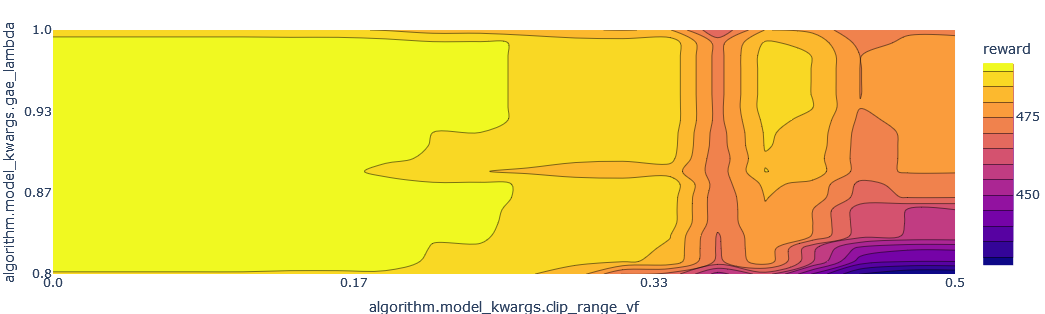}
    \includegraphics[width=0.47\textwidth]{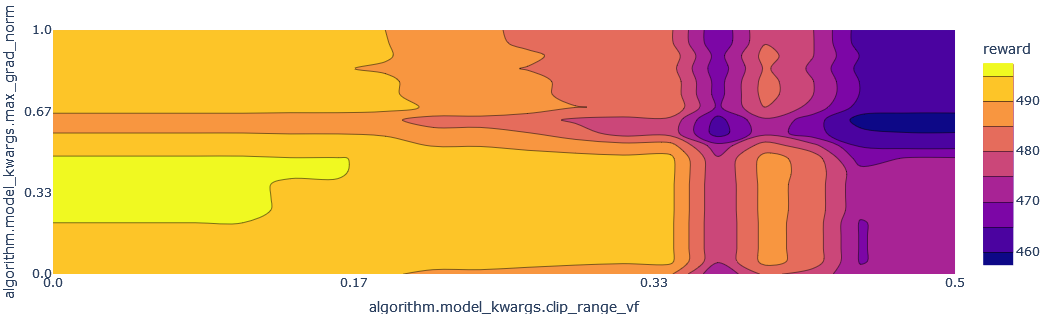}
    \includegraphics[width=0.47\textwidth]{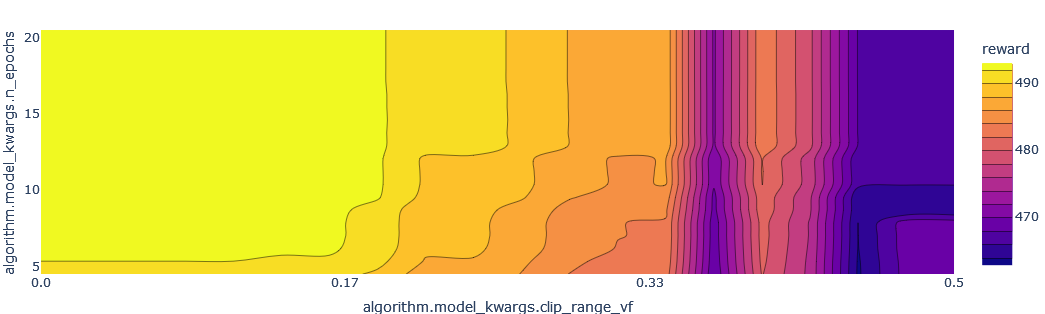}
    \includegraphics[width=0.47\textwidth]{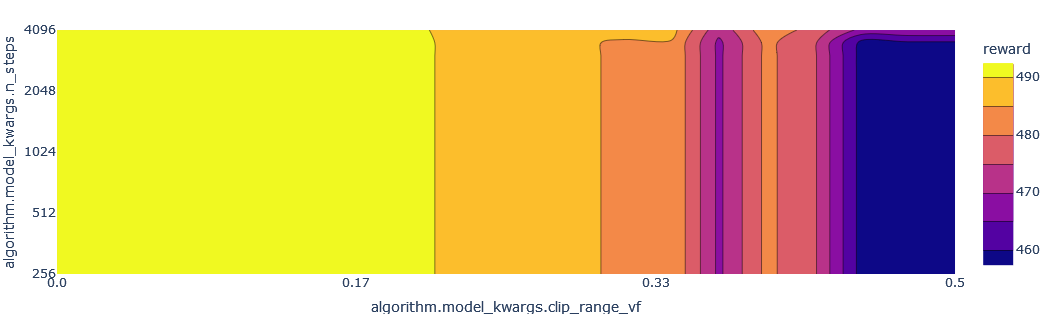}
    \includegraphics[width=0.47\textwidth]{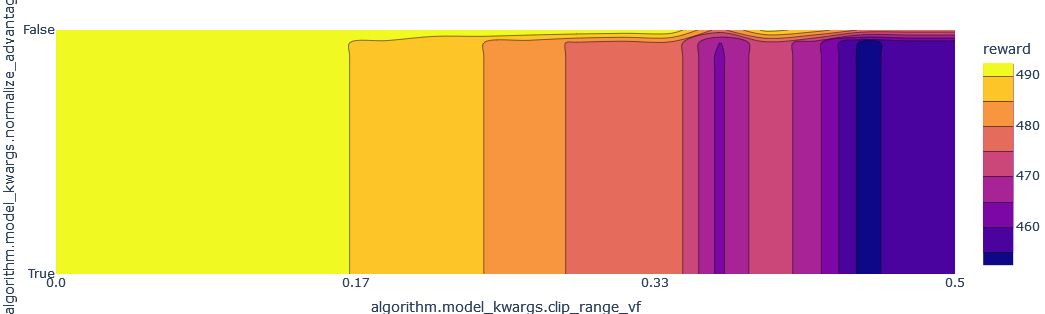}
    \includegraphics[width=0.47\textwidth]{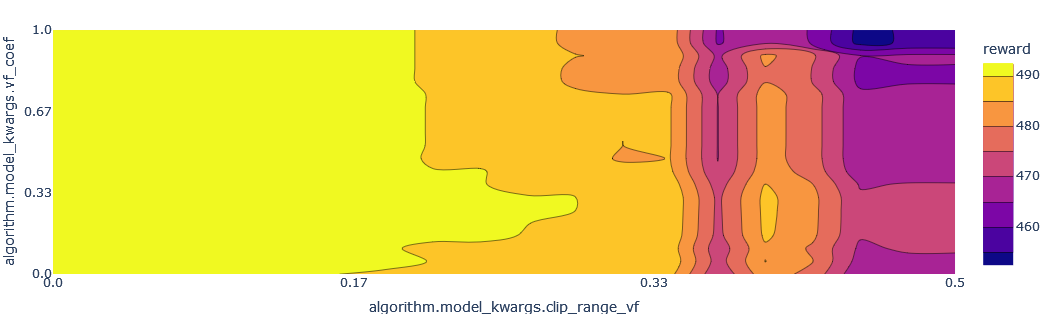}
    \includegraphics[width=0.47\textwidth]{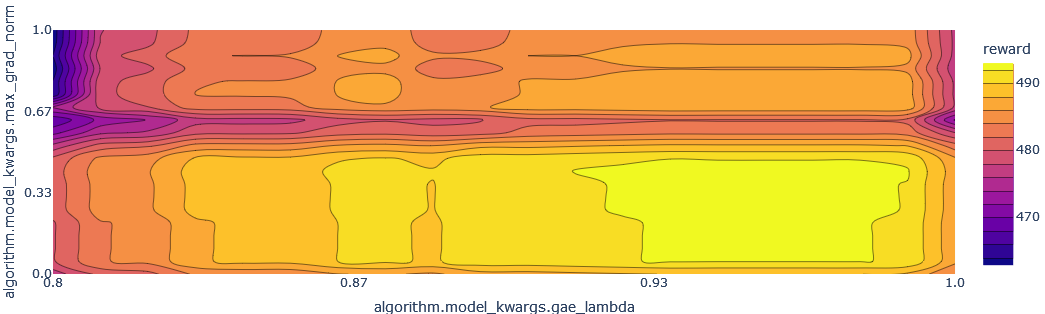}
    \includegraphics[width=0.47\textwidth]{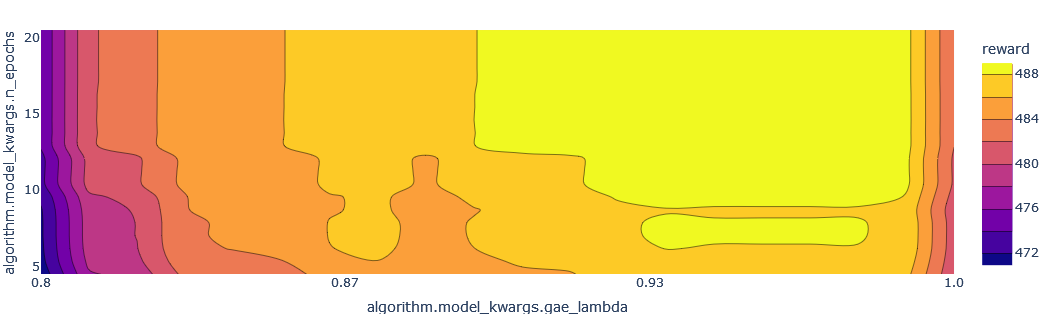}
    \includegraphics[width=0.47\textwidth]{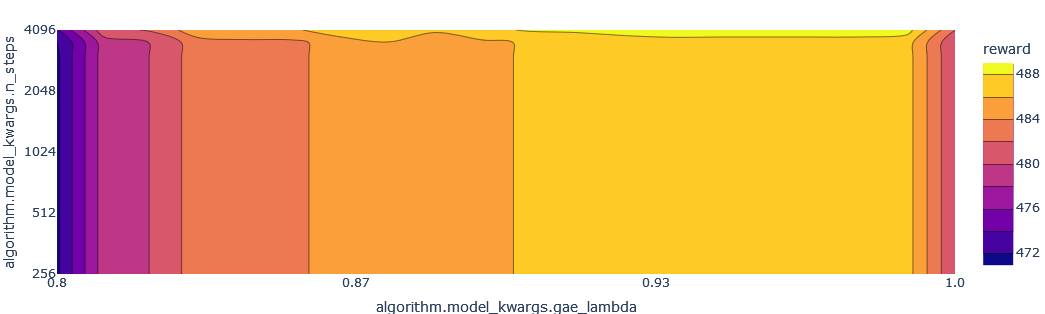}
\end{figure}

\begin{figure}
    \centering
    \includegraphics[width=0.47\textwidth]{figs/pdps/ppo_gae_n_steps.png}
    \includegraphics[width=0.47\textwidth]{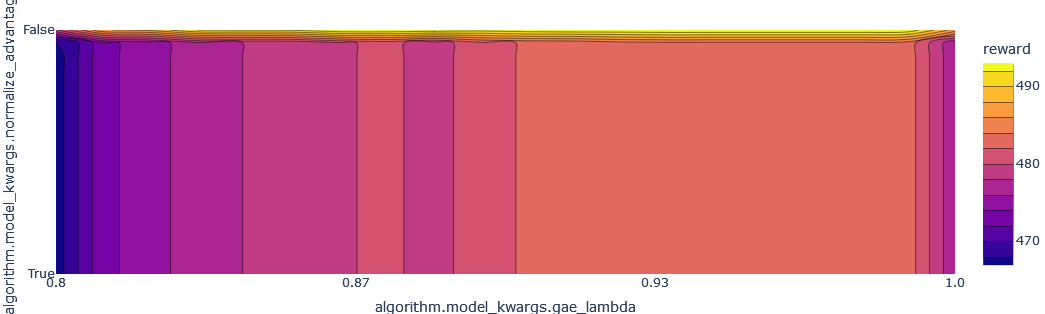}
    \includegraphics[width=0.47\textwidth]{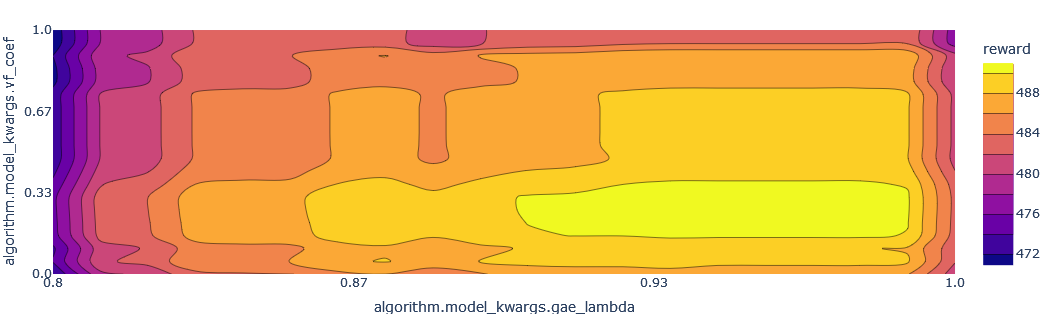}
    \includegraphics[width=0.47\textwidth]{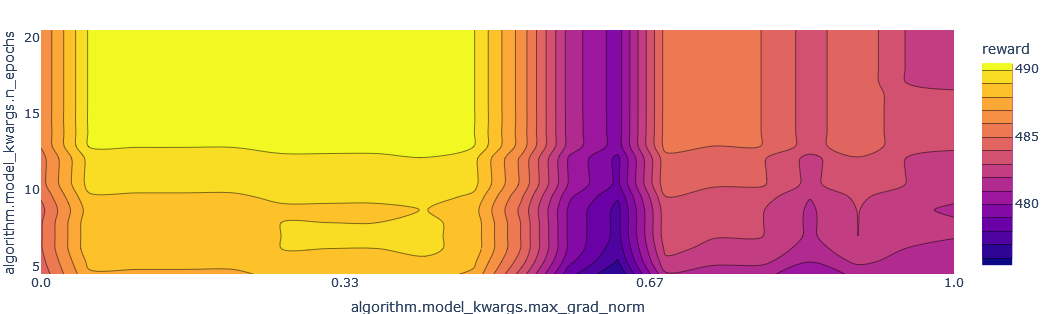}
    \includegraphics[width=0.47\textwidth]{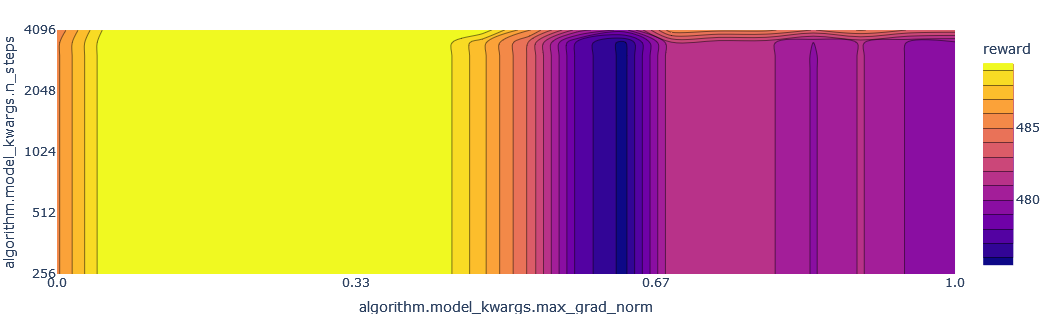}
    \includegraphics[width=0.47\textwidth]{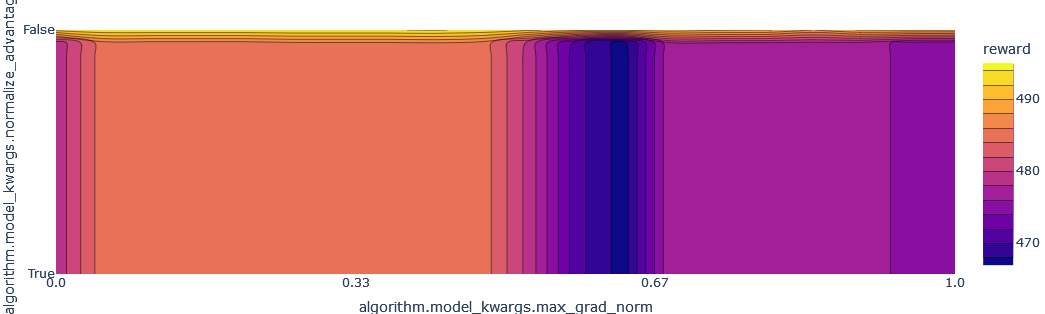}
    \includegraphics[width=0.47\textwidth]{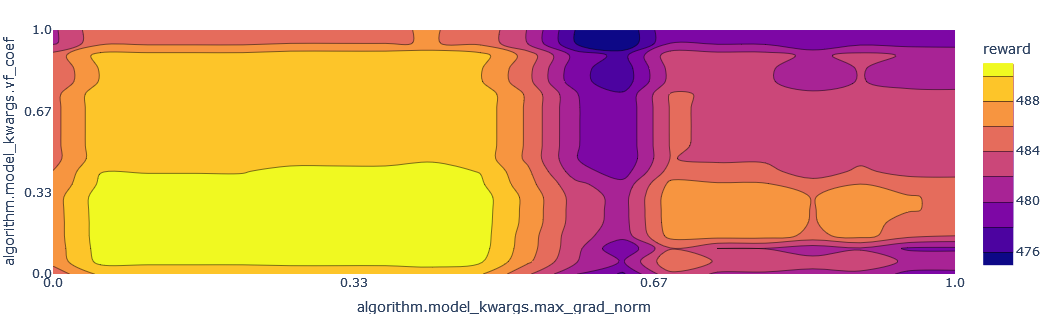}
    \includegraphics[width=0.47\textwidth]{figs/pdps/ppo_n_epochs_n_steps.png}
    \includegraphics[width=0.47\textwidth]{figs/pdps/ppo_n_epochs_norm_adv.png}
    \includegraphics[width=0.47\textwidth]{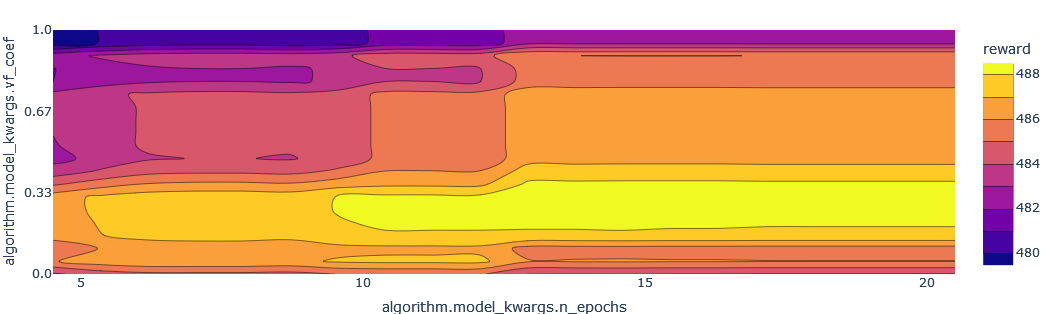}
    \includegraphics[width=0.47\textwidth]{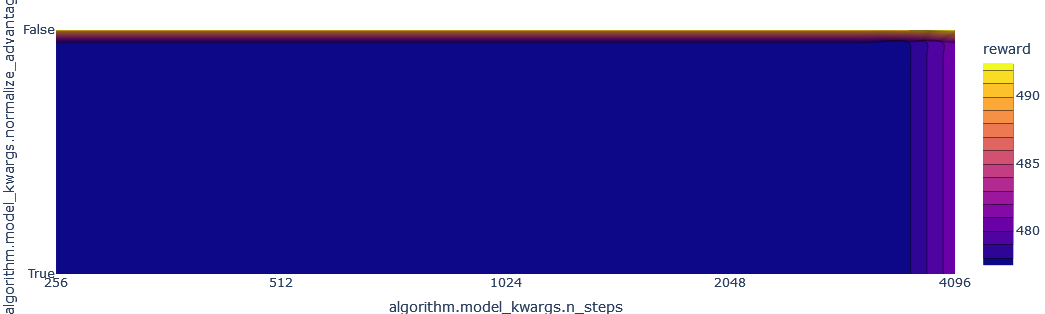}
    \includegraphics[width=0.47\textwidth]{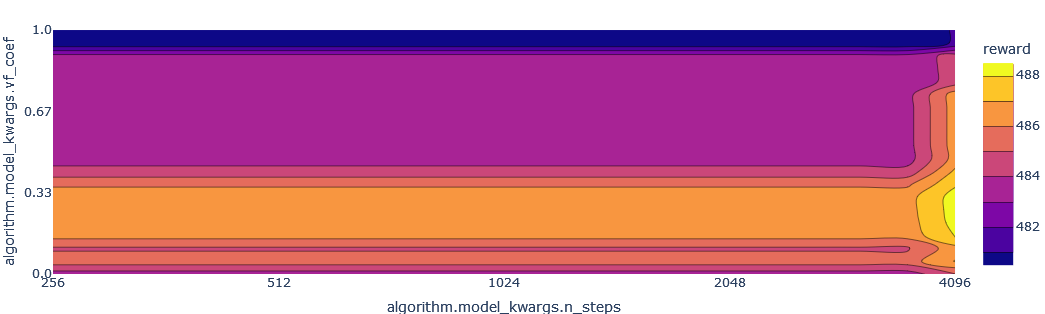}
    \includegraphics[width=0.47\textwidth]{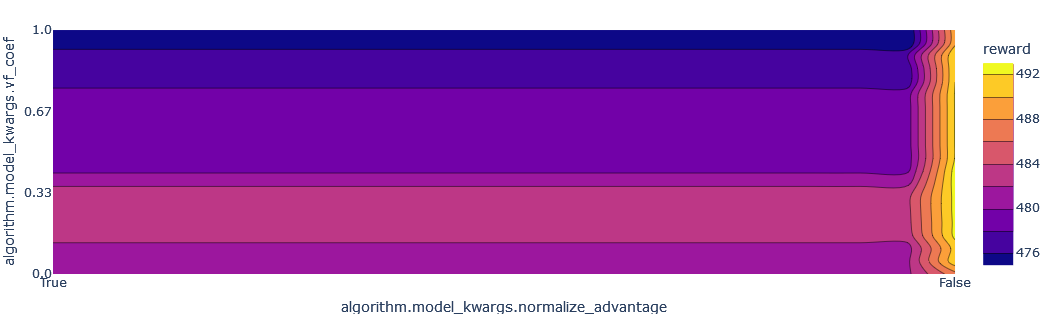}
\end{figure}
   

\end{document}